\documentclass[usenames, dvipsnames]{dissertation}
\usepackage{verbatim}
\usepackage[mathcal]{euscript}
\RequirePackage{fix-cm}
\usepackage[numbers]{natbib}
\usepackage{graphicx}
\usepackage{calrsfs}
\usepackage{bibentry}
\DeclareMathAlphabet{\pazocal}{OMS}{zplm}{m}{n}
\usepackage{setspace}
\usepackage{times}
\usepackage[normalem]{ulem}
\usepackage{enumerate}
\usepackage{paralist}
\usepackage{fancyvrb}
\usepackage{amsmath,mathtools}
\usepackage{amssymb}
\usepackage{amsthm}
\newtheorem{theorem}{Theorem}
\newtheorem{definition}{Definition}
\newtheorem{case}{Case}[section]

\newtheorem{lemma}{Lemma}
\newtheorem{example}{Example}
\usepackage{bm}
\usepackage{subfig}
\usepackage{rotating}
\usepackage{graphics}
\usepackage{url}
\usepackage{algorithm}
\usepackage{algpseudocode}
\usepackage{array}
\usepackage{listings}
\usepackage{transparent}
\usepackage{wasysym}
\usepackage{longtable}
\usepackage{collcell}
\usepackage{booktabs}
\usepackage[toc, acronyms]{glossaries}

\usepackage{supertabular}
\usepackage{enumerate}
\usepackage{enumitem}
\usepackage{tl_macros}
\definecolor{light-gray}{gray}{0.1}
\definecolor{tck-grey}{cmyk}{0,0,0,0.34}
\usepackage{tikz}
\usetikzlibrary{arrows,backgrounds,calc,chains,intersections,matrix,positioning,scopes}

\usepackage{verbatim}
\usepackage{natbib}

\usepackage{afterpage}
\usepackage{tikz}
\usetikzlibrary{arrows,calc,chains,matrix,positioning,scopes,shapes}

\colorlet{hgreen}{black!30!green}
\colorlet{hred}{black!10!red}
\colorlet{hgrey}{gray!10!white}

\newcommand{\hlp}[2][pink]{ {\sethlcolor{#1} \hl{#2}} }
\newcommand{\hlg}[2][hgreen]{ {\sethlcolor{#1} \hl{#2}} }
\newcommand{\hlr}[2][hred]{ {\sethlcolor{#1} \hl{#2}} }

\colorlet{discharged}{black!30!green}
\colorlet{violated}{black!10!red}

\usepackage{bibentry}

\usepackage{fixme}
\fxsetup{
    status=draft,
    author={Note},
    layout=inline, 
    theme=color
}
\FXRegisterAuthor{tck}{atck}{TCK}
\usepackage{collectbox}


\usepackage{soul}
\usepackage{color}

\DeclareMathAlphabet{\mathpzc}{OT1}{pzc}{m}{it}
\DeclareMathAlphabet{\mathcalligra}{T1}{calligra}{m}{n}

\newcommand{\GovStruct}{Multi-level Governance Institution}
\newcommand\myalign[1]{\begin{align*}#1\end{align*}}

\usepackage{thmtools}
\renewcommand\thmcontinues[1]{Continued}

\newsavebox\mytempbib 

\DeclareCaptionFormat{cont}{#1 (cont.)#2#3\par}
\begin{document}

\title[A Formal Framework for Analysing Institutional Design and Enactment Governance]{Governing Governance}
\author{Thomas Christopher}{King}


\begin{titlepage}

\begin{center}

\vspace*{2\bigskipamount}

{\makeatletter
\titlestyle\bfseries\LARGE\@title
\makeatother}

{\makeatletter
\ifx\@subtitle\undefined\else
    \bigskip
    \titlefont\titleshape\Large\@subtitle
\fi
\makeatother}

\end{center}

\cleardoublepage
\thispagestyle{empty}

\begin{center}


\vspace*{2\bigskipamount}

{\makeatletter
\titlestyle\bfseries\LARGE\@title
\makeatother}

{\makeatletter
\ifx\@subtitle\undefined\else
    \bigskip
    \titlefont\titleshape\Large\@subtitle
\fi
\makeatother}

\vfill


{\Large\titlefont\bfseries Proefschrift}

\bigskip
\bigskip

ter verkrijging van de graad van doctor

aan de Technische Universiteit Delft,

op gezag van de Rector Magnificus Prof. Ir. K.Ch.A.M. Luyben,

voorzitter van het College voor Promoties,

in het openbaar te verdedigen op donderdag 27 oktober 2016 om 10.00 uur

\bigskip
\bigskip

door

\bigskip
\bigskip

\makeatletter
{\Large\titlefont\bfseries\@firstname\ {\titleshape\@lastname}}
\makeatother

\bigskip
\bigskip

Bachelor of Science in Computer Science from King's College London, United Kingdom \\
geboren te Londen, Verenigd Koningkrijk

\vspace*{2\bigskipamount}

\end{center}

\clearpage
\thispagestyle{empty}

\noindent This dissertation has been approved by the promotors:

\medskip\noindent
\begin{tabular}{l}
    Promotor: Prof.\ dr.\ C. M.\ Jonker \\
    Copromotor: Dr.\ M. V. \ Dignum \\
    Copromotor: Dr.\ M. B. \ van Riemsdijk
\end{tabular}

\bigskip
\noindent Composition of the doctoral committee:

\medskip\noindent
\begin{tabular}{p{5cm}l}
    Rector Magnificus, & chairman \\
    Prof.\ dr.\ C. M. \ Jonker, & Delft University of Technology, promotor \\
    Dr.\ M. V. \ Dignum, & Delft University of Technology, copromotor \\
    Dr.\ M. B. \ van Riemsdijk, & Delft University of Technology, copromotor \\
    Independent members: \\
    Prof.\ dr.\ K. \ Atkinson, & University of Liverpool \\
    Prof.\ dr.\ F. \ Brazier, & Delft University of Technology \\
    Prof.\ dr.\ J. \ Pitt, & Imperial College London \\
    Prof.\ dr.\ L. \ van der Torre, & University of Luxembourg \\
    Reserve: \\
    Prof. dr. C. Witteveen & Delft University of Technology
\end{tabular}


\vfill
\begin{center}
    \includegraphics[scale=0.8]{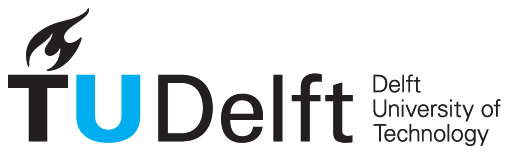}
    \hspace{2em}
    \includegraphics[scale=0.2]{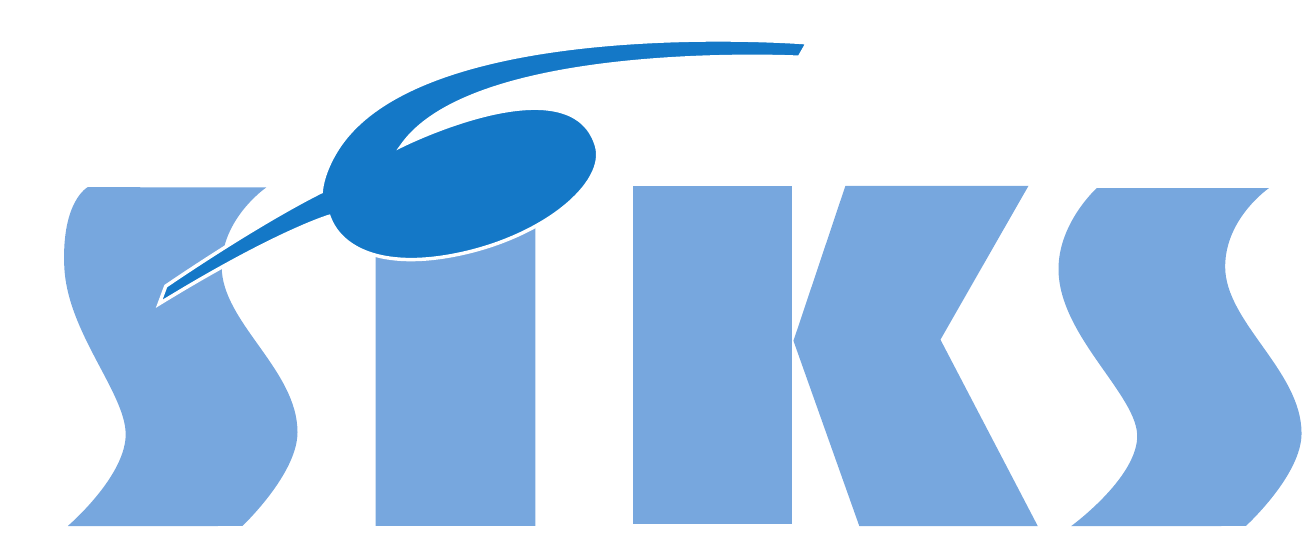}
    \hspace{2em}
    \includegraphics[scale=0.3]{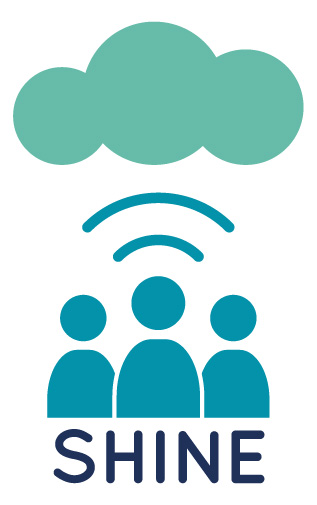}
\end{center}
\vfill

\noindent SIKS Dissertation Series No. 2016-41 \\
The research reported in this thesis has been carried out under the auspices of SIKS, the Dutch Research School for Information and Knowledge Systems. \\
The research reported in this thesis was funded by the SHINE project of TU Delft.

\vspace{4\bigskipamount}

\noindent Copyright \textcopyright\ 2016 by Thomas~C.~King


\medskip
\noindent ISBN 978-94-6186-726-1

\medskip
\noindent An electronic version of this dissertation is available at \\
\url{http://repository.tudelft.nl/}

\end{titlepage}

\cleardoublepage
\pagenumbering{roman}
\chapter*{Summary}
\addcontentsline{toc}{chapter}{Summary}
\setheader{Summary}


This dissertation is motivated by the need, in today's globalist world, for a precise way to enable governments, organisations and other regulatory bodies to evaluate the constraints they place on themselves and others. An organisation's modus operandi is enacting and fulfilling contracts between itself and its participants. Yet, organisational contracts \textit{should} respect external laws, such as those setting out data privacy rights and liberties. Contracts \textit{can} only be enacted by following contract law processes, which often require bilateral agreement and consideration. Governments need to legislate whilst understanding today's context of national and international governance hierarchy where law makers shun isolationism and seek to influence one another. Governments \textit{should} avoid punishment by respecting constraints from  international treaties and human rights charters. Governments \textit{can} only enact legislation by following their own, pre-existing, law making procedures. In other words, institutions, such as laws and contracts are designed and enacted under constraints.

The common thread shared by these examples is that institution designers, such as organisations and governments, are constrained in two senses. They are loosely tied in how the institutions they enact \textit{should} be designed and strongly tied in how and when they \textit{can} enact institutional changes. We can clearly see such constraints exist in written form, even though we cannot physically see how the machinery that applies and tightens those constraints around the institutional designers works. It is consequently hard to grasp exactly how institution designers need to operate under these constraints.

This dissertation addresses this issue by contributing a formal framework for analysing institutional design and enactment governance. Through formalisation, the framework provides a mathematically rigorous account of constraints placed on institutional designs and enacting institutional changes. Hence, the unseen constraints become seeable not as physical bindings, but as symbols on a page defining general institutional reasoning. From a conceptualisation standpoint, the main benefit is that we are able to understand the institutional constraints and identify any conceptual flaws. Or, in the words of Leibniz, ``The only way to rectify our reasonings is to make them as tangible as those of the Mathematicians, so that we can find our error at a glance''. 

There are three main practical benefits. Firstly, the formal reasoning is disseminated in clear mathematical language. Institution designers and judiciaries apply exactly the same reasoning, making judgements predictable, providing they share the same legal rule bases. Moreover, citizens can have common knowledge of when regulatory changes are enacted by applying the same institutional reasoning. Secondly, the framework facilitates automation of otherwise cognitively difficult tasks in understanding the constraints placed on institution designers. Thirdly, the framework deals with the pragmatics with applications to real world case studies, in order to capture the meaning of institutional constraints as they are used.

In summary, this dissertation introduces the PARAGon framework for \textbf{P}ractical \textbf{A}utomated \textbf{R}easoning for \textbf{A}ssessing \textbf{Go}ver\textbf{n}ance of institution design and change. 

Central to the PARAGon framework is Searle's well known institutional constitutive counts-as rules of the form ``A counts-as B in context C''. For example, ``a piece of paper with certain European Union symbols counts-as money in the context of the European Union''. These rules build a social reality from brute facts and regulate social reality change when the brute facts change. PARAGon makes three main contributions founded on counts-as rules.

PARAGon contributes formalised reasoning for a governance architecture called \textit{multi-level governance} where institutions operate at different governance levels. The lowest-level institutions prescribe concrete regulations to govern societies (e.g. national legislation), whilst higher-level level institutions impose abstract regulations on the concrete regulatory outcomes of lower-level institutions (e.g. directives, human rights charters and supranational agreements). The PARAGon framework derives from counts-as rules whether concrete regulations at lower governance levels can be applied in social contexts such that they violate abstract regulations as set out in directives, human rights charters, etcetera.

PARAGon contributes a computational mechanism for finding explanations for non-compliant institution designs used to rectify non-compliance. The explanations are minimal counts-as rule additions, deletions and modifications that ensure compliance. Moreover, the explanations keep institution designs as close as possible to designers' original intentions.

PARAGon formalises constraints called secondary legal rules, which define when institutions and institutional changes can be enacted. PARAGon formalises such secondary rules as \textit{rule-modifying} counts-as rules, which ascribe rule change at various points in time. For example, a government voting for a rule change \textit{counts-as} enacting a rule change. Determining rule change legality is difficult, since changing counts-as rules alters social contexts, which rule changes are conditional on. PARAGon contributes formal reasoning for determining when rule changes count-as legal rule changes.

The PARAGon framework was developed under the SHINE (Sensing Heterogeneous Information Network Environment) project\footnote{\url{http://shine.tudelft.nl}}, which aims to form large-scale heterogeneous sensor networks, using existing sensors in the environment belonging to external stakeholders. PARAGon aims to support automated governance for large scale heterogeneous `SHINE' sensor networks. Firstly, by automating the application of multi-level governance to forming SHINE sensor super-systems of sub-systems, comprising a thin SHINE institution layer abstractly governing the design of sub-system institutions towards coordinated regulations for collecting environmental data collection. Secondly, PARAGon supports sensor network stakeholders with automated institution re-design recommendations to ensure the institutions are designed compliantly. Thirdly, PARAGon supports automated reasoning for how and when sensor system regulations are changed over time as different governance needs arise (e.g. as the data needs change or as it emerges existing rules are inadequate).


The PARAGon framework makes both formal and practical contributions. From the formal side, this dissertation aims to formalise previously informal notions provided by political science and legal philosophy.
Particular attention is paid to understanding the new conceptualisations and testing them against a number of case studies to assess whether the formal contributions provide `correct' inferences. From the practical side, the contributed reasoning is either coupled with an implementation or a computational characterisation of the formal concepts providing necessary details for implementation. This dissertation lies at the intersection of \textit{legal philosophy and symbolic artificial intelligence}.

\chapter*{Samenvatting}
\addcontentsline{toc}{chapter}{Samenvatting}
\setheader{Samenvatting}

Ten grondslag aan deze dissertatie ligt de behoefte in de hedendaagse globalistische wereld aan een precieze methode om overheden, organisaties en andere regelgevende instanties in staat te stellen de beperkingen die ze opleggen aan zichzelf en anderen te evalueren. De modus operandi van een organisatie bestaat uit het instantiëren en uitvoeren van verbintenissen tussen zichzelf en haar deelnemers. Organisationele verbintenissen dienen echter wel de externe wetgeving te respecteren, zoals de wetgeving waarin de rechten en vrijheden op het gebied van gegevensbescherming zijn neergelegd. Verbintenissen kunnen alleen worden geïnstantieerd door verbintenissenrechtelijke processen te volgen, die vaak bilaterale overeenstemming en afweging vereisen. Overheden moeten wetten opstellen en daarnaast inzicht hebben in de hedendaagse context van nationale en internationale governance-hiërarchie waarin wetgevers isolationisme uit de weg gaan en elkaar proberen te beïnvloeden. Overheden dienen sancties te voorkomen door zich te houden aan de beperkingen die zijn opgelegd via internationale verdragen en mensenrechtenhandvesten. Overheden kunnen wetten alleen maar instantiëren door hun eigen, reeds bestaande wetgevingsprocedures te volgen. Met andere woorden: het ontwerpen en instantiëren van instituties, zoals wetten en verbintenissen, is onderworpen aan beperkingen.

De rode draad bij deze voorbeelden is dat ontwerpers van instituties, zoals organisaties en overheden, in twee opzichten beperkt zijn. Ze zijn lichtelijk gebonden in hoe de instituties die ze instantiëren moeten worden ontworpen, en sterk gebonden in hoe en wanneer ze institutionele veranderingen kunnen instantiëren. We zien duidelijk dat dergelijke beperkingen op schrift bestaan, ook al kunnen we niet fysiek zien hoe de machinerie werkt die de beperkingen toepast en ze rond de institutionele ontwerpers aantrekt. Derhalve is het moeilijk precies te begrijpen hoe ontwerpers van instituties moeten opereren te midden van deze beperkingen.

Deze dissertatie gaat in op deze problematiek door een formeel raamwerk aan te reiken voor het analyseren van institutioneel ontwerp en instantiatie-governance. Door formalisering biedt het raamwerk een mathematisch rigoureus overzicht van de beperkingen die opgelegd zijn aan institutionele ontwerpen en instantiatie van institutionele veranderingen. Daarmee worden de onzichtbare beperkingen zichtbaar, niet in fysiek opzicht, maar als symbolen op een pagina waarop algemeen institutioneel redeneren wordt gedefinieerd. Vanuit het oogpunt van conceptualisatie is het grootste voordeel dat we in staat zijn de institutionele beperkingen te begrijpen en eventuele conceptuele tekortkomingen te identificeren. Oftewel, in de woorden van Leibniz, “De enige manier om onze redeneringen te corrigeren, is ze net zo tastbaar te maken als die van de Mathematici, zodat we een fout van ons in één oogopslag kunnen ontwaren”.

Er zijn drie grote praktische voordelen. In de eerste plaats wordt formeel redeneren alom verspreid in heldere mathematische taal. Institutie-ontwerpers en rechters passen exact dezelfde redeneringen toe, waardoor uitspraken voorspelbaar worden, mits ze uitgaan van dezelfde juridische grondslagen. Bovendien kunnen burgers door toepassing van dezelfde institutionele redeneringen gemeenschappelijke kennis hebben van het moment waarop wijzigingen in regelgeving worden geïnstantieerd. In de tweede plaats faciliteert het raamwerk automatisering van anderszins cognitief lastige taken bij het begrijpen van de beperkingen die zijn opgelegd aan ontwerpers van instituties. In de derde plaats gaat het raamwerk in op de pragmatica middels toepassingen op casestudies uit de praktijk, om zo de betekenis van gehanteerde institutionele beperkingen te kunnen begrijpen.

Samenvattend, introduceert deze dissertatie het PARAGon-raamwerk voor \textbf{P}ractical \textbf{A}utomated \textbf{R}easoning for \textbf{A}ssessing \textbf{G}overnance met betrekking tot het ontwerpen en wijzigen van instituties.

Een centrale plaats in het PARAGon-raamwerk wordt ingenomen door Searle’s welbekende institutionele constitutieve "geldt als"-regels in de vorm “A geldt in context C als B”. Voorbeeld: “een stuk papier met bepaalde aanduidingen van de Europese Unie geldt in de context van de Europese Unie als geld”. Deze regels bouwen een sociale realiteit van brute feiten op en reguleren veranderingen in de sociale realiteit wanneer de brute feiten veranderen. PARAGon levert drie hoofdbijdragen die gebaseerd zijn op geldt-als-regels.
PARAGon biedt geformaliseerd redeneren voor een governance-architectuur, multilevel governance genaamd, waarbij instituten op meerdere governance-niveaus opereren. De instituties op het laagste niveau schrijven concrete reguleringen voor om samenlevingen te besturen (bijv. nationale wetgeving), terwijl instellingen op hoger niveau abstracte reguleringen opleggen met betrekking tot de concrete regulerende uitkomsten van lagere instituties (bijv. richtlijnen, mensenrechtenhandvesten en supranationale overeenkomsten). Het PARAGon-raamwerk leidt van geldt-als-regels af of concrete reguleringen op lagere governance-niveaus zodanig kunnen worden toegepast in sociale contexten dat ze een inbreuk vormen op abstracte reguleringen als neergelegd in richtlijnen, mensenrechtenhandvesten, enzovoort.
PARAGon reikt een berekeningsmechanisme aan voor het vinden van verklaringen voor niet-compliante institutie-ontwerpen die gebruikt worden om \\non~--compliantie te herstellen. De verklaringen zijn ten opzichte van geldt-als-regels minimale toevoegingen, verwijderingen en aanpassingen die zorgen voor compliantie. Verder houden de verklaringen de institutie-ontwerpen zo dicht mogelijk bij de oorspronkelijke bedoelingen van de ontwerpers.

PARAGon formaliseert beperkingen, de zogenaamde secundaire rechtsregels, die \\definiëren wanneer instituties en institutionele wijzigingen geïnstantieerd kunnen worden. PARAGon formaliseert dergelijke secundaire regels als rule-modifying geldt-als-regels, die op verschillende momenten wijzigingen van regels toekennen. Een stemming in het parlement om een regel te wijzigen geldt bijvoorbeeld als het instantiëren van een regelwijziging. Het bepalen van de legaliteit van regelwijzigingen is lastig aangezien een verandering van geldt-als-regels leidt tot een verandering van sociale contexten, waar regelwijzigingen afhankelijk van zijn. PARAGon biedt een formele redenering voor het bepalen wanneer regelwijzigingen gelden als wijzigingen van rechtsregels.

Het PARAGon-raamwerk is ontwikkeld in het kader van het project SHINE (Sensing Heterogeneous Information Network Environment)\footnote{\url{http://shine.tudelft.nl}}, dat beoogt grootschalige heterogene sensornetwerken te vormen met behulp van bestaande sensoren in de omgeving die toebehoort aan externe belanghebbenden. PARAGon beoogt ondersteuning van geautomatiseerde governance voor grootschalige heterogene SHINE-sensornetwerken. In de eerste plaats door automatisering van de toepassing van multi-level governance op de vorming van SHINE sensor-supersystemen van subsystemen, bestaande uit een dunne SHINE institutielaag die op abstracte wijze het ontwerp van subsysteem-instituties aanstuurt richting gecoördineerde regulering voor het verzamelen van milieugegevens. In de tweede plaats ondersteunt PARAGon belanghebbenden in het sensornetwerk met aanbevelingen voor het geautomatiseerd herontwerpen van instituties om zo te waarborgen dat de instituties op compliante wijze worden ontworpen. In de derde plaats ondersteunt PARAGon geautomatiseerde redenering voor hoe en wanneer regels van sensorsystemen in de loop der tijd worden gewijzigd naarmate andere governance-behoeften ontstaan (bijv. als de data gewijzigd moeten worden of als blijkt dat bestaande regels ontoereikend zijn).

Het PARAGon-raamwerk levert zowel formele als praktische bijdragen. In formeel opzicht streeft deze dissertatie naar het formaliseren van voorheen informele noties die vanuit de politieke wetenschap en de rechtsfilosofie werden aangeleverd. Bijzondere aandacht wordt besteed aan het begrijpen van de nieuwe conceptualisaties en het toetsen ervan aan een aantal casestudies teneinde te beoordelen of de formele bijdragen resulteren in ‘correcte’ inferenties. In praktisch opzicht wordt de aangedragen redenering gekoppeld aan ofwel een implementatie ofwel een rekenkundige karakterisering van de formele concepten die de noodzakelijke details aanleveren voor implementatie. Deze dissertatie bevindt zich op het raakvlak van rechtsfilosofie en symbolische artificiële intelligentie.




\tableofcontents
\listoffigures
\listoftables
\cleardoublepage
\pagenumbering{arabic}





\chapter{Introduction}
\label{chapter_1}

\epigraph[0pt]{The only way to rectify our reasonings is to make them as tangible as those of the Mathematicians, so that we can find our error at a glance, and when there are disputes among persons, we can simply say: Let us calculate [calculemus], without further ado, to see who is right.\cite{Leibniz1685}}{Gottfried Wilhelm Leibniz}

\newpage

In today's increasingly connected world, governments, organisations and other regulatory bodies do not operate in isolate free from control. Let us take a look at three examples. Our first example concerns the SHINE\footnote{\url{http://shine.tudelft.nl}} project under which this dissertation's research was conducted. SHINE aims to form systems comprising heterogeneous environmental sensors (e.g. cellphone cameras, weather radars) contracted from the sensor owners to collect environmental data. However, constraints are placed on forming contracts, in the sense that a sensor owner is only likely to agree to a contract if it meets their own policies stating what rights the contract should confer and what liberties it should not take away. Our second example concerns European Union (EU) law. In the EU, member states' governments enact legislation to meet governance aims. When the EU council wishes to coordinate legislation across the union, an EU directive is issued. Directives constrain and direct member states to implement regulations that meet cross--national aims. For example, retaining communications data for EU--wide criminal investigations \cite{EUDRD}. In turn, national legislation and EU directives are also required to confer rights and uphold liberties specified in the EU Charter of Fundamental Rights \cite{EU2000}. Our third example concerns the rules that make enacting and changing regulations possible. An organisation can only enact a legally valid and binding contract if the rules and processes that state how and when contractual regulations are enacted are followed \cite[p.96]{Hart1961}. The EU council is only able to legislate directives according to the rules that give it rule--making powers \cite[Art. 288]{EuropeanUnion2008a}. The United States government has the power to enact laws by following law making procedures, but laws cannot be created that apply to the past \cite[Art. 1 Sec. 9 Cl. 3]{USConstitution}. In other words, governments, organisations, contract writers and regulatory bodies are constrained in their regulatory activities.

These examples are all about institutions and the constraints placed on institution designers. Institutions are sets of rules and regulations, such as national legislation and contracts \cite{Boella2004c,Ruiter1997}. Institution designers are governments, contract writers, organisations and other regulatory bodies. Institutions regulate, organize and guide individuals' behaviour in a society towards collaboratively meeting societal aims \cite{Ostrom2005}. Societal participants are \textit{autonomous agents}, such as people or software agents, which are liable to act in their own self interests. Hence, institutions pair regulations with penalties to ensure it is in an individual's own self--interest to comply for the greater good of the governed society \cite{Axelrod1986,Cialdini1990}\cite[p.80]{Conte1995}. Institution designers craft institutions to regulate agents towards societal aims, such as people contractually participating in an organisation collecting and providing data, or citizens participating in a society whilst respecting one another's rights. By regulating rather than regimenting (forcing certain) agent behaviour, institution designers can achieve societal aims whilst preserving agents' autonomy.

Institution designers are also autonomous agents and hence are liable to act in their own self--interests. Consequently, regulations are placed on institution designers to reign their behaviour in. The first two examples showed how institution designers are governed in the institutions they design. Some institutions are designed to govern societies, such as national legislation. But, such institutional designs are in turn governed by other institutions, such as EU directives and human rights charters. The third example shows how institution designers are regulated in their ability to enact institutional changes by establishing new institutions or change existing institutions. By enactment, we mean in the sense of passing of a new law or a change to legislation, making a contract legally binding or more generally the social action of making an institutional change legally valid and imposed on agents. For example, enacting contracts or changing existing institutional rules to meet new aims. Governing regulatory change is defined by rules that stipulate when institutional changes are enacted. Hence, institution designers autonomously design and enact institutions, but institution designers are also governed in the institutions they \textit{design} and the institutional \textit{enactment} process.

In today's increasingly technological world, operationalising institutions by applying rules and regulations is not just left up to opaque human reasoning. Instead, the fields of \textit{Normative Multi--Agent Systems} and \textit{AI and Law} are concerned with automated reasoning for institutions by contributing formal frameworks that interpret and apply institutions (we discuss background on implemented systems later in Chapter~\ref{chapter_2}, and a literature survey is provided by \cite{Dagstuhl4}).  Formalisation, in general, takes informal reasoning, that is ambiguous and unclear, and exposes it in plain sight with precise mathematically rigorous definitions.


Yet, thus far the reasoning involved in the governance placed on institution designers has not been formalised. We can see a common thread is shared by our examples, in that institution designers are constrained loosely in how the institutions they enact \textit{should} be designed and strongly in how and when they \textit{can} enact institutional changes. Yet, although we can clearly see such constraints exist in their written form, we cannot physically see precisely how the machinery that applies and tightens those constraints around the institutional designers works. It is consequently hard to grasp how exactly institution designers need to operate under institutional design and enactment constraints. This dissertation addresses this knowledge gap, by contributing \textit{a formal framework for analysing institutional design and enactment governance}. To understand more specifically why such formal reasoning is important, let us first take a closer look at the concepts behind institutional design and enactment governance.

\textbf{From the institutional \textit{design} governance perspective}, institution designs are governed by other institutions in what is called \textit{multi--level governance} \cite{Liesbet2003}. In this dissertation, multi--level governance is conceived as higher--level institutions designed to govern and guide the institution \textit{designs}, enacted by autonomous institution designers, operating at lower governance levels. Multi--level governance facilitates institution designs in being coordinated, when viewed as being related in a wider multi--institution system (e.g. \cite{Zuckerman1979}). For example, at a national level a government enacts institutions to govern a nation. At a cross--national level institutions are designed to guide national governments in enacting institutions with coordinated regulations. Moreover, designers are guided in ensuring their designs do not take away rights and liberties, such as by human rights charters. In multi--level governance, institution designers have autonomy to design institutions according to their aims, but their institution designs are also subject to being governed and guided by higher--level institutions.

Multi--level governance creates the possibility for institution designs to be non--compliant. A non--compliant institution design is problematic. From the higher governance levels' perspective non--compliant institution designs are uncoordinated with other institutions governing separate jurisdictions, do not uphold rights and/or do take away liberties. From an institution designer's perspective, they are liable to face punishment for non--compliant design. When societal members act in a non--compliant way they are liable to being penalised in order to guide society towards compliance\cite[p. 279]{Anderson1994}. Hence, when an institution designer designs a non--compliant institution, they are liable to being fined or having institution designs annulled \cite{Scharpf1997} by judiciaries. Non--compliance in multi--level governance should be detected and avoided before institution enactment by an institution designer in order to avoid punishment, and detected after institution enactment by a judiciary to issue punishment and thus incentivise institution designers in enacting compliant institutions.

\textbf{From the institutional change \textit{enactment} governance perspective}, institutional rules define the legislative actions that constitute a valid institutional change enactment conditional on the social context \cite{Biagioli1997}. In a simple case, a legislature voting by majority on an institution enactment change suffices. In other cases, there are further constraints on valid institution change enactment, such as there being no valid way to change an institution in the past (retroactively) \cite[Art. 1 Sec. 9 Cl. 3]{USConstitution} ``No Bill of Attainder or ex post facto Law shall be passed''. In general, there are rules that state how and when rules are changed. According to the influential legal philosopher Hart, these are \textit{secondary} institutional rules and they create the \textit{possibility} for governments to change legislation and citizens to create contracts \cite[p.81]{Hart1961}. When an institution designer designs or changes an institution outside of the secondary rules the ``enacted'' changes are \textit{invalid}. For example, enactment does not occur merely by physically writing institutional rules without following the necessary voting procedure set out by secondary rules. From an institution's perspective, invalid enactments or changes do not take place. Hence, the system or society the institution governs should ignore such invalid enactments in order to have the correct and shared view of an institution's rules.

Formal reasoning is important for the governance of institution design and enactment. Specifically, for precision and automation:

\begin{itemize}
\item[] \textbf{Precision}. In general, formalisation removes ambiguity. Without ambiguity, rigorous interrogation of the underlying reasoning is possible. Hence, reasoning flaws can be found and fixed, which can have real positive and negative consequences for agents being rewarded or punished. Moreover, formalisation communicates in clear mathematical terms the reasoning involved. Hence, given an institution and the facts of a case, agents can independently come to the same conclusions by applying identical reasoning. 
\item[] From a multi--level governance perspective, an institution designer can predict whether a judiciary would find the institution design non--compliant. \textit{With one caveat, the designer and judiciary must have the same legal rule base, including any unwritten rules such as conceptual interpretations}. Prediction benefits an institution designer who can choose to not enact a non--compliant institution in order to avoid punishment.
\item[] In the case of governing the institutional change enactment process with secondary rules, all agents governed by an institution are able to come to the same understanding over the changes made to the rules they are governed by. A common understanding benefits two agents in neither having a different understanding of the regulations in place nor misunderstanding what they ought to do according to the institutions governing them.
\end{itemize}

\begin{itemize}
\item[] \textbf{Automation}. Formalisation removes ambiguity, which is a necessary pre--requisite for automation since it means, potentially, the formal reasoning can be implemented in a computational language as a program for a computer to execute.
\item[] From the multi--level governance perspective, automation lowers compliance checking costs. A judiciary does not face the dilemma over either arduously determining if an institutional design is compliant or forego judgement and face the possibility that institution designs are allowed to be uncoordinated or take away agents' rights and liberties without punishment. Automation also lowers compliance checking costs for institution designers and helps rectify non--compliance. Fixing non--compliance is not necessarily easy. There can be multiple possible explanations for why an institution is non--compliant in there there is often a space of many compliant institution designs. For example, EU member states are at liberty to comply with an EU directive in many ways \cite[p.5]{Folsom1996}. Yet, some compliant designs will meet an institution designers' objectives more than others. By automating compliance checking, searching for explanations for non--compliance that can be used to rectify the underlying causes is also automatable in a way that meets an institution designer's aims the most.
\item[] It is non--trivial to determine how an institution is changed according to secondary rules governing rule change. For example, a new rule might be enacted stating rules can be changed by majority vote. Enacting such a rule affects which future rule changes can be enacted. More generally, without wishing to delve into the temporal details at this point, rule changes can also be applied to the past, present or future and have many complex interacting affects with other rule changes. Automation takes over the cognitively difficult task for a human of determining when rule change enactments are legally valid.
\end{itemize}

In order to address the need for precision and automation, institutional design and enactment governance should be formalised, but  there is a lack of formal work in this area. This dissertation addresses this problem by proposing the PARAGon\footnote{Just as an institution defines ideality, a paragon is an example of a person or thing regarded as a perfect example of such an ideal \cite{oxforddictionariesonlineparagon}} formal framework for \textbf{P}ractical \textbf{A}utomated \textbf{R}easoning for \textbf{A}ssessing \textbf{Go}ver\textbf{n}ance of institution design and enactment. The formalism comprises precise definitions for compliance in multi--level governance, a mechanism to determine explanations and rectifications for non--compliance, and precise definitions for institution enactment validity according to secondary rules.

Moreover, this dissertation adopts a practical approach in two senses. Firstly, it provides definitions that are either coupled with a corresponding computational implementation or are defined in such a way as to make it obvious how the reasoning can be implemented. Thus, the formalisation provides \textit{automated} reasoning. Secondly, the conceptualisations are aimed at capturing a number of real--world case studies in a realistic setting that includes factors such as time and change. The framework allows an institution designer to specify the institutions in a formal grammar and determine using the reasoning whether the design is compliant in multi--level governance and rectify any non--compliance. Moreover, institution designers and agents governed by institutions alike can apply the framework to determine which rule changes are validly enacted and when. The framework users need not understand the underlying mechanics, in terms of how it is \textit{decided} what the regulatory effects of an institution are. Rather, users only need to understand what the concept of an institution being used by the system is and correctly provide as input the various institutions that are governing/being governed and the actions of various agents that are occurring. Consequently, the reasoning burden can be delegated to a computer to mechanically determine institution design compliance, rectifications for non--compliance and valid rule change enactment.

This chapter proceeds to introduce the research questions in section~\ref{secIntroRQs}. We describe how this dissertation fits into the SHINE project in section~\ref{secIntroSHINEProject}. Then, the research approach is described in section~\ref{secIntroResearchApproach}. Finally, the outline of this dissertation is given in section~\ref{secIntroDissertationOutline}.

\section{Research Questions}
\label{secIntroRQs}

As we discussed, there is a lack of formal and practical reasoning for governing institution design and enactment that needs addressing. This leads to the main research question:

\begin{quote}
How can institutional design and enactment governance be supported with formal reasoning?
\end{quote}

The main research question is broken down into five sub--questions.

\vspace{4mm}

The idea is to contribute formal reasoning to support human stakeholders in understanding the constraints placed on institution designers. Stakeholders need a way to represent institutions and their governance relationships. This leads to our first research question.

\begin{quote}
\textbf{Sub--research question 1:} What is a suitable representation to specify institutional design and enactment governance?
\end{quote}

It is important for the representation to be natural, by which we mean with a clear correspondence to how written and verbal institutions are represented, for two reasons. Firstly, from the precision perspective we are interested in formalising institutional governance to make it clear what we mean by the relevant concepts. The idea being, the concepts are exposed and can be interrogated, and discussed and debated for `correctness'. A natural representation supports this aim by demonstrating to what extent the formalised reasoning is doing all the work of coming to correct conclusions and to what extent it is actually highly dependent on the way an institution is written. The latter case is far from desirable. For example, if we have to represent an institution in a very convoluted or procedural way to get the correct reasoning results, it is reasonable to assume that the informal reasoning has hardly been formalised at all. From the automation perspective, a `user' \footnote{a user of the PARAGon framework can be a human agent interested in reasoning about institutional design and change enactment governance, such as to avoid punishment for non--compliant design. A user can also be a researcher wishing to understand the precise meaning of the concepts involved.} needs to face as little burden as possible, meaning they should be able to specify an institution close to how they would a written law without thinking about how a computer might interpret it. To give a contrasting example, we could simply provide the user with an abstract computational machine (e.g. a Turing machine \cite{Turing1936}) and give the user all the flexibility to specify institutions in whichever way they want as procedural code. However, by giving users an `unnatural' language to specify institutions, we have contributed nothing to automating institutional governance reasoning. Hence, a natural representation is central to being able to critique the PARAGon framework, and the framework itself providing the benefits of formalisation for institutional design and enactment governance.

The PARAGon framework answers this question by providing different ways to represent institutions depending on whether we are interested in the governance of institution design or change enactment. The first step in answering the question is to gather background knowledge on the kinds of institutional rules and regulations we wish to represent and to what extent there are already suitable grammars to represent those concepts in Chapter~\ref{chapter_2}. Then, in Chapter~\ref{chapter_3} and \ref{chapter_4} we provide a representation for multi--level governance and apply it to European Union laws. A simpler representation for governing institutional designs is provided in Chapter~\ref{chapter_5} and applied to a crowdsourced mobile sensing scenario inspired by the SHINE project. In Chapter~\ref{chapter_6} we provide a further representation language, this time focussed on secondary rules used to govern institutional change enactments. Once we have a way to represent institutions, we need a way to reason about them, leading to our next four sub--research questions.

\vspace{4mm}

The first aspect we look at is \textit{compliance} of institution designs in multi--level governance. Although it is expected that institution designers enact compliant institutions, due to designers' autonomy it is not guaranteed. Compliance of institution designs should be evaluated to enable institution designers to understand whether their designs are compliant before enactment, and higher governance levels' judiciaries in punishing institution designers for offending institution designs. Compliance should be assessed in a precise and consistent way, such that the same notion of compliance is shared by institution designers and judiciaries. Hence, compliance requires formalisation. Since this dissertation takes a practical approach, addressing real--world needs, compliance should be formalised to capture its real--world informal notion. This leads to our next sub--research question:

\begin{quote}
\textbf{Sub--research question 2:} How can we formalise compliance in multi--level governance?
\end{quote}

In general, compliance involves adhering to regulatory requirements \cite{merriamwebCompliance}. In multi--level governance, compliance represents whether from the perspective of institutions designed at higher governance levels (e.g. \cite[p.5]{Folsom1996}), lower--level institutions regulate in a `good' way to meet particular higher--level governance aims. There is a difference between societal compliance and institution design compliance. Societal compliance focuses on agent behaviour, which is already widely formalised \cite{Dagstuhl4}, such as agents being compliant if they do not murder one another according to institutions that prohibit murder. On the other hand, institution design compliance focuses on whether the effects of regulations are good, such as by being coordinated with other institutions' regulatory effects (e.g. controlling carbon emissions through high--level international institution design coordination \cite{Ostrom2011}), conferring human rights (e.g. prohibiting murder) and respecting liberties. So, compliant institution design is formalised by defining when the regulatory effects at lower governance levels are compliant with the regulations at higher governance levels.

The formal notion of compliance in multi--level governance must, however, also take into account \textit{abstraction}. In multi--level governance, there are conceptual differences between institutions operating at different governance levels. Institutions embody institutional concepts defined with context--sensitive conceptualising rules \cite{Searle1995}. For example, a rule stating that a soldier killing another person who is not a member of an opposing army is murder. Such conceptual rules serve an important role, since they allow regulations to be succinctly defined over high--level concepts \cite{Breuker1997}. For example, a regulation prohibiting murder as opposed to a regulation prohibiting a soldier from doing such and such. Separate institutions can potentially define the same concepts differently or define different concepts altogether. Moreover, since institutions define regulations over separate concepts, the regulations themselves have different, context--sensitive, meanings. In multi--level governance, we can see higher governance levels in fact use more abstract concepts. For example national legislation might concretely require communications data to be stored (e.g. \cite{UK2009}). Yet, human rights charters describe abstract notions such as the right to a private and family life \cite{ECHR1953}. Therefore, compliance should be defined in terms of concrete lower--level governance regulations being compliant with abstract higher--level governance regulations. Moreover, compliance should account for the meanings of the regulations potentially being entirely different depending on the social context. 

The PARAGon framework provides a formal definition of compliance in multi--level governance in  Chapter~\ref{chapter_3}, reconciling the different, context--sensitive, regulation meanings from a conceptually concrete lower--level institution compared to a conceptually abstract higher--level institution. By defining compliance in multi--level governance, we are able to unambiguously determine compliance with mathematical rigour and regularised predictable results for a real world setting.

\vspace{4mm}

Having a formal definition of compliance does not necessarily reveal how to practically reason about compliance in a computational way. This leads us to the third sub--research question:

\begin{quote}
\textbf{Sub--research question 3:} How can institutional design compliance in multi--level governance be computationally verified?
\end{quote}

Computationally verifying institution design compliance in a practical way is desirable to reduce compliance checking burdens placed on the various actors involved in multi--level governance. Whilst each institution designer at each level of governance must provide the institutions being governed and doing the governing as input, there should be no burden on those designers to implement or understand the mechanics of compliance checking. As Sergot exemplifies using the law \cite{Sergot1988}, institutions are sets of declarative rules. Whilst rules can influence each other, and rules can define procedures which should be followed, the ordering of the rules themselves should not impact their meaning. What we can take from this is that institutions are not algorithmic and any algorithmic reasoning applied to an institution is not visible in an institutional specification. Consequently, institution designers are not particularly concerned with the reasoning mechanisms used to operationalise institutions. So, automated compliance checking is needed to ensure institution designers do not need to take extra effort in applying the formalised reasoning. That is, by providing a computational framework which hides the mechanics (i.e. algorithms or semantics) for compliance checking. So, what we aim for when answering this research question, is a computational framework that takes a natural representation of institutions as input and determines if an institution design is compliant in a sound and complete way (i.e. equivalent) with respect to our formal notion of compliance.

The PARAGon framework contributes computational compliance checking in multi--level governance in Chapter~\ref{chapter_5}, which is sound and complete with respect to our formal notion of compliance. This supports judiciaries in determining penalties issued on institution designers for non--compliant designs. Moreover, it supports institution designers in avoiding penalties by indicating a non--compliant institution design should perhaps not be enacted.

\vspace{4mm}

For an institution designer it might be simplistic to state that they can simply avoid punishment by not enacting a non--compliant institution design. Firstly, in the case that an institution designer is actually obliged to enact an institution in order to implement the imperatives issued by a higher--level institution, as is required in complying with EU Directives and other cross--national institutions. Secondly, in the case that an institution designer prefers to enact an adjusted and compliant institution rather than not enacting an institution at all. This leads us to our fourth sub--research question:

\begin{quote}
\textbf{Sub--research question 4:} How can non--compliant institution designs be explained in order to rectify non--compliance according to the institution designer's objectives?
\end{quote}

Institutions are designed with a purpose, to guide societies towards an ideal envisaged by the institution designer or guide and coordinate other institution designs towards an ideal. We are interested in resolving non--compliant designs. Mitchell \cite{Mitchell1996}, provides a number of reasons for (non--)compliance in general, such as the non--compliance if agentive actions in society. Pertinent to institution design, non--compliance can occur due to preference, in our case because institution designers prefer a non--compliant institution design due to its positive governance effects, even in the face of penalties for non--compliance. Non--compliance can also occur due to incapacity, in our case simply because the institution designers do not know how to design a compliant institution.

These reasons for non--compliance should be taken into account when rectifying non--compliant institution designs. From the non--compliance due to designer preference perspective, any non--compliance resolution should balance the requirement to successfully adjust the design to be compliant against the objectives of the institution designer. Assuming the non--compliant design was crafted to achieve particular governance aims, resolving non--compliance must remain as closely as possible to the original design's regulatory effects. From the non--compliance due to designer incapacity perspective, non--compliance resolution should take into account the fact that incapacity can be due to institutions being complicated, comprising many interrelated rules and regulations \cite[p.14]{Governatori2005} \cite[p.2]{Meneguzzi2012}. This means it can be non--obvious to an institution designer which rules and their interactions are causing non--compliance. 

Assuming an institution designer wishes to resolve non--compliance in a way they understand, non--compliance resolution should seek the simplest and most general explanations for non--compliance, to support the institution designer in understanding and remedying the underlying problems. For example, the following case of non--compliance could have multiple explanations, which resolve the non--compliance -- a lower governance level institution is governed by a higher--level institution in only obliging agents that are adults in providing personal information, but the lower--level institution obliges both children and adults to provide personal information. One possible explanation for non--compliance is that it is due to there existing a rule that obliges people to provide personal data and that rule should be removed to ensure compliance. A second explanation for non--compliance that gives less drastic institution re--design advice and is, is that the rule obliging people to provide personal information is too general and it should be modified to only applying to adults. In summary, non--compliance resolution should adhere to an institution designer's own objectives and recommend the simplest explanations and rectifications for non--compliance.

The PARAGon framework answers this question in Chapter~\ref{chapter_5} with a computational non--compliance resolution mechanism.

\vspace{4mm}

The previous research questions address institution design governance, our fifth and last research question addresses institution change enactment governance:

\begin{quote}
\textbf{Sub--research question 5:} How can we formally define when legally valid institutional change enactments occur?
\end{quote}

Institutional change enactment is constrained in its validity according to secondary legal rules. We need a formal definition of when, for a set of secondary legal rules, particular physical behaviour such as `the thing we call' voting on a rule change causes the social action of rules actually being changed. Such secondary rules create the possibility to take the social action of changing rules \cite{Biagioli1997}. These rules making rule change possible are also themselves changeable \cite{Suber1990}. Formalising the  legal validity of rule change enactment must take into account the fact that changing rules can affect further rule change enactments. Since we are dealing with real--world case studies the temporal aspect of secondary rules is a factor. It is possible, for example, for secondary rules to make past rule change enactments possible. Institution designers have enacted past rule changes in order to `undo' the consequences of `bad' institution design decisions \cite{FinanceAct2008}. The implication is that, since changing rules can affect rule change enactments, changing rules in the past can affect rule changes at various other points in time, including in the more recent past, the present or the future. So, what this research question requires is that we formalise rule change enactment validity whilst taking into account complex temporal interdependencies between secondary rules and rule changes.

The PARAGon framework answers this research question in Chapter~\ref{chapter_6} with a formal framework for determining when rule changes count--as legal rule change enactments. In answering this question, special attention is paid to formalising in such a way that demonstrates the computational mechanism to decide whether a rule enactment is valid. Moreover, an account of the temporal aspects is given, demonstrated against a number of real and imagined case studies.

By addressing these research questions the following argument is made:

\begin{quote}
This dissertation formalises governing governance by giving a rigorous mathematical definition for institutions being governed in how they \textit{should} be designed and how institutional change enactments \textit{can} be made.
\end{quote}

Now we proceed to the approach we take to answering the research questions.

\section{Research Approach}
\label{secIntroResearchApproach}

This dissertation's result is the PARAGon framework for formally reasoning about governing institution design and enacting changes. A research approach was followed to develop the PARAGon framework and its constituent parts. The approach is to start with a literature review and then \textit{per research question} gather case studies, develop a formal framework and apply model checking to assess the framework against the case studies. Each step is described as follows.

\begin{itemize}
\item[] \textbf{Literature review:} this step achieves three aims. This research started out in the broad area of institutional and normative reasoning. The first aim was to understand informal definitions of the main concepts involved, namely institutions and norms, governance and closely related concepts. The second aim was to identify an area of research which was both useful to the SHINE project of governing large--scale sensor systems (and systems of systems) and had not been looked at previously from an Artificial Intelligence perspective. Namely, the governance of institution design and enacting changes. The third aim was to identify a single underlying formalism providing more primitive concepts (institutions and regulations) on which the PARAGon framework can use as foundations.
\item[] \textbf{Framework development cycle:} The next step was, per research question, to iterate the following three sub--steps. The iterative cycle terminates when applying the PARAGon framework to case studies results in a natural representation and judgements that correspond to real--world judgements or what we would intuitively expect. Although, `natural representation' and `intuitive expectations' are subjective, the representation and reasoning are made precise so that they can be argued for and against.
\begin{itemize}
\item \textbf{Gather case studies:} Each research question addressed a real--world governmental process to formalise and automate. To answer the research questions, the first step was to gather or synthesise relevant case studies against which the framework was developed, to ensure it was grounded in a realistic setting.
\item \textbf{Formalisation:} The next step was to take the relevant case studies and understand the specific reasoning involved in order to come to the intuitive outcome for each case study. Then, to provide a more general account of the reasoning through \textit{formalisation}, defining a formal syntax and semantics, potentially coupled with a corresponding computational mechanism. Following Hansson's argument \cite{Hansson2013a}, the purpose of formalisation is not to produce an empirically supported theory, but rather to precisely define previously informal concepts (in our case secondary rules, governance of institution design and so on).
\item \textbf{Model checking case studies:} given a formal language comprising a representation and semantics, it is possible to formally represent the case studies and construct \textit{models} for those case studies. By a model, we mean a logical model, which for a logical theory is a structure that \textit{satisfies} the theory (if such a structure exists). By constructing a model, it is possible to check various properties of the theory. In our case, the logical theory is an institutional specification and a series of events. The events either occur hypothetically as a part of an offline institution design check (e.g. checking how the institution behaves when an agent decides to collect data) or events which occur in reality (e.g. agents voting on rule enactments). The properties being checked are whether an institution design is compliant, ways to rectify non--compliance and which rule changes have validly been enacted and when. During this step, it was often revealed that there were counter--intuitive results for the developed semantics and given case studies, consequently further refinement of the formalisation was required and the previous steps repeated.
\end{itemize}
\item[] \textbf{Putting it all together:} The final step in the research approach was to apply the results and reflect on open questions. This step takes the application further to show its practical relevance by implementing part of the PARAGon framework in a prototype system described in Chapter~\ref{chapter_7}. Each component of the PARAGon framework has its own implications. In this step those implications are compared and synthesised into an overall conclusion and set of unanswered questions in the final chapter, Chapter~\ref{chapter_8}.
\end{itemize}

\section{The SHINE Project}
\label{secIntroSHINEProject}

This research was initiated and supported by the SHINE project of TU Delft. SHINE was a large interdisciplinary research project. It aimed to develop techniques for acquiring and coordinating large numbers of heterogeneous data resources (e.g. cellphone sensors, radars and people). The idea was to use these sensors to gather a wide range of detailed environmental data (e.g. rainfall and pollution levels). In so doing, various stakeholders (e.g. citizens, municipalities) can gain answers to questions pertaining the environment (e.g. `how do I get from A to B whilst avoiding flooding?'). SHINE looked at the problem from many different angles, such as algorithms for configuring sensors, user modelling, governing and coordinating resources, and visualising the acquired data to help answer questions. 

This dissertation contributes techniques for formalising governance, with a focus on formalising governance which is, arguably, particularly suitable for governance and coordination of `SHINE--like' sensor systems of various types. For example, governing and coordinating systems of cellphones gathering geospatial audio data to determine crowding levels or systems of weather radars used to determine rainfall levels. The formalisation is applied to a mixture of governance case studies, from national and international data and human rights laws to imagined SHINE sensor system regulations. The idea is to formalise compliance in multi--level governance and institutional enactment validity as found in the social world to support automated `SHINE sensor system' governance.

Multi--level governance is relevant to SHINE according to the argument that it is a necessary governance architecture for governing heterogeneous sociotechnical systems (e.g. what is also called polycentric governance \cite{Pitt2014,Pitt2015}). We will exemplify why using examples concerning SHINE systems. On the one hand, a homogeneous sensor system comprising users crowdsourced into donating their cellphone sensors can, arguably, be governed with a single set of related regulations. For example, regulations defining a communal economy with a single incentive for users to contribute data (i.e. receiving data from a common pool in return). On the other hand, SHINE aims to form heterogeneous sensor systems which, arguably, are unsuitable for a `one--size--fits--all' set of regulations. For example, a system comprising weather radars contributed by organisations could operate best as a market economy, where organisations are incentivised to join the system in order to trade data. In this case, the regulations for a crowdsourcing sensor system operating as a communal economy are entirely inappropriate. One possibility is to view each set of sensors as a separate sensor \textit{sub--system} and write regulations to govern those sub--systems separately to form a super system of sub--systems. However, this places a burden on the SHINE--system institution designer in writing appropriate regulations for each sensor sub--system.

Multi--level governance offers an architecture to move the burden of writing specific regulations from the SHINE--system institution designer to the sensor sub--system stakeholders. The architecture proposes a solution, in the same vein as existing proposals of poly--centric governance for smart--grids comprising heterogeneous energy sub--systems \cite{Diaconescu2014}. The idea being, that the stakeholder wishing to form a heterogeneous SHINE sensor system crafts an institution at a second governance level which governs the design of institutions governing the separate sub--systems. The `SHINE institution' is a thin governance layer, comprising abstract regulations requiring that sub--sensor--systems are governed by institutions which regulate resources towards collecting useful environmental data and punishing sensor owners for contributing erroneous data. Appealing to the principle of subsidiarity, what can be done at the local level should be left up to the local level, the SHINE institution would give space for the sensor sub--system stakeholders to determine which data is collected and what the incentives are (e.g. a market economy or a communal economy). In return, sensor sub--system stakeholders can design institutions to govern those sub--systems in order to join the SHINE super--system and gain data from other sub--systems in return. The PARAGon framework helps to operationalise multi--level governance for forming SHINE systems governed by a higher--level SHINE institution, by automating compliance checking and non--compliance rectification. 

The automated reasoning for compliance checking is also applicable to the SHINE project's aim of crowdsourcing the existing sensors people already own and their time. For example, crowdsourcing people in donating their cellphone audio sensors in order to detect geo--spatial crowding, or crowdsourcing people into taking photographs of the sky when requested to determine pollution levels. This type of sensing is dubbed by the SHINE project as `request driven social sensing' and a key idea behind it is that people are offered contracts for use of their devices and time. Since people are ideally offered many contracts to address many data needs as and when they arise, it is important cellphone users can automatically accept or reject contracts on the basis of policies they define stating how, when and for whom their cellphone sensors can be used. In this situation, a policy stating sensor usage governs offered contracts. The automated multi--level governance reasoning can be used to automate such contract rejection and is implemented in a prototype simulated crowdsourced mobile sensing system described in Chapter~\ref{chapter_7}.

Secondary rules governing institutional enactment are relevant to SHINE from the perspective that they provide sensor system stakeholders a flexible and automated way to govern sensor systems' regulatory change. Flexibility is meant in two senses. Firstly, stakeholders are able to make how regulatory change legally operates flexible. This is important since the sensing aims or dynamics of the system are liable to changing over time and hence regulatory changes need to be enacted to meet stakeholders new aims or system participant's changing behaviour. Secondly, the way in which the regulatory change enactment process is defined is flexible. For example, one sensor system can define regulatory change as requiring sensor system participant's democratic vote, this might be suitable for a system of crowdsourced cellphone users that donate their cellphone sensors to the system partly due to having a say in how that system is run. In another case, where designing regulations requires highly--technical knowledge of sensors' operation, a sensor system's regulatory change may be defined on the basis of elected technocrats coming to an agreement. Hence, secondary rules allow a diverse range of governance and regulatory change enactment styles to be defined for different sensor systems in a way that enables those system's regulations to adapt, as deemed appropriate, to new aims and needs.

The PARAGon framework supports realising these benefits by contributing formalisation to automatically determine how and when regulatory changes are legally enacted. For example, such as due to a vote to move a system from a market economy where data is traded to a communal economy where data is contributed to a common pool and shared. Moreover, formalisation means the way in which changes are enacted is automatically changed according to legal rule changes, such as moving from a directed democracy to an elected technocracy. Automation means system stakeholders can operate a flexible governance system at lower cost and therefore makes systems where flexible governance is necessary, such as for diverse SHINE sub--systems, more viable.

\section{Dissertation Outline}
\label{secIntroDissertationOutline}

\begin{figure}[h]
\centering
\begin{tikzpicture}[xscale=0.8, yscale=0.8, every node/.style={xscale=0.8, yscale=0.8}]
\draw  (-2.25,4.75) rectangle (1.25,3) node[pos=.5, text width=4cm, align=center] {\small \textbf{Chapter \ref{chapter_1}} \\Introduction};
\draw  (-2.25,2.5) rectangle (1.25,0.75) node[pos=.5, text width=4cm, align=center] {\small \textbf{Chapter \ref{chapter_2}} \\Background};
\draw  (3,-1) rectangle (6.25,-2.75) node[pos=.5, text width=4cm, align=center] {\small \textbf{Chapter \ref{chapter_6}} \\ Institution\\ Change and Enactment \\ Validity};
\draw (-3.25,0.25) rectangle (2.25,-4.25);
\draw (-3.25,0) node [text width=7cm, anchor=west] {Institutional Design Compliance};
\draw  (-2,-0.25) rectangle (1,-2) node[pos=.5, text width=3.5cm, align=center] {\small \textbf{Chapter \ref{chapter_3}} \\ Formalising \\ Compliance};
\draw  (-2,-2.25) rectangle (1,-4) node[pos=.5, text width=3.5cm, align=center] {\small \textbf{Chapter \ref{chapter_4}} \\ Computational \\ Compliance Checking};
\draw  (-7.25,-0.75) rectangle (-4,-2.75) node[pos=.5, text width=3.5cm, align=center] {\small \textbf{Chapter \ref{chapter_5}} \\ Rectifying \\ Institutional Design \\Non-Compliance};
\draw  (-2,-4.6) rectangle (1,-6.35) node[pos=.5, text width=3.5cm, align=center] {\small \textbf{Chapter \ref{chapter_7}} \\ Application};
\draw  (-2,-6.75) rectangle (1,-8.5) node[pos=.5, text width=3.5cm, align=center] {\small \textbf{Chapter \ref{chapter_8}} \\ Conclusions};
\draw [very thick, ->, dashed] (-5.75,1.5) node (v1) {} -- (-5.75,-0.75);
\draw [very thick, ->, dashed] (-0.5,3) -- (-0.5,2.5);
\draw [very thick, dashed] (1.25,1.5) -- (4.5,1.5);
\draw [very thick, ->, dashed] (-0.5,0.75) -- (-0.5,0.25);
\draw [very thick, dashed] (-2.25,1.5) -- (v1);
\draw [very thick, ->, dashed] (4.75,1.5) -- (4.75,-1);
\draw [very thick, ->, dashed] (-0.5,-4.25) -- (-0.5,-4.6);
\draw [very thick, ->, dashed] (-0.5,-6.35) -- (-0.5,-6.75);
\draw [very thick, ->, dashed] (4.75,-8) -- (1,-8);
\draw [very thick, ->, dashed] (-5.75,-8) -- (-2,-8);
\draw [very thick,  dashed] (-5.75,-2.75) -- (-5.75,-8);
\draw [very thick,  dashed] (4.75,-2.75) -- (4.75,-8);
\draw [very thick,  dashed] (1.7,-4.3) -- (1.7,-7.5);
\draw [very thick, ->, dashed] (1.7,-7.5) -- (1,-7.5);
\end{tikzpicture}
\caption[Dissertation overview]{An overview of the dissertation with suggested reading orders.}
\label{figDissertationOverview}
\end{figure}
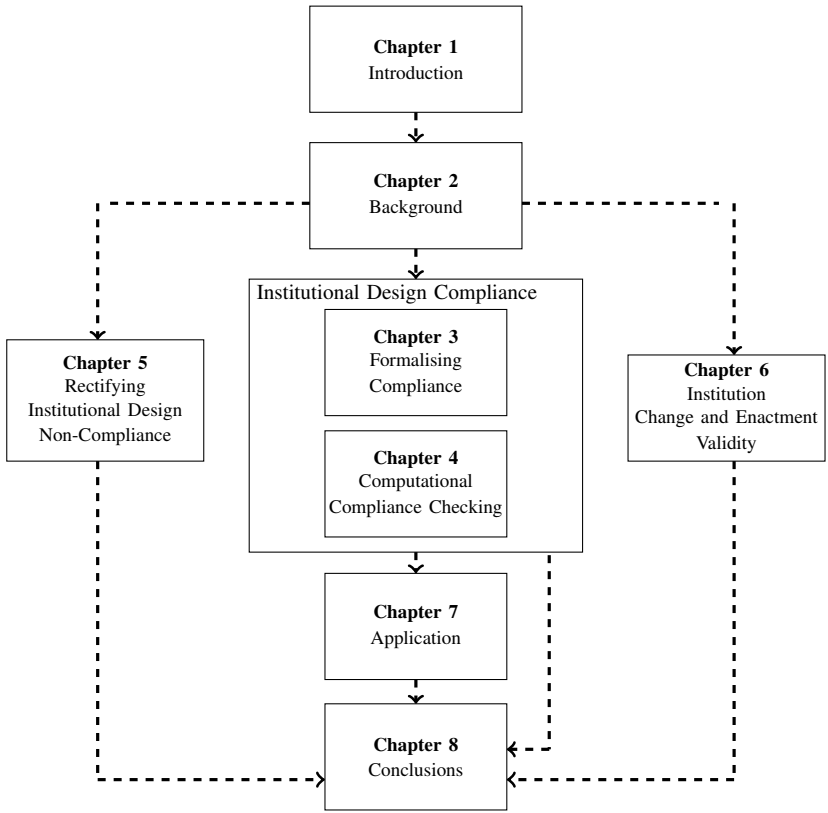

The dissertation outline is depicted in figure~\ref{figDissertationOverview}. This dissertation is broken up into the introduction (Chapter~\ref{chapter_1}) and the background (Chapter~\ref{chapter_2}), then the main contributions of formal reasoning for governing institution design and change enactment (chapters~\ref{chapter_3}~to~\ref{chapter_6}), and finally an illustration of the framework's application (Chapter~\ref{chapter_7}) and the conclusions (Chapter~\ref{chapter_8}). The main technical contributions begin by looking at \textit{soft constraints}. These are constraints that create the possibility for (non--)compliance and, respecting agents' autonomy, cannot be guaranteed to be complied with. In our case, the soft constraints specify how an institution \textit{should} be designed (chapters~\ref{chapter_3} and \ref{chapter_4}). We subsequently contribute a mechanism for revising an institution design to be compliant with such soft constraints (Chapter~\ref{chapter_5}). Then, we move to reasoning about governance in the form of \textit{hard institutional constraints}. These are constraints that are not violatable. In our case, the hard constraints are secondary institutional rules defining when institutional change enactment is \textit{possible} (Chapter~\ref{chapter_6}).

The following chapters are described in more detail:

\begin{itemize}
\item[] \textbf{Chapter~\ref{chapter_2}} provides background starting with an overview of the informal concepts we wish to formalise and reason about. Then, proceeding to analyse the existing knowledge and its gaps in formalising governance of institution design and change enactment. Next, suitable frameworks that provide preliminary formalisation on which to build the PARAGon framework are compared. Finally, suitable foundational formalisations are selected on which this dissertation builds.
\item[] \textbf{Chapter~\ref{chapter_3}} contributes a formalisation of compliance in multi--level governance, providing a way to precisely determine compliant institution designs in a predictable way.
\item[] \textbf{Chapter~\ref{chapter_4}} contributes the computational means to go about detecting compliance in multi--level governance, furthermore this chapter contributes a computational framework which is sound and complete with respect to the more theoretical formalism in the previous chapter.
\item[] \textbf{Chapter~\ref{chapter_5}} contributes a computational mechanism to automatically rectify non--compliant institution designs.
\item[] \textbf{Chapter~\ref{chapter_6}} contributes a practical formalisation of institution design validity, looking at when rule changes count--as legal rule change enactments in the face of secondary rules governing the rule change enactment process.
\item[] \textbf{Chapter~\ref{chapter_7}} describes a prototype application of compliance checking institution designs (contracts in this case) to forming networks of crowdsourced users and their cellphones in order to collect and aggregate weather data whilst giving users autonomy over how, when and for whom their devices are used.
\item[] \textbf{Chapter~\ref{chapter_8}} concludes with discussion on the contributions, implications and directions for future work.
\end{itemize}

\section{List of Publications}
\label{secIntroPublications}

The chapters in this dissertation are based on the following publications:

\begin{itemize}
\item[] \textbf{King, T. C.}, Li, T., De Vos, M., Dignum, V., Jonker, C. M., Padget, J., \& van Riemsdijk, M. B. (2016, July). Automated Multi--level Governance Compliance Checking. \textit{Journal of Autonomous Agents and Multiagent Systems (JAAMAS)}. International Foundation for Autonomous Agents and Multiagent Systems. (\textbf{In Submission})
\item[] \textbf{King, T. C.}, Dignum, V., \& Jonker, C. M. (2016). When Do Rule Changes Count--as Legal Rule Changes? In \textit{Proceedings of the 22nd European Conference on Artificial Intelligence (ECAI 2016). Frontiers in Artificial Intelligence and Applications.} Vol 285. (pp. 3 – 11). IOS Press. http://doi.org/10.3233/978--1--61499--672--9--3
\item[] \textbf{King, T. C.}, Li, T., Vos, M. De, Jonker, C. M., Padget, J., \& Riemsdijk, M. B. Van. (2016). Revising Institutions Governed by Institutions for Compliant Regulations. \textit{Coordination, Organizations, Institutions, and Normes in Agent Systems XI: COIN 2015 International Workshops, COIN@ AAMAS, Istanbul, Turkey, May 4, 2015, COIN@ IJCAI, Buenos Aires, Argentina, July 26, 2015, Revised Selected Papers.}, 9628, 191 – 208.
\item[] \textbf{King, T. C.}, Li, T., De Vos, M., Dignum, V., Jonker, C. M., Padget, J., \& Riemsdijk, M. B. Van. (2015). A Framework for Institutions Governing Institutions. In \textit{Proceedings of the 2015 International Conference on Autonomous Agents and Multiagent Systems (AAMAS 2015)} (pp. 473–481). Istanbul, Turkey: International Foundation for Autonomous Agents and Multiagent Systems.
\item[] \textbf{King, T. C.}, Liu, Q., Polevoy, G., Weerdt, M. de, Dignum, V., Riemsdijk, M. B. van, \& Warnier, M. (2014). Request Driven Social Sensing (Demonstration). In A. Lomuscio, P. Scerri, A. Bazzan, \& M. Huhns (Eds.), In \textit{Proceedings of the 2014 International Conference on Autonomous Agents and Multiagent Systems (AAMAS 2014)} (pp. 1651 – 1652). Paris, France: International Foundation for Autonomous Agents and Multiagent Systems.
\item[] \textbf{King, T. C.}, Riemsdijk, M. B. Van, Dignum, V., \& Jonker, C. M. (2015). Supporting Request Acceptance with Use Policies. Coordination, Organizations, Institutions, and Norms in Agent Systems X: COIN 2014 International Workshops, COIN@ AAMAS, Paris, France, May 6, 2014, COIN@ PRICAI, Gold Coast, QLD, Australia, December 4, 2014, Revised Selected Papers (pp. 114 – 131). Springer.
\end{itemize}
\chapter{Background}
\label{chapter_2}

\epigraph[0pt]{There is a lot of work which tries to do sophisticated statistical analysis. You know, Bayesian so on and so forth. Without any concern for the actual structure of language. As far as I'm aware that only achieves success in a very odd sense of success [...] I don't know anything like it in the history of Science. \cite{Pinker2016a}}{Noam Chomsky}

\newpage

This chapter makes the following contributions:

\begin{itemize}
\item An overview of the informal background knowledge from multi--agent systems, philosophy of language and law, and political science. The focus is on institutions, and institutional design and enactment governance.
\item Formalisation requirements for the PARAGon framework.
\item A comparison of existing formal approaches and the extent to which they fulfil the PARAGon framework's requirements.
\end{itemize}

This dissertation formalises reasoning for understanding institutional design and enactment governance. The formalisation supports institution designers in designing compliant institutions and enacting institutional changes in a valid way. The same formalisation allows judiciaries to determine if punishments should be imposed for non--compliant institution designs. Moreover, the formalisation supports agents in understanding when institutional rules are changed. Generally, the objective is to \textit{formalise} the \textit{informal} concepts involved in governing institution design and rule change enactment to the extent that the reasoning can be executed by a computer.

In this chapter, we give background on the research contributed by this dissertation. First, we situate the research within the wider field of autonomous agents and multi--agent systems in Section~\ref{secBackAgents}. Then, Section~\ref{secBackGoverningMAS} gives a broad overview of institutions and norms from an informal point of view. Section~\ref{secBackFound} summarises the formal foundations required for this dissertation to build on and the new formal building blocks this dissertation develops in order to formalise institutional design and enactment governance. Section~\ref{secBackFormal} outlines a number of relevant formalisations of institutions on which the contributions of this dissertation can be based. Special attention is paid to identifying to what extent existing approaches can be built on by this dissertation and where there are knowledge gaps. We summarise the knowledge gap in Section~\ref{secBackApproach} and outline the approach we take.

\section{Agents and Multi--agent Systems}
\label{secBackAgents}

The research presented in this dissertation overlaps with the field of Agents and Multi--agent systems. Examples include socio--technical systems and human societies. To situate this dissertation, it is important to give some background on the general concepts used by the field.

When we talk about an agent, we mean an entity which has the ability to act on the world, or in other words possesses agency \cite{Schlosser2015}. Philosophy contributes many agent concepts, primarily dealing with human agents and action. Agency as the ability to act intentionally is a view taken by  Bratman \cite{Bratman1987}, Goldman \cite{Goldman1970} and Mele \cite{Mele2003}. Under this definition agents are able to bring about changes of their own volition to realise their intentions. The view that agency is the ability to initiate action spontaneously regardless of intention is taken by Ginet \cite{Ginet1990}, Lowe \cite{Lowe2008} and O'Connor \cite{OConnor2000}, among others. A more complicated agent, called a `person' is described by \cite{Frankfurt1971} as being able to act both on the world and also on their own internal mental states (e.g. realising second--order desires). Agents can be significantly simpler -- Dennett \cite{Dennett1987} questions the conceptual utility in requiring mental attitudes as pre--requisites for agency. He undercuts arguments for requiring mental attitudes in the agent concept with Occam's razor -- they should only be ascribed to agents if they help with predicting action. Agents need not be human or biological at all, Wooldridge \cite{Wooldridge1995,Wooldridge2002} offers a frequently used definition of an artificial autonomous agent in the MAS community. According to Wooldridge, an agent is software able to act in an environment to meet its own goals, often those given to it at design--time. Without attempting to resolve the differences between these stances the key point subscribed to by this dissertation and the wider field is that agents are able to act on their environment, by extension act in ways that affect one another and importantly act \textit{autonomously} in ways that cannot be controlled directly.

A system comprising agents is called a multi--agent system (MAS). Such systems can be biological (e.g. a nation state comprising humans), artificial (e.g. a software system of agents controlling sensors) or mixed (e.g. a system comprising teams of humans and robots). An MAS is often formed with a goal in mind or an idea of how the agents should behave, from the perspective of the system's stakeholders (e.g. the members of a nation state or the designer of a software system). For example, a nation state may have the societal goal to minimise the number of car accidents. Another example is a system comprising software agents on people's cellphones for collecting and aggregating location and sound level data to determine large crowds of people \cite{Lu2009}. To achieve the MAS' ideal and system goals as envisioned by the relevant stakeholders, the agents in an MAS need to act collectively. For example, driving on the same side of the road to avoid collisions or configuring sensors appropriately (e.g. ensuring a microphone is on) for the data being gathered. An MAS is a system comprising agents which is typically designed with an aim or ideal for agents to collectively achieve.

The autonomy of agents in an MAS makes it difficult to ensure agents collectively achieve the MAS' goals. Agents are liable to acting in their own self--interests (e.g. turning cellphone sensors off to conserve cellphone energy rather than contributing data) and if agents are left to their own devices their actions are liable to be uncoordinated (e.g. driving on different sides of the road). Mechanisms for system--wide agent control are required to ensure agents achieve the MAS' goals.

Different mechanisms to control and coordinate agents are available. We split these mechanisms into \textit{bottom--up control}, \textit{top--down governance} and \textit{bottom--up compliance with top--down control}. Which approach is appropriate depends on the level of control a system stakeholder (e.g. software designer, national government) has over the agents and conversely, the level of agency an agent has.

\begin{itemize}
\item \textbf{Bottom--up control} -- this approach relies on being able to control artificial agents directly. Specifically, by programming agents to coordinate their behaviour and collectively achieve the MAS' goals. Even basic software agents only capable of reacting to a single percept with no memory of past percepts and actions have been shown to exhibit complex emergent coordination such as decentralised gathering and collecting of objects \cite{Gauci2014}. For more complex cases agent programming languages have been proposed, often based on an architecture where reasoning is compartmentalised into mental constructs such as beliefs, desires and intentions (BDI). In these architectures agents can communicate and therefore be programmed to coordinate their behaviour (e.g. \cite{Dastani2008a, DeBoer2001, Hindriks2000, Rao2009, R.H.Bordini2007}), in the coo--BDI agent architecture for BDI (Belief, Desires, Intentions) agents' cooperation and coordination is supported with specific constructs for sharing plans \cite{Ancona2003}. Overall, the \textit{bottom--up control} approach works by coordinating an MAS from the \textit{bottom} level comprising the agents and their internal reasoning by regimenting the behaviour of each individual agent, such as by programming their behaviour in being coordinated and collaborative.
\item \textbf{Top--down governance} -- this approach deals with cases where agents cannot be controlled directly. For example, when the agents are biological (e.g. humans). Since such agents cannot be controlled directly they are liable to act in their own self--interests. A common example is the tragedy of the commons \cite{Hardin1968}, where given no constraints it is optimal for agents to act in their self--interests based on the premise that other agents will be doing the same. A top--down approach is also required when an \textit{open artificial multi--agent system} \cite{Hewitt1986} is designed comprising heterogeneous agents that are unknown \textit{a--priori}. In an open artificial MAS agents are liable to changing over time and are contributed by various programmers such that they are not (easily) reprogrammable and hence should be guided in behaving in the right way rather than being directly controlled. In top--down governance regulations prescribe how agents should behave (e.g. agents should drive on the left) coupled with rewards/punishments to give agents reason to comply (e.g. fines). Whilst in this situation direct control of agents cannot be automated, the top--down governance can. For example, formal and/or automated reasoning for: monitoring agents' compliance \cite{Bulling2013,Chesani2012,Faci2008,Fornara2010,Modgil2009}; determining which regulations apply and when and which agents are compliant and should be rewarded or punished \cite{Boella2003d, Boella2004, Boella2004c, Boella2009a, Chopra2016, Cliffe2006, Cliffe2007, Dignum2002, DInverno2012, Governatori2005a, Governatori2005, Governatori2005a, Governatori2005b, Governatori2007, Governatori2009, Governatori2010, Governatori2013, Jiang, Jianga, Jiang2011, Lopez2003, Lopez2006}; synthesising institutions \cite{Morales2011,Morales2013,Morales2014}; verifying institution designs meet certain correctness properties and do not enable agents to exploit regulations \cite{Knobbout2014,MaxKnobbout2016}. In this approach societal coordination is realised by \textit{top--down governance} in the sense that it operates at a level sitting above the entire system and governs agents rather than directly controls them.
\item \textbf{Bottom--up compliance with top--down governance} -- this approach lies in between the two previously mentioned approaches as a way to automate compliance with top--down governance. In this approach, norm--aware agents are programmed. A norm--aware agent could be an artificial agent contributed to an open MAS with top--down governance. Here, the agent can act appropriately to comply with the top--down governance with no assumptions that the regulations placed on the agent will be known \textit{a--priori}. Research in this area has focussed on agent programming languages, semantics and reasoning for agents acting in a compliant or preferable way way with the regulations and organizations the agents might join (e.g. \cite{Alechina2012, Criado2015, Dastani2008a, Hubner2007, Jensen2014, Jensen2015, Lee2014, Meneguzzi2009, Shams2016, Tinnemeier2009}).
\end{itemize}

This dissertation provides reasoning for both top--down governance and bottom--up compliance with top--down governance. Top--down governance reasoning is provided for determining whether institutions are being designed in a bad way (i.e. the designs themselves are non--compliant in multi--level governance) or institutional changes are enacted without following procedure (i.e. secondary legal rules are not adhered to and thus institutional changes are invalid). Bottom--up compliance with top--down governance is provided for institution designers themselves, so that they can design institutions in a compliant way and validly enact institutional changes.  In the former case, bottom--up compliance is supported with automated reasoning to find explanatory rectifications for non--compliant institutions. In the next sections we will give background on institutions and governing institutional design and enactment.

\section{Governing Multi--Agent Systems}
\label{secBackGoverningMAS}

In this section we discuss governance of multi--agent systems from a conceptual point of view. This dissertation builds on concepts of institutions, discussed in Section~\ref{secBackGoverningMASInstitutions} and norms discussed in Section~\ref{secBackGoverningMASNorms}. This dissertation contributes new formal conceptualisations involved in institutional design and enactment governance. Background informal concepts for institution design and enactment governance are described in Section~\ref{secBackLegAndInstDesi}.

\subsection{Institutions}
\label{secBackGoverningMASInstitutions}

Institutions provide top--down governance of agents. A definition from economics is given by North \cite{North1990}, as a set of rules regulating behaviour enforced with rewards and punishments. For example, fining agents or simply deviating from certain rules being considered a societal taboo and therefore affecting trust. Ostrom \cite{Ostrom1990} provides a similar definition, put best in her own words as ``the prescriptions that humans use to organize all forms of repetitive and structured interaction''. In legal theory Ruiter \cite{Ruiter1997} agrees with North and also analogously refers to institutions as a kind of ``social agent'', realised as agents practising the institution with behaviour that complies with the institution's rules. John Austin \cite{Austin1832} viewed the law (legal institutions) as rules laid down by one individual who has power over another. Alternatively, Ruiter \cite{Ruiter1997} also views contracts between agents as legal institutions. In legal positivism, legal rules' validity depends on the source (i.e. whether legislated by a legislator) \cite{Gardner2001}. According to these definitions regulative rules are an essential part of an institution and when an institution is legislated, such as by a parliament, it is a \textit{legal} institution.

In philosophy \cite{Miller2014} an institution is viewed as comprising regulative rules but also rules establishing the language and concepts with which agents and a social reality are talked about. For example, a rule stating `I do' establishes what we call `marriage'. Such rules come about through social acceptance. Concepts such as marriage established by an institution, according to Searle \cite{Searle1969,Searle2005}, are institutional facts and come about only because an institution makes social actions (e.g. marrying) \textit{possible}. These institutional facts describe a \textit{social reality} ascribed by the institution, giving an institutional interpretation of the \textit{brute} (actual) reality. Searle refers to the facts that are independent of an institutional interpretation as \textit{brute facts}, referring to a ground truth. For example, an institutional fact might be `paper' which refers to a brute fact `the thing we call paper'. Searle proposes constitutive institutional counts--as rules of the form ``A counts--as B in a context C'', to which from brute facts (As) ascribe more abstract institutional facts (Bs) in a particular social context entailed by the social reality (Cs). For example, the thing we call paper counts--as paper. In turn institutional facts may count--as further institution facts. For example, an agent does not just possess paper with 10 euros written on it, various watermarks and an identification number. Rather, that piece of paper is collectively agreed as counting--as money in the context of a particular nation state and so the agent possesses money in the right context. Counts--as rules establish a concise social description of reality, which can feasibly be talked about at a more abstract, social level, than would otherwise be possible (e.g. `money' is far more concise than a fine--grained description of everything that constitutes money).

Searle  argues counts--as rules also ascribe special statuses and  deontic powers to various concepts \cite{Searle2005}. For example, a piece of paper that counts--as money gives an agent possessing it the special deontic power to purchase goods. But, only because there is a social institution ascribing a special status to the money. On the deontic dimension of institutions, Searle says that various obligations are entailed by an institutional fact having a certain status. For example, a police person has a socially recognised status where they are obliged to uphold the law. In fact, Searle states that counts--as rules ascribing institutional facts and assigning statuses to those facts are necessary for there to be a deontology -- ``\textit{No language, no status functions. No status functions, no institutional deontology}'' \cite[p.14]{Searle2005}. Searle states institutions contain constitutive counts--as rules, which create a social reality from brute facts, and counts--as rules are necessary for a deontology.

Ricciardi \cite{Ricciardi1997} views related constitutive rules (e.g. all the rules ascribing meaning and status to money) as \textit{constituting} the institution they belong to (e.g. the `money' institution). A classic example from Ricciardi is the rules of chess constituting the game of chess and the \textit{institution} of chess. Moves can physically be made on a chess board that do not conform to chess' rules, but then the game of chess is not being played and the chess institution does not recognise those moves. Hence, an institution (e.g. chess) \textit{is} its set of rules.

The key points in these definitions subscribed to by this dissertation are:

\begin{itemize}
\item Institutions can be created by law makers (legislation), between agents (contracts) or through social acceptance. This dissertation mostly focuses on legal case--studies, but we do not tie ourselves to any particular institution source.
\item Institutions \textit{are} a set of related counts--as rules, which ascribe a social meaning to brute facts by building an institutional reality comprising institutional facts (n.b. this differs from the more complicated notion of an institution as a social agent). Chapter~\ref{chapter_3}, and Chapter~\ref{chapter_6} focus on formalising constitutive rules.
\item Institutions are used to govern and guide agents towards an ideal with rules and regulations (a deontology), rewards and punishments. Chapters \ref{chapter_3} to \ref{chapter_5} focus on this regulatory aspect of institutions.
\item Counts--as rules and a social reality are necessary for an institution to have a deontology.
\end{itemize}

In the next section we will describe in more detail the deontic aspect of institutions and its informal representation.

\subsection{Norms}
\label{secBackGoverningMASNorms}

In this section we discuss institutional regulations, called norms, further. Norms state what the social reality \textit{should} be (e.g. an agent should pay for a train ticket before boarding a train). With a suitable mechanism to ensure norms are abided by, norms guide agents towards an ideal. Two mechanisms to ensure norms are abided by are discussed in the literature: regimentation and enforcement.

\textbf{Regimentation} \cite{Jones1993} forces norm compliance on agents by controlling their behaviour directly. For example, by programming agents so they cannot board a train unless they have paid for a ticket or by constructing a barrier at the train station that does not let agents onto the train platform unless they have a ticket. One drawback of regimentation is that it assumes  hard constraints can be implemented (e.g. an agent can be directly programmed or there are physical barriers that can be constructed). Another drawback is that regimentation does not acknowledge that there are exceptional circumstances under which norms should not be followed -- for example when an agent needs to drive faster than the speed limit in the case of an emergency. Regimented norms act as hard constraints on agents' actions, which is not always possible or in the best interests of the agents and the system. We do not consider norm regimentation in this dissertation.

\textbf{Enforcement} \cite{Grossi2007a} gives agents the option to comply with norms that  act as soft constraints on agents' behaviour. To give reasons for agents to comply, secondary reward and punishment norms come into force if a primary norm is violated. For example, a norm stating an agent should pay a fine for boarding a train without a ticket. In turn, the secondary norms can have further enforcing norms. For example, a train guard that finds an agent who refuses to pay a fine should throw the non--compliant agent off the train. This dissertation considers norms that can be violated and are supported through enforcement.

Norms take two common forms in the literature. Firstly, an \textit{evaluative} representation lacking an explicit deontic modality (obligation/permission/prohibition). Secondly, a \textit{modal} representation where the deontic modality is explicit.

\textbf{Evaluative} norms are found in some formal work \cite{Anderson1958,Grossi2008a}. In an evaluative representation norms are represented as counts--as rules denoting that ``A counts--as being good/bad/compliant/a violation in a context C'' with no reference to obligation/permission/prohibition. To give an example based on the UK's data retention regulations \cite{UK2009} -- a communications provider (e.g. telephonics company) retaining all of a person's communications meta--data (e.g. the time a phone call was made) counts--as compliance in the context that the communications provider is operating in the United Kingdom. An evaluative norm creates evaluative institutional facts in the social reality. For example, creating `compliance' if meta--data is stored. The social reality does not represent which norms are in force (what should be done). If we want to know whether the UK's data retention regulations require meta--data to be stored then the query is ``do the rules state storing meta--data counts--as compliance in a context that holds in the present social reality?''. Determining which evaluative norms are in force requires assessing the constitutive rules for whether they ascribe a concept counting--as compliance/violation in a context that holds in the social reality.

\textbf{Modal} norms are counts--as statements ascribing explicit deontic (obligation/prohibition/permission) positions in the social reality. Specifically, as Searle proposed in \cite[p. 63]{Searle1969} ``A counts--as undertaking an obligation to do B in context C''. For example, communicating using a communications provider's services establishes an obligation for the communications provider to store your meta--data in the context that they are operating in the United Kingdom. A modal representation creates a deontic position in the social reality. For example, an obligation to store communications meta--data. The social reality explicitly represents what should be the case or should be done. If we want to know whether the UK's data retention regulations require meta--data to be stored then we can look at the social reality for whether storing meta--data obliged. Determining which modal norms are in force requires inspecting the built social reality for which deontic modal statements hold.

In this dissertation, one objective is to reason about norms that govern other norms. Expressing and determining that particular norms should (not) be imposed differs between modal and evaluative norms. There are two ways for a norm to express that it is \textit{required that other norms do not require} storing communications metadata in the context that a user has not consented to their communications metadata being stored. 

In the evaluative form one possible representation is to nest a counts--as rule within another counts--as rule. For example, ``the rule (storing metadata counts--as compliance in a context C) counts--as a violation if context C is somehow compatible with the user not consenting)''. There may be other evaluative representations, but it appears to be the simplest that fully captures a norm stating the requirement for other norms to not require a specific social fact. In a modal form an unconditional constitutive rule seems to express the same thing -- ``it is prohibited to oblige a user's metadata to be stored in the context that they have not consented''.

The modal norm appears simpler and with little difference between prohibiting an obligation to store metadata and  the modal norm obliging metadata to be stored. We would expect that determining compliance is more or less the same for a modal norm obliging storing metadata compared to a modal norm prohibiting obliging storing metadata. In comparison, the evaluative version seems far more complex and is not a simple generalisation of evaluative norms about events. This dissertation considers a modal norm representation for simplicity of reasoning about norms that regulate other norms.


Norms prescribe a social reality and are often used together with counts--as rules ascribing non--normative institutional facts. According to Sergot \cite{Sergot1982,Sergot1988}, and Jones and Sergot \cite{Jones1993},  a legal system (what we call an institution) largely consists of rules that define concepts (what we call counts--as rules ascribing institutional facts). Secondary in occurrence to rules defining concepts are the institution's norms. Definitional rules are used for succinctness, according to Ross \cite{Ross1956}. For example, we can define storing communications metadata or content data as counting--as storing personal data. If we want to describe a new concept relating to personal data then we can refer to personal data rather than everything that constitutes personal data. For example, disrespecting someone's privacy is storing their personal data without their consent, rather than storing their metadata or content data or ... without their consent. Such definitional rules, according to Grossi et al. \cite{Grossi2005} and Breuker et al. \cite{Breuker1997}, allow norms to be issued more succinctly over abstract concepts. For example a norm stating ``you should not disrespect someone's privacy'' is more succinct than ``you should not store content data without consent'', ``you should not store metadata without consent'', etcetera. Institutions largely comprise concept defining rules, which ascribe an abstract social reality, over which norms are concisely defined.

The key points in these definitions subscribed to by this dissertation are:

\begin{itemize}
\item Norms can take an evaluative or a modal form. An evaluative form does not make explicit that there is an obligation/prohibition/permission in force, it simply states a social action will cause compliance or a violation. A modal form makes explicit that something is obliged/prohibited/permitted in the social reality. The modal form offers a simpler way to represent norms governing other norms and a straightforward way to evaluate such norms. Chapters \ref{chapter_3} to \ref{chapter_5} formalise modal norms.
\item Two ways to ensure norms are abided by are found in the literature -- regimentation and enforcement. Regimentation assumes agents can be directly controlled. Enforcement does not assume agents' actions can be controlled, instead it aims to guide agents' behaviour. This dissertation does not make assumptions that agents can be controlled directly. In chapters  \ref{chapter_3} to \ref{chapter_5} we look at soft constraints which guide, rather than regiment, agents in their actions within an MAS and how they design institutions. However, in Chapter~\ref{chapter_6} we look at a kind of hard constraint in the form of rules stating which institutional rule changes are possible. These are not hard constraints in the traditional sense (e.g. physically preventing an agent to board a train or directly controlling an agent's reasoning to comply with norms). So, we call these hard  \textit{institutional} constraints.
\item Norms are defined in terms of an abstract social reality created by counts--as rules. Counts--as rules and norms are closely tied, counts--as rules by building an abstract social reality provide a pragmatic means to author concise norms. The link between constitutive rules ascribing abstract concepts and norms defined over these abstract concepts is focussed on in Chapter~\ref{chapter_3} and Chapter~\ref{chapter_4}. Counts--as rules that make the social action of changing rules possible are addressed in Chapter~\ref{chapter_6}.
\end{itemize}

\subsection{Governing Institutional Design and Enactment}
\label{secBackLegAndInstDesi}

In this section we discuss how institution designers, such as legislators, are governed. The need to govern institution designers is motivated by the fact that they are autonomous agents. Being autonomous agents, institution designers are liable to behaving in ways that are sub--ideal. For example, designing an institution that takes away agents' rights or enacting institutional rule changes without following relevant procedure, such as obtaining a majority vote from a parliament. Hence, we look at how institutional designers are governed in their institutional designs and institutional change enactments, which we describe further from an informal point of view in this section.

\subsubsection{Governing Institutional Design: Multi--level Governance}
\label{secBackLegAndInstDesiMLG}

The problem with agency is that agents are liable to acting undesirably and hence need institutional governance; the problem with institutional governance is that autonomous agents can design subjectively `bad' institutions. For example, an institution can lack coordinated regulations with institutions governing other jurisdictions or place unacceptable limits on agents' rights. One solution is to override an institution designers' regulations. But this would violate their autonomy and the principle of subsidiarity -- what can be legislated at the local level should be left up to the local level. Hence, institution designers need to be guided in designing subjectively good institutions. In MAS agents are guided towards good behaviour with institutions governing their actions. In order to guide institution designers, institutions are used to govern institution \textit{designs}.

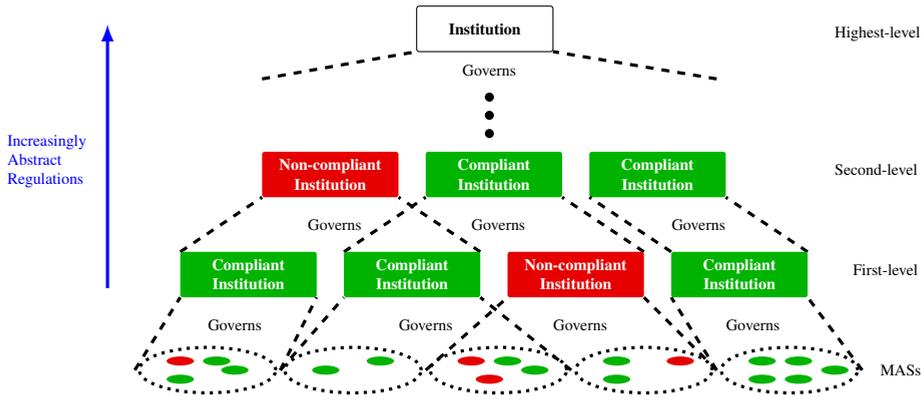
\begin{figure}
\centering
\begin{tikzpicture}[scale=0.6, every node/.style={scale=0.6}]
\node at (-16.8,1.8) {\parbox{20mm}{\color{blue}Increasingly Abstract Regulations}};
\draw[very thick, color=blue, >=latex,->] (-15.6,-1) -- (-15.6,4.8);

\draw [rounded corners=1pt] (-8.8,5.2) rectangle (-5.8,4.2) node[pos=.5] {\textbf{Institution}};
\draw[very thick, dashed] (-12.2,3.6) -- (-8.8,4.2);
\node at (-7.2,3.8) {Governs};
\draw[very thick, dashed] (-2.2,3.6) -- (-5.8,4.2);

\fill (-7.2,3.2) circle (1mm);
\fill (-7.2,2.8) circle (1mm);
\fill (-7.2,2.4) circle (1mm);

\fill[rounded corners=1pt, black!10!red] (-12.2,2) rectangle (-9.2,1) node[pos=.5] {\parbox{30mm}{\centering \color{white}{\textbf{Non-compliant Institution}}}};
\fill[rounded corners=1pt, black!30!green] (-8.6,2) rectangle (-5.6,1) node[pos=.5] {\parbox{30mm}{\centering \color{white}{\textbf{Compliant Institution}}}};
\fill[rounded corners=1pt, black!30!green] (-5,2) rectangle (-2,1) node[pos=.5] {\parbox{30mm}{\centering \color{white}{\textbf{Compliant Institution}}}};

\draw[very thick, dashed] (-14,-0.2) -- (-12.2,1);
\node at (-10.6,0.4) {Governs};
\draw[very thick, dashed] (-7.4,-0.2) -- (-9.2,1);

\draw[very thick, dashed] (-10.4,-0.2) -- (-8.6,1);
\node at (-7,0.4) {Governs};
\draw[very thick, dashed] (-3.8,-0.2) -- (-5.6,1);

\draw[very thick, dashed] (-3.2,-0.2) -- (-5,1);
\node at (-2.6,0.4) {Governs};
\draw[very thick, dashed] (-0.2,-0.2) -- (-2,1);

\fill[rounded corners=1pt, black!30!green] (-14,-0.2) rectangle (-11,-1.2) node[pos=.5] {\parbox{30mm}{\centering \color{white}{\textbf{Compliant Institution}}}};
\fill[rounded corners=1pt, black!30!green] (-10.4,-0.2) rectangle (-7.4,-1.2) node[pos=.5] {\parbox{30mm}{\centering \color{white}{\textbf{Compliant Institution}}}};
\fill[rounded corners=1pt, black!10!red] (-6.8,-0.2) rectangle (-3.8,-1.2) node[pos=.5] {\parbox{30mm}{\centering \color{white}{\textbf{Non-compliant Institution}}}};
\fill[rounded corners=1pt, black!30!green] (-3.2,-0.2) rectangle (-0.2,-1.2) node[pos=.5] {\parbox{30mm}{\centering \color{white}{\textbf{Compliant Institution}}}};

\draw[very thick, dashed] (-15,-2.8) -- (-14,-1.2);
\node at (-12.8,-1.8) {Governs};
\draw[very thick, dashed] (-11.8,-2.8) -- (-11,-1.2);
\draw[very thick, dashed] (-11.8,-2.8) -- (-10.4,-1.2);
\node at (-8.6,-1.8) {Governs};
\draw[very thick, dashed] (-5.4,-2.8) -- (-7.4,-1.2);
\draw[very thick, dashed] (-8.6,-2.8) -- (-6.8,-1.2);
\node at (-5.2,-1.8) {Governs};
\draw[very thick, dashed] (-2.2,-2.8) -- (-3.8,-1.2);
\draw[very thick, dashed] (-2.2,-2.8) -- (-3.2,-1.2);
\node at (-1.4,-1.8) {Governs};
\draw[very thick, dashed] (1,-2.8) -- (-0.2,-1.2);

\draw[very thick, dotted] (-13.4,-2.8) ellipse (1.5 and 0.5);
\fill[black!30!green] (-14,-3) ellipse (0.3 and 0.1);
\fill[black!30!green] (-12.8,-2.8) ellipse (0.3 and 0.1);
\fill[black!10!red] (-14,-2.6) ellipse (0.3 and 0.1);
\fill[black!30!green] (-13.2,-2.6) ellipse (0.3 and 0.1);

\draw[very thick, dotted] (-10.2,-2.8) ellipse (1.5 and 0.5);
\fill[black!30!green] (-10.8,-2.8) ellipse (0.3 and 0.1);
\fill[black!30!green] (-9.6,-2.6) ellipse (0.3 and 0.1);

\draw[very thick, dotted] (-7,-2.8) ellipse (1.5 and 0.5);
\fill[black!30!green] (-6.2,-2.8) ellipse (0.3 and 0.1);
\fill[black!10!red] (-7.2,-3) ellipse (0.3 and 0.1);
\fill[black!10!red] (-7.6,-2.6) ellipse (0.3 and 0.1);
\fill[black!30!green] (-6.8,-2.6) ellipse (0.3 and 0.1);

\draw[very thick, dotted] (-3.8,-2.8) ellipse (1.5 and 0.5);
\fill[black!30!green] (-4.4,-3) ellipse (0.3 and 0.1);
\fill[black!30!green] (-4.4,-2.6) ellipse (0.3 and 0.1);
\fill[black!10!red] (-3,-2.6) ellipse (0.3 and 0.1);

\draw[very thick, dotted] (-0.6,-2.8) ellipse (1.5 and 0.5);
\fill[black!30!green] (-1.2,-3) ellipse (0.3 and 0.1);
\fill[black!30!green] (0.4,-2.8) ellipse (0.3 and 0.1);
\fill[black!30!green] (-1.2,-2.6) ellipse (0.3 and 0.1);
\fill[black!30!green] (-0.4,-2.6) ellipse (0.3 and 0.1);
\fill[black!30!green] (-0.4,-3) ellipse (0.3 and 0.1);

\node at (2.4,4.6) {\parbox{40mm}{Highest-level}};
\node at (2.4,1.6) {\parbox{40mm}{Second-level}};
\node at (2.8,-0.6) {\parbox{40mm}{First-level}};
\node at (3.4,-2.8) {\parbox{40mm}{MASs}};
\end{tikzpicture}
\caption[High--level multi--level governance overview]{A high--level overview of multi--level governance. At the bottom level there exists MASs (e.g. nation states, sensor networks). The first level comprises institutions governing MASs (e.g. national legislation, or institutions governing artificial societies such as sensor networks). The second--level comprises institutions governing the first--level institutions (e.g. EU directives, cross--MAS objectives). The highest--level comprises institutions governing lower--level institutions (e.g. human rights charters and constitutions).}
\label{figMLGBackground}
\end{figure}

\textit{Multi--level governance} is an umbrella term for governance styles where institution designs themselves are governed. We depict it abstractly in Figure~\ref{figMLGBackground}. The term multi--level governance is taken from political science \cite{Liesbet2003} (alternatively called multi--tier and polycentric governance). An early conceptualisation by Marks \cite{Marks1993} views multi--level governance as negotiation between state actors in the European Union to achieve coordinated goals through regulatory change. The negotiated outcome being binding agreements/institutions between EU member states to design their institutions (national legislation) in such a way that they implement those binding agreements. In effect, the negotiations establish a hierarchy of institutions. The negotiated binding agreement being a higher--level institution governing the design of member states' legislation, which are lower level institutions. More recently, Börzel and Risse \cite{Borzel2000} conceive multi--level governance as existing at sub--national, national and supranational levels. Conceptually, elected institutional designers (i.e. local councils, national governments, EU governing bodies) create institutions to govern institutions at lower governance levels and comply with institutions enacted at higher governance levels. Hooghe and Marks \cite{Liesbet2003} classify multi--level governance into two types. In the first type, an institution at a higher--level governs a jurisdiction of institutions nested at a level below. In this type lower--level institutions do not belong to any other jurisdiction and are only governed by a single institution at the level above. In the second type institutions at each level can belong to multiple jurisdictions and therefore be governed by multiple higher--level institutions. More generally, multi--level governance comprises institutions at higher governance levels (e.g. EU directives) governing the institution designs operating at lower governance levels (e.g. national legislation) which in turn might govern institution designs at even lower governance levels (e.g. sub--national regulations) etcetera.

Multi--level governance guides institution designers towards enacting ideal regulations. Eichener \cite{Eichener1997} showed multi--level governance in the European Union successfully resulted in cooperating legislation on occupational safety. By issuing higher--level institutions, an EU directive, the EU established workplace health and safety regulations across the EU. The health and safety regulations were coordinated in the sense that no country had more relaxed regulations allowing them to out--compete other member states on labour costs. Hence coordinated regulations ensured no `race to the bottom' was triggered. From a human rights point of view, multi--level governance has ensured legislation that violated certain rights of agents is revoked, such as the right to privacy set out by EU human rights law \cite{ECHR1953,ECJDRD}. In Artificial Intelligence, Diaconescu and Pitt \cite{Diaconescu2014}, Pitt and Diaconescu \cite{Pitt2014}, and Jiang, Pitt and Diaconescu \cite{Jiang2015}, view the multi--level governance paradigm as being applicable to artificial societies such as smart grids. In \cite{Diaconescu2014} sub--systems of energy users are governed by institutions, each sub--system (e.g. a small housing community) has its own governance to manage its own energy production and usage. Sub--systems can cooperate, requiring coordinated regulations, such as to ensure greater energy stability by sharing energy. Coordinated regulations can be achieved with hierarchical governance, where an overarching institution ensures institution's regulations at lower--levels are compatible. Multi--level governance helps to ensure that regulations are subjectively ideal in the sense of being coordinated across jurisdictions and upholding agents' rights.

In multi--level governance, institution designs are governed with regulations that govern other regulations. Since multi--level governance is regulatory it implies compliance is possible but not guaranteed. Hence, regulations governing regulations also give the possibility for non--compliant institution designs. In order to ensure institution designers comply, non--compliance can result in legal action being brought about \cite{Scharpf1997}, which Smith \cite[p. 15]{Smith2004} argues is necessary for effective multi--level governance.

From a philosophical tradition, Von Wright argues that \cite{Wright1983} regulation governing regulations (or in his words, higher-order norms) are really there to \textit{transmit will}. By which we mean, for example, a regulation requiring another regulation to oblige communications meta-data is stored, is really issued because the issuer wishes not just that there is an obligation for meta-data to be stored, but also that meta-data is \textit{actually} stored. That is, the issuer is transmitting their will for meta-data to be stored via an intermediary regulator, operating at a lower governance level, acting as a conduit for their will. Following this idea, it has been shown by Boella and van der Torre \cite{Boella2006a} that in order to transmit will via regulation governing regulations there must be regulation \textit{enforcement} with reward and/or punishment. Two types of enforcement required for effective regulation governing regulations, need to be distinguished. First, enforcement of the regulation governing regulations, such as punishing EU member states for not implementing an EU directive, which helps to ensure the will is transmitted from the higher-authority to a lower-authority. Secondly, the enforcement of the regulations governing society, which are designed to be compliant with regulation governing regulations. For example, the Data Retention Directive \cite{EUDRD} requires other regulations to not just oblige meta-data to be stored but also to enforce that obligation by punishing for not storing meta-data. Without enforcement, a regulation could be compliant with a directive by obliging meta-data to be stored and yet that obligation would be meaningless since it is unenforced. Moreover, enforcement is required so that the will of the  higher-authority (e.g. an issuer of an EU directive) is transmitted all the way down to the societal level. Philosophically, regulation governing regulations only make sense if firstly they are enforced through reward/punishment for (non--)compliance and secondly if they also require the enforcement of the regulations that they require to be implemented (if any).

Determining such (non--)compliance in multi--level governance is different from determining the compliance of societal members with an individual institution. The main difference is that in multi--level governance, when compared to single--levelled societal governance, regulations operate at different levels of abstraction. Regulations are the most abstract in the highest--level institutions and the most concretely defined in the lowest--level institutions. To give an example, at the (typically) highest level of governance human rights charters use abstract terminology such as `fairness' or `privacy' which can have many different interpretations. At a slightly lower--level, such as supranational agreements or EU directives, the terminology is more precise but countries can comply in different concrete ways. For example, the Data Retention Directive \cite{EUDRD} states that member states should legislate for communications metadata (e.g. the time of a phone call) to be stored between 6 and 24 months. The directive's regulation is far clearer than human rights regulations, but does not provide the precise data retention time. At a slightly lower--level, such as at the level of nation--states, regulations are more concrete. For example, providing a precise time in which data should be stored. In multi--level governance increasingly abstract regulations are prescribed at increasingly higher levels of governance which can be interpreted in many different ways, making determining compliance of institutional designs different from determining compliance of a society's members.

Institutional design compliance is determined by legal monitors such as courts, which interpret concrete regulations to determine if they violate more abstract regulations. To give an example, the European Court of Justice \cite{ECJDRD} determined that the Data Retention Directive's relatively concrete requirement for communications metadata to be stored violated the EU Human Rights Charter's requirement for personal data to be processed fairly \cite{EU2000}. The judgement was based on the interpretation that storing metadata was the same as storing personal data, and storing personal data without someone's consent was the same as processing data unfairly. In a different context, where someone has given consent, storing personal data would not be unfair data processing. Hence, a relationship between concrete concepts having a context--sensitive abstract meaning is used to determine compliance between concrete and abstract regulations. According to the Searlian institution concept we adopt, the context--sensitive rules linking concrete and abstract social facts are constitutive rules. Hence, the relation between concrete and abstract \textit{norms} as found in multi--level governance is derived from constitutive rules.

To summarise, institutions govern the design of other institutions in what is known as multi--level governance, the main points are:

\begin{itemize}
\item Constitutive rules provide context--sensitive links between concrete and abstract concepts and through a derivation, concrete and abstract norms. Chapter~\ref{chapter_3} and Chapter~\ref{chapter_4} adopt constitutive rules to determine links between concrete and abstract norms in different social contexts.
\item In multi--level governance abstract regulations at higher--levels are used to govern concrete regulations at lower--levels. Non--compliant concrete regulations results in punishment. Chapter \ref{chapter_3} and Chapter~\ref{chapter_4} look at formalising and automatically detecting (non--)compliance in multi--level governance. The objective is to flag problems to institution designers before they enact non--compliant institutions (and hence are subject to punishment) and to support higher--levels of governance in monitoring the compliance of lower--level institution designs. In order to support institutional designers in enacting compliant designs, Chapter~\ref{chapter_5} looks at finding explanatory rectifications for non--compliant institutional designs.
\item Enforcement should be ensured for regulation governing regulations in two senses. Firstly, the regulation governing regulations should be supported with rewards and/or punishments. We do not look specifically at formalising reward and punishment for regulation governing regulations, but the fact that it should and does exist motivates the contribution of automated compliance checking in Chapter~\ref{chapter_3} and Chapter~\ref{chapter_4}. Secondly, regulations governing regulations issued by higher-level authorities really require that lower-level authorities do not just implement certain regulations in order to be compliant, but also enforce those implemented regulations. We do not formalise the general principle of both implementation and enforcement required by regulation governing regulations, but we do formalise the \textit{specific} requirement of the Data Retention Directive \cite{EUDRD} to be supported with punishments for non-compliance in Chapter~\ref{chapter_3} and Chapter~\ref{chapter_4}.
\end{itemize}

\subsubsection{Governing the Institution Design Process: Constitutive Secondary Legal Rules Ascribing Rule Change}

Constitutive rules define how a social reality grows out of brute facts describing the ground truth. Moreover, constitutive rules make certain social actions, such as marriage, possible. Since brute facts (the ground truth) are liable to changing, it follows that the social reality is liable to changing too and so too the social actions that are possible. Consequently, constitutive rules both describe how the social reality is built and how it evolves over time. Biagoli \cite{Biagioli1997} views an institution itself as having similar dynamic qualities, referring to institutions (legal systems) as organic sets of rules that in themselves change over time. Constitutive rules regulate the change of the social reality but are also subject to changing themselves.

According to Hart \cite{Hart1961} the complexity of institutions necessitates institutional rules with the sole function to regulate rule change. Hart refers to these rule--change regulating rules as \textit{secondary rules}. Secondary rules state how and when rules can be changed, modified, repealed and by whom. To give an example, the Italian constitution states that after a law is voted to be enacted by one of the houses of government it is \cite{RepublicOfItaly1947}[Art. 73] ``promulgated by the President of the Republic within one month of their approval''. Such rules  regulate rule change by describing the institution design \textit{process} that must be followed for rule changes to take place. In other words, rules exist with the exclusive function to state what actions constitute rule changes and thereby make the social actions of changing institutional rules possible.

Biagoli \cite{Biagioli1997} views rules regulating rule change as \textit{constitutive} rules that ascribe what constitutes a rule change. From this perspective, a physical rule change does not necessarily count--as a rule change. Physically changing a rule in the Italian rule book (i.e. writing it down) does not count--as changing the rule on its own. Rather, only the president approving a law previously voted for by one of the government houses counts--as a rule change, making the social action of rule change possible.

Following this idea, just as constitutive rules ascribe a social reality, making the social reality possible, constitutive rules ascribing rule change make rule changes possible. An analogy can be made to Ricciardi's argument \cite{Ricciardi1997} of  the rules of chess constituting the game of chess where moves made outside of chess' rules can physically be made but then the game of chess is not being played. Likewise, physical rule modifications can be made that do not follow an institution's constitutive rule change rules, but then according to the institution the rule change has not actually taken place. Taking this view to its logical conclusion -- an institution's rule--modifying counts--as rules constitute its own rule--change system and ascribe the rule changes that \textit{can} be made.

The philosopher Suber \cite{Suber1990} famously describes changing institutional rules as a game commonly known as Nomic. In Nomic, a move is proposing a rule change, debating it and then applying it according to the constitutive rules of the game (e.g. by majority vote or some other mechanism). Changing the system's rules can affect which rule changes are possible in the first place. For example, changing which agents are constituted as being able to change rules by participating in a vote. Suber observes many paradoxes can also arise, in the simplest case by modifying rules that make rule modifications possible. It follows that an agent wishing to change institutional rules to meet their goals, must understand how rule changes affect the built social reality and similarly which rule changes are possible.

The effects of rule change on the social reality and possible rule changes also takes a temporal dimension. At the very least, changing rules in the present affects the social reality and possible rule changes in the future. In some institutions it is only possible to change rules from the present onwards -- such as ascribed by the United States Constitution \cite[Art. 1 Sec. 9 Cl. 3]{USConstitution} ``No Bill of Attainder or ex post facto Law shall be passed''. In other institutions rule modifications in the past (retroactive modifications) are possible, such retroactive modifications have been made in the United Kingdom \cite[Sec. 58]{FinanceAct2008}. Given that changing rules affects which rule changes are possible, changing rules in the past, present and future can affect which rule modifications are possible or even happened up until the present.

In other cases, constitutive rules ascribe rule changes conditional on the effects of the rule changes. An example is found in the European Convention on Human Rights \cite[Art. 7]{ECHR1953}, which explicitly states that retroactive changes to rules are only possible if they do not criminalise formerly innocent people in the past. In order for an agent to determine what rule changes they can make or what rule changes actually took place according to the institution, the hypothetical effects of rule changes must be accounted for.

The main points are:

\begin{itemize}
\item Institutional rules are dynamic and subject to being changed over time. In order to determine what the social reality is at any given point we need to understand how rules have been modified and the effects of modification. Chapter~\ref{chapter_6} provides reasoning for rule modifications.
\item Constitutive rules state what rule changes \textit{can} be made, that is, the rule changes that are recognised by the institution as being valid. Chapter~\ref{chapter_6} adopts these kinds of constitutive rules, earlier chapters \ref{chapter_3} to \ref{chapter_5} adopt constitutive rules in general but not rule--modifying constitutive rules.
\item Rule changes are conditional on the built social reality, the rule changes that have and will take place and the potential effect of the rule changes on the social reality. Chapter~\ref{chapter_6} pays special attention to an interdependency between rule changes that are conditional on the built social reality and are able to change the social reality.
\end{itemize}


\section{Formal Foundations}
\label{secBackFound}

In the previous section we saw that institution designers are governed in what institutions they \textit{should} design and institutional rule enactments they \textit{can} make. The idea of this dissertation is to formalise practical reasoning in a \textit{single framework}: to determine which institutions should be designed, how institutional design non--compliance can be explained to support rectification and what rule modifications can be made. Several foundations for the framework are elicited based on the requirement for a practical framework, and our overview of the informal concepts we wish to formalise given in the previous section. These foundations are:

\begin{itemize}
\item A temporal setting -- the physical reality is not static, brute facts are subject to change, consequently the social reality is also subject to change. When a currency is decommissioned, the paper that counted--as money no longer counts--as money. If you are an academic, you are obliged to submit papers to conferences before their paper submission deadlines. Practical reasoning implies reasoning for realistic institutions and consequently reasoning for a temporal setting.
\item Constitutive rules -- institutions do not govern over a brute reality, but rather a social reality established by constitutive rules. Constitutive rules are a theory of institutional language and to paraphrase Searle \cite[p. 13]{Searle2005} if there is no language (e.g. represented as constitutive rules) then there can be no deontology.
\item Modal norms -- institutions govern and guide agents towards an ideal with norms. It is simpler to reason about and represent regulations governing regulations using a modal rather than evaluative norms, as this dissertation sets out to do.
\end{itemize}

On top of these foundations, this dissertation lays the following building blocks for formalising the governance of institutional design and enactment:

\begin{itemize}
\item Automatically detecting non--compliance in multi--level governance:
\begin{itemize}
\item Regulations governing regulations -- in multi--level governance institutions act to govern other institution designs. The instruments governing institution designs are regulations governing other regulations.
\item Abstraction based on constitutive rules -- in multi--level governance institutions are designed at different levels of abstraction. Lower--level institution's concrete regulations are interpreted for whether they comply with the abstract regulations at higher--levels of governance. The interpretation between concrete and abstract concepts is based on constitutive rules. In turn, the interpretation of concrete norms in terms of the abstract norms that govern them is derived from relationships between concrete and abstract concepts defined by constitutive rules.
\end{itemize}
\item Automatic resolution of non--compliance -- when an institution is designed that is non--compliant, punishments can be issued. In order to avoid punishment an institution designer should rectify the underlying causes of non--compliance rather than enacting a non--compliant institution. The assumption is that it is preferable to rectify non--compliance whilst remaining as closely to the institution's original design goals rather than not enact an institution in the first place.
\item Determining which rule--changes \textit{can} be made:
\begin{itemize}
\item Rule--modifying constitutive rules -- institution designers are constrained in which rule modifications they can make by constitutive rules. These constitutive rules ascribe rule modifications based on the context in which the modification takes place, including the hypothetical effects of rule modification.
\item Modifiable institutions in the past, present and future -- institution designers can make modifications to institutions. These modifications can be at any point in time made possible by the rule--modifying constitutive rules.
\end{itemize}
\end{itemize}

In summary, this dissertation builds on foundational concepts comprising institutional reasoning in a temporal setting, constitutive rules and modal norms. Building on these concepts, this dissertation contributes novel formalisations of institutional notions for institutions governed in multi--level governance, explanations for non--compliant institution designs and institutions comprising rule--modifying constitutive rules, where the institution can be modified over time. Collectively, these novel formal building blocks allow us to reason about institutional design and enactment governance.

\section{Formal Approaches}
\label{secBackFormal}

In this section existing formal approaches for normative and institutional reasoning are compared. We focus on approaches that provide useful historical and conceptual context or provide foundations on which to base our framework. In formal philosophy, deontic logic is the field dedicated to the study of `ought' and other normative statements relevant to institutional reasoning. This section overviews a few systems of deontic logic and related developments.

\subsection{Standard Deontic Logic}

Von Wright's Standard Deontic Logic (SDL) \footnote{n.b. Standard is a moniker and does not denote that SDL is by any means \textit{the} standard \cite[p. 39]{Gabbay2013}, hence we do not restrict our search for an appropriate formalism to just SDL.} \cite{VonWright1951} is the first deontic logic widely considered to be a viable formalisation of ought (see \cite[p. 5]{Gabbay2013} for a historical overview dating back to medieval times). SDL is situated in a propositional setting. It introduces modal operators over propositional formulae $p$ to express the deontic modalities of obligation ($\textbf{O}p$), prohibition/forbidden ($\textbf{F}p$) and permission ($\textbf{P}p$).

Von Wright introduced an axiomatisation for the modal deontic operators which, among other axioms, provided equivalences between the deontic modalities that have been used in many other systems of deontic logic. These equivalences are, what is forbidden is obliged to the contrary ($\textbf{F}p = \textbf{O}\neg p$) and what is permitted is not obliged to the contrary/not forbidden ($\textbf{P}p = \neg \textbf{O} \neg p$) (i.e. permission is the dual of obligation). This scheme gives us the well known deontic square of opposition depicted in Figure~\ref{figDeonticOpposition}, succinctly summarising the relationships between the different modalities.
\begin{figure}
\centering
\begin{tikzpicture}
\draw[ultra thick, >=latex,->] (-4,3) -- (-4,1.4);
\draw[ultra thick] (-4,1.4) -- (-4,0);
\draw[ultra thick, >=latex,->] (-1,3) -- (-1,1.4);
\draw[ultra thick] (-1,1.4) -- (-1,0);
\draw[ultra thick, color=black!30!green, ->>] (-2.4,0) -- (-1,0);
\draw[ultra thick, color=black!30!green, <<-] (-4,0) -- (-2.4,0);
\draw[ultra thick, color=blue, dashed] (-4,3) -- (-1,3);

\draw[ultra thick, color=black!10!red, dotted] (-4,0) -- (-1,3);
\draw[ultra thick, color=black!10!red, dotted] (-4,3) -- (-1,0);
\node at (-4.2,3.4) {Obligatory $\textbf{O}p$};
\node at (-0.8,3.4) {Prohibited $\textbf{F}p$};
\node at (-4.2,-0.4) {Permissible $\textbf{P}p$};
\node at (-0.8,-0.4) {Omissible $\textbf{P}\neg p$};
\end{tikzpicture}
\caption[The deontic square of opposition]{The Deontic Square of Opposition \cite{sep-logic-deontic}. Contraries are denoted with \tikz[baseline=-0.5ex, very thick, color=blue, dashed]{ \draw [-] (0,0) -- (5ex,0); }, contradictories are denoted with \tikz[baseline=-0.5ex, very thick, color=black!10!red, dotted]{ \draw [-] (0,0) -- (5ex,0); }, implications are denoted with \tikz[baseline=-0.5ex, very thick]{ \draw [-, >=latex,->] (0,0) -- (3.5ex,0); \draw [-] (3.5ex,0) -- (5ex,0); } and sub-contraries are denoted with \tikz[baseline=-0.5ex, >=latex,->, very thick]{ \draw [very thick, color=black!30!green, ->>] (0,0) -- (5ex,0); }. For further geometric analyses of deontic and related logics see \cite{Moretti2009a}.}
\label{figDeonticOpposition}
\end{figure}

SDL is not without its problems, rather it is susceptible to a number of so--called paradoxes. In SDL conditional norms, such as if there is a fence ($f$) it should be a white fence ($w$) are represented with formulae mixing propositional sentences and deontic operators ($f \rightarrow \textbf{O}w$). A problem arises if we extend the previous example with a prohibition on there not being a fence ($\textbf{F}f$), a fact that there is a fence ($f$) and the implication that a painted fence implies there is a fence ($p \rightarrow f$). The problem is that it leads to a contradiction in SDL -- there ought to be a fence and not a fence ($\textbf{O}f \wedge \neg f$) \cite{Prakken1996}. However, in principle, there should not be a contradiction. In general, a problem arises where there is a primary norm (forbidden for there to be a fence) that is supported with a secondary norm that when applied represents a level below what is ideal (the fence should be painted white). Potentially, such a secondary norm can also be used to represent a punishment for sub--ideal behaviour or circumstances, for example, you should not speed but if you do then you should pay a fine. Such statements are known as contrary--to--duty norms (CTDs). They are viewed as being an important aspect of the law where norms are commonly used to define punishments for non--compliance or state what the sub--ideal circumstances are \cite{Jones1992,Jones1993}. Hence, it is important contrary--to--duty norms are handled adequately.

Several proposals address the problems caused by CTDs in SDL. A common view is to distinguish between \textit{prima facie} oughts, the oughts that on the face of it hold, and \textit{all things considered} oughts \cite[p. 256]{Gabbay2013}. For example, \textit{prima facie} it is forbidden for there to be a fence and obliged the fence is painted white. All things considered, it is only obliged the fence is painted white. Proposals addressing CTDs thus derive `all things considered' oughts from prima facie oughts. For example, by using defeasible reasoning \cite{ryu1994} to exclude \textit{prima facie} oughts that are violated, in favour of secondary oughts. Other significant approaches are preference--based formalisations \cite{Torre1997,Torre1999} which interpret ought as ideal and CTDs as representing sub--ideal circumstances, then separating contradictory oughts into separate worlds of ideality. For example, in the ideal world it is forbidden for there to be a fence, in a sub--ideal world the fence ought to be painted white and both do not hold in the same ideality (hence no contradiction). In all of these approaches the semantics of classical implication are replaced with non--classical semantics.

Relevant to our dissertation is the fact that, as expected, such a modal representation for norms supports nesting. Nested deontic modalities such as $\textbf{OO}p$ are grammatical in SDL. Unfortunately, SDL is a strictly non--temporal logic and hence unsuitable on its own for the purposes of this dissertation. Furthermore, SDL lacks constitutive rules; with material implication as the only conditional of which context is not a part.

\subsection{Anderson's Reduction}

Anderson proposed a logic that replaces SDL's deontic operators in favour of evaluative norms \cite{Anderson1957,Anderson1958}. In Anderson's proposal SDL is reduced to alethic modal\footnote{If this is Greek to you, alethic modalities are modalities denoting truth, in comparison to deontic modalities, which denote normativity.} logic by replacing $\textbf{O}p$ with a formula stating that it is necessary ($\square$) that going against the norm ($\neg p$) materially implies a violation ($V$): $\square(\neg p \rightarrow V)$. In Anderson's reduction, norms are rules that ascribe violations.

Grossi \cite{Grossi2008a} developed Anderson's idea further by proposing a logic for evaluative norms expressed as \textit{constitutive} rules. In Grossi's proposal a logic of context and ascriptions is proposed. In his proposal implications operating on a context $i$ are introduced, denoted as $\Rightarrow^{\textit{cl}}_{i}$. Evaluative norms stating that $p$ is a duty are represented as a formula ascribing a violation $V$ in a context $i$: $\neg p \Rightarrow^{\textit{cl}}_{i} V$. In Grossi's proposal, Anderson's reduction is realised in a logic of constitutive rules. Hence, Grossi realises the context--sensitive ascription of abstract institutional concepts, including violation, in a deontic logic.

Aldewereld et al. \cite{Aldewereld2010a} build on these proposals with a formalism for reasoning about evaluative norms at different abstraction levels. They combine constitutive rules to ascribe abstract institutional facts from more concrete brute or institutional facts. Constitutive rules define norms by ascribing evaluative statements (compliant and violation in their case). To give their example, the rule ``transferring money with a credit card counts--as payment'' ascribes the abstract institutional fact of payment from the more concrete fact of using a credit card. A norm is then ``paying counts--as fulfilment'. The norm is abstract in that it states someone should make a payment but it does not concretely define, on its own, what payment exactly is. By providing a semantics for counts--as statements that includes a form of transitivity, norms can also be concretised. For example, through transitivity, using a credit--card counts--as fulfilment. In their proposal, constitutive rules are contextual and hence transitivity only holds between constitutive rules if their contexts are compatible. To summarise, Aldewereld et al. take the principles behind Grossi's reduction to constitutive rules in order to reason about concretisation of abstract norms.

Developments in the reductionist approach to norms are relevant to this dissertation. First, they offer a way to reason about abstract norms, in this case by concretising abstract norms. Hence, it seems that the same approach can be taken to multi--level governance. Perhaps, abstract regulations governing other regulations at higher--levels of governance can be reified to determine their concrete meaning and therefore whether concrete norms violate the abstract norms. Conversely, perhaps the same reasoning can be reversed to take concrete norms and abstract them to determine if they are compliant with more abstract norms. However, a major stumbling block is that by reducing norms to evaluative constitutive rules it is no longer straightforward to represent and reason about regulation governing regulations, as we argued previously. Consequently, this dissertation does not take a reductionist approach to norms.

\subsection{Temporal Deontic Logics}

The previously described approaches are situated in a static setting with no consideration for time. Hence, they deal with non--temporal norms. For example the norm `you should not murder' has no temporal element. However, in the real world duties are often temporal and contain deadlines. For example `you should submit your paper before the submission deadline'. Thus, there are deontic logics that deal explicitly with time, by which we mean a temporal ordering of states containing formulae, including temporal obligations.

Several temporal logics independent from deontic logic already exist. The prominent ones are Linear Temporal Logic (LTL) \cite{Pnueli1977}, Computational Tree Logic (CTL) \cite{Emerson1985} and CTL* \cite{Emerson1990}, which combines both. There are also logics of actions as transitions between states. Dynamic logic \cite{Harel1984} being a prominent action logic, which combines modal logic operators of necessity and possibility with actions to express and reason about statements such as `it is possible that action $a$ will cause $p$ to hold'. A common approach to temporalising deontic logic is to combine a deontic logic and one of the aforementioned temporal or action logics.

To name a few. Broersen et al. \cite{Broersen2004} combine SDL and CTL. They introduce dyadic deontic modalities ($O(\rho \leq \delta)$) representing a propositional formula $\rho$ should hold before or at the same time as a formula $\delta$ representing the deadline. In Broersen et al.'s formalisation if $\delta$ occurs before $\rho$ then a proposition denoting violation holds. Conversely, $\rho$ holding before or at the same time as $\delta$ causes a proposition denoting norm fulfilment to hold. An obligation holds from one state to the next until it is fulfilled or violated, that is, a norm persists by default.

F. Dignum et al. \cite{Dignum1996} combine SDL and dynamic logic. Obligations can take the form $O(\alpha < \rho < \delta)$ representing an obligation conditional on $\alpha$ that requires action $\rho$ to be performed before $\delta$. States also have time indices. The special index $\textit{now}$ denotes the time of the current state. In F. Dignum et al's proposal instantaneous norms that must be fulfilled immediately are expressed as $O(\alpha < \rho < \textit{now} + 1)$ representing that $\alpha$ causes an obligation for $\rho$ to be performed before the next state.

F. Dignum and Kuiper \cite{Dignum1998} combine SDL with a logic of `dense' time in which actions are not instantaneous. Rather, actions are performed over a time period. An obligation $O(\alpha)$ represents that an action $\alpha$ should be continuously performed until it is done. It is interesting to note that violations persist from one state to the next until the norm that has been violated is `repaired' (e.g. by performing some punishing action, such as paying a fine).

To summarise, temporal deontic logics allow expressing that there should be an ordering in which propositions hold in states or actions are performed. A temporal obligation holding in a state typically states that at the present time an $\alpha$ should be done/hold before a $\delta$. Obligations persist from one state to another if not discharged. Hence, if $\alpha$ and $\delta$ do not occur, then $\alpha$ should still be performed/hold before $\delta$. Instantaneous norms then just state that an $\alpha$ should be done/hold before the next state. In this dissertation we adopt similar notions in Chapter~\ref{chapter_3}, Chapter~\ref{chapter_4} and Chapter~\ref{chapter_5} -- a deontic statement persists until discharged, and either represents one thing should be done before another or that something must be done instantaneously.

\subsection{Seinsollen and Tunsollen: Ought--to--be and Ought--to--do}

There is a difference between ought--to--be statements about what should hold (seinsollen) and ought--to--do statements about what should be done (tunsollen) \cite{Castaneda1970}. In SDL \cite{VonWright1951} the two are not distinguished; SDL abstracts away from whether a proposition denotes an action or a state--of--affairs. Subsequent developments have focussed on one or the other, or both.

\begin{itemize}
\item \textbf{Ought--to--be} -- logics of this type are characterised by obligations that hold in states and are about propositions that hold in states. Examples include \cite{Broersen2004,Dastani2008,Governatori2007}. It is interesting to note that ought--to--be can represent two types of obligation explored in \cite{Governatori2007}. The first is an achievement obligation: a state of affairs should be achieved before a deadline. The second is a maintenance obligation: a state of affairs should be maintained up until a deadline.
\item \textbf{Ought--to--do} -- logics of this type are characterised by obligations that hold in states and are about events or actions that occur when transitioning between states. Examples include \cite{Broersen2004,Dignum1996,Dignum2002,Horty1995}. It is interesting to note that Broersen et al. \cite{Broersen2004}, Horty and Belnap \cite{Horty1995}, and V. Dignum et al. \cite{Dignum2002} use a stit (see--to--it--that) operator (or a version thereof) to represent the action of an agent bringing about a particular state of affairs. For example $\textit{stit}\phi$ is an action to bring about $\phi$ in a state. Consequently, ought--to--do statements are framed as ought--to--bring--about a particular state, abstracting away from labelled actions whilst still remaining ought--to--do statements.
\end{itemize}

In this dissertation, we look at obligations about agents' actions (or events if we view actions as requiring intention). This is because they seem to be more concrete and are frequently found in the law -- you should not \textit{murder}, you should not \textit{steal}, if you do either you should \textit{go} to prison. We also look at obligations about obligations to reason about regulations governing other regulations. Yet, obligations are normative propositions that  \textit{hold} in states and so obligations about obligations are seinsollen ought statements about a state of affairs. As Von Wright points out, if obligations are about actions then nested modalities are nonsensical \cite[p. 91]{VonWright1968}. Where does this leave us, do we look at both ought--to--do (obligations about actions) and ought--to--be (obligations about obligations which hold in states)?

One approach would be only to look at ought--to--do. Here, the idea is to follow Wansing's suggestion \cite{Wansing1998} that nested obligations are obligations to see--to--it--that obligations hold. The idea being that a nested obligation is an obligation to perform a locutionary act to establish an obligation, for example, if we are an authority with legal power, by saying ``I command you to do X'' we can see--to--it--that there is an obligation to do X (see Searle's account of how to go from is to ought, or obligations from locutionary promises \cite{Searle1964}). The problem is, this means imposing obligations requires agency. However, in this dissertation we set out to look at institutions with regulations that govern other regulations -- an institution is not a type of agent, it belongs to a class with instantiations that include the written law, promises, contracts etcetera.

Hence, in this dissertation we combine ought--to--do for obligations that govern agents' actions and ought--to--be for norms governing other norms. The combination of both was also looked at by D`Altan et al. \cite{DAltan1996}. Chapter~\ref{chapter_3}, Chapter~\ref{chapter_4} and \ref{chapter_5} combine both types of ought. Other aspects, of ought--to--be statements such as maintenance and achievement obligations are not considered in this dissertation.

\section{Practical Formal Governance}

In this section a number of frameworks for practical formal governance are proposed. There are many links between deontic logics and the following research described and hence the two cannot be easily disentangled. Rather, the following frameworks are characterised firstly by drawing from a wider informal sphere such as legal and organisational theory, whilst deontic logics come from a philosophical tradition. Secondly, by being pragmatic, such as by having a low computational complexity (e.g. by talking about obligations over single propositions rather than arbitrary propositional formulae) or providing clear links to implementation. It is important to note that this dissertation falls firmly in practical formal governance, since it provides a formal account that is characterised in such a way that the decision procedure is obvious and/or is coupled with a computational framework (i.e. an implementation).

\subsection{Organisational Frameworks}

Whilst this dissertation focuses on institutional reasoning a similar coordination mechanism, multi--agent organisations, is also widely studied for formalisation. Like institutions, organisations comprise a normative dimension. In contrast with institutions, organisations also comprise constructs such as roles agents play and interaction patterns to achieve various organisational objectives. If an institution is legislation, contracts and promises, then an organisation is a university or a corporation within which institutions can play a part in the form of contracts and other bodies of rules and regulations. Consequently, formal reasoning between the two areas has much overlap (e.g. norms), institutions tending to focus on legal concepts (e.g. legal empowerment to affect an institution) whilst organisations focus on cooperation concepts and processes to achieve goals (e.g. interaction patterns).

\textbf{HarmonIA} \cite{Vazquez-Salceda2003} is an organisational framework that focuses on conceptualising `electronic organisations' and methodologies for their design. In HarmonIA, organisations comprise different levels of abstraction. These are an abstract level, concrete level, rule level and a procedure level. At the abstract level, the organisation defines abstract norms such as statutes. At the concrete level the organisation refines the abstract level's norms to concrete policies. At the rule level further refinements are made to the concrete norms as rules for agents to follow in order to comply with the concrete norms. The procedure level comprises the actual computational implementation of the rules. In HarmonIA norms are modal and their refinement/concretisation is based on the counts--as relation. However, a semantics for norm refinement is not investigated. Instead the \textit{approach} is given \cite[p.94]{Vazquez-Salceda2003} and tools to automate refinement/concretisation are left for future work \cite[p.168]{Vazquez-Salceda2003}. HarmonIA's focus is on the conceptualisation and design methodology for organisations in MAS at different abstraction levels, but not the formal reasoning.

\textbf{MOISE+} \cite{Hubner2007} provides a conceptualisation for designing organisations comprising roles, dependencies between roles and norms. In MOISE+ norms have a modal representation. In contrast to much of the work in formal philosophy and elsewhere, MOISE+ does not provide a formal semantics for norms or other organisational concepts. Rather, it focuses on programming organisations at an agent level and an organisation--control level. At the agent level, the J--MOISE+ (sub--)framework provides a way to program agents to enact roles in a MOISE+ organisation. At the organisation level, the S--MOISE+ \cite{Hubner2005} (sub--)framework provides an interface for agents to join a MOISE+ organisation. Through the S--MOISE+ interface organisational constraints are regimented on agents as \textit{hard} constraints such that agents are forced to comply. MOISE+ is an agent--organisation framework focussed on programming agents and organisations, there are many aspects relevant to institutions that are not considered (e.g. constitutive rules, abstraction and rule change).

\textbf{OperA} \cite{Dignum2003} is a framework providing an organisation design methodology and formal organisational reasoning. Conceptually, an organisation in OperA is analogous to a human organisation. An organisation comprises a social structure defining organisation objectives, agent roles and relations between roles; an interaction structure describing scenes for agents to play roles in and landmark objectives for the organisation to achieve; a normative structure describing the norms agents must abide by when fulfilling roles and interacting in scenes; and a communicative structure describing the communicative acts agents can make. An organisational logic formalises OperA organisations. The logic combines deontic logic and temporal logic to define normative reasoning for the norms agents adopt when enacting roles and reasoning about the responsibilities and capabilities endowed on agents for enacting those roles. The logic also determines when organisational landmarks are achieved. The formal theory provided by OperA is implemented with the Operetta tool \cite{Aldewereld2010} for checking organisational properties. In OperA organisations are the formal counterpart to human organisations and a formal semantics provides organisational reasoning, but OperA lacks many aspects found in the legal/institutional sphere (again, constitutive rules, abstraction and rule change).

\textbf{OperA+} \cite{Jiang2015} by Jiang builds on the OperA framework with additional organisation design concepts and methodology, and novel computational mechanisms for organisational reasoning. Enhancing OperA's organisation design methodology, OperA+ proposes multiple levels of abstraction. The most abstract level comprises organisational objectives. From here, the design process enters a contextual level where the organisation design is concretised by decomposing the organisation objectives into sub--domains or situations. Jiang uses a top--level objective of train maintenance as an example where in this case the sub--domains are planned and unplanned maintenance. Such sub--domains are then concretised further at a solution level, by instantiating the sub--domains as social structures and normative structures. Like in OperA, the social structures comprise roles agents play and the normative structure comprises norms. Unlike OperA, OperA+ also includes constitutive rules for establishing institutional actions from brute facts. Opera+ also introduces an operational level, comprising groups of agents enacting roles represented as group preferences for particular actions. OperA+ organisations are operationalised using Coloured Petri Nets (CPNs), a graph--based formalism for describing distributed systems and computing their states over time. Norms are represented as CPN graphs, akin to an evaluative norm representation. Since norms have an evaluative representation, it is unclear how to extend OperA+'s representation and reasoning to norms governing other norms. Other aspects missing from OperA+ but focussed on in the dissertation you are reading now include a semantic definition of links between concrete and abstract norms and governing rule change.

\subsection{Institutional Frameworks}

Institutional frameworks in MAS can be understood as analogous to their informal counterparts (e.g. legislation). Typical elements are constitutive rules, norms, and the ability for the institution to evolve from one institutional state to the next according to the change of brute facts and the institution's constitutive rules. Research on institutional reasoning falls largely into two spheres, legal reasoning (e.g. for legislation) and open multi--agent systems (e.g. artificial markets). We describe three frameworks in what is by no means an exhaustive overview, instead we focus on the frameworks that are closest to what this dissertation is trying to achieve.

\textbf{OCeAN} \cite{Fornara2007b,Fornara2008,Fornara2009} (\textbf{Ontology CommitmEnts Authorizations Norms}) is a high--level institution specification language and an operationalisation in the discrete event calculus. In OCeAN an ontology specifies institutional actions, events and roles agents play (hence there is an organisational flavour to OCeAN). OCeAN adopts constitutive rules (``A counts--as B in a context C'') in order to define an agent communication language. Here, the communicative act `A' counts--as an institutional action in a context where the communicative agent is playing a role that authorises that agent to realise the institutional action. For example, an auctioneer telling everyone the auction is open counts--as opening the auction, where playing the role of auctioneer represents the agent's authorisation. OCeAN also contains a normative component, adopting the usual deontic modalities in a modal representation. OCeAN's focus is on ontologies, social commitments and organisational concepts such as roles.

\textbf{TMDL} \cite{Governatori2005a,Governatori2007} (Temporal Modal Defeasible Logic) is a non--monotonic logic for reasoning about institutions over time. The institutional concepts captured are the deontic modalities, constitutive rules, normative power (the ability for an agent to establish a norm through a communicative act), and positive and negative rights. The logic is temporal in the sense that propositions (including modal statements) hold at points in time when they are initiated by an action according to a rule, and persist in holding (inertia) until they are terminated. The logic's modus operandi is \textit{defeasibility}, the idea that a proposition is defeasibly proven but a proof to the contrary \textit{defeats} the proposition and makes it disproven. In TMDL there are three types of rule in horn--clause form. Firstly, strict rules $\phi_0, ..., \phi_n \rightarrow \psi$ where the consequent is strictly provable when the antecedent is provable and cannot be disproven if the antecedent is proven. Secondly, defeasible rules $\phi_0, ..., \phi_n \Rightarrow \psi$ where proving the antecedent means the consequent is defeasibly proven, but proving contrary propositions from the consequents of other rules defeats $\psi$ or the rule's defeasible premises and makes $\psi$ disproven. Thirdly, defeater rules $\phi_0, ..., \phi_n \rightsquigarrow \psi$, which do not make their conclusion proven, but rather defeat contrary premises and conclusions of other rules that are defeasibly proven. To determine which rules can defeat other rules, a superiority relation between rules is specified.

Defeasible logic, such as TMDL, defines a proof procedure that can be implemented in Prolog \cite{Antoniou2001}. The idea is that a derivation asserts the conclusion of a rule, then all the ways to attack the derivation and its conclusion are found, all the attacks of the attacks are found, and so on. This process continues until no more attacks can be applied and it is determined whether there remain any undefended attacks of the original assertion, in which case it is not proven, and otherwise it is proven. TMDL, through its proof procedure, makes it possible to reason about exceptions as found in the law and also exceptions to exceptions etcetera.

\textbf{InstAL} \cite{Cliffe2006,Cliffe2007} (INSTitution Action Language) is a framework for temporal institutional reasoning. An institution specification in InstAL comprises events that can occur and fluents that can hold in the institution, as well as constitutive rules. The events can be observable corresponding to a brute fact, or institutional such as `payment' or denoting a norm is violated. Fluents represent institutional facts about a domain, the deontic statements of obligation and permission in a modal--form, and institutional empowerment denoting an event is \textit{empowered} to affect the institution. In InstAL, anything not permitted is prohibited and hence a prohibitive society is reasoned about. The constitutive rules ascribe institutional events from observable events (corresponding to the notion of a brute fact changing) and other institutional events. The constitutive rules also ascribe the effects of institutional events on the institution's state, causing institutional fluents to be initiated or terminated from one state to the next. Regulations are represented as constitutive rules where the consequents are the initiation or termination of obligation, permission and empowerment fluents.

InstAL's semantics are defined with a formal framework based on set theory and mathematical functions. Complementing the formal framework is a computational framework that executes institution specifications in response to a trace of observable events. The computational framework uses Answer--Set Programming (ASP), a non--monotonic logic programming language \cite{Gelfond2008,Gebser2011} based on the stable model semantics \cite{Gelfond1988}. The output of the framework is a state--transition system. Each state contains fluents representing institutional facts and normative positions (obligations and permissions). Transitions between states are the events produced by observable events and the transitive closure of constitutive institutional event generation rules. Fluents persist by default from one state to the next unless terminated, capturing the common--sense law of inertia using Event--Calculus--like constructs. InstAL is complemented with a framework for finding rule changes to meet certain properties, in this case resolving inconsistencies between norms, based on Inductive Logic Programming in ASP \cite{Li2014,Li2015}.

\section{Knowledge Gaps and Approach}
\label{secBackApproach}

\def\tick{\tikz\fill[scale=0.35](0,.35) -- (.25,0) -- (1,.7) -- (.25,.15) -- cycle;}
\begin{table}[t!]
\centering
{\renewcommand{\arraystretch}{1.25}
\begin{tabular}{p{50mm} l l l l l l l l l l}
    & \rotatebox{90}{Modal Norms} & \rotatebox{90}{Temporal Setting} & \rotatebox{90}{Constitutive Rules} & \rotatebox{90}{\textbf{Regulations Governing Regulations}} & \rotatebox{90}{\parbox{50mm}{\textbf{Abstract Norms}}} & 
\rotatebox{90}{\textbf{Automatic Rule Change for Compliance}} & \rotatebox{90}{\textbf{Rule-modifying Constitutive Rules}} & \rotatebox{90}{\textbf{Temporal Rule Modifications}} &
\rotatebox{90}{Implementation}  \\
Standard Deontic Logic \cite{VonWright1951} & \tick & & & \tick & & & & & \\
Anderson's Reduction \cite{Anderson1957,Anderson1958} and Subsequent Developments \cite{Aldewereld2010a,Grossi2008a} & & & \tick & & \tick & & & & \\
Temporal Deontic Logics \cite{Broersen2004,Dignum1996,Dignum1998} & \tick & \tick & & & & & & & \\
HarmonIA \cite{Vazquez-Salceda2003} & {\huge \textasciitilde} & {\huge \textasciitilde} & {\huge \textasciitilde} & & {\huge \textasciitilde} & & & & \\
MOISE+ \cite{Hubner2007} & \tick & \tick & & & & & & & \tick \\
OperA \cite{Dignum2003} & \tick & \tick & & & & & & & \tick \\
OperA+ \cite{Jiang2015} & & \tick & \tick & & & & & & \tick \\
OCeAN \cite{Fornara2007b,Fornara2008,Fornara2009} & \tick & \tick & \tick & & & & & & \tick \\
TMDL \cite{Governatori2005a,Governatori2007} & \tick & \tick &  \tick & & & & & & \tick \\
InstAL \cite{Cliffe2006,Cliffe2007,Li2014,Li2015} & \tick & \tick &  \tick & & & {\huge \textasciitilde} & & & \tick \\
\end{tabular}} \quad
\caption[Existing knowledge comparison]{Comparison between the state--of--the--art knowledge provided by different frameworks. A tick denotes the framework provides the necessary representation and reasoning constructs. A {\huge \textasciitilde} denotes the framework only provides a conceptualisation sans semantics in the case of HarmonIA or technologies that can feasibly be adjusted for our purposes in the case of InstAL. Knowledge in a regular font denotes foundational concepts that need formalising, knowledge in \textbf{bold} represents the building block concepts that need formalising to reason about institutional design and enactment governance.}
\label{tabInstALComp}
\end{table}

In this section we compare the frameworks described previously, depicted in table~\ref{tabInstALComp}, in terms of their support for PARAGon's objective of institutional design and enactment governance reasoning. The comparison is split between foundational representation and reasoning requirements for our formalisation aims, and the necessary conceptual building blocks (shown in \textbf{bold}) we require to formalise institutional design and enactment governance. We also denote those frameworks with a corresponding implementation, although the degree to which there exists an implementation varies from one framework to another (see the previous discussion).

The \textit{foundational} representation and reasoning constructs for this dissertation are modal norms, since they can be extended to nested modal norms and therefore regulations governing regulations; a temporal setting for real--world relevance and the reasoning about rule--change over time; and constitutive rules for reasoning about institutions operating at different levels of abstraction. The conceptual \textit{building blocks} we require are: rule--modifying constitutive rules for ascribing which rules an agent can modify and when; temporal rule modifications; regulations governing regulations for reasoning about multi--level governance; norm abstraction or concretisation based on constitutive rules for interpreting institutions in multi--level governance; and automatic rule change for compliance in order to help institution designers in multi--level governance enact institutions whilst avoiding punishment for non--compliance. As we can see, the foundational components are adequately supported by some frameworks, however the conceptual building blocks are not and hence it is those building blocks which the PARAGon framework addresses.

Organisation frameworks support some necessary foundational institutional reasoning constructs, such as norms and potentially constitutive rules. However, they lack the semantics for the governance of institutional design and change. In the case of the HarmonIA framework, whilst significant attention is paid to norm concretisation it is done by way of example and not with a general semantics. The same is true for the other aspects of HarmonIA. What HarmonIA does offer, is an indication of on what basis norm abstraction should be reasoned about. Namely, constitutive rules. In fact, that is the same intuition this dissertation takes, albeit we provide a formal semantics for reasoning about abstracting rather than concretising norms using constitutive rules. MOISE+, OperA and OperA+ all offer some foundational aspects that could potentially be built on by this dissertation. However, the institutional reasoning frameworks, focussing on institutions as we do, offer further institutional reasoning foundations suitable for our aims.

As for the institutional reasoning frameworks -- OCeAN, TMDL and InstAL all share institutional concepts such as norms, constitutive rules and institutional empowerment. Some differences are that OCeAN also comprises organisation concepts, TMDL incorporates defeasible reasoning and InstAL has been used as a basis for Inductive Logic Programming based norm revision. In all of these frameworks norms are specified at a single level of abstraction, constitutive rules establish an institutional reality but not rule changes, and detecting and rectifying compliant institution \textit{designs} is not captured. Hence, all frameworks provide necessary formal foundations, but do not close the knowledge gap this dissertation addresses.

The question arises as to the suitability of these frameworks for our purposes. In this dissertation InstAL is used as a basis for the institutional reasoning we contribute. The main reason is that in InstAL, due to having an ASP computational framework, it is already possible to apply existing ASP--based Inductive Logic Programming techniques in order to resolve problems with institutions. In the case of \cite{Li2015} the problems resolved are normative conflicts. In this dissertation we seek to resolve non--compliance in multi--level governance and hence Inductive Logic Programming techniques are used and developed towards our aims.

The following summarises how the PARAGon framework fills the gap in the knowledge. Determining compliance according to institutional design governance in multi--level governance requires: regulations governing regulations, and norm abstraction or concretisation based on constitutive rules, and is addressed with a formal framework in Chapter~\ref{chapter_3} and a computational framework in Chapter~\ref{chapter_4}. Automatically resolving non--compliant institution designs to support designers in avoiding punishment requires automatic rule change for compliance and is addressed with a computational framework in Chapter~\ref{chapter_5}. Determining when rule changes are legally made requires rule--modifying constitutive rules and temporal rule modifications, and is addressed with a formal framework in Chapter~\ref{chapter_6}.

\section{Discussion}

In this chapter we began with an overview of the informal concepts behind multi--agent systems, institutions, governance and governing governance. Two aspects stand out. First, there are many parallels between the need to govern agents (due to their autonomy and liability to deviate from the ideal), and governing institution designers (for the same reason, since institution designers are also agents). Second, the informal work looked at in the area of institutions is, in fact, fairly formal (e.g. in the case of Searle's counts--as rules). On the other hand, work on governing governance appears distinctly informal, where we largely drew on work in political science and examples from real--world legislation.

These aspects led us to identifying the main notions that the PARAGon framework needs to formalise. First, we elicited the foundational notions to formalise (institutions, norms). Then, we identified the building blocks required to formalise governing governance. For example, we elicited the need to rely on modal norms over evaluative norms to reason about regulations governing regulations. Likewise, we argued that it is clear constitutive rules play an important role in governing of governance both for providing links between concrete and abstract concepts and governing institutional enactment of new rules or changes to existing rules. What was unclear at this point was to what extent existing approaches provided the foundations or the building blocks required.

Hence, we provided a comparison between existing formal approaches. We paid particular attention to where the gaps in the existing knowledge lie that need to be filled. Based on this analysis we selected the InstAL framework from which we can use various foundational formal components on which to build the PARAGon framework this dissertation contributes.
\chapter{Formalising Compliance in Multi--level Governance}
\label{chapter_3}

\epigraph[0pt]{For a large class of cases--though not for all--in which we employ the word ``meaning'' it can be defined thus: the meaning of a word is its use in the language\cite[Sect. 43]{Wittgenstein1958a}}{Ludwig Wittgenstein}

\blfootnote{\color{tck-grey}This chapter is based on the following papers:\\
King, T. C., Li, T., De Vos, M., Dignum, V., Jonker, C. M., Padget, J., \& van Riemsdijk, M. B. (2016, July). Automated Multi--level Governance Compliance Checking. \textit{Journal of Autonomous Agents and Multiagent Systems (JAAMAS)}. International Foundation for Autonomous Agents and Multiagent Systems. (\textbf{In Submission})\\
Which extends the following paper:\\
\textbf{King, T. C.}, Li, T., De Vos, M., Dignum, V., Jonker, C. M., Padget, J., \& Riemsdijk, M. B. Van. (2015). A Framework for Institutions Governing Institutions. In Proceedings of the 2015 International Conference on Autonomous Agents and Multiagent Systems (AAMAS 2015) (pp. 473–481). Istanbul, Turkey: International Foundation for Autonomous Agents and Multiagent Systems. \cite{King2015a}}

\newpage

In this chapter we look at \textit{soft constraints} placed on institution designers, which prescribe the institution designs they \textit{should} enact. The contributions of this chapter are:

\begin{itemize}
\item A formal representation language for institutions governing other institution designs in multi--level governance.
\item A formal semantics for determining when institutions are compliant in multi--level governance, accounting for concrete regulations at lower--levels of governance, governed by increasingly abstract regulations at higher--levels of governance.
\end{itemize}

Institutions (e.g. legislation) govern societies towards ideal and coordinated behaviour with rules and regulations coupled with the means to detect compliance and issue rewards and punishments. Increasingly, institutional reasoning is formalised and computerised with automated normative and institutional reasoning frameworks (see \cite{Dagstuhl4} for a review). Such frameworks can support governing bodies in punishing agents and support agents in understanding their legal duties.

However, typical institutions are not written in a vacuum. Rather, institutions are constrained and regulated by higher--level institutions. This is what is called \textit{multi--level governance} \cite{Liesbet2003}. In multi--level governance, institution designers design institutions comprising rules and regulations, but whose design is also subject to regulation. In 2006 the European Union issued the Data Retention Directive \cite{EUDRD} for harmonising member states' data retention regulations, in 2009 the UK implemented the directive with the Data Retention Regulations \cite{UK2009} in order to avoid fines. Yet, in 2014 the European Court of Justice ruled \cite{ECJDRD} that the directive was non--compliant with the Charter of Fundamental Rights \cite{EU2000}, and annulled the directive. Legislating in multi--level governance exposes institution designers to risks of punishment and wasted legislating time and burdens a judiciary with checking compliance.

So far, institutional reasoning frameworks have focussed on single--levelled societal governance. Typically, automated institutional reasoning deals with regulations operating at the level of institutions governing agents and/or corporations. For example, the UK's Data Retention Regulations \cite{UK2009} obliges communications providers to store communications metadata. There lacks reasoning for cases where regulations themselves are regulated by higher--level institutions in multi--level governance. For example, how EU directives govern national legislation but where EU directives are in turn governed by human rights charters. In this chapter we look at how institutions themselves are regulated by higher--level institutions.

In particular, we look at increasingly abstract regulations at higher--levels of governance that govern more concrete regulations at lower--levels of governance. Such abstraction sets multi--level governance apart from single--levelled governance of societies. In multi--level governance at the highest--level, such as human rights charters, regulations are intentionally abstract and open to interpretation. Such abstract regulations provide many ways in which to (non--)comply. At a lower--level, such as EU directives, regulations are more concrete and less open to interpretation. At the lowest level, such as national or sub--national legislation, regulations are concrete and the least ambiguous. Despite the differences in abstraction between levels, each level's institution design must somehow be demonstrated to be compliant with relatively more abstract regulations at higher--levels.

To give an example, the European Charter of Fundamental Rights \cite{EU2000} contains vague regulations requiring that people's private and family life is respected. The EU's data retention directive \cite{EUDRD} contains a more concrete regulation requiring communications service providers (e.g. ISPs and telephonic companies) to store people's communications metadata (e.g. a phonecall's time and place) within a fixed time frame. The directive's communications metadata regulation must be shown to be compliant with the Charter of Fundamental Right's more abstract right to a private and family life, or else the directive will be annulled. In fact, the directive was annulled due to violating privacy \cite{ECJDRD}. At the same time, the directive itself governs the design of institutions -- member state's legislation. Member states must implement the directive in a compliant way in order to avoid fines. The directive gives some scope for member states to implement a compliant institution differently, allowing the data retention period to be between 6 and 24 months. The UK's data retention regulations \cite{UK2009} are more concrete and must be shown to ensure communications metadata is stored within the required time frame, no shorter and no longer. In fact, they do just that, concretely requiring communications metadata is stored for 13 months which complies with the abstract requirement of the directive to store data between 6 and 24 months. The UK's institution must be shown to be compliant with the more abstract regulation of the directive in order to avoid fines. In turn, the data retention directive must be shown to be compliant with the more abstract charter of fundamental rights in order to avoid annulment.

To this end, this chapter contributes two main formal building blocks that together contribute a system for detecting non--compliance in multi--level governance. These novel elements are:

\begin{itemize}
\item A way to represent and reason about regulations governing regulations -- specifically, combining norms about agents' actions with norms that oblige and prohibit the imposition of other norms in different social contexts. In this chapter we look explicitly at norms, rules that state what should occur and when. In particular, we look at norms that state what the effects of other norms should be.
\item Abstraction based on constitutive rules -- specifically, taking concrete norms at lower--levels of governance and defining a semantics that re--interprets those concrete norms by \textit{abstracting} them to the same level of abstraction the higher--levels of governance govern them at. The semantics define an abstracting relation between concrete norms and their abstract definitions based on context--sensitive links provided by constitutive rules between concrete and abstract concepts. For example, the European Data Retention Directive \cite{EUDRD} requires member states to store communications metadata. The semantics determine that storing communications metadata without someone's consent is, abstractly, unfair data processing.
\end{itemize}

This chapter is written to be as self--contained as possible. Thus, we start by describing the individual components of institutions and the approach we take to reasoning about them in Section~\ref{secMLGFormApproach}. The representation we use is given in Section~\ref{secMLGSyntax}. The formal reasoning is presented in Section~\ref{secMLGSemantics}. We describe related work in Section~\ref{secMLGRelatedWork}. Conclusions are presented in Section~\ref{secMLGDiscussion}.

\section{Approach}
\label{secMLGFormApproach}

In this section we describe the approach we take to automatically determining compliance in multi--level governance. Since we are reasoning about institutions in multi--level governance, we build on an existing institutional reasoning framework. Our proposal requires representation and reasoning for: constitutive rules, modal regulatory rules, higher--order norms, connections between institutions and reasoning about regulation abstraction. The InstAL (Institution Action Language) framework \cite{Cliffe2006,Cliffe2007} provides constitutive rules and modal regulatory rules. Hence, we base our proposal on the InstAL framework and extend it to multi--level governance with higher--order and abstract norm representation and reasoning. We also modify the InstAL reasoning from being about prohibitive societies (where anything not permitted is forbidden) to permissive societies (everything is permitted unless explicitly prohibited), adding explicit prohibitions and removing explicit permissions. Whilst in InstAL brute facts are observable events or properties which hold in a state (fluents), we only consider brute facts to be observable events. For simplicity, all properties which hold in a state are institutional and not brute in our framework. Based on InstAL \cite{Cliffe2006,Cliffe2007}, an institution in our framework specifies six elements.

Firstly, following InstAL, events. Events can represent observable changes to reality, corresponding to the notion of brute fact. Events can represent changes to the social reality, corresponding to the notion of institutional fact. For example, the brute fact we call storing metadata is an observable event, whilst storing metadata and storing personal data are institutional events.

Secondly, following InstAL, fluents which describe institutional facts holding in a social reality and are subject to changing over time. For example, a user consenting to processing their data causes a fluent to hold stating they have consented, which is removed if they revoke their consent. Some fluents represent the deontic positions that hold. The deontic positions denote institutional \textit{empowerments}, and \textit{obligations and prohibitions}. Empowerments represent that an institutional event is \textit{empowered} to affect the institution, meaning an institutional event \textit{can} occur in the institution. For example, ``storing communications metadata is empowered to occur''. Empowerments represent hard constraints on the events that can (in general) occur (for further discussion see \cite{Cliffe2006,Cliffe2007}). Fluents representing obligations and prohibitions are normative fluents. For example, ``an obligation to pay a fine''. Higher--order normative fluents can also be specified, for example an obligation to oblige paying a fine. We deal with institutions in a temporal setting, so the various deontic positions (normative fluents) express that something should be done before a deadline. For example, an obligation to pay a fine within one month. 

Thirdly, following InstAL, constitutive rules which cause institutional events to occur when observable or other events occur in a given context. For example, ``the observable event of storing metadata counts--as the institutional event of storing metadata''. An example of institutional events causing further institutional events to occur is ``storing personal data counts--as unfair data processing in the context that a user has not consented''.

Fourthly, following InstAL, constitutive rules which initiate and terminate fluents in holding due to institutional events. For example, ``a user consenting to storing their data counts--as initiating the fluent stating the user has consented''. Constitutive rules establishing what we call normative fluents are norms. For example, ``a user using a communications device counts--as initiating an obligation for their communications' metadata to be stored''. Higher--order norms impose higher--order normative fluents. Once a fluent is initiated by such a rule it holds until it is terminated by another rule. That is, these rules initiate and terminate inertial fluents.

Modal norms, which are regulative rules, are constitutive rules which initiate and terminate fluents due to institutional events. For example ``a user using electronic communications initiates an obligation for the communications provider to store their communications metadata''. Thus, norms are not a distinctive element of institutions, but rather a type of constitutive rule.

Fifthly, \textit{extending} InstAL, constitutive rules which derive fluents from other fluents in a particular context. For example, ``an obligation to store personal data non--consensually derives (counts--as) unfair data processing in all social contexts''. Viewed as counts--as rules, these rules ascribe a special meaning B to a fluent A in a context C. For example, an obligation to store personal data non--consensually has the special meaning of being unfair data processing. So long as the fluent `A' holds in a context `C' then its special meaning `B' also holds. But, unlike constitutive rules which initiate and terminate fluents the special meaning `B' does not hold until terminated, rather, it holds when `A' holds in the context `C'. That is the `Bs' in these types of rule are \textit{non--inertial} fluents, since the Bs do not persist from over time by default until terminated (i.e. they do not possess inertia). Unlike the previous rules, constitutive rules which derive fluents from other fluents are not present in the InstAL framework.

Sixthly, following InstAL, a set of initial inertial fluents which hold in the institution's first state and continue to hold until terminated. The set of initial inertial fluents can be the empty set. To summarise, an institution specifies events, fluents and constitutive rules which ascribe institutional events or institutional fluents.

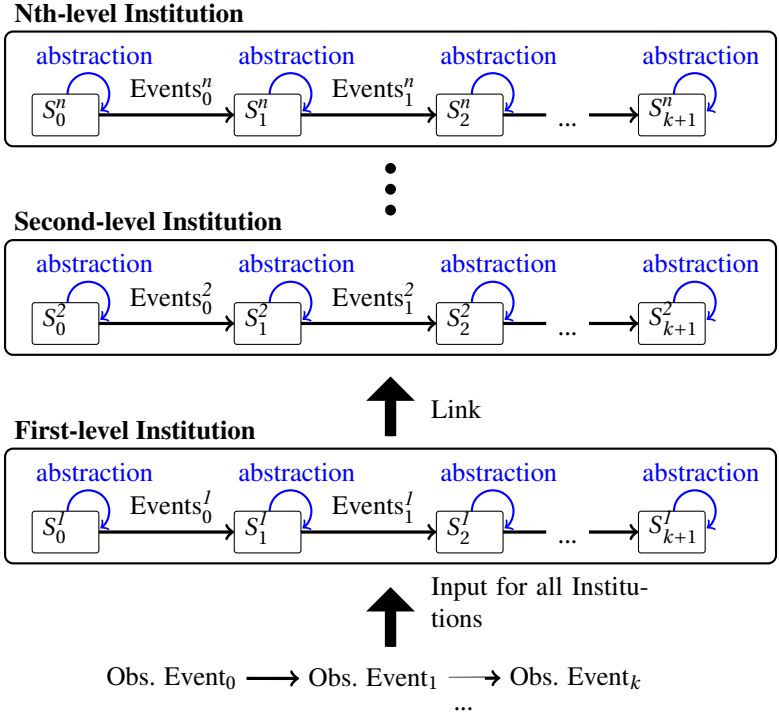
\begin{figure}[t!]
\centering
\def\State at (#1,#2,#3,#4){
\coordinate (Pos) at (#1+0.085,#2+0.1);
\draw 		[thick, ->, color=blue] (Pos) arc (180:-55:0.05);
\node [above, color=blue] at (#1+0.15,#2+0.15) {abstraction};

\draw (#1,#2) [black][rounded corners=1pt] ++(0,0) rectangle +(0.16,0.1);
\node [above right] at (#1,#2) {$S_{#4}^{\textit{#3}}$};
}

\def\InstEvo at (#1,#2,#3,#4){

\coordinate (S0) at (#1+-0.185,#2+0.6);
\State at (#1+-0.185,#2+0.6,#3,0)
\draw (S0) [->, very thick] +(0.16,0.05) -- +(0.485,0.05);
\node [above] at (#1+0.15,#2+0.65) {Events$_{0}^{\textit{#3}}$};

\coordinate (S1) at (#1+0.3,#2+0.6);
\State at (#1+0.3,#2+0.6,#3,1)
\draw (S1) [->, very thick] +(0.16,0.05) -- +(0.485,0.05);
\node  [above] at (#1+0.635,#2+0.65) {Events$_{1}^{\textit{#3}}$};

\coordinate (S2) at (#1+0.785,#2+0.6);
\State at (#1+0.785,#2+0.6,#3,2);
\draw (S2) [->, very thick] +(0.16,0.05) -- +(0.485,0.05);
\draw (#1,#2) [fill][white] +(1.05,0.6) rectangle +(1.15,0.7);
\node [above, font=\bf] at (#1+1.1,#2+0.6) {$...$};

\State at (#1+1.27,#2+0.6,#3,k+1);

\node [above right,font=\bf] at (#1+-0.25, #2+0.85) {#4};
\draw (#1,#2)[black][thick,rounded corners=3pt] +(-0.25,0.575) rectangle +(1.6,0.85);
}

\def\ObsEvent at (#1,#2,#3,#4){
\coordinate (Pos) at (#1+0.085,#2+0.1);
\draw 		[thick, ->] (Pos) arc (180:-55:0.05);
\node [above] at (#1+0.15,#2+0.15) {abstraction};

\draw (#1,#2) [black][rounded corners=1pt] ++(0,0) rectangle +(0.16,0.1);
\node [above right] at (#1,#2) {$S_{#4}^{\textit{#3}}$};
}

\def\ObsTrace at (#1,#2){

\coordinate (S0) at (#1+-0.185,#2+0.6);

\node [above] at (#1+0.15,#2+0.65) (ObEv0) {Obs. Event$_{0}$};
\node  [above] at (#1+0.635,#2+0.65) (ObEv1) {Obs. Event$_{1}$};
\node  [above] at (#1+1.12,#2+0.65) (ObEvk) {Obs. Event$_{k}$};
\draw [->, very thick] (ObEv0.east) -- (ObEv1.west);
\draw [->, very thick] (ObEv1.east) -- (ObEvk.west);

\draw (#1,#2) [fill][white] +(0.8,0.6) rectangle +(0.9,0.7);
\node [above, font=\bf] at (#1+0.85,#2+0.6) {$...$};

}
\begin{tikzpicture}[xscale=5.5,yscale=5.5]
\InstEvo at (0,1,n,Nth-level Institution)
\InstEvo at (0,0.5,2,Second-level Institution)
\InstEvo at (0,0,1,First-level Institution)

\ObsTrace at (0,-0.4)

\fill (0.675,1.522) circle (0.13mm);
\fill (0.675,1.472) circle (0.13mm);
\fill (0.675,1.422) circle (0.13mm);

\draw [line width=1.5mm,
preaction={-triangle 90,line width=0.8mm,draw,shorten >=-1mm}] (0.675,0.880) -- (0.675,1.005);
\node [above right, text width = 3cm] at (0.75, 0.9) {Link};

\draw [line width=1.5mm,
preaction={-triangle 90,line width=0.8mm,draw,shorten >=-1mm}] (0.675,0.375) -- (0.675,0.5);
\node [above right, text width = 3cm] at (0.75, 0.4) {Input for all Institutions};

\end{tikzpicture}
\vspace{-35mm}
\caption[Multi--level Governance reasoning overview]{Multi--level Governance reasoning overview}
\label{figMLGReasOverview}
\end{figure}

Multi--level governance is operationalised with a semantics. This semantics defines how each institution evolves from one state to the next in response to a trace of observable events. These events can be real events occurring in the MAS, or hypothetical  events if a pre--runtime check for compliance is performed. An institution's evolution is  schematically depicted in Figure~\ref{figMLGReasOverview} and described as follows.

The institution starts in an initial state in which its initial set of inertial fluents holds. State transitions are driven by observable events occurring in the MAS (potentially hypothetically). During a state transition, further events occur in an institution according to its constitutive rules, building up an institutional interpretation of reality based on the observable events that have occurred. Further events signifying there is (non--)compliance also occur, for example if there is an obligation to store communications' metadata within one month and the data is not stored within one month, then a norm violation occurs. If it is prohibited to oblige storing communications' metadata, then a higher--order norm violation occurs. That is, norm violations are \textit{institutional events} denoting non--compliance. A newly transitioned to state can contain different fluents from the previous state, based on each institution's constitutive rules variously initiating and terminating fluents from one state to the next. Thus, each institution evolves over time from one state to the next transitioned by events. 

But, how are concrete regulations imposed at lower levels of governance determined to be (non--)~compliant with abstract regulations at higher levels of governance? The approach we take is to firstly, link each institutional level such that concrete normative fluents holding in lower level institutions are `passed up' to the corresponding state in higher level institutions. For example, an obligation to oblige storing communications metadata in the EU--DRD is `passed up' to the EU--CFR. Likewise, so too are norm compliance events. Then, in each institutional state of a higher level institution the concrete normative fluents coming from lower level institutions are re--interpreted and \textit{abstracted}.

In order to abstract concrete normative fluents, we use constitutive rules on which to base the abstraction. Recall that constitutive rules establish links between concrete and abstract concepts, from which we can determine links between concrete and abstract norms. For example, the obligation to oblige storing communications metadata is re--interpreted as an obligation to oblige non--consensual data processing. In turn, from these abstractions any further abstractions are also derived. For example, the obligation to oblige non--consensual data processing is abstracted simply to being unfair data processing. Thus, each institutional state contains concrete normative fluents from lower levels and the state contains the closure of all abstractions on these concrete normative fluents based on constitutive rules. 

Given that normative fluents are abstracted, they can be determined for compliance with abstract higher--order normative fluents that govern them. For example, whether the abstract obligation to oblige storing data non--consensually is in itself prohibited, or perhaps has an even more abstract meaning (e.g. unfair data processing) which is prohibited/obliged. So, concrete normative fluents from lower levels are re--interpreted as more abstract normative fluents at higher levels to determine whether in their abstract incarnation they cause non--compliance.

\begin{figure}[t!]
\begin{tikzpicture}[xscale=0.7, yscale=0.9, every node/.style={xscale=0.9}]

\draw [ultra thick] (-4,17.1556) rectangle (1.5,8.8814);
\draw [ultra thick] (1.5,8) rectangle (-4,6.5);
\draw [ultra thick] (5,6.5) rectangle (10.5,8);
\draw [ultra thick] (5,17.1556) rectangle (10.5,9);

\node [above right, text width = 4cm] at (-4,16.1) {Prohibited to process data unfairly};
\node [above right, text width = 4cm] at (-4,15.3) {Unfair data processing};
\node [above right, text width = 4cm] at (5,14.5) {Prohibited to process data unfairly};

\node [above right, text width=4cm] at (-4,13.1814) {Obligation to oblige processing data non-consensually};
\node [above right, color=blue] at (-1.65,12.45) {\textbf{Abstraction}};
\node [above right, text width=4cm] at (-4,11.1814) {Obligation to oblige storing personal data};
\node [above right, color=blue] at (-1.65,10.35) {\textbf{Abstraction}};
\node [above right, text width=4cm] at (-4,9.1814) {Obligation to oblige storing metadata};
\draw [-triangle 60, color=blue](-1.95,10.2) -- (-1.95,11.25);
\draw [-triangle 60, color=blue](-1.95,12.3) -- (-1.95,13.2);

\node [above right, text width=4cm] at (5,13) {Ada has consented to storing data};
\node [above right, text width=4cm] at (5,11.1814) {Obligation to oblige storing personal data};
\node [above right, text width=4cm] at (5,9) {Obligation to oblige storing metadata};
\draw [-triangle 60, color=blue](7.05,10.05) -- (7.05,11.1);

\draw [-triangle 60, dashdotted](-1.5,8) -- (-1.5,8.9);
\node [fill=white] at (-1.5,8.4) {Link};

\draw [-triangle 60, dashdotted](7.5,8) -- (7.5,9);
\node [fill=white] at (7.5,8.4) {Link};

\node [above right, text width=2.5cm] at (1.5,11.7136) {\textbf{Ada consents to storing data\\ Violated prohibition to process data unfairly}};
\draw [-triangle 60,ultra thick](1.5,11.2136) -- (5,11.2136);

\node [above right, text width=4cm] at (-4,6.5) {Obligation to oblige storing metadata};
\node [above right, text width=4cm, color=blue] at (7.2,10.2) {\textbf{Abstraction}};
\node [above right, text width=4cm] at (5,6.5) {Obligation to oblige storing metadata};
\node [above right, text width=2.5cm] at (1.5,7) {\textbf{Ada consents to storing data}};
\draw [-triangle 60,ultra thick](1.5,7) -- (5,7);\textbf{Violation of prohibition to oblige processing data unfairly}\newline \newline

\node [above right, text width = 2.5cm] at (-7.5,12.5) {\textbf{Charter of Fundamental Rights of the EU}};
\node [above right, text width = 2.5cm] at (-7.5,6.5) {\textbf{Data Retention Directive}};

\draw [-triangle 60, color=blue](-1.95,14.7) -- (-1.95,15.3);

\node [above right, color=blue] at (-1.65,14.7) {\textbf{Abstraction}};

\end{tikzpicture}
\caption[Multi--level governance abstraction semantics example]{An example of abstracting normative fluents at different levels of governance based on the context. Normative fluents oblige/prohibit an aim $a$ occurs before or at the same time as a deadline $d$. We use $<$ to denote one thing occurring before another and $\leq$ to denote one thing occurring before or at the same time as another.}
\label{figAbstrExample}
\end{figure}
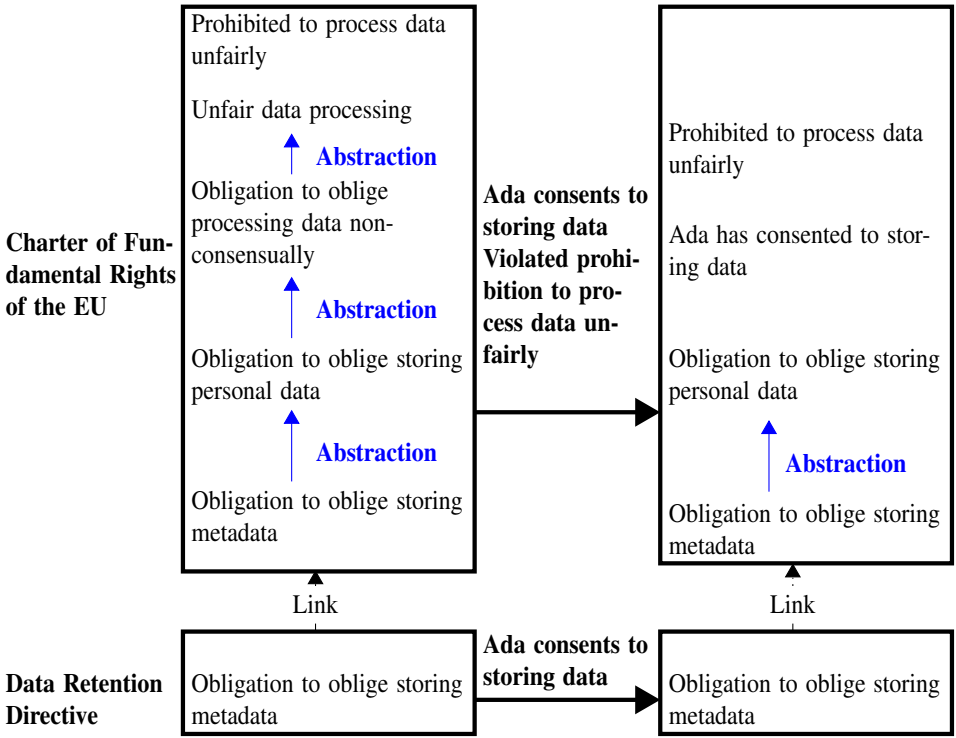

An example is depicted in Figure~\ref{figAbstrExample} based on the running case study. In the EU--DRD's first state there is an obligation to oblige storing communications' metadata, which is passed up to the EU--CFR. As described, concrete normative fluents are abstracted based on whether the prescribed event \textit{counts--as} a more abstract event in a context entailed by the state. Storing metadata counts--as storing personal data. Storing personal data counts--as non--consensual data processing in the context where an agent has not consented. In the EU--CFR's first state the obligation to oblige storing metadata is abstracted to an obligation to oblige storing personal data. Then, to an obligation to oblige processing data without consent. An obligation to oblige processing data non--consensually is abstracted further to `unfair data processing'. Unfair data processing is prohibited and thus a norm violation event occurs in the transition to the EU--CFR's next state. 

In the EU--CFR institution the next state lacks an obligation to oblige processing data without consent because a user has consented. So, unfair data processing also does not hold. That is, the abstract meaning of concrete normative fluents evolves as the context evolves. Consequently, compliance of normative fluents is context sensitive because normative fluents' abstraction is context sensitive.

To summarise, our semantics for multi--level governance defines the evolution of each institution over time in response to a sequence of events. Specifically, whether concrete normative fluents imposed by lower level institutions have an abstract interpretation in higher level institutions. Moreover, whether these abstract interpretations of concrete norms violate abstract higher--order norms in higher level institutions. Such non--compliance can be determined by inspecting the sequence of events in higher level institutions for higher--order norm compliance events. The semantics provide a mechanism acting as a kind of legal monitor for multi--level governance.

\section{Formal Representation}
\label{secMLGSyntax}

In this section we present the syntax for representing multi--level governance. We begin with representing normative fluents. These are fluents which represent temporal obligations and prohibitions, meaning they have an aim which should be achieved before a deadline. In contrast, InstAL considers at temporal obligations and non--temporal permissions, where anything not permitted is prohibited (a prohibitive society). We look at temporal obligations and temporal prohibitions, where anything not prohibited is permitted (a permissive society).

\begin{figure}
\centering
\colorlet{discharged}{black!30!green}
\colorlet{violated}{black!10!red}

\begin{tikzpicture}[xscale=0.9, yscale=0.9, every node/.style={xscale=0.9}]

\node at (-7,-0.5) {\large{$\textit{obl(a,d)}$}};
\node at (-2,-0.5) {\large{$\textit{pro(a,d)}$}};
\node at (-7,-4.5) {\large{Event/fluent $a$}};
\node at (-2,-4.5) {\large{Event/fluent $d$}};

\draw [-triangle 60, ultra thick, color=violated](-6.5,-4) -- (-2.5,-1);
\fill [-triangle 60, fill=white] (-4,-1.3332)rectangle (-3,-2);
\node [text width = 1cm] at (-3.806,-1.5552) {Violates if $< d$};

\draw [-triangle 60, ultra thick, color=discharged](-7,-4) -- (-7,-1);
\fill [-triangle 60, fill=white] (-7.5,-2)rectangle (-6.5,-3);
\node [text width = 1cm] at (-7,-2.5) {Discharges if $\leq d$};

\draw [-triangle 60, ultra thick, color=discharged](-2,-4) -- (-2,-1);
\fill [-triangle 60, fill=white] (-2.5,-2)rectangle (-0.5,-3);
\node [text width = 1cm] at (-2,-2.5) {Discharges if $\leq a$};

\draw [-triangle 60, ultra thick, color=violated](-2.5,-4) -- (-6.5,-1);
\fill [-triangle 60, fill=white] (-6.0556,-1.3888)rectangle (-5,-2);
\node [text width = 1cm] at (-5.5,-1.5552) {Violates if $< a$};

\node at (-7,2.5) {\large{$\textit{obl(obl(a,} d\textit{),}d^{\prime})$}};
\draw [-triangle 60, ultra thick, color=discharged](-7,0) -- (-7,2);
\fill [-triangle 60, fill=white] (-7.5,1.5)rectangle (-6.5,0.5);
\node [text width = 1cm] at (-7,1) {Discharges if $\leq d^{\prime}$};

\node at (-2,2.5) {\large{$\textit{obl(pro(a,} d\textit{),}d^{\prime})$}};
\draw [-triangle 60, ultra thick, color=discharged](-2,0) -- (-2,2);
\fill [-triangle 60, fill=white] (-2.5,1.5)rectangle (-1.5,0.5);
\node [text width = 1cm] at (-2,1) {Discharges if $\leq d^{\prime}$};

\node at (-10,2.5) {\large{$\textit{pro(obl(a,} d\textit{),}d^{\prime})$}};
\draw [-triangle 60, ultra thick, color=violated](-8,0) -- (-10,2);
\fill [-triangle 60, fill=white] (-9.5,1.5)rectangle (-8.5,0.5);
\node [text width = 1cm] at (-9,1) {Violates if $< d^{\prime}$};

\node at (1,2.5) {\large{$\textit{pro(pro(a,} d\textit{),}d^{\prime})$}};
\draw [-triangle 60, ultra thick, color=violated](-1,0) -- (1,2);
\fill [-triangle 60, fill=white] (-0.5,1.5)rectangle (0.5,0.5);
\node [text width = 1cm] at (0,1) {Violates if $< d^{\prime}$};
\end{tikzpicture}
\caption[Modal norm discharge and violation]{Discharge and violation (higher--order) normative fluent conditions. $< X$ denotes the event/fluent holding or occurring strictly before X causes a violation. $\leq X$ denotes the same, but the condition is non--strictly before.}
\label{figDischViolOverview}
\end{figure}
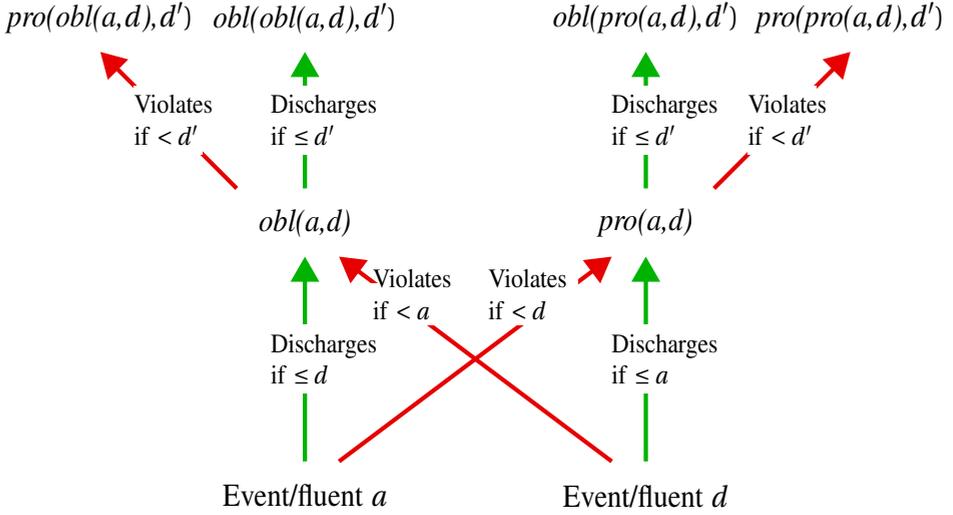

Obligation and prohibition fluents are respectively represented as $\textit{obl(aim, deadline)}$ and $\textit{pro(aim, deadline)}$. The aims and deadlines can be events, fluents or other normative fluents to represent higher--order normative fluents. Two special events are used in aims and deadlines, \textit{now} and \textit{never}. The event `now' occurs immediately to represent something should (not) be done immediately. For example, $\textit{obl(aim, now)}$ means the aim should occur `now'. Our representation is inspired by the formalisation of instantaneous norms in a variant of dynamic logic \cite{Dignum1996}, which similarly use `now' to refer to the present state. An aim or deadline event \textit{never} represents that the aim/deadline never occurs, potentially meaning the normative fluents lasts indefinitely. For example $\textit{pro(aim, never)}$ means it is always prohibited for the aim to occur or in other words the aim should `never' occur. 

The informal semantics for normative fluents' is described in terms of when obligations/prohibitions are discharged and violated, overviewed in Figure~\ref{figDischViolOverview}. An obligation fluent, of the form $\textit{obl(aim, deadline)}$, represents that the aim should occur/hold before or at the same time as the deadline to \textit{discharge} the obligation (fulfil all duties). If the deadline occurs/holds strictly before the aim then the obligation is \textit{violated}. Prohibition fluents, of the form $\textit{pro(aim, deadline)}$, are the inverse of obligations. Prohibitions represent that the aim should not occur/hold strictly before the deadline. When a normative fluent $n$ is discharged it causes the event $\textit{disch}(n)$ to occur. If $n$ is violated then the event $\textit{viol}(n)$ occurs.

Higher--order norms impose higher--order normative fluents. A higher--order normative fluent obliges/prohibits another normative fluent is imposed before a deadline. The deadline is an event or another normative fluent. Compliance--focussed higher--order norms can also be expressed, which oblige/prohibit compliance with a norm (norm discharge -- $\textit{disch}(n)$) or violation ($\textit{viol}(n)$) occurs before/after an event occurs or another normative fluent is imposed (e.g. it is obliged a norm is violated before a fine is imposed). 

A grammar to specify normative fluents is formalised as:

\begin{definition}{\textbf{Normative Fluents}}\label{defNormFl} Let $P$ be a set of propositions denoting events with typical element $p$. The set of normative fluents $\pazocal{N}\vert _{P}$ is the set of all normative fluents $n$ expressed as:
\begin{alignat*}{3}
& \textit{aim} & \; \; \; ::= & \; \; \; p \; \mid n \; \mid \textit{now} \; \mid \textit{never} \; \mid \textit{disch(n)} \; \mid \textit{viol(n)} \\
& \textit{deadline} & \; \; \; ::= & \; \; \; p \; \mid n \; \mid \textit{now} \; \mid \textit{never} \; \mid \textit{disch(n)} \; \mid \textit{viol(n)} \\
& \textit{n} & \; \; \; ::= & \; \; \; \textit{obl}(\textit{aim}, \textit{deadline}) \; \mid \textit{pro}(\textit{aim}, \textit{deadline})
\end{alignat*}
\end{definition}

We give some examples concerning two agents, a law enforcement officer called Charles and  an internet communications user called Ada, and an internet communications provider colloquially called an ISP. The UK--DRR implements the EU--DRD. They state that if a law enforcement official (e.g. $\textit{charles}$) requests the data stored by a communications provider (e.g. $\textit{isp}$) of a user (e.g. $\textit{ada}$) then the communications provider is obliged to provide the data within one month ($\textit{m1}$):
\begin{equation*}
\textit{obl(provideData(isp, charles, ada), time(m1))}
\end{equation*}

Instantaneous normative fluents express that something should (not) be done or a normative fluent should (not) be imposed \textit{now}. One way an institution designer might use instantaneous norms is to express that as soon as something happens a normative fluent should be imposed. For example, as soon as a norm is violated it is obliged that there is an obligation to punish the violator. The EU--DRD as we formalise it, requires that any implementing institution should impose punishment as soon as regulations are violated. Thus, when there is a violation it imposes a normative fluent obliging an obligation to punish the violator is imposed immediately:

\begin{equation*}
\textit{obl(obl(punish(isp), time(m6)), now)}
\end{equation*}

Compliance--focussed normative fluents can be used to express that an agent should discharge/violate a normative fluent before another normative is imposed that rewards/punishes the agent. For example, in our previous work \cite{King2015b}, an obligation expressed ``it is obliged that a norm is violated before a fine is imposed''. Such compliance focussed normative fluents do not state that a normative fluent being discharged should \textit{cause} a reward/punishment. Rather, they state that discharge/violation should occur before the reward/punishment is imposed. Following this chapter's case--study -- it is obliged that the communications provider $\textit{isp}$ violates the obligation to provide $\textit{charles}$ with data which concerns $\textit{ada}$ before any obligation to punish the communications provider $\textit{isp}$ is imposed.

\begin{equation*}
\textit{obl(viol(obl(provideData(isp, charles, ada), time(m1))), obl(punish(isp), time(m6)))}
\end{equation*}

Normative fluents can also be explicitly first--order, but implicitly higher--order by obliging/prohibiting fluents that abstractly represent other normative fluents. Recall that various obligations in the EU--DRD can abstractly be interpreted as unfair data processing. Hence, the following is an example of an abstract first--order norm that indirectly governs other norms. The EU--CFR states that it is prohibited to process Ada's data unfairly (indefinitely):

\begin{equation*}
\textit{pro(unfairDataProcessing(ada), never)}
\end{equation*}

We now proceed to representing individual institutions. In short, institutions are specified as a tuple, extending the formal specification of an institution in the InstAL framework \cite{Cliffe2006}. Generally, speaking, an individual institution describes the things that can occur (events) and hold (fluents) in the institution as well as the institution's rules causing events to occur and fluents to hold. An institution's constitutive rules -- cause institutional events to occur in response to other events (``an event A counts--as an event B in context C''), fluents to hold in response to events (``an event A counts--as establishing/removing a fluent B in context C''), and further, more abstract, fluents to be derived from other fluents (``a fluent A counts--as a fluent B in context C''). Rules stating fluents are derived are not present in InstAL but we introduce them since they provide an abstracting relation \textit{between} fluents and thus contribute to our goal of reasoning about abstraction in multi--level governance. Regulative rules are just modal norms represented as constitutive rules which establish normative fluents, ``an event A counts--as establishing an obligation/prohibition in context C''. 

Specifically, institutions comprise the following elements:

\begin{itemize}
\item[] \textbf{Events} -- a set of propositions ($\pazocal{E}$) denoting events that can occur in the institution, s.t. $\textit{now}, \textit{never} \not \in \pazocal{E}$, meaning the institution cannot define when the \textit{brute} events $\textit{now}$ and $\textit{never}$ occur. The set of events comprises:
\begin{itemize}
\item Observable events ($\pazocal{E}_{\textit{obs}}$) that are exogenous to the institution corresponding to the notion of a \textit{brute fact} denoting an event.
\item Internal institutional events ($\pazocal{E}_{\textit{inst}}$) representing an institutional description of reality.
\item Compliance events ($\pazocal{E}_{\textit{norm}} = \{ \textit{disch}(n), \textit{viol}(n) \mid \; n \in \pazocal{F}_{\textit{cnorm}} \cup \pazocal{F}_{\textit{anorm}} \}$) indicating a normative fluent (in the set of concrete and abstract normative fluents $\pazocal{F}_{\textit{cnorm}} \cup \pazocal{F}_{\textit{anorm}}$) has been discharged or violated.
\end{itemize}
\item[] \textbf{Fluents} -- a set of propositions ($\pazocal{F}$) denoting fluents which can hold in the institution, comprising:
\begin{itemize}
\item \textit{Domain fluents} ($\pazocal{F}_{\textit{dom}}$) providing an institutional description of the state of reality (e.g. an agent has consented to their data being processed).
\item \textit{Empowerment fluents} ($\pazocal{F}_{\textit{pow}} \subseteq \{ \textit{pow(e)} \mid e \in \pazocal{E}_{\textit{inst}} \}$) denoting an event is \textit{recognised} by the institution in a state and has the power to affect the institution (i.e. is empowered).
\item \textit{Normative fluents} ($\pazocal{F}_{\textit{norm}} = \pazocal{F}_{\textit{cnorm}} \cup \pazocal{F}_{\textit{anorm}}$) comprising mutually disjoint sets of \textit{concrete normative fluents} ($\pazocal{F}_{\textit{cnorm}} \subseteq \pazocal{N} \vert_{\pazocal{E} \cup \pazocal{F}_{\textit{dom}}}$) and \textit{abstract normative fluents} ($\pazocal{F}_{\textit{anorm}} \subseteq \pazocal{N} \vert_{\pazocal{E} \cup \pazocal{F}_{\textit{dom}}}$):
\begin{itemize}
\item \textit{Concrete normative fluents} denote obligations and prohibitions imposed by the institution about events or domain fluents. These normative fluents are concrete in the sense of being explicitly imposed by an institutional norm, rather than being abstract interpretations of other normative fluents that have been imposed.
\item \textit{Abstract normative fluents} denote obligations and prohibitions imposed by the institution about events or domain fluents. These are abstract in the sense of not being imposed by the institution, but rather represent an abstract interpretation of other \textit{more} concrete normative fluents. For example, an obligation to store personal data is a more abstract interpretation of an obligation to store communications metadata. 

\end{itemize}
\item \textit{Inertial and non--inertial fluents},
We assume that fluents are either inertial or non--inertial represented as mutually disjoint sets of \textit{inertial} fluents ($\pazocal{F}_{\textit{inert}}$) and \textit{non--inertial} fluents ($\pazocal{F}_{\textit{ninert}}$) such that $\pazocal{F} = \pazocal{F}_{\textit{inert}} \cup \pazocal{F}_{\textit{ninert}}$ and $\pazocal{F}_{\textit{inert}} \cap \pazocal{F}_{\textit{ninert}} = \emptyset$. Institutions define fluents which can be initiated by the institution's state consequence function and then persist from one state to the next by default until they are terminated. That is, some fluents are \textit{inertial}. Other fluents hold due to constitutive rules stating more abstract fluents are derived from more concrete fluents. These abstract fluents hold whenever the concrete fluents hold and do not persist from state to state by default. That is, they are \textit{non--inertial} fluents. Concrete normative fluents are inertial, since an institution explicitly imposes them by initiation and termination according to the state consequence function  ($\pazocal{F}_{\textit{cnorm}} \subseteq \pazocal{F}_{\textit{inert}}$). Abstract normative fluents are non--inertial since they are derived from other normative fluents and do not persist from state to state by default ($\pazocal{F}_{\textit{anorm}} \subseteq \pazocal{F}_{\textit{ninert}}$).
\end{itemize}
\item[] \textbf{Contexts} -- these characterise a condition on a state and denote the social context each rule is conditional on. A context is a set of positive and weakly negative fluents, which holds in a state if all the positive fluents hold and none of the negative fluents hold. Formally, the set of all contexts is $\pazocal{X} = 2^{\pazocal{F} \cup \neg \pazocal{F}}$ s.t. $\neg \pazocal{F} = \{ \neg f \mid f \in \pazocal{F} \}$ is the set containing the negation of all elements in the set $\pazocal{F}$.
\item[] \textbf{State change rules} ($\pazocal{C} : \pazocal{X} \times \pazocal{E} \rightarrow 2^{\pazocal{F}_{\textit{inert}}} \times 2^{\pazocal{F}_{\textit{inert}}}$), described as a state consequence function. They specify that due to the occurrence of events conditional on a context holding in a state, inertial fluents are initiated and terminated from one state to the next. State change rules can be descriptive (e.g. a user consenting to their data being stored initiates a fluent stating that they have consented) and regulative rules by initiating and terminating normative fluents (e.g. someone using electronic communications initiates an obligation for the communications provider to store their communications' metadata).
\item[] \textbf{Event generation rules} -- ($\pazocal{G} : \pazocal{X} \times \pazocal{E} \rightarrow 2^{\pazocal{E}_{\textit{inst}}}$), described as an event generation function. These rules are only descriptive. They specify that when an exogenous or institutional event occurs conditional on a context holding in a state another institutional event occurs.
\item[] \textbf{Fluent derivation rules} -- ($\pazocal{D} : \pazocal{X} \times \pazocal{F} \rightarrow 2^{\pazocal{F}_{\textit{ninert}}}$), described as a fluent derivation function. These rules state that a fluent holding in a state derives a \textit{non--inertial} fluent holding in the same state, conditional on a social context.
\end{itemize}

According to these notions, an individual institution is formally defined as:

\begin{definition}{\textbf{Individual Institution}}\label{def:inst}  An institution is a tuple $\pazocal{I} = \langle \pazocal{E}, \pazocal{F}, \pazocal{C}, \pazocal{G}, \pazocal{D}, \Delta \rangle$ such that:
\begin{itemize}
\item $\pazocal{E} = \pazocal{E}_{\textit{obs}} \cup \pazocal{E}_{\textit{inst}} \cup \pazocal{E}_{\textit{norm}}$ is the set of events.
\item $\pazocal{F} = \pazocal{F}_{\textit{dom}} \cup \pazocal{F}_{\textit{norm}}$ is the set of fluents.
\item $\pazocal{C} : \pazocal{X} \times \pazocal{E} \rightarrow 2^{\pazocal{F}_{\textit{inert}}} \times 2^{\pazocal{F}_{\textit{inert}}}$ is the state consequence function.
\item $\pazocal{G} : \pazocal{X} \times \pazocal{E} \rightarrow 2^{\pazocal{E}_{\textit{inst}}}$ is the event generation function.
\item $\pazocal{D} : \pazocal{X} \times \pazocal{F} \rightarrow 2^{\pazocal{F}_{\textit{ninert}}}$ is the fluent derivation function.
\item $\Delta \subseteq \pazocal{F}_{\textit{inert}}$ is a set of inertial fluents which hold in the institution's initial state and persist from one state to the next if they are not terminated.
\end{itemize}
Some further useful constructs are:
\begin{itemize}
\item $\Sigma = 2^{\pazocal{F}}$ to denote the set of all states for $\pazocal{I}$.
\item Given a context $X \in \pazocal{X}$ and an event $e \in \pazocal{E}$ we denote the result of the consequence function as $\pazocal{C}(X, e) = \langle \pazocal{C}^{\uparrow}(X, e), \pazocal{C}^{\downarrow}(X, e) \rangle$ s.t. the set of initiated fluents is $\pazocal{C}^{\uparrow}(X, e)$ and the set of terminated fluents is $\pazocal{C}^{\downarrow}(X, e)$. 
\item For readability if an institution is denoted with a superscript, such as $\textit{id}$ then all its elements have the same superscript, such as $\pazocal{I}^{\textit{id}} = \langle \pazocal{E}^{\textit{id}}, \pazocal{F}^{\textit{id}}, \pazocal{C}^{\textit{id}}, \pazocal{G}^{\textit{id}}, \pazocal{D}^{\textit{id}}, \Delta^{\textit{id}} \rangle$, the set of states being $\Sigma^{\textit{id}}$ and the set of contexts being $\pazocal{X}^{\textit{id}}$.
\end{itemize}
\end{definition}

We exemplify using institutional specification fragments where for clarity we use a superscript denoting the name of each institution. The EU--CFR \cite[Art. 8.2]{EU2000} states that a person's data must be processed fairly. For an agent called `ada', the set of inertial fluents initially holding in the EU--CFR institution includes:

\begin{equation*}
\textit{pro(unfairDataProcessing(ada), never)} \in \Delta^{\textit{cfr}}
\end{equation*}

A communications provider, called `isp`, storing metadata is by default an event \textit{empowered} to affect the Data Retention Regulations:

\begin{equation*}
\textit{pow(storeData(isp, ada, metadata))} \in \Delta^{\textit{drd}}
\end{equation*}

According to the European Court of Justice's (ECJ) judgement \cite{ECJDRD} on the EU--DRD, with respect to the EU--CFR, storing communications metadata counts--as storing personal data. If an agent's, Ada's, metadata is stored in any context (the empty set) then the event of storing the agent's, Ada's, personal data is generated. Additionally, if Ada's personal data is stored in the context that Ada has not consented then the event of non--consensually processing Ada's data occurs. The following rules are a part of the EU--CFR, incorporating the ECJ's judgement.
\begin{subequations}
\begin{align*}
\pazocal{G}^{\textit{cfr}}(\emptyset, \textit{storeData(isp, ada, metadata)}) \ni \textit{storeData(isp, ada, personal)}
\end{align*}
\begin{align*}
\pazocal{G}^{\textit{cfr}}(\{ & \neg \textit{consentedDataProcessing(ada, isp)} \}, \textit{storeData(isp, ada, personal)}) \ni \\ &
\textit{nonConsensualDataProcessing(ada)}
\end{align*}
\end{subequations}

The EU--DRD \cite[Art. 8]{EUDRD} requires data concerning people is transmitted to authorities on request before any undue delay. A fluent initiation rule is conditional on an agent, Charles, requesting the data from a communications provider, ISP, of another agent, Ada. In the context that Charles is a law enforcement official the rule initiates an obligation to immediately oblige that ISP provides Charles with data concerning Ada before any undue delay.
\begin{align*}
\pazocal{C}^{\textit{drd} \uparrow}(\{ & \textit{is(charles, lawEnforcement)} \}, \textit{requestData(ada, isp, charles)}) \ni \\ & 
\textit{obl(obl(provideData(isp, charles, ada), undue\_delay), now)} 
\end{align*}

According to the ECJ's interpretation of the EU--DRD \cite{ECJDRD} with respect to the EU--CFR. Obliging that personal data is processed non--consensually counts--as unfair data processing. We represent the ECJ's interpretation as a fluent derivation rule in the EU--CFR institution. It states that obliging an agent, Ada's, personal data is processed without consent in any social context (the empty set) derives the fluent of (counts--as) unfair data processing.
\begin{align*}
\pazocal{D}^{\textit{cfr}}(\textit{obl(nonConsensualDataProcessing(ada), now)}, \emptyset) \ni \textit{unfairDataProcessing(ada)}
\end{align*}

In multi--level governance, institutions are related in the sense that institutions designed at lower levels of governance are governed by institutions designed at higher levels of governance. In our approach, this means that if a lower level institution imposes an obligation or a prohibition, then the same obligation/prohibition holds in any higher level institution which governs it. Institutions are linked in this way in what we call a \textit{multi--level governance institution}, where institutions are ordered by the level they operate at and linked with a relation between lower level and higher level institutions. The relations linking institutions are expressed as a set of directed edges $R$ between lower level institutions and higher level institutions they are governed by. Each relation is restricted such that higher levels cannot be governed by lower levels, such that collectively the relations are always acyclic. Formally, a multi--level governance institution is:

\begin{definition}{\textbf{\GovStruct{}}} A \GovStruct{} is a directed graph  $\langle \pazocal{T}, R \rangle$. The vertices are represented as a tuple $\pazocal{T} = \langle \pazocal{I}^{1}, ..., \pazocal{I}^{n} \rangle$ of institutions. The arrows are a set of pairs $R = 2^{[1,n] \times [1,n]}$ of institution indexes in $\pazocal{T}$ such that $\forall \langle {i}, {j} \rangle \in R : i < j$.
\end{definition}

According to these definitions, we can represent the three main aspects of multi--level governance we focus on in this chapter. Firstly, regulations which regulate other regulations in higher level institutions with higher--order normative fluents in prescriptive rules. Secondly, the links between lower level institutions governed by higher level institutions by composing multi--level governance institutions. Thirdly, the interpretation of concrete concepts and normative fluents as more abstract concepts and normative fluents.

In our framework abstraction can occur in a multi--level governance institution in two ways. Firstly with constitutive rules which state a concrete concept counts--as a more abstract concept in a particular context. These abstracting constitutive rules are represented as the event generation function -- a concrete event counts--as a more abstract event in a particular social context; the state consequence function -- a concrete event counts--as initiating/terminating a fluent that abstractly describes the social reality in a particular context; and the fluent derivation function -- a concrete fluent counts--as a more abstract fluent in a particular social context. The second method of abstraction is implicit abstraction of concrete normative fluents regulating concrete concepts to more abstract normative fluents regulating abstract concepts. Normative fluent abstraction is defined semantically, based on constitutive rules between concrete and abstract concepts, and requires no explicit representation.

\section{Case Study Formalisation}
\label{secCaseStudy}

To demonstrate the representation in full, our case study and its formalisation in the formal framework are subsequently summarised. We look at three legal institutions. These institutions operate at different levels of abstraction. Higher--levels interpret the regulatory effects of lower--levels in terms of more abstract normative fluents and concepts based on constitutive rules. Providing the constitutive rules to reinterpret and abstract regulations are the interpretations provided by courts. For example, the lower--level Data Retention Directive's interpretation by the European Court of Justice (ECJ) with respect to the higher--level Charter of Fundamental Rights. We view constitutive rules provided by court interpretations of a lower--level institution as being a part of the higher--level institution which governs the lower--level. For example, the Charter of Fundamental Rights comprises its rules and regulations set out in writing and also the interpretations of the Data Retention Directive's concepts provided by the ECJ.

Formally, the multi--level governance institution is the tuple $\pazocal{ML} = \langle \pazocal{T}, R \rangle$ where the institutions are $\pazocal{T} = \langle \pazocal{I}^{\textit{drr}}, \pazocal{I}^{\textit{drd}}, \pazocal{I}^{\textit{cfr}} \rangle$ and the governance relation is $R = \{ \langle \textit{drr}, \textit{drd} \rangle, \langle \textit{drd}, \textit{cfr} \rangle \}$. Each of the institutions are formalised in tables \ref{tabCFRFormalExample}, \ref{tabDRDFormalExample} and \ref{tabDRRFormalExample} where for brevity we leave out empowerment fluents and upper--case terms represent variables acting as shorthand for their ground instantiations. For clarity, we use labels rather than indexes for each institution. Each institution is described and formalised as follows.

\begin{itemize}
\item[] \textbf{The EU Charter of Fundamental Rights} \cite{EU2000} (EU--CFR, Formalised in Table~\ref{tabCFRFormalExample}), a third--level institution which through the most abstract regulations governs the Data Retention Directive. The specific fragments of legislation we look at are:
\begin{itemize}
\item Article 7 -- ``\textit{Everyone has the right to respect for his or her private and family life, home and communications.}'' (rule \ref{exConCFR2}, and \ref{exInSCFR1})
\item Article 8 -- ``\textit{Everyone has the right to the protection of personal data concerning him or her.}'' (rule \ref{exConCFR3}, and \ref{exInSCFR1}) and ``\textit{Compliance with these rules shall be subject to control by an independent authority.}'' (rule \ref{exConCFR5}, and \ref{exInSCFR1})
\end{itemize}
Partly based on the European Court of Justice's interpretation of the Data Retention Directive \cite{ECJDRD} we also take the following legal interpretations as holding in the EU--CFR:
\begin{itemize}
\item Storing communications content or metadata counts--as storing personal data (rules \ref{exGenCFR1} and \ref{exGenCFR2}).
\item Storing personal data if the person it concerns has not consented counts--as non--consensual data processing. (rule \ref{exGenCFR3}).
\item There is unfair data processing whenever there is an obligation for there to be non--consensual data processing (rule \ref{exDerCFR1}).
\item Data is unprotected if there is an obligation to store data in the context that the data is not anonymised (rule \ref{exDerCFR4}).
\item Privacy is disrespected when there is an obligation to store personal data (rule \ref{exDerCFR8}).
\end{itemize}
It is interesting to note that the Charter of Fundamental Rights regulates regulations but in an implicit manner, that is, nowhere does it state that the regulations regulate other regulations. That is, not obliging and prohibiting obligations or prohibitions, but rather obliging and prohibiting abstract concepts which can represent obligations and prohibitions. For example, rather than prohibiting obligations for data to be processed unfairly, the EU--CFR prohibits unfair processing itself. In this chapter, we assume that whenever the EU--CFR governs an abstract concept, such as unfair data processing, then it views obliging that abstract concept as counting--as that abstract concept. For example, processing data unfairly has many interpretations, one of which is obliging data is processed unfairly. Following this idea, if the EU--CFR prohibits processing data unfairly, then an obligation to process data unfairly is reduced to the abstract concept of unfair data processing itself, which violates the EU--CFR's prohibition. Through transitivity, it follows that an obligation to oblige unfair data processing is also reduced in the same way resulting in a violation of a prohibition to process data unfairly. Interpreting abstract normative fluents as the abstract concepts they oblige/prohibit is realised with fluent derivation rules:
\begin{itemize}
\item Obliging unfair data processing counts--as unfair data processing (rule \ref{exDerCFR2}).
\item Obliging data is anonymised counts--as data being anonymised (rule \ref{exDerCFR3}).
\item Obliging data is unprotected counts--as data being unprotected (rule \ref{exDerCFR5}).
\item Obliging privacy is disrespected counts--as privacy being disrespected \\(rule \ref{exDerCFR7}).
\item Obliging data is processed counts--as data being processed (rule \ref{exDerCFR9}).
\end{itemize}
\end{itemize}

\newcounter{rowno}
\setcounter{rowno}{0}

\begin{spacing}{0.1}
\begin{longtable}[l]
{%
     >{\collectcell\myalign}%
      m{0.05\textwidth}%
     <{\endcollectcell}%
     >{\collectcell\myalign}%
      m{0.2\textwidth}%
     <{\endcollectcell}%
}
\caption[Charter of Fundamental Rights formalisation]{Charter of Fundamental Rights of the European Union Formalisation}
\label{tabCFRFormalExample}
\endfirsthead
\endhead
{
& \pazocal{G}^{\textit{cfr}}(\emptyset, \textit{storeData(CommProv0, Agent0, content)}) \ni \\
& \textit{storeData(CommProv0, Agent0, personal)}
} & 
{
\tag{CFRG.1}\label{exGenCFR1}
} \\
{
& \pazocal{G}^{\textit{cfr}}(\emptyset, \textit{storeData(CommProv0, Agent0, metadata)}) \ni \\
& \textit{storeData(CommProv0, Agent0, personal)}
} &
{
\tag{CFRG.2}\label{exGenCFR2}
} \\
{
& \pazocal{G}^{\textit{cfr}}( \{ \neg \textit{consentedDataProcessing(Agent0, CommSerProv)} \}, \\ &\textit{storeData(CommProv0, Agent0, personal)}) \ni \\ & \textit{nonConsensualDataProcessing(Agent0)}
} &
{
\tag{CFRG.3}\label{exGenCFR3}
} \\
{
& \pazocal{G}^{\textit{cfr}}(\{ \neg \textit{jurisdiction(Location0, eu)} \}, \\
& \textit{storeDataAt(CommProv0, Agent0, Location0)}) \ni \\ & \textit{storeDataOutsideEU}
} &
{
\tag{CFRG.4}\label{exGenCFR4}
} \\
{
& \pazocal{C}^{\textit{cfr}\uparrow}(\emptyset, \textit{consent(CommSerProv0, Agent0}) \ni \\ & \textit{consentedDataProcessing(Agent0, CommSerProv0)}
} &
{
\tag{CFRC.1}\label{exConCFR1}
} \\
{
& \pazocal{C}^{\textit{cfr}\uparrow}(\emptyset, \textit{viol(pro(privacyDisrespected(Agent), never))}) \ni \\ & \textit{pro(privacyDisrespected(Agent), never)}
} &
{
\tag{CFRC.2}\label{exConCFR2} 
} \\
{
& \pazocal{C}^{\textit{cfr}\uparrow}(\emptyset, \textit{viol(\textit{pro(dataUnprotected(Agent0, personal), never)})}) \ni \\
& \textit{pro(dataUnprotected(Agent0, personal), never)}
} &
{
\tag{CFRC.3}\label{exConCFR3}
} \\
{
& \pazocal{C}^{\textit{cfr}\uparrow}(\emptyset, \textit{viol(pro(unfairDataProcessing(Agent0), never))}) \ni \\ & \textit{pro(unfairDataProcessing(Agent0), never)}
} &
{
\tag{CFRC.4}\label{exConCFR4}
} \\
{
& \pazocal{C}^{\textit{cfr}\uparrow}(\emptyset, \textit{viol(pro(uncontrolByIndepAuth, never))}) \ni \\ & \textit{pro(uncontrolByIndepAuth, never)}
} &
{
\tag{CFRC.5}\label{exConCFR5} 
} \\
{
& \pazocal{D}^{\textit{cfr}}(\emptyset, \textit{obl(nonConsensualDataProcessing(Agent0), now)}) \ni \\ & \textit{unfairDataProcessing(Agent0)}
} &
{
\tag{CFRD.1}\label{exDerCFR1} 
} \\
{
& \pazocal{D}^{\textit{cfr}}(\emptyset, \textit{obl(unfairDataProcessing(Agent0), now)}) \ni \\ & \textit{unfairDataProcessing(Agent0)}
} &
{
\tag{CFRD.2}\label{exDerCFR2}
} \\
{
& \pazocal{D}^{\textit{cfr}}(\emptyset, \textit{obl(dataAnonymised(CommSerProv0, Agent0), now))}) \ni \\ & \textit{dataAnonymised(CommSerProv0, Agent0)}
} &
{
\tag{CFRD.3}\label{exDerCFR3}
} \\
{
& \pazocal{D}^{\textit{cfr}}(\{ \neg \textit{dataAnonymised(CommSerProv0, Agent0)} \}, \\ & \textit{obl(storeData(CommSerProv0, Agent0, Data0), now)} \ni \\
& \textit{obl(dataUnprotected(Agent0, CommSerProv0), now))} &
} &
{
\tag{CFRD.4}\label{exDerCFR4}
} \\
{
& \pazocal{D}^{\textit{cfr}}(\emptyset, \textit{obl(dataUnprotected(Agent0, CommSerProv0), now)} \ni \\ & \textit{dataUnprotected(Agent0, CommSerProv0)}
} &
{
\tag{CFRD.5}\label{exDerCFR5}
} \\
{
& \pazocal{D}^{\textit{cfr}}(\emptyset, \{ \textit{obl(storeData(CommSerProv0, Agent0, personal), now)} \}) \ni \\ &  \textit{privacyDisrespected(Agent0)}
} &
{
\tag{CFRD.6}\label{exDerCFR6}
} \\
{
& \pazocal{D}^{\textit{cfr}}(\emptyset, \{ \textit{obl(privacyDisrespected(Agent0), now)} \}) \ni \\ & \textit{privacyDisrespected(Agent0)}
} &
{
\tag{CFRD.7}\label{exDerCFR7}
} \\
{
& \pazocal{D}^{\textit{cfr}}(\emptyset, \textit{obl(storeData(CommSerProv0, Agent0, personal), now)}) \ni \\ & \textit{dataProcessed}
} &
{
\tag{CFRD.8}\label{exDerCFR8}
} \\
{
& \pazocal{D}^{\textit{cfr}}(\emptyset, \textit{obl(dataProcessed, now)}) \ni \textit{dataProcessed}
} &
{
\tag{CFRD.9}\label{exDerCFR9}
} \\
{
& \pazocal{D}^{\textit{cfr}}(\{ \neg \textit{pro(storeDataOutsideEU, never)} \}, \textit{dataProcessed}) \ni \\ & \textit{uncontrolByIndepAuth}
} &
{
\tag{CFRD.10}\label{exDerCFR10}
}
\end{longtable}
\end{spacing}

\begin{spacing}{0.1}
\begin{longtable}[l]
{%
     >{\collectcell\myalign}%
      m{0.05\textwidth}%
     <{\endcollectcell}%
     >{\collectcell\myalign}%
      m{0.15\textwidth}%
     <{\endcollectcell}%
}
{
\Delta^{\textit{cfr}} = \{ & \textit{pro(privacyDisrespected(Agent), never)}, \\ & 
\textit{pro(dataUnprotected(Agent0, personal), never)}, \\ & \textit{pro(unfairDataProcessing(Agent0), never)}, \\ & \textit{pro(uncontrolByIndepAuth, never)}, \textit{is(charles, lawEnforcement)} \}
} &
{
\tag{CFRIS.1}\label{exInSCFR1}
}
\end{longtable}
\end{spacing}

\begin{itemize}
\item[] \textbf{The EU Data Retention Directive} \cite{EUDRD} (EU--DRD) is a second--level institution. It consists of less abstract regulations than the Charter of Fundamental Rights which governs it. On the other hand, it has more abstract regulations than the institution it governs, the UK's implementation of the EU--DRD called the UK's Data Retention Regulations. Whilst both the EU--CFR and the EU--DRD govern regulations, a directive by nature is written specifically for that purpose as a regulation regulating legal institution. Consequently, for the EU--DRD we adopt an explicit form for regulations governing regulations.

 The specific fragments of the Data Retention Directive that we formalise are:
\begin{itemize}
\item Article 5 -- ``\textit{Member states shall ensure that the following categories of data are retained under this Directive:'' (paraphrased) the data necessary to trace and identify the source, destination, date, time and duration, and type of a communication. Also, the users' communication equipment and its location} (rule \ref{exDRDC1}, and \ref{exDRDIS1}) and ``\textit{No data revealing the content of the communication may be retained pursuant to this Directive}'' (\ref{exDRDIS1}).
\item ``\textit{Member States shall ensure that the categories of data specified in Article 5 are retained for periods of not less than six months and not more than two years from the date of the communication}'' (rule \ref{exDRDC2}).
\item Article 8 -- ``\textit{Member States shall ensure that the data specified in Article 5 are retained in accordance with this Directive in such a way that the data retained and any other necessary information relating to such data can be transmitted upon request to the competent authorities without undue delay}'' (rule \ref{exDRDC3}).
\item Article 13 (clause 1) -- ``[...]\textit{sanctions are fully implemented with respect to the processing of data under this Directive.}'' (rule \ref{exDRDC4}).
\end{itemize}
We also take the following legal interpretations as holding in the EU--DRD:
\begin{itemize}
\item Something happening after one month counts--as undue delay (rule \ref{exDRDG1}).
\item Paying a fine counts--as being punished (rule \ref{exDRDG2}).
\item Data retention is ensured for a period of between six and twenty--four months whenever there is a prohibition on deleting the data before 12 months and an obligation to delete it after 13 months (rules \ref{exDRDD1} and \ref{exDRDD2}).
\end{itemize}
\end{itemize}

\begin{spacing}{0.1}
\begin{longtable}[l]
{%
     >{\collectcell\myalign}%
      m{0.05\textwidth}%
     <{\endcollectcell}%
     >{\collectcell\myalign}%
      m{0.1\textwidth}%
     <{\endcollectcell}%
}
\caption[Data Retention Directive formalisation]{EU Data Retention Directive Formalisation}
\label{tabDRDFormalExample}
\endfirsthead
\endhead
{
\pazocal{G}^{\textit{drd}}(\emptyset, \textit{time(m1)}) \ni \textit{undue\_delay}
} &
{
\tag{DRDG.1}\label{exDRDG1}
} \\
{
\pazocal{G}^{\textit{drd}}(\emptyset, \textit{payFine(CommSerProv0, Agent0)}) \ni \textit{punish(CommSerProv0)}
} &
{
\tag{DRDG.2}\label{exDRDG2}
} \\
{
& \pazocal{C}^{\textit{drd}\uparrow}(\emptyset, \textit{useElectronicCommunication(Agent0, CommServProv0)}) \ni \\ &
\textit{obl(obl(storeData(CommServProv0, Agent0, metadata), now), now)}
} &
{
\tag{DRDC.1}\label{exDRDC1}
} \\
{
& \pazocal{C}^{\textit{drd}\uparrow}(\emptyset, \textit{storeData(CommSerProv0, Agent0, Data0})) \ni \\ &
\textit{obl(ensure\_data\_retention\_period(Agent0,} \\ & \; \; \; \; \; \; \; \; \; \; \textit{CommSerProv0, Data0, m6, m24), now)}
} &
{
\tag{DRDC.2}\label{exDRDC2}
} \\
{
& \pazocal{C}^{\textit{drd}\uparrow}(\{ \textit{is(Agent0, lawEnforcement}) \}, \\ & 
\; \; \; \; \textit{requestData(Agent0, CommSerProv0, Agent1})) \ni \\ &
\textit{obl(obl(provideData(CommSerProv0, Agent0, Agent1), undue\_delay), now)}
} &
{
\tag{DRDC.3}\label{exDRDC3}
} \\
{
& \pazocal{C}^{\textit{drd}\uparrow}(\emptyset, \textit{viol(obl(provideData(Co, Ag0, Ag1), undue\_delay)))} \ni \\ &
\textit{obl(obl(punish(CommServProv0), time(m6)), now)}
} &
{
\tag{DRDC.4}\label{exDRDC4}
} \\
{
& \pazocal{D}^{\textit{drd}}(\{ \textit{pro(deleteData(CommSerProv0, Agent0, Data0), time(m12))} \}, \\ & 
\textit{obl(deleteData(CommSerProv0, Agent0, Data0), time(m13))}) \ni \\ &
\textit{ensure\_data\_retention\_period(Agent0, CommSerProv0, Data0, m6, m24)}
} &
{
\tag{DRDD.1}\label{exDRDD1}
} \\
{
& \pazocal{D}^{\textit{drd}}(\{ \textit{obl(deleteData(CommSerProv0, Agent0, Data0), time(m13))} \}, \\ & 
\textit{pro(deleteData(CommSerProv0, Agent0, Data0), time(m12))}) \ni \\ &
\textit{ensure\_data\_retention\_period(Agent0, CommSerProv0, Data0, m6, m24)}
} &
{
\tag{DRDD.2}\label{exDRDD2}
}
\\
{
\Delta^{\textit{drd}} = \{ & \textit{obl(pro(storeData(CommProv0, Agent0, content), never), now)}, \\ & \textit{is(charles, lawEnforcement)} \}
} &
{
\tag{DRDD.2}\label{exDRDIS1}
}
\end{longtable}
\end{spacing}

\begin{itemize}
\item[] \textbf{The UK's Data Retention Regulations} \cite{UK2009} (UKDRR) is a first--level institution which governs communications service providers in the United Kingdom. The fragments of the legislation we formalise are:
\begin{itemize}
\item Article 4 -- ``\textit{It is the duty of a public communications provider to retain the communications data specified in the following provisions of the Schedule to these Regulations}'' obliges metadata is stored and prohibits content data is stored (rule \ref{exDRRC1}, and \ref{exDRRIS1})
\item Article 5 -- ``\textit{The data specified in the Schedule to these Regulations must be retained by the public
communications provider for a period of 12 months from the date of the communication in question}'' (rules \ref{exDRRC2} and \ref{exDRRC3}).
\item Article 8 -- ``\textit{The data retained in pursuance of these Regulations must be retained in such a way that it can be transmitted without undue delay in response to requests}'' (rule \ref{exDRRC4}).
\item We assume the regulations are enforced with fines (e.g. rule \ref{exDRRC4}).
\end{itemize}
\end{itemize}

\begin{spacing}{0.1}
\begin{longtable}[l]
{%
     >{\collectcell\myalign}%
      m{0.05\textwidth}%
     <{\endcollectcell}%
     >{\collectcell\myalign}%
      m{0.08\textwidth}%
     <{\endcollectcell}%
}
\caption[Data Retention Regulations formalisation]{UK Data Retention Regulations Formalisation}
\label{tabDRRFormalExample}
\endfirsthead
\endhead
{
& \pazocal{C}^{\textit{drr} \uparrow}(\emptyset, \textit{useElectronicCommunication(Agent0, CommProv0)}) \ni \\ &
\textit{obl(storeData(CommProv0, Agent0, metadata), now)}
} &
{
\tag{DRRC.1}\label{exDRRC1}
} \\
{
& \pazocal{C}^{\textit{drr} \uparrow}(\emptyset, \textit{storeData(CommProv0, Agent0, metadata)})) \ni \\ &	\textit{pro(deleteData(CommProv0, Agent0, metadata), time(m12))} 
} &
{
\tag{DRRC.2}\label{exDRRC2}
} \\
{
& \pazocal{C}^{\textit{drr} \uparrow}(\emptyset, \textit{storeData(CommProv0, Agent0, metadata)}) \ni \\ & \textit{obl(deleteData(CommProv0, Agent0), time(m13))}
} &
{
\tag{DRRC.3}\label{exDRRC3}
} \\
{
& \pazocal{C}^{\textit{drr} \uparrow}(\{ \textit{is(Agent0, lawEnforcement)} \}, \\ & 
\textit{requestData(Agent0, CommProv0, Agent1, metadata)}) \ni \\ &
 \textit{obl(provideData(Agent0, CommProv0, Agent1, metadata), time(m1))}
} &
{
\tag{DRRC.4}\label{exDRRC4}
} \\
{
& \pazocal{C}^{\textit{drr} \uparrow}(\emptyset, \textit{viol(obl(provideData(CommProv0, Agent0, Agent1), time(Length0)))}) \ni \\ &
\textit{obl(payFine(CommProv0, secrOfState), time(m6))}
} &
{
\tag{DRRC.5}\label{exDRRC5}
} \\
{
\Delta^{\textit{drr}} = \{ & \textit{pro(storeData(CommProv0, Agent0, content), never)}, \\ &
\textit{is(charles, lawEnforcement)} \}
} &
{
\tag{DRRIS.1}\label{exDRRIS1}
}
\end{longtable}
\end{spacing}

\section{Semantics}
\label{secMLGSemantics}

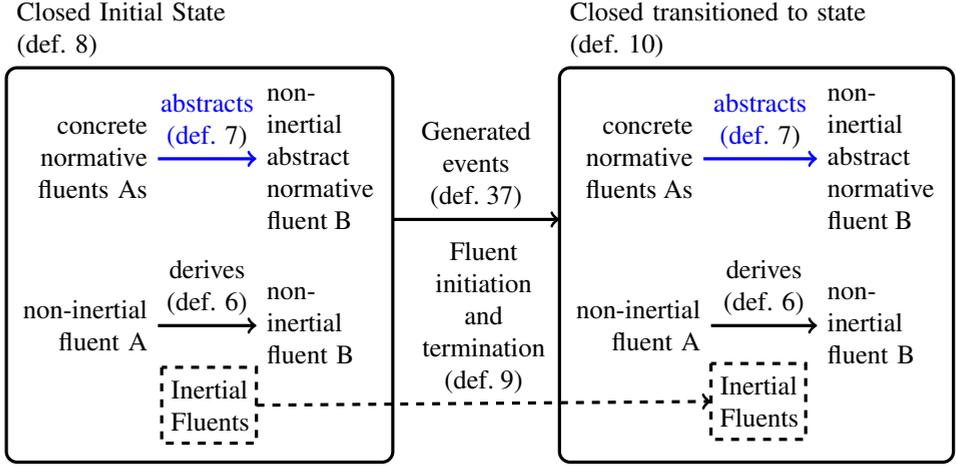
\begin{figure}[t!]
\begin{tikzpicture}
\draw  [very thick, rounded corners] (-5.9,3.7) node [anchor=south west, text width=5cm] (v1) {Closed Initial State \\ (def.~\ref{defInitialStates})} rectangle (-0.8,-1.5);
\draw  [very thick, rounded corners] (1.4,3.7) node [anchor=south west, text width=5cm] {Closed transitioned to state \\ (def.~\ref{defST})} rectangle (6.6,-1.5);
\draw [->, very thick] (-0.8,1.7) -- node [anchor=south,  align=center] {Generated \\ events \\ (def.~\ref{defGR})} (1.4,1.7);

\draw  [very thick, dashed] (-3.85,-0.25) rectangle (-2.6,-1.25);
\node [anchor=west, text width=3cm] at (-3.85,-0.75) {Inertial \\ Fluents};

\draw  [very thick, dashed] (3.4,-0.2) rectangle (4.65,-1.2);
\node [anchor=west, text width=3cm] at (3.4,-0.7) {Inertial \\ Fluents};
\draw [very thick, dashed, ->] (-2.6,-0.75) -- node [anchor=south,  align=center] {Fluent \\ initiation \\ and \\ termination \\ (def.~\ref{defIndFlInTe})}  (3.4,-0.7);

\draw [very thick, ->] (-3.9,2.5) node [anchor=east,  align=right, text width = 1.8cm] {concrete \\ normative \\ fluents As} [color=blue] -- node [align=center, above, text width = 3cm] {abstracts \\ (def.~\ref{defDC})} (-2.6,2.5) node [anchor=west, color=black, align=left] {non- \\inertial \\abstract \\ normative \\ fluent B};
\draw [very thick, ->] (-3.9,0.3) node [anchor=east, align=right, text width=1.8cm] {non-inertial \\ fluent A} -- node [align=center, above, text width = 3cm] {derives \\ (def.~\ref{defFlDer})} (-2.6,0.3) node [anchor=west, text width=4cm] {non- \\inertial \\ fluent B};

\draw [very thick, ->] (3.3,2.5) node [anchor=east, align=right] {concrete \\ normative \\ fluents As} [color=blue] -- node [align=center, above, text width = 3cm] {abstracts \\ (def.~\ref{defDC})} (4.8,2.5) node [anchor=west, color=black, align=left] {non- \\ inertial \\ abstract \\ normative \\ fluent B};
\draw [very thick, ->] (3.4,0.3) node [anchor=east, align=right, text width=1.8cm] {non-inertial \\ fluent A} -- node [align=center, above, text width = 3cm] {derives \\ (def.~\ref{defFlDer})} (4.8,0.3) node [anchor=west, align = left, text width = 4cm] {non- \\ inertial \\ fluent B};

\end{tikzpicture}
\caption[Multi--level governance semantics definitional overview]{An overview of the semantics, depicting the transition from the initial state to the next state and state closure.}
\label{figSemanticsIdea}
\end{figure}

In this section we present the formal semantics for multi--level governance. Given a multi--level governance institution specification the semantics define a \textit{model}, comprising for each institution states transitioned between by events, in response to a supplied trace of observable events. The key idea behind the semantics, depicted in figure~\ref{figSemanticsIdea} is to transition from one state to another, driven by generated events, by initiating and terminating \textit{inertial} fluents. Then each state is closed by deriving \textit{non--inertial} fluents according to an institution's fluent derivation function and abstracting concrete fluents to non--inertial abstract normative fluents according to normative fluent abstraction. Given a multi--level governance institution model it can be determined whether individual institutions are compliant with the institutions that govern them in different contexts. The formal semantics provide a mechanism for automated compliance--checking in multi--level governance.

In order to reduce repetition the following definitions are with respect to several common objects. First, a multi--level governance institution $\pazocal{ML} = \langle \pazocal{T}, R \rangle$ where $\pazocal{T} = \langle \pazocal{I}^{1}, ..., \pazocal{I}^{n} \rangle$ is a tuple of institutions with typical elements being $\forall i \in [1, n] : \pazocal{I}^{i} = \langle \pazocal{E}^{i}, \pazocal{F}^{i}, \pazocal{C}^{i}, \pazocal{G}^{i}, \pazocal{D}^{i}, \Delta^{i} \rangle$. Second, a tuple of states, representing the state of each institution for a single point in time $j$ -- $\langle S^{1}_j, ..., S^{n}_j \rangle$. Third, a tuple of event sets, representing the events occurring in each institution for a single point in time $j$ -- $\langle E^{1}_j, ..., E^{n}_j \rangle$.

\subsubsection{State Conditions}

Institutions in a multi--level governance specification contain rules which are conditional on states and the occurrence of events. Therefore, determining if a rule is `fired' requires determining in part if its state condition, a social context, holds in a state. We begin by defining when contexts are \textit{modelled} by (hold in) a state.  

Informally, a state formula is modelled by a state if for each positive fluent in the formula there is an equivalent fluent that is a member of the state and for each negative fluent in the formula there is not an equivalent fluent that is a member of the state. Rather than defining modelling a state formula in terms of whether the positive/negative fluent is \textit{in} the state, we use equivalence. This is because two normative fluents can have an equivalent meaning whilst being syntactically different -- this is not unusual, in `Standard Deontic Logic' \cite{VonWright1951} forbidden X is defined as obliged not X and likewise for much subsequent work. 

In our case, we define equivalences between two fluents based on whether they are are syntactically identical and two normative fluents based on whether their discharge and violation coincide. Referring again to Figure~\ref{figDischViolOverview}, an event/fluent $a$ is obliged to occur/hold before or at the same time as some $d$ by the obligation fluent $\textit{obl(a, d)}$ and is prohibited to occur/hold strictly before $d$ by the prohibition fluent $\textit{pro}(a, d)$. Given two normative fluents $\textit{obl(a, d)}$ and $\textit{pro}(a^{\prime}, d^{\prime})$ where $a$ is equivalent to $d^{\prime}$ and $d$ is equivalent to $a^{\prime}$, the obligation's and prohibition's discharge and violation coincide, and therefore they are equivalent. The equivalences ($\equiv$) of obligations and prohibitions according to their discharge and violation is summarised as $ \textit{obl}(a, d) \equiv \textit{pro}(a^{\prime}, d^{\prime})$ if  $a \equiv d^{\prime}$ and $d \equiv a^{\prime}$, a definition that generalises to higher--order normative fluents.\footnote{An example of higher--order equivalence generalisation is $\textit{obl}(\textit{obl}(a, d), d^{\prime}) \equiv \textit{obl}(\textit{pro}(d, a), d^{\prime}) \equiv \textit{pro}(d^{\prime}, \textit{obl}(a, d))$, etc.}


Accordingly, we define modelling a state formula as:

\begin{definition}{\textbf{State Formulae}}\label{defStateForm} 
Let $f \in \pazocal{F}^{i}$ be a fluent. We define $\equiv$ and $\models$ for all contexts $X \in \pazocal{X}^{i}$ as:
\begin{equation*}
\begin{array}{r c l r}
f \equiv f \; \; \; & & \; \; \; \\
\textit{obl}(a, d) \equiv \textit{pro}(a^{\prime}, d^{\prime}) \; \; \; & \Leftrightarrow & \; \; \; a \equiv d^{\prime} \textbf{ and } d \equiv a^{\prime} \\
S^{i} \models f \; \; \; & \Leftrightarrow & \; \; \; \exists f^{\prime} \in S^{i} : f \equiv f^{\prime} \\
S^{i} \models \neg f \; \; \; & \Leftrightarrow & \; \; \; \nexists f^{\prime} \in S^{i} : f \equiv f^{\prime} \\
S^{i} \models X \; \; \; & \Leftrightarrow & \; \; \; \forall x \in X : S^{i} \models x
\end{array}
\end{equation*}
\end{definition}

\subsubsection{Events}

In this section we semantically define the events occurring in an institution, in response to other events in specific contexts. Precisely, an event generation operation $\textit{GR}^{i}$ defines for an institution $\pazocal{I}^{i}$ in a multi--level governance institution which events occur in a state $S^{i}$ in response to a set of events $E^{i}$. An event occurs in an institution if it is generated by the institution's event generation function $\pazocal{G}^{i}$, or if it represents the discharge/violation of a discharged/violated normative fluent holding in the institution's state or that of a lower--level institution the institution governs. The event generation operation is formalised below and explained subsequently:

\begin{definition}{\textbf{Event Generation Operation}}\label{defGR} The event generation operation $\textit{GR}^{i} : \Sigma^{i} \times 2^{\pazocal{E}^{i}} \rightarrow 2^{\pazocal{E}^{i}}$ is defined for each institution $\pazocal{I}^{i}$ w.r.t. the tuple of multi--level governance states $\langle S^{1}_j, ..., S^{n}_j \rangle$ and event sets $\langle E^{1}_j, ..., E^{n}_j \rangle$. The operation is defined as $\textit{GR}^{i}(S^{i}, E^{i}) = E^{\prime}$ iff $E^{\prime}$ \textit{minimally} (w.r.t. set inclusion) satisfies all of the following conditions:
\begin{subequations}\\
\allowdisplaybreaks[1]
\begin{align*}
& \textit{now} \in E^{\prime} & \tag{D\ref{defGR}.1}\label{eqGR10}\\
& E^{i} \subseteq E^{\prime} & \tag{D\ref{defGR}.2}\label{eqGR11} \\
& \exists X \in \pazocal{X}^{i}, e \in E^{\prime}, e^{\prime} \in \pazocal{G}^{i}(X, e) : S^{i} \models X \wedge S^{i} \models \textit{pow}(e^{\prime}) \Rightarrow e^{\prime} \in E^{\prime} & \tag{D\ref{defGR}.3}\label{eqGR13} \\
& S^{i} \models \textit{obl}(a, d) \wedge (a \in E^{\prime} \vee S^{i} \models a) \Rightarrow \textit{disch}(\textit{obl}(a, d)) \in E^{\prime} & \tag{D\ref{defGR}.4}\label{eqGR14} \\
& S^{i} \models \textit{obl}(a, d) \wedge (d \in E^{\prime} \vee S^{i} \models d) \wedge \textit{disch}(\textit{obl}(a, d)) \not \in E^{\prime} \Rightarrow \textit{viol(obl(a, d))} \in E^{\prime} & \tag{D\ref{defGR}.5}\label{eqGR15} \\
& S^{i} \models \textit{pro}(a, d) \wedge (d \in E^{\prime} \vee S^{i} \models d) \Rightarrow \textit{disch}(\textit{pro}(a, d)) \in E^{\prime} & \tag{D\ref{defGR}.6}\label{eqGR16} \\
& S^{i} \models \textit{pro}(a, d) \wedge (a \in E^{\prime} \vee S^{i} \models a) \wedge \textit{disch}(\textit{pro}(a, d)) \not \in E^{\prime} \Rightarrow \textit{viol}(\textit{pro}(a, d)) \in E^{\prime} & \tag{D\ref{defGR}.7}\label{eqGR17} \\
& \exists \langle h, i \rangle \in R, e \in \pazocal{E}^{h}_{\textit{norm}} \cap \pazocal{E}^{i}_{\textit{norm}} \Rightarrow e \in E^{\prime} \tag{D\ref{defGR}.8}\label{eqGR18}
\end{align*}
\end{subequations}
\end{definition}

In more detail:
\begin{itemize}
\item \ref{eqGR10} -- the event of $\textit{now}$ always occurs. 
\item \ref{eqGR11} -- events that have already occurred still occur (monotonicity).
\item \ref{eqGR13} -- an event generated by the institution's event generation function in response to another event, conditional on a social context modelled by the state and the event being empowered to occur. 
\item \ref{eqGR14} to \ref{eqGR17} -- a compliance event occurring signifying a normative fluent is discharged or violated in a state, by an obliged/prohibited event, fluent or another normative fluent. Compliance events do not need to be empowered in order to occur.
\item \ref{eqGR18} -- norm compliance events occurring in lower level institutions linked to this institution, also occur in this institution.
\end{itemize}

Note that $\textit{GR}^{i}$ is increasingly monotonic and a well--defined partial function. The function $\textit{GR}^{i}$ is partial if there is a fault in the institutional specification or the set of events passed are inconsistent. Specifically, if an institution is defined such that violating a normative fluent causes an event that discharges the same normative fluent via the event generation function $\pazocal{G}$ (either directly or transitively).


\subsubsection{Derived Fluents}

In this section we semantically define deriving fluents from other fluents in a given state. We define a fluent derivation operation $\textit{FD}^{i}$ which, operating on an institutional state, extends the state to include derived fluents based on fluent derivation rules of the form ``fluent A derives fluent B in context C'' described by the fluent derivation function $\pazocal{D}^{i}$. These derived fluents are the `Bs' from fluent derivation rules where the context `C' holds and the fluent `A' is modelled by the state. By deriving fluents from other fluents in a state, it is possible further fluents should be derived. Thus, the fluent derivation operation $\textit{FD}^{i}$ is defined to close a state by producing an extended state which includes all derived fluents \textit{with respect to} the extended state itself. The fluent derivation operation is formally defined as:

\begin{definition}{\textbf{Fluent Derivation Operation}}\label{defFlDer} The fluent derivation operation $\textit{FD}^{i} : \Sigma^{i} \rightarrow \Sigma^{i}$ is defined for each institution $\pazocal{I}^{i}$ and a state $S^{i} \in \Sigma^{i}$ such that $\textit{FD}^{i}(S^{i}) = S^{\prime}$ iff $S^{\prime}$ \textit{minimally} (w.r.t. set inclusion) satisfies all of the following conditions:
\begin{align*}
& S^{i} \subseteq S^{\prime} \tag{D\ref{defFlDer}.1}\label{eqDepFl1} \\
& \exists X \in \pazocal{X}, f \in S^{\prime}, f^{\prime} \in \pazocal{D}^{i}(X, f) : S^{\prime} \models X \Rightarrow f^{\prime} \in S^{\prime} \tag{D\ref{defFlDer}.2}\label{eqDepFl2}
\end{align*}
\end{definition}

In more detail:

\begin{itemize}
\item \ref{eqDepFl1} -- Closure of the state does not remove any fluents from the input state.
\item \ref{eqDepFl2} -- A fluent derived from another fluent conditional on a social context modelled by the state according to the institution's fluent derivation function is a member of the closed state.
\end{itemize}

Note that the fluent derivation operation is undefined if an institution's fluent derivation function has an output that is inconsistent with its input. For example $\pazocal{D}(\{ \neg B \}, A) \ni B$ or in words ``A counts--as B in the context that B does not hold''. In other cases, the fluent derivation operation is multi--valued if at least two rules defined by the institution's fluent derivation function are mutually inconsistent. For example $\pazocal{D}(\{ \neg \textit{B2} \}, A) \ni \textit{B1}$ and $\pazocal{D}(\{ \neg \textit{B1} \}, A) \ni \textit{B2}$, or in words ``A counts--as B1 in the context that B2 does not hold'' and vice versa ``A counts--as B2 in the context that B1 does not hold''. The occurrence of these issues indicates an institution design problem, which should be resolved by the institution designer.

\subsubsection{Abstracting Normative Fluents}

\begin{figure}[t!]
\begin{tikzpicture}[scale=0.8]

\node[rectangle, thick, draw=black, dashed, right, text width = 1.5cm] at (-27.525,-5.95) {$\textit{obl}(a_0, d)$\\ {\color{blue}\textit{counts-as}} \\ $\textit{obl}(b, d)$};
\node[rectangle, thick, draw=black, dashed, right, text width = 1.5cm] at (-24.9,-5.95) {$\textit{obl}(a_1, d)$\\ {\color{blue}\textit{counts-as}} \\ $\textit{obl}(b, d)$};
\node[right] at (-22.47,-6.125) {\textbf{...}};
\node[rectangle, thick, draw=black, dashed, right, text width = 1.5cm] at (-21.695,-5.925) {$\textit{obl}(a_n, d)$\\ {\color{blue}\textit{counts-as}} \\ $\textit{obl}(b, d)$};

\node[rectangle, thick, draw=black, dashed, right, text width = 1.5cm] at (-18.5,-5.2) {$\textit{pro}(a_0, d)$, \\$\textit{pro}(a_1, d)$, ... and $\textit{pro}(a_n, d)$ \\{\color{blue}\textit{counts-as}} \\$\textit{pro}(b, d)$};

\node[rectangle, thick, draw=black, dashed, right] at (-26.4,-9.705) {$a_0$ {\color{blue}\textit{counts-as}} $b$};
\node[rectangle, thick, draw=black, dashed,right] at (-23.1,-9.705) {$a_1$ {\color{blue}\textit{counts-as}} $b$};
\node[right] at (-20.2,-9.705) {\textbf{...}};
\node[rectangle, thick, draw=black, dashed, right] at (-19.6,-9.705) {$a_n$ {\color{blue}\textit{counts-as}} $b$};

\draw [->, thick, dashed] (-25.2,-9.4) -- (-26.3,-6.9);
\draw [->, thick, dashed] (-21.9,-9.4) -- (-23.9,-6.9);
\draw [->, thick, dashed] (-18.6,-9.4) -- (-20.4,-6.9);

\draw [thick, dashed] (-26.55,-10.305) rectangle (-16.4,-9.105);
\draw [->, thick, dashed] (-17.4,-9.105) -- (-17.4,-6.975); 

\node[right, text width = 3cm] at (-31.8,-6) {Deontological counts-as between normative fluents based on $a_i$ counting as $b$};
\node[right, text width = 4cm] at (-31.8,-10.1) {An event, fluent or normative fluent $a_i$ counts-as an event, fluent or normative fluent $b$ in a state according to event generation, derived fluents and deontological counts-as functions};

\end{tikzpicture}
\caption[Deontological counts--as semantics overview]{Overview for deontological counts--as semantics between concrete and abstract normative fluents, based on counts--as relations between the elements they prescribe holding in a context entailed by a single state.}
\label{figAbstrGeneralIdea}
\end{figure}
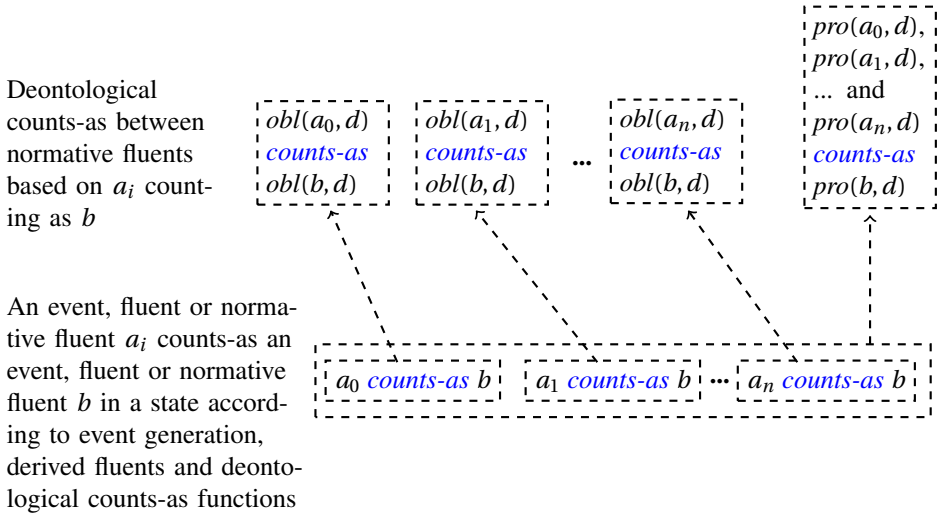

This section presents a semantics for abstracting concrete normative fluents. The basic idea, depicted in Figure~\ref{figAbstrGeneralIdea}, is to establish new counts--as relations between concrete normative fluents and abstract normative fluents, based on the concrete concepts they prescribe counting--as more abstract concepts. In other words, counts--as relations between concrete and abstract normative fluents are based on other counts--as relations between the concrete and abstract concepts they prescribe. Before we go into the actual semantics for abstracting concrete normative fluents, we describe the intuition and general semantics, then give numerous examples and finally the formalisation.

We call the relation between concrete and abstract normative fluents \textit{deontological counts--as} and derive it based on three types of counts--as rules (referring again to figure~\ref{figAbstrGeneralIdea}). Firstly, based on counts--as between events according to an institution's event generation function. Here, we derive relations stating concrete normative fluents about events count--as an abstract normative fluent about an event. Secondly, based on counts--as between fluents according to an institution's fluent derivation function. Here, we derive relations stating concrete normative fluents about fluents count--as an abstract normative fluent about a fluent. Thirdly, based on counts--as between normative fluents themselves according to the deontological counts--as relation we define. Here, we derive relations stating \textit{higher--order} concrete normative fluents prescribing normative fluents count--as a more abstract \textit{higher--order} normative fluent prescribing a normative fluent. So, a \textit{deontological} counts--as relation between concrete and abstract normative fluents is derived from primitive \textit{ontological} counts--as relations according to an institution's event generation and fluent derivation functions, and deontological counts--as itself in order to derive deontological counts--as between higher--order normative fluents.

How the deontological counts--as relations between concrete and abstract normative fluents are derived is described as follows. The intuition is that concrete normative fluents count--as a more abstract normative fluent if and only if complying with the concrete normative fluents (i.e. discharging or not violating) by causing an event to occur or fluent to hold counts--as a certain institutional event or fluent to hold which \textit{guarantees} the abstract normative fluent is also complied with (i.e. discharged or not violated).

Following this intuition, we start by describing deontological counts--as for obligations. In reference to figure~\ref{figAbstrGeneralIdea}, whenever any of $a_0$ to $a_n$ occur or hold we are guaranteed $b$ occurs/holds. If there is a concrete obligation imposed on one of $a_0, ..., a_n$ to occur/hold before a deadline $d$, then it is guaranteed that complying with the concrete obligation (discharging or not violating) means a more abstract obligation for $b$ to hold before the same deadline $d$ is also guaranteed to be discharged or not violated. Therefore, we derive a deontological counts--as relation stating that a concrete obligation on any of $a_0, ..., a_n$ before $d$ \textit{counts--as} a more abstract obligation for $b$ to occur before $d$.

Prohibitions are different. If $a_0, ..., a_n$ count--as $b$, then unlike obligations we cannot apply modus ponens and say that prohibiting $a_0$ before $d$ counts--as prohibiting $b$ before $d$. The reason being, $a_0$ not occurring/holding \textit{does not mean} $b$ does not occur/hold. Thus, prohibiting $a_0$ on its own does not mean $b$ should not occur. In other words. discharging or not violating a prohibition on $a_0$ before $d$ does not guarantee that a prohibition on $b$ before $d$ is discharged or not violated. The reason is $b$ can occur due to any of $a_1,...,a_n$ occurring/holding (all counting--as $b$) and thus violate a prohibition on $b$ before $d$. We might be tempted to apply modus tollens and say that $b$ not occurring/holding means $a_0, ..., a_n$ do not occur/hold, therefore prohibiting $b$ before $d$ counts--as prohibiting $a_0, ..., a_n$ before $d$. However, this would be \textit{concretisation} since $a_0, ..., a_n$ are more concrete than $b$ (recall that concrete concepts count--as abstract concepts, and $a_0, ..., a_n$ count--as $b$). On the other hand, we are interested in \textit{abstraction}. To summarise, unlike obligations modus ponens is inappropriate to base counts--as between prohibitions on, whilst modus tollens is inappropriate since it concretises rather than abstracts.

Instead, we derive a deontological counts--as relation between prohibitions stating that prohibiting all of $a_0, ..., a_n$ from occurring/holding before $d$ counts--as a prohibition on a more abstract event/fluent $b$ occurring/holding before $d$. This is based on the fact that counts--as is \textit{ascriptive} defining all ways an abstract institutional concept can occur/hold when more concrete concepts occur/hold. Since the institutional events/fluents are \textit{ascribed} by an institution's counts--as relations, if none of $a_0, ..., a_n$ occur/hold then $b$ is not ascribed and therefore does not occur/hold. Thus, complying with (discharging or not violating) all prohibitions on $a_0, ..., a_n$ occurring/holding before $d$ guarantees that a prohibition on $b$ before $d$ is also complied with (discharged or not violated).

These informal semantics abstract concrete normative fluents with different concrete aims to an abstract normative fluent with a more abstract aim. Normative fluents' deadlines are also abstracted. However, as we observed when defining equivalences between normative fluents, the aim of an obligation is by definition obliged, whilst the deadline is prohibited and vice versa for prohibitions. Thus, the abstraction of obligation fluents' deadlines should be under the same semantics as prohibitions' aims and vice versa for prohibitions. So, given that $a_0$ counts--as $b$, a prohibition for $z$ to occur before $a_0$ counts--as a prohibition for $z$ to occur before $b$. Alternatively, we can just apply the equivalences between normative fluents such that we have an obligation for $a_0$ to occur before $z$ which counts--as an obligation for $b$ to occur before $z$, which is equivalent to a prohibition for $z$ to occur before $b$. Since a state with a prohibition fluent also \textit{models} an equivalent obligation fluent and vice versa, we define \textit{deontological counts--as} based on the normative fluents a state \textit{models} and obtain the abstraction of normative deadlines `for free'.

This summarises the intuition behind deontological counts--as. More formally, deontological counts--as is defined as the function $\textit{DC}^{i} : \Sigma^{i} \rightarrow 2^{\pazocal{F}^{i}_{\textit{norm}}} \times \pazocal{F}^{i}_{\textit{norm}}$. The function specifies for a state ($S$) a relationship ($\langle N, n \rangle \in \textit{DC}(S^{i})$) between sets of relatively concrete normative fluents ($N$) that count--as an abstract normative fluent ($n$) in the state $S^{i}$. 

We exemplify the deontological counts--as function using our running case study. We focus on the EU--DRD's prescriptions formalised as an institution $\pazocal{I}^{\textit{drd}}$. Counts--as between events according to the EU--DRD's event generation $\pazocal{G}$ function state that a communications provider ($\textit{isp}$) storing the content of a user's ($\textit{ada}$) communications data \\($\textit{storeData(isp, ada, content)}$) counts--as (causes the institutional event of) storing personal data ($\textit{storeData(isp, ada, personal)}$). Likewise, storing communications' metadata \\($\textit{storeData(isp, ada, metadata)}$) counts--as storing personal data.

Storing metadata or content data counts--as storing personal data. Thus, obliging metadata \textit{or} obliging content data is stored immediately counts--as obliging personal data is stored immediately, since if a communications provider stores metadata or content data then it also stores personal data:
\begin{align*}
& \langle\{\textit{obl(storeData(isp, ada, content), now)}\}, \\ & \; \; \; \textit{obl(storeData(isp, ada, personal), now}) \rangle \in \textit{DC}^{i}(S^{i}) \\ &
\langle \{\textit{obl(storeData(isp, ada, metadata), now)}\}, \\ & \; \; \; \textit{obl(storeData(isp, ada, personal), now}) \rangle \in \textit{DC}^{i}(S^{i})
\end{align*}

Prohibiting storing both content and metadata indefinitely counts--as prohibiting storing personal data indefinitely:
\begin{equation*}
\begin{aligned} \langle \{ & \textit{pro(storeData(isp, ada, content), \textit{never})}, \\ & \textit{pro(storeData(isp, ada, metadata), \textit{never})} \}, \\ & \textit{pro(storeData(isp, ada, personal), \textit{never})} \rangle \in \textit{DC}^{i}(S^{i}) \end{aligned}
\end{equation*}
Higher--order normative fluents are abstracted using the same intuitions as first--order normative fluents, but with abstraction based on \textit{deontological counts--as}. According to our case study, obliging an obligation to store content data counts--as obliging an obligation to store personal data. Likewise, obliging an obligation to store metadata counts--as obliging an obligation to store personal data.

\begin{align*}
\begin{aligned} \langle \{ & \textit{obl(obl(storeData(isp, ada, content), \textit{now}), \textit{now})} \}, \\ & \textit{obl(obl(storeData(isp, ada, personal), \textit{now}), \textit{now})} \} \rangle \in& \textit{DC}^{i}(S^{i}) \end{aligned}\\
\begin{aligned} \langle \{ & \textit{obl(obl(storeData(isp, ada, metadata), \textit{now}), \textit{now})} \}, \\ & \textit{obl(obl(storeData(isp, ada, personal), \textit{now}), \textit{now})}\} \rangle \in& \textit{DC}^{i}(S^{i}) \end{aligned} 
\end{align*}

Likewise, but for prohibitions, prohibiting storing metadata and prohibiting storing content data counts--as prohibiting storing personal data. Thus, obliging to immediately prohibit storing metadata \textit{and} obliging to immediately prohibit content data counts--as obliging to immediately prohibit storing personal data. Only obliging to prohibit storing content data, does not mean it is obliged to prohibit storing personal data:
\begin{equation*}
\begin{aligned} \langle & \{\textit{obl(pro(storeData(isp, ada, content), \textit{never}), \textit{now})}, \\ & \; \; \textit{obl(pro(storeData(isp, ada, metadata), \textit{never}), \textit{now})} \}, \\ &
\textit{obl(pro(storeData(isp, ada, personal), \textit{never}), \textit{now})} \rangle \in \textit{DC}^{i}(S^{i}) \end{aligned}
\end{equation*}
Higher--order prohibition abstraction semantics generalises the intuition of deontological counts--as for first--order prohibitions, but based on deontological counts--as itself. Prohibiting \textit{all} concrete normative fluents that count--as a more abstract normative fluent, counts--as prohibiting the more abstract normative fluent.

According to our case study, indefinitely prohibiting obliging storing content data and prohibiting to oblige storing metadata, counts--as indefinitely prohibiting obliging storing personal data. Likewise, for prohibiting prohibitions.
\begin{align*}
& \begin{aligned} & \begin{aligned} \langle \{ & \textit{pro(obl(storeData(isp, ada, content), \textit{now}), \textit{never})}, \\ & \textit{pro(obl(storeData(isp, ada, metadata), \textit{now}), \textit{never})} \}, \end{aligned} \\ & \; \; \;\textit{pro(obl(storeData(isp, ada, personal), \textit{now}), \textit{never})} \} \rangle  \in \textit{DC}^{i}(S^{i}) \end{aligned} \\ 
& \begin{aligned} & \begin{aligned} \langle \{ & \textit{pro(pro(storeData(isp, ada, content), \textit{never}), \textit{never})}, \\ & \textit{pro(obl(storeData(isp, ada, metadata), \textit{never}), \textit{never})} \}, \end{aligned} \\ & \; \; \; \textit{pro(pro(storeData(isp, ada, personal), \textit{never}), \textit{never})} \rangle \in \textit{DC}^{i}(S^{i}) \end{aligned}
\end{align*}

Abstracted normative fluents can also be further abstracted. To give an example, in the EU--DRD the \textit{event} of storing personal data without someone's consent counts--as a non--consensual data processing event. Hence in the context that the agent Ada has not consented ($S \models \neg \textit{consentedDataProcessing(ada)}$) we have the following deontological counts--as relation. It states the EU--DRD is effectively obliging an obligation for data to be processed non--consensually:

\begin{align*}
\begin{aligned} \langle \{ & \textit{obl(obl(storeData(isp, ada, personal), \textit{now}), \textit{now})} \}, \\ & \textit{obl(obl(nonConsensualDataProcessing(ada), \textit{now}), \textit{now})} \} \rangle \in& \textit{DC}^{i}(S^{i}) \end{aligned}
\end{align*}

Deontological counts--as relations are also derived from the fluent derivation function $\pazocal{D}^{i}$. To exemplify, we take the previous example where we have an abstract obligation obliging Ada's data is stored non--consensually. Loosely speaking, the ECJ judged \cite{ECJDRD} that the EU--DRD, by obliging an obligation for non--consensual data processing, violated the EU--CFR's prohibition on \textit{unfair data processing} (e.g. $\textit{pro}(\textit{unfairDataProcessing(ada), never})$). But how do we go from a second--order obligation for data to be processed non--consensually to violating a first--order prohibition on unfair data processing? One possibility is that the EU--CFR's fluent derivation function ($\pazocal{D}^{\textit{cfr}}$) states that obliging non--consensual data processing counts--as unfair data processing, such that \\ $\pazocal{D}^{\textit{cfr}}(\emptyset, obl(\textit{nonConsensualDataProcessing(ada)), \textit{now}})) \ni \textit{unfairDataProcessing(ada)}$. Thus we have the following relation stating the second--order obligation for non--consensual data processing deontologically counts--as, more abstractly, obliging data is processed unfairly:

\begin{align*}
\begin{aligned} \langle \{ & \textit{obl(obl(nonConsensualDataProcessing(ada)), \textit{now}), \textit{now})} \}, \\ & \textit{obl(unfairDataProcessing(ada), \textit{now})} \rangle \in \textit{DC}^{i}(S^{i}) \end{aligned}
\end{align*}

However, obliging data is processed unfairly does not violate the EU--CFR prohibition on unfair data processing, $\textit{pro}(\textit{unfairDataProcessing(ada), never})$. This is unsurprising, the EU--CFR does not impose an explicit second--order prohibition, or contain any explicit higher--order norms for that matter (both in reality and in our formalisation). Unfair data processing is somehow derived from an obligation to oblige non--consensual data processing. One possibility is as follows: \begin{inparaenum}
\item according to the fluent derivation function obliging non--consensual data processing counts--as unfair data processing, thus \item obliging an obligation to process data non--consensually counts--as obliging unfair data processing. \item The EU--CFR considers whether data is processed unfairly or obliged to be processed unfairly as irrelevant, both are viewed as unfair data processing. \item Thus, an obligation to process data unfairly counts--as unfair data processing according to the fluent derivation function, $\pazocal{D}^{\textit{cfr}}(\emptyset, obl(unfairDataProcessing(ada)), \textit{now})) \ni \textit{unfairDataProcessing(ada)}$. \end{inparaenum} That is, normative fluents about abstract concepts are reduced to (ascribed as) those abstract concepts, in this way first--order norms can indirectly govern other norms.

The idea here does not mean what ought to be the case is the case (unfair data processing). Rather, unfair data processing is an abstract concept which has many meanings, including obliging unfair data processing itself. Note that this means not only is an obligation to process data unfairly reduced to unfair data processing, but so is a second--order obligation, a third--order obligation, etcetera. Formally:

\begin{align*}
& \begin{aligned} \langle \{ & \textit{obl(obl(unfairDataProcessing(ada), \textit{now}), \textit{now})} \}, \\ & \textit{unfairDataProcessing(ada)} \rangle \in \textit{DC}^{i}(S^{i}) \end{aligned} \\
& \begin{aligned} \langle \{ & \textit{obl(obl(obl(unfairDataProcessing(ada), \textit{now}), \textit{now}), now)} \}, \\ & \textit{unfairDataProcessing(ada)} \rangle \in \textit{DC}^{i}(S^{i}) \end{aligned}\\
& ...
\end{align*}

Following this discussion, we formally define deontological counts--as, based on counts--as relations which hold in a state according to the event generation function, fluent derivation function and deontological counts--as itself. For convenience, we collect the event generation and fluent derivation counts--as relations into a single set $\pazocal{A}^{i}$ which form the function's \textit{base cases}. Since deontological counts--as is also defined based on its own counts--as relations (in order to generalise to higher--order normative fluents), deontological counts--as is also defined recursively. Formally, deontological counts--as is defined as:

\begin{definition}{\textbf{Deontological Counts--as}}\label{defDC} Given a state $S^{i}$, the deontological counts--as function $\textit{DC}^{i} : \Sigma^{i} \rightarrow 2^{\pazocal{F}^{i}_{\textit{norm}}} \times \pazocal{F}^{i}_{\textit{norm}}$ is defined for the state $S^{i} \in \Sigma^{i}$ such that $\textit{DC}^{i}(S^{i})$ is the minimal (w.r.t. set inclusion) set of all pairs $\langle N^{\prime}, n^{\prime} \rangle$ where $N^{\prime} \neq \emptyset$ that satisfy the following:
\begin{subequations}
\begin{alignat*}{3}
N^{\prime} = \{ \textit{obl}(a, d) \mid \; a \in A \} s.t. \langle A, b \rangle \in \pazocal{A}^{i}(S^{i}) \cup \textit{DC}^{i}(S^{i}) \wedge n^{\prime} = \textit{obl}(b, d) \in \pazocal{F}^{\prime}_{\textit{norm}} && \quad \textbf{or} \tag{D\ref{defDC}.1}\label{eqDC1} \\
N^{\prime} = \{ \textit{pro}(a, d) \mid \; \langle A, b \rangle \in \pazocal{A}^{i}(S^{i}) \cup \textit{DC}^{i}(S^{i}) \wedge a \in A \} \wedge n^{\prime} = \textit{pro}(b, d) \in \pazocal{F}^{i}_{\textit{norm}} && \tag{D\ref{defDC}.2}\label{eqDC2}
\end{alignat*}
\end{subequations}
Where the set of abstracting counts--as relations $\pazocal{A}^{i}(S^{i})$ for the state $S^{i}$ is defined as:
\begin{subequations}
\begin{align*}
\pazocal{A}(S^{i}) = & \{ \langle \{ a \}, b \rangle \mid X \in \pazocal{X}^{i}, a \in \pazocal{E}^{i}, b \in \pazocal{G}^{i}(X, a) \wedge S^{i} \models \textit{pow}(b) \} \cup \tag{D\ref{defDC}.3}\label{eqCA1}  \\
& \{ \langle \{ a \}, b \rangle \mid X \in \pazocal{X}^{i}, a \in \pazocal{F}^{i}, b \in \pazocal{D}^{i}(X, a) \} \tag{D\ref{defDC}.4}\label{eqCA2}
\end{align*}
\end{subequations}
A state closed under deontological counts--as function is the function $\overline{\textit{DC}}^{i} : \Sigma^{i} \rightarrow \Sigma^{i}$, such that $S^{\prime} = \overline{\textit{DC}}^{i}(S^{i})$ iff it \textit{minimally} (w.r.t. set inclusion) satisfies all of the following conditions:
\begin{subequations}
\begin{alignat*}{3}
S^{i} \subseteq S^{\prime}  && \quad & \quad \tag{D\ref{defDC}.6}\label{eqDCCl1} \\
\exists \langle N^{\prime}, n^{\prime} \rangle \in \textit{DC}^{i}(S^{i}) : N^{\prime} \subseteq S^{\prime} \wedge n^{\prime} \in \pazocal{F}^{i}_{\textit{anorm}} \Rightarrow n^{\prime} \in S^{\prime} && \quad  & \quad \tag{D\ref{defDC}.7}\label{eqDCCl2}
\end{alignat*}
\end{subequations}
\end{definition}

In more detail. Concrete obligations count--as a more abstract obligation according to \ref{eqDC1}. Concrete prohibitions count--as a more abstract prohibition according to \ref{eqDC2}. These counts--as relations are derived from relations between concrete concepts counting--as an abstract concept defined by the event generation function and fluent derivation function according to \ref{eqCA1} -- \ref{eqCA2} (the base cases) and with respect to deontological counts--as itself since it is defined recursively.

Deontological counts--as does not describe whether normative fluents in a state $S^{i}$ are abstracted, but rather whether they \textit{could be}. Closing a state under deontological counts--as is according to the operation $\overline{\textit{DC}}^{i}$. Condition \ref{eqDCCl1} ensures any fluents already in the state remain in the state. Condition \ref{eqDCCl2} ensures if concrete normative fluents, should they hold in a state are abstracted to a normative fluent, and they do indeed hold, then the abstracted normative fluent also holds. Note that in \ref{eqDCCl2} it is ensured only normative fluents which belong to the abstract set of normative fluents can hold in a state due to being derived from concrete normative fluents. Consequently, deontological counts--as only adds non--inertial abstract normative fluents to a state.

Note that $\overline{\textit{DC}}^{i}$ is a partial function if there is a fault in the institutional specification. For example, if an institution obliges an event $a$ to occur in some state, and the event $a$ generates the event $b$ in that state, then $b$ is also obliged to occur in that state. However, if $a$ generates the event $b$ \textit{conditional} on $b$ not being obliged then there is a problem. We have that $b$ is obliged since $a$ is obliged. But, if $b$ is obliged then $a$ does not count--as $b$, thus obliged $a$ does not count--as obliged $b$ and so there is no obligation for $b$ to occur. Again, in principle there is nothing wrong with the possibility of this paradox occurring since it is an institutional design fault. If we have $\overline{\textit{DC}}^{i}(S) = \bot$ then we have detected an institutional design problem for the institution designer to rectify.

\subsubsection{Models}

In this section we define a \textit{model} which describes how each institution in a multi--level governance institution evolves from one state to the next, driven by observable events which generate institutional events in state transitions. A model is defined in response to a \textit{trace} of observable (exogenous) events. 

The approach we take is to put together all the previous operations according to the following description. An institution starts at an initial state which includes the institution's initial set of inertial fluents ($\Delta^{i}$), and the state closed under the fluent derivation and deontological counts--as operations. The institution transitions between states with a set of events generated by the event generation operation in response to an observable event in the event trace. Each state transitioned to contains the fluents that held in the previous state that were not terminated, any newly initiated fluents as well as closing the state under the fluent derivation and deontological counts--as operations. Additionally, an institution's evolution is affected by the evolution of other institutions it governs. This means that a higher level institution's state includes normative fluents from lower level institutions it governs. These normative fluents are `passed up' to the higher level institution in order to abstract the lower levels normative fluents and determine if they are compliant in their abstract interpretation.

We begin by defining the initial state of each individual institution. Formally and described subsequently:

\begin{definition}{\textbf{Initial States}}\label{defInitialStates} The initial state $S^{i}_{0}$ for each individual institution $\pazocal{I}^{i}$ w.r.t. $\pazocal{ML} = \langle \pazocal{T}, R \rangle$ and a tuple of initial states $\langle S^{1}_0, ..., S^{n}_0 \rangle$ is the set $S^{i}_{0}$ if and only if $S^{i}_{0}$ \textit{minimally} (w.r.t. set inclusion) satisfies the following:
\begin{align*}
& S^{i}_0 \subseteq \Delta^{i} \tag{D\ref{defInitialStates}.1}\label{eqIS1} \\
& \exists \langle h, i \rangle \in R, n \in (\pazocal{F}^{h}_{\textit{cnorm}} \cup \pazocal{F}^{h}_{\textit{anorm}}) \cap \pazocal{F}^{i}_{\textit{ninert}} : n \in S^{h}_{0} \Rightarrow n \in S^{i}_{0} \tag{D\ref{defInitialStates}.2}\label{eqIS2} \\
& S^{i}_0 = \textit{FD}^{i}(S^{i}_0) \tag{D\ref{defInitialStates}.3}\label{eqIS3} \\
& S^{i}_0 = \overline{\textit{DC}}^{i}(S^{i}_0) \tag{D\ref{defInitialStates}.4}\label{eqIS4}
\end{align*}
\end{definition}

\begin{itemize}
\item \ref{eqIS1} -- an institution's initial set of inertial fluents is included in the institution's initial state.
\item \ref{eqIS2} -- if the institution governs a lower level institution then it contains any normative fluents holding in that lower level institution's initial state.
\item \ref{eqIS3} -- the initial state is closed under the fluent dependency operation, such that all derived fluents are included.
\item \ref{eqIS4} -- the initial state is closed under deontological counts--as such that all abstracted normative fluents are included.
\end{itemize}

Now we define which fluents are initiated and terminated from one state to the next in response to a generated set of events (i.e. by the event generation operation). The set of fluents that are initiated ($\textit{INIT}^{i}$) and terminated ($\textit{TERM}^{i}$) from one state to the next are formally defined as and subsequently described:

\begin{definition}{\textbf{Fluent Initiation and Termination}}\label{defIndFlInTe} The sets of all initiated and terminated fluents for $\pazocal{I}^{i}$ are respectively defined with the functions $\textit{INIT}^{i} : \Sigma^{i} \times 2^{\pazocal{E}^{i}} \rightarrow 2^{\pazocal{F}^{i}}$ and $\textit{TERM}^{i} : \Sigma^{i} \times 2^{\pazocal{E}^{i}} \rightarrow 2^{\pazocal{F}^{i}}$:
\begin{subequations}
\begin{align*}
\textit{INIT}^{i}(S^{i}, E^{i}) = \{ f \mid \; \exists e \in E^{i}, \exists X \in \pazocal{X}^{i} : f \in \pazocal{C}^{i \uparrow}(X, e)\wedge S^{i} \models X \} & \tag{D\ref{defIndFlInTe}.1.1}\label{eqFIT11}
\end{align*}
\begin{align*}
\textit{TERM}^{i}(S^{i}, E^{i}) = \{ f \mid \;  \exists e \in E^{i}, X \in \pazocal{X}^{i} : & S^{i} \models f \wedge f \in \pazocal{C}^{i \downarrow}(X^{i}, e) \wedge S^{i} \models X & \textbf{or} & \tag{D\ref{defIndFlInTe}.2.1}\label{eqFIT21} \\
& S^{i} \models f \wedge (\textit{viol}(f) \in E^{i} \vee \textit{disch}(f) \in E^{i}) \} & \tag{D\ref{defIndFlInTe}.2.2}\label{eqFIT22}
\end{align*}
\end{subequations}
\end{definition}

Condition \ref{eqFIT11} specifies the set of initiated inertial fluents according to the institution's consequence function. An inertial fluent is initiated by the state consequence function conditional on an event occurring and a social context holding in the state. Conversely, \ref{eqFIT21} specifies the set of terminated inertial fluents includes any inertial fluents terminated according to the institution's consequence function. Condition \ref{eqFIT22} states any discharged or violated inertial (concrete) normative fluents are also terminated, meaning discharged/violated normative fluents do not persist by default\footnote{Meaning, if you discharge or violate an obligation you are no longer obliged and likewise for prohibitions. In some cases, it can make sense for a discharged/violated normative fluent to persist. For example, if you violate a prohibition on murder, it is still usually the case that you are still prohibited from committing murder. For an extensive discussion on when it does and does not make sense for obligations and prohibitions to persist after discharge/violation see \cite{Governatori2007}.}.

A state transition operation ($\textit{TR}^{i}(S^{i}, E^{i})$) produces a new institutional state based on the previous state ($S^{i}$) due to the occurrence of events ($E^{i}$). The new state includes any inertial fluents that held in the previous state and have not been terminated, any newly initiated fluents, normative fluents holding, and the state's closure under the fluent derivation and deontological counts--as operations. It is formally defined as and subsequently described:

\begin{definition}{\textbf{State Transitions}}\label{defST} The state transition operation $\textit{TR}^{i} : \Sigma^{i} \times  2^{\pazocal{E}^{i}} \rightarrow \Sigma^{i}$ is defined for each institution $\pazocal{I}^{i}$, a state $S^{i}$ and a set of events $E^{i}$ w.r.t. the states of other institutions $\langle S^{1}_{j}, ..., S^{n}_{j} \rangle$ holding at the same time and $\pazocal{ML} = \langle \pazocal{T}, R \rangle$, such that $\textit{TR}^{i}(E^{i} S^{i}) = S^{\prime}$ iff $S^{\prime}$ \textit{minimally} (w.r.t. set inclusion) satisfies all of the following conditions:
\begin{subequations}
\begin{align*}
& \forall f \in (S^{i} \cap \pazocal{F}^{i}_{\textit{inert}}) \backslash \textit{TERM}^{i}(S^{i}, E^{i}) \Rightarrow f \in S^{\prime} \tag{D\ref{defST}.1}\label{eqST11} \\
& \forall f \in \textit{INIT}^{i}(S^{i}, E^{i}) \Rightarrow f \in S^{\prime} \tag{D\ref{defST}.2}\label{eqST12} \\
& \exists \langle h, i \rangle \in R, n \in (\pazocal{F}^{h}_{\textit{cnorm}} \cup \pazocal{F}^{h}_{\textit{anorm}}) \cap \pazocal{F}^{i}_{\textit{ninert}} : n \in S^{h}_{j} \Rightarrow n \in S^{\prime} \tag{D\ref{defST}.3}\label{eqST13} \\
& S^{\prime} = \textit{FD}^{i}(S^{\prime}) \tag{D\ref{defST}.4}\label{eqST14} \\
& S^{\prime} = \overline{\textit{DC}}^{i}(S^{\prime}) \tag{D\ref{defST}.5}\label{eqST15}
\end{align*}
\end{subequations}
\end{definition}

\begin{itemize}
\item \ref{eqST11} -- non--terminated inertial fluents persist from one state to the next, following the common sense law of inertia. 
\item \ref{eqST12} initiated fluents hold in the next state. 
\item \ref{eqST13} a higher level institution's state contains all normative fluents that hold in the same state of a lower level institution the higher level governs. 
\item \ref{eqST14} the newly transitioned to state includes all normative fluents that can be derived according to the fluent derivation operation. 
\item \ref{eqST15} the newly transitioned state contains all normative fluent abstractions according to deontological counts--as.
\end{itemize}

We now proceed to event traces. The trace a model defined in response to is a sequence of observable events recognised by the institutions involved in a multi--level governance relationship. That is, it is a trace of only those events that can affect the institutions involved, driving their evolution and the multi--level governance institution's evolution as a whole. Each event in a trace needs to be recognised by at least one institution to drive its evolution over time. We call such a trace, a \textit{composite event trace}, formally:

\begin{definition}\textbf{Composite Event Trace}\label{defCompTrace} Let $\pazocal{ML} = \langle \pazocal{T}, \pazocal{B} \rangle$ be a multi--level governance institution where $\pazocal{T} = \langle \pazocal{I}^{1}, ..., \pazocal{I}^{n} \rangle$. $\textit{ctr} = \langle e_0, ..., e_k \rangle$ is a composite trace for $\pazocal{ML}$ iff $\forall j \in [0,k], \exists i \in [n] : \textit{e}_j \in \pazocal{E}^{i}_{\textit{obs}}$
\end{definition}

Synchronisation issues can arise between institutions. These issues occur if a composite trace includes an event recognised by one institution, therefore driving its state forward, but not recognised by another meaning its state does not evolve. If an event in a composite trace is not recognised by an institution, then the institution should still transition to a new state to ensure it is evolving at the same rate as other institutions. We replace unrecognised events by the event of no change, the null event, in a \textit{synchronised trace} for each institution derived from a composite trace. Formally:

\begin{definition}\textbf{Synchronised Trace}\label{defSynchTrace} Let $\pazocal{I}$ be an institution, and $\textit{ctr} = \langle e_0, ..., e_k \rangle$ be a composite event trace. A trace $\textit{str} = \langle \textit{se}_0, ..., \textit{se}_k \rangle$ is a \textit{synchronised trace} of $\textit{ctr}$ for $\pazocal{I}$ iff $\forall i \in [0, k] : $ if $e_k \in \pazocal{E}_{\textit{obs}}$,  $ \textit{se}_k = e_k$ and $\textit{se}_k = e_{\textit{null}}$ otherwise.
\end{definition}

We now define a \textit{multi--level governance institution model}. A model comprises sequences of states ($S$) and events ($E$). One state sequence for each individual institution ($S^{i}$) and one sequence of event sets for each individual institution ($E^{i}$) driving its state transitions. A model is defined in response to a composite trace such that the corresponding synchronised trace for each institution drives its evolution over time, causing events to occur and driving state transitions forward. Each state and set of transitioning events is defined for each institution assuming the states and set of transitioning events exist for every other institution. Formally:

\begin{definition}{\textbf{Multi--level Governance Institution Model}}
\label{defMLGModel} Let $M = \langle M^{1}, ..., M^{n} \rangle$ be a tuple of state and event sequence pairs for each institution $\pazocal{I}^{i}$ with typical element $M^{i} = \langle S^{i}, E^{i} \rangle$ where $S^{i} = \langle S^{i}_{0}, ..., S^{i}_{k+1} \rangle$ and $E^{i} = \langle E^{i}_0, ..., E^{i}_k \rangle$. Let $\textit{ctr}$ be a composite trace for $\pazocal{ML} = \langle \pazocal{T}, R \rangle$ and $\textit{str}^{i} = \langle \textit{se}^{i}_0, ..., \textit{se}^{i}_k \rangle$ be a synchronised trace of $\textit{ctr}$ for each institution $\pazocal{I}^{i}$. Let $\forall i \in [1, n], \forall j \in [0, k] : \textit{GR}^{i}(S^{i}_{j}, E^{i}_{j})$ be the event generation operation for $\pazocal{I}^{i}$ w.r.t. $\langle S^{1}_{j}, ..., S^{n}_{j} \rangle$ and $\langle E^{1}_{j}, ..., E^{n}_{j} \rangle$. Let $\forall i \in [1, n], \forall j \in [0, k] : \textit{TR}^{i}(S^{i}_{j}, E^{i}_{j})$ be the state transition operation for each institution $\pazocal{I}^{i}$  w.r.t. $\langle S^{1}_{j}, ..., S^{n}_{j} \rangle$. The tuple $M$ is a model of $\pazocal{ML}$ w.r.t. $\textit{ctr}$ if and only if:
\begin{subequations}
\begin{align*}
& \forall i \in [1, n] : S^{i}_0 \text{ is the initial state of each institution } \pazocal{I}^{i} \text{ w.r.t. } \langle S^{1}_{0}, ..., S^{n}_{0} \rangle & \tag{D\ref{defMLGModel}.1}\label{eqMLGM2} \\
& \forall i \in [1,n], \forall j \in [0,k] : E^{i}_j = \textit{GR}^{i}(S^{i}_j, \{ \textit{se}^{i}_{j} \}) & \tag{D\ref{defMLGModel}.2}\label{eqMLGM3} \\
& \forall i \in [1,n], \forall j \in [0,k] : S^{i}_{j+1} = \textit{TR}^{i}(S^{i}_j, E^{i}_j) & \tag{D\ref{defMLGModel}.3}\label{eqMLGM4}
\end{align*}
\end{subequations}
\end{definition}

\begin{itemize} 
\item \ref{eqMLGM2} -- the initial state of each individual institution, which is defined with respect to the initial state of every other institution (meaning a higher--level institution includes normative fluents from a lower--level institution).
\item \ref{eqMLGM3} -- each institution's set of events transitioning to a new state comprises all events generated from the corresponding event in the synchronised trace and the previous state according to the event generation operation. The event generation operation is also defined with respect to the states and events from every other institution, such that norm compliance events are `passed up' between governance levels.
\item \ref{eqMLGM4} -- the next state transitioned from the previous state by the set of transitioning events. The state transition operation is also defined with respect to the states and events from every other institution, such that normative fluents are `passed up' between governance levels.
\end{itemize}

This concludes multi--level governance institution semantics.

\subsubsection{Compliance Monitoring}

A multi--level governance institution model monitors the compliance of other institutions' regulations and their outcomes. A model determines if the concrete regulatory effects of one institution are non--compliant with the more abstract regulations of a higher level institution in a particular context. This is by `passing up' any concrete normative fluents from a lower level institution to the higher level institution which governs it. Then, abstracting those concrete normative fluents in the higher level institution according to the higher level institution's abstracting constitutive rules (i.e. under the semantics of deontological counts--as). Then, taking the more abstract interpretation of the lower levels' concrete normative fluents, generating any discharge and violation events of the higher level institution's higher--order norms that oblige/prohibit the abstracted lower level institution's concrete norms. All that is needed to determine if there is non--compliance is to collect a set of violation events from the multi--level governance model for each institution. Formally, the set of sets of violation events for each individual institution denoting non--compliance is:

\begin{definition}{\textbf{Multi--level Governance Violations}}\label{defMLGV} Let $\pazocal{ML} = \langle \pazocal{T}, R \rangle$ be a multi--level governance institution and $M = \langle M^{1}, ..., M^{n} \rangle$ a model of $\pazocal{ML}$ w.r.t. a composite trace $\textit{ctr}$ such that $\forall i \in [n] : M^{i} = \langle S^{i}, E^{i} \rangle, S^{i} = \langle S^{i}_{0}, ..., S^{i}_{k+1} \rangle, E^{i} = \langle E^{i}_0, ..., E^{i}_k \rangle$. The tuple $V = \langle V_1, ..., V_n \rangle$ is the set of multi--level governance violations for $\pazocal{ML}$ w.r.t. $\textit{ctr}$ if and only if:
\begin{align*}
\forall i \in [1,n] : V_{i} = \{ e \mid \; \exists f, j : f \in \pazocal{F}^{i}_{\textit{cnorm}} \cup \pazocal{F}^{i}_{\textit{anorm}}, j \in [k] \wedge \textit{viol}(f) \in E^{i}_{j} \wedge e = \textit{viol}(f) \} \tag{D\ref{defMLGV}.1}\label{eqMLGV1}
\end{align*}
\end{definition}

Non--compliance is found if the set of violation events is not the empty set. For an institution governing a society this implies the society is non--compliant (either in reality if compliance checking is performed before run--time or hypothetically if not). For a higher level institution governing a lower level institution non--compliance denotes the regulatory effects are non--compliant if the violated norms belong to the higher level institution. Such non--compliant regulatory effects can be due to having a more abstract, non--compliant, meaning.

\section{Related Work}
\label{secMLGRelatedWork}

In this chapter we presented a formal framework for reasoning about multi--level governance. Specifically, determining compliance where higher--level institutions which impose abstract regulations governing lower--level institutions which impose concrete regulations. The purpose of this chapter is to contribute a rigorous formalisation, with particular attention paid to the philosophical aspects. Closely related works contribute formalisms for reasoning about hierarchical governance and abstract regulations.

\subsection{Hierarchical Governance}

There appears to be little on hierarchical governance and the regulation of regulations. In \cite{Lopez2003} Lopez and Luck propose a framework for reasoning about norms governing agents, created from a top--down governance perspective. Their framework, based on the Z specification language, gives a precise specification language of a normative system/institution rather than like ours a specification language and operationalisation (semantics). Like our framework, theirs offers similar expressivity with temporal norms, rewards, punishments, etcetera. In particular Lopez and Luck formalize what they call \textit{legislative norms} which are special norms filling the role of, viewing norms as being dynamic and subject to creation, deletion and modification, governing the act of norm changes. This still presents a substantial difference to the method of hierarchical governance and regulation governing regulations we propose, since we use higher--order norms that govern the \textit{outcome} of other norms from which (non--)compliance is determined (typically pre--runtime). Lopez and Luck's legislative norms on the other hand govern the changes to the norms (rules) themselves.

Boella and van der Torre \cite{Boella2003d} offer a conceptual formalisation of \textit{hierarchical} normative systems in Input/Output Logic (a logic aimed at studying conditional norms \cite{Makinson2003}). Like our formalisation, they formalise governance hierarchies, but their formalised concepts have nuanced differences with ours. One of their foci is the role of permissions in hierarchical normative systems. They look at permissions from two perspectives: firstly permissions in static normative systems where authorities do not change norms, secondly permissions in dynamic normative systems where authorities are liable to change norms. In the first case of static normative systems, permissions are issued by higher authorities (e.g. existing in higher level institutions) and act to \textit{derogate} (except) obligations to the contrary (prohibitions) issued by lower level authorities. Conversely, a higher-level authority issuing an obligation is essentially preventing a lower-level authority from issuing a permission that derogates the obligation. In the second case of dynamic normative systems, permissions issued by higher-level authorities act to \textit{enable} lower-level authorities to \textit{change} the normative system in some way, such as permitting a norm to be implemented. The same goes for obligations and prohibitions, a higher-level authority may prohibit a lower-level from implementing a permission that derogates other norms. In contrast, we do not look at derogation, similar concepts would require a kind of defeasibility, which as discussed in the relation to defeasible logic is left for future work. In terms of norms that enable/disable changes to a \textit{dynamic} normative system, we do not look at dynamics of norms in this chapter, but later in Chapter~\ref{chapter_6} we formalise counts-as conditionals that regulate norm change in a \textit{temporal setting}.

Subsequent work by Boella and van der Torre \cite{Boella2006a} formalises higher-order norms in the context of security policies. A security governance hierarchy involves global policy makers obliging/permitting/prohibiting local policy makers to/from obliging/permitting/prohibiting individual agents to/from sharing information and knowledge. In the formalisation agents impose norms by ascribing violations to the actions of other named agents. For example, if agent $A$ prohibits agent $B$ from doing $x$ then agent $A$ ascribes, according to an internal rule, a violation on $B$ for doing $x$. Hence, the general approach is to adopt Anderson's \cite{Anderson1958} reduction of deontic logic to alethic modal logic, where the modalities are replaced by rules with violation as the necessary consequent of prohibited actions/states of affairs. In order to impose higher-order norms, agents ascribe nested violations. So for example agent $C$ prohibiting agent $A$ from prohibiting $B$ to do $x$, is reduced to an ascription by agent $C$ to the action of agent $A$ ascribing a violation to $B$ doing $x$ (a violation violates a norm). The formalisation is in contrast to ours. A higher-order norm according to Boella and van der Torre's reduction to ascribed violations, is violated only when the norm it prescribes is violated. Consequently, if there are no violations of first-order norms then there are no violations of higher-order norms, owing to one focus being on the enforcement of norms which requires violations being ascribed to agents. In contrast, higher-order norms in our formalisation are discharged/violated by the mere imposition of obligations and prohibitions by lower-level institutions, although we show examples where violations themselves are obliged/prohibited, moreover we specifically address the different level of abstractions norms operate at.

Aside from these different approaches and foci, Boella and van der Torre's formalisation in \cite{Boella2006a} provides an argument for von Wright's ``transmission of will'' principle requiring norms of all orders are enforced. As we discussed briefly in the background section (specifically \ref{secBackLegAndInstDesiMLG}), it is shown in formalised scenarios that von Wright's `transmission of will' \cite{Wright1983} does not merely involve an obligation to oblige being fulfilled, but also the lower-order obligation being enforced. Moreover, higher-order norms should be enforced, or else lower authority agents acting in their own self interests may (not) impose certain norms that contravene higher-order norms. Central to demonstrating these two enforcement requirements is the use of violations being ascribed to agents' actions (including the action of ascribing violations). In our formalisation, it is grammatical for higher-order norms to be enforced but in our context of institutions governing institutions higher-order norm enforcement is meaningless. That is, in our formalisation violation occurs as a part of a pre-runtime check (e.g. to check whether national legislation implements an EU directive before a particular deadline) and hence it makes no sense to enforce higher-order norms when violations occur. Rather we assume rewards/punishments for (non-)compliant institution designs are imposed by some kind of external process or logic after a pre-runtime check has determined (non-)compliance of the institution design as a whole. In comparison, since Boella and van der Torre are dealing with agents imposing norms on other agents, enforcement of higher-order norms is meaningful. As for transmission of will requiring obliged lower-order norms to be enforced, we do not have a general formalisation of this requirement. However, we do formalise a specific component the EU Data Retention Directive, which that when obliged obligations are violated then it is obliged there is an obligation for an offender to be punished. In summary, our work agrees with Boella and van der Torre's that both higher and lower-order norms should be enforced, but in the former case we assume an external process handles enforcement and in the latter case we formalise specific scenarios but provide no generalised reasoning for the transmission of will.

\subsection{Abstracting Norms}

As we discussed in the background section, there has already been a reduction of Standard Deontic Logic \cite{VonWright1951} to a logic of counts--as conditionals representing evaluative norms \cite{Anderson1958} (as studied in \cite{Grossi2008,Grossi2011}). For example, `B counts--as a violation in a context C'. Following this idea, Aldewereld et al. \cite{Aldewereld2010a} propose implemented reasoning for concretising abstract norms. This is by representing abstract norms as counts--as statements such as `B counts--as a violation in a context C' and so B is forbidden in C. Then, making use of the fact that more concrete concepts count--as more abstract concepts (e.g. `A counts--as B in context C'). Finally, applying transitivity to concretise abstract norms (e.g. `A counts--as a violation in context C', since A counts--as B and B counts--as a violation). Alderwereld et al. provide a computational approach to the normative reasoning with a rule--based computational language. The same warning against this approach for multi--level governance that we make in the background, applies to what differentiates it from our work. Specifically, by ignoring deontic modalities it is difficult to describe and reason about higher--order norms. Although concretisation of norms is possible, higher--order normative reasoning (regulation governing regulations) is not and neither is the abstraction of higher--order norms.

A description--logic based mechanism for reasoning about abstract institutional concepts is also proposed by Grossi et al. \cite{Grossi2006d}. Unlike in our work, Grossi et al. do not propose abstraction of norms themselves (in fact, they propose concretising concepts), since normative reasoning is not a part of their proposal. Rather, they offer guidance on how normative reasoning can be incorporated, either by the reduction of norms to counts--as, which like us they acknowledge does not support nesting of deontic modalities and therefore higher--order norms. They also offer an alternative path to normative reasoning. Namely, assigning different descriptive concepts a `role' of ideal thereby designating their normativity together with axioms for normative reasoning. The idea being, norms can then be nested and normative statements concretised. However, this part of the proposal is not formalised. Furthermore, our work still differs in that we are interested in abstracting rather than concretising norms in a temporal setting.

In comparison a series of papers by Fornara and
Colombetti \cite{Fornara2010}, Fornara \cite{Fornara2011} and Fornara, Okouya and Colombetti \cite{Fornara2012} combine the semantic--web focussed description logic OWL2DL with normative reasoning. In their proposal obligations are about events with a time--indexed deadline. Time is not integrated within the underlying logic, rather it is reasoned using an external process which adds facts to the knowledge--base (e.g. that an action has occurred, time has passed, etc.). Like our proposal and many others, the deadline of an obligation occurring before the aim triggers violations and potentially causes punishing obligations to be imposed. In comparison to the work of Grossi et al. they do explicitly look at representing and reasoning about norms in description logic but do not aim to reason about the relationship between concrete and abstract concepts or the concretisation/abstraction of norms. The same differences apply when compared to our own work with the additional difference that we do not restrict norms to being about events. Rather, in our proposal normative fluents can be higher--order and about events or other fluents.

Criado et al. \cite{Criado2013a} look at agent reasoning for fulfilling agent desires about abstract institutional concepts. Such desires may come about due to the presence of regulative norms (e.g. an obligation to be married), but their focus is on the concretisation of these abstract institutional concepts (e.g. if an agent wants to get married, what are the brute facts that need to be realised?). In relation to our work, Criado et al. also view counts--as, as providing interpretive primitives in which abstract institutional concepts can be reasoned about. However, they do not explicitly look at how to transform abstract norms into concrete ones, or as we do concrete (higher--order) norms into abstract ones to check compliance. Rather, their focus is on the interpretation of the abstract \textit{concepts} in order to fulfil agents' desires.

To summarise, work proposing ways to reason about abstract and concrete norms or using techniques that can be extended to do so is quite different from our own. Whilst some work does look at the concretisation of abstract norms, there is apparently no work that looks at the abstraction of concrete, potentially \textit{higher--order}, norms. Furthermore, the aforementioned work that explicitly looks at concretisation is not in a temporal setting. Rather, our proposal focuses on the temporal aspects where as the institutional context evolves so does the abstract meaning of concrete norms and thus their compliance with abstract norms at higher governance levels.

\section{Discussion}
\label{secMLGDiscussion}

In this chapter, we answered the question ``\textit{How can we formalise compliance in multi--level governance?}'' with a formal framework. The general answer being that a lower--level institution is compliant in multi--level governance when the effects of its regulations (obligations and prohibitions) are compliant with higher--order regulations at higher--levels of governance. In particular, taking into account the fact that lower--level institutions operate at a different levels of abstraction to higher--levels. Accommodating for this fact, compliance is determined based on the abstract meaning of concrete regulations. We adopted the usual notion of \textit{counts--as} between concrete and abstract institutional concepts. Based on the \textit{counts--as} ontological primitive we semantically defined the abstraction of norms, first and higher--order, with the notion of \textit{deontological counts--as}. Under the semantics, concrete regulations from lower--level institutions are abstracted to the same level of abstraction in higher--level institutions. In their abstract form, the effects of concrete regulations (obligations and prohibitions) are determined for compliance with abstract regulations in higher--level institutions. It is important to note that the abstract meaning of concrete regulations is context--sensitive. Therefore, whether concrete regulations are compliant depends on the contexts they are applied in. To summarise, institutions are compliant in multi--level governance if in different social contexts the abstract meaning of their concrete regulations is compliant with abstract higher--order norms at higher--levels of governance.

The strength of the semantics is in formalising the abstraction of concrete norm effects (obligations and prohibitions), potentially abstracting higher--order concrete norm effects. By paying particular attention to the norm abstraction semantics, with lengthy discussion and the application to our case study, we argued the semantics correctly abstract concrete normative fluents. However, in general the semantics proposed in this section are not without their weaknesses. In particular, as we discussed multiple times, many of the operations are undefined for cases where an inconsistent institution design causes contradiction. Arguably, the fault is with the institution design and not the semantics. On the other hand, inconsistent information is a fact of life. In this chapter, we do not propose an inconsistency tolerant semantics. But, it is important for future work to address this weakness. Particular attention should be paid to resolving inconsistencies arising under the novel semantics provided by deontic abstraction (whereas current work in the literature deals with inconsistency occurring according to a semantics where abstraction is not modelled).
\chapter{Computational Multi--level Governance Compliance Checking}
\label{chapter_4}

\epigraph[0pt]{Science is knowledge which we understand so well that we can teach it to a computer; and if we don't fully understand something, it is an art to deal with it.}{Donald Knuth}

\blfootnote{\color{tck-grey}This chapter is based on the following papers:\\
King, T. C., Li, T., De Vos, M., Dignum, V., Jonker, C. M., Padget, J., \& van Riemsdijk, M. B. (2015, May). Automated Multi--level Governance Compliance Checking. \textit{Journal of Autonomous Agents and Multiagent Systems (JAAMAS)}. International Foundation for Autonomous Agents and Multiagent Systems. (\textbf{In Submission}) \\
Which extends the following paper: \\
\textbf{King, T. C.}, Li, T., De Vos, M., Dignum, V., Jonker, C. M., Padget, J., \& Riemsdijk, M. B. Van. (2015). A Framework for Institutions Governing Institutions. In Proceedings of the 2015 International Conference on Autonomous Agents and Multiagent Systems (AAMAS 2015) (pp. 473–481). Istanbul, Turkey: International Foundation for Autonomous Agents and Multiagent Systems. \cite{King2015a}}

\newpage

This chapter makes the following contributions:

\begin{itemize}
\item A computational framework for representing institutions in multi--level governance and automatically detecting compliance in multi--level governance.
\end{itemize}

The previous chapter gave a formal definition for when institutions are compliant in multi--level governance. Special attention was paid to the increasing abstraction of regulations at higher--levels of governance that govern the relatively concrete regulations of lower--levels of governance. Hence, one objective of this dissertation was achieved. Namely, to formalise the informal notions of compliance in multi--level governance.

In this chapter the focus moves from formalising compliance to automated compliance checking with a computational framework that corresponds to the formal framework. The approach we take is to represent institutions and their semantics as Answer--Set Programming (ASP) programs. By using ASP we can produce models of a multi--level governance institution for a trace of events. Thus we can automatically determine compliance by inspecting the generated models for non--compliant lower--level institutions. This follows the approach of InstAL \cite{Cliffe2006,Cliffe2007}, which provides a representation in Answer--Set Programming for individual institutions governing MAS. However, we need to provide a novel representation for institutions governing other institutions in multi--level governance institutions. Thus, we need to provide a corresponding Answer--Set Program for reasoning about abstract norms governing more concrete norms.

To this end, this chapter contributes the following novel elements for practical reasoning about compliance in multi--level governance:

\begin{itemize}
\item Regulations governing regulations computational reasoning. By taking a formal representation of multi--level governance institutions and producing, through a transformation, a set of answer--set programming rules. These rules capture norms governing norms' representation and reasoning. The previous chapter contributed a representation and semantics for norms governing norms. This chapter contributes the implementation, such that in our computational framework when a norm at a higher--level of governance is violated this indicates that the regulatory effects (imposed obligations and prohibitions) of lower--levels are non--compliant.
\item Norm abstraction reasoning. By taking a formal representation of multi--level governance institutions and producing through a transformation a set of answer--set programming rules. These rules capture the semantics for abstracting concrete norms based on whether the concrete prescribed concepts count--as more abstract concepts.
\item A computational framework based on answer--set programming, including the two aforementioned representations, which \textit{corresponds} to the formal framework. That is, firstly an answer--set program such that for every answer--set program it produces there is an equivalent model in the formal framework for the same multi--level governance institution (soundness). Secondly, an answer--set program such that for every model in the formal framework there is an equivalent answer--set produced by the answer--set program for the same multi--level governance institution (completeness).
\end{itemize}

In addition to these contributions, the following application is contributed:

\begin{itemize}
\item An implementation of the aforementioned transformations as a parser and compiler, extending the InstAL implementation. The implementation provides a high--level representation language for representing multi--level governance institutions and outputs an answer--set programming for checking compliance in those institutions. In this chapter the contributions are applied to an extensive formalisation of EU law for assessing compliance in our running case study.
\end{itemize}

In the rest of this chapter we begin by introducing Answer--Set Programming (ASP) in \ref{secASPIntro}. We describe the overall approach we take to representing multi--level governance institutions in answer--set programming in \ref{secASPApproach}.
We give the representation in ASP in \ref{secMLGCompMapping}. We demonstrate the proposed computational framework we have implemented as a compiler which takes a high--level description of a multi--level governance institution. The output is an ASP program that operationalises a multi--level governance institution. We give an overview of the implementation and results of executing the resulting ASP program for our running case study in \ref{secASPCompiler}. To show that the computational framework provides a practical implementation of the formal framework we provide \textit{soundness} and \textit{completeness} properties between the two frameworks in section~\ref{secMLGCompTheorems}. The properties are proven in appendices. We finish with conclusions in section~\ref{secMLGCompDiscussion}.

\section{Preliminaries -- Answer Set Programming}
\label{secASPIntro}

Answer--Set Programming is a non--monotonic logic--programming language \cite{Baral2003,Gelfond1988}, for declaring problems according to the syntax of AnsProlog as a set of first--order rules. AnsProlog is fully declarative in the sense that the ordering of logical formulae (horn clauses) makes no semantic difference. Executing an AnsProlog program solves the declared problems by first running a grounder which grounds all rules, replacing variables with ground terms, and then running a solver against the ground program. A solver computes the set of \textit{answer--sets}, where each answer--set is a \textit{model} of the AnsProlog program and a solution to the problem declared. Answer--sets are computed according to the stable--model semantics \cite{Gelfond1988}.

We use AnsProlog for two main reasons. Firstly, it provides a natural representation of individual and multi--level governance institutions, where institutions' functions are represented as AnsProlog rules. Secondly, it supports meeting the goal of our framework: automatically checking different contexts, or traces of exogenous events, for whether lower--level institutions are non--compliant. Using AnsProlog, a single trace of events can be supplied to check for compliance, but we can also specify a partial trace and that all variants of that trace must be used to check compliance or even all possible traces up to a specific length must be checked for compliance. It is also possible to declare that each answer--set produced must have a particular property, such as `there must be at least one violation of a norm in a higher level institution'. In this case the property implies that if no answer--sets are produced then there is full compliance for all traces up to a certain length. In summary, Answer--Set Programming provides a natural representation of multi--level governance institutions and an easy way to perform a contextual search for compliance.

There are many answer--set solvers available (e.g. \cite{Eiter1999,Gebser2011}). We briefly reintroduce the main definitions to give context for what follows, focussing on the syntax of the CLINGO \cite{Gebser2011} grounder and solver making use of a number of its unique constructs. In more detail, an AnsProlog program is built from atoms and predicates. Predicates can be ground, such as \verb;lays_eggs(slinky); or non--ground predicates containing variables representing the ground instance schemas, such as \verb;bird(X);. Atoms and predicates can be weakly negated, such as \verb;not;. \footnote{we ignore the case of strong negation since it is unnecessary in our use of AnsProlog.} A rule $r$ is typically of the form \verb;p_0 :- p_1, ..., p_n; comprising a head atom denoted $H(r)$ and a set of body literals denoted $B(r)$, which can be delineated into the positive body atoms $B^{+}(r)$ and atoms appearing negated in the body $B^{--}(r)$. A rule $r$ can also be a fact by having an empty body such that $B(r) = \emptyset$ containing only a single head atom such as \verb;lays_eggs(slinky);. To give an example adapted from \cite{Baral2003}, the following program declares that a bird is an animal that lays eggs which is not a reptile, a reptile is an animal that lays eggs that is not a bird and slinky is an animal that lays eggs:

\begin{verbatim}
bird(X) :- lays_eggs(X), not reptile(X).
reptile(X) :- lays_eggs(X), not bird(X).
lays_eggs(slinky).
\end{verbatim}

A (total) interpretation of an answer--set program is a truth--assignment to literals, comprising a set of the atoms assigned the value of `true'. An answer--set is a \textit{minimal} interpretation containing all atoms that are \textit{justified} in being true. Precisely, for a rule $r$, the head atom denoted $H(r)$ is \textit{justified} in being true if all positive body atoms, denoted $B^{+}(r)$, are true, and none of the weakly negated body atoms, denoted $B^{--}$, are true. This implies facts are always justified in being true (e.g. \verb;lay_eggs(slinky);). Looking at the previous example there can be more than one answer--set. If \verb;bird(slinky); is in an interpretation then \verb;reptile(slinky); cannot be in the interpretation for it to be an answer--set, and vice versa. These answer--sets are:

\begin{itemize}
\item \{ \verb;bird(slinky), lays_eggs(slinky); \}
\item \{ \verb;reptile(slinky), lays_eggs(slinky); \}
\end{itemize}

Determining if an interpretation is an answer--set requires knowing which atoms are justified according to the program's rules. In the presence of weak negation this means we should only consider the rules that do not contain weakly negated atoms that are in the answer--set. Furthermore, for those rules that remain we do not need to consider their weakly negated literals to determine if the head is justified. Removing all rules in a program with weakly negated literals that are in an interpretation and all weakly negated literals from the remaining rules is called the \textit{reduct} of the program, formally from \cite{Gelfond2008}:

\begin{definition}{\textbf{Reduct}} Let $\Pi$ be an Answer--Set Program and $X$ an interpretation of $\Pi$, the reduct denoted $\Pi^{X}$ is the set:
\begin{equation*}
\{ H(r) \leftarrow B^{+}(r) \mid \; r \in \Pi \text{ and } B^{--}(r) \cap X = \emptyset \}
\end{equation*}
\end{definition}

We want to determine for a reduct and a set of atoms, whether that set of atoms is closed under the program (containing all justified atoms) and whether it is minimal (containing no atoms that are not justified). To give an example, if we have a reduct \\$\Pi = \{ \verb; lays_eggs(slinky). bird(slinky) :- lays_eggs(slinky). ; \}$, \\ then the set $\{ \verb;lays_eggs(slinky), bird(slinky), some_atom; \}$ is closed since \\ \verb;lays_eggs(slinky); and \verb;bird(slinky); are justified but it is not minimal due to the presence of \verb;some_atom;. Formally adapted from \cite{Gelfond2008}:

\begin{definition} Let $\Pi$ be a reduct and $X$ a set of atoms. The set of atoms $X$ is closed under $\Pi^{X}$ if for all $r \in \Pi^{X}$, we have $H(r) \in X$ iff $B^{+}(r) \subseteq X$. The smallest set of atoms closed under $\Pi^{X}$ is denoted $\textit{Cn}(\Pi^{X})$.
\end{definition}

An answer--set is simply a minimal interpretation of a reduct of the program for the interpretation:

\begin{definition}{\textbf{Answer--Set}} Let $\Pi$ be an Answer--Set Program and $X$ be an interpretation of $\Pi$ and $\Pi^{X}$ be the reduct of $\Pi$ w.r.t.  $X$. $X$ is an answer--set of $\Pi$ iff $X = \textit{Cn}(\Pi^{X})$.
\end{definition}

In addition to the Answer--Set Programming semantics given above we use three concise ways to represent useful constructs present in CLINGO \cite{Gebser2011}. Namely, \textit{constraints}, \textit{choice rules} and \textit{conditional literals}. Constraints are a special type of rule of the form \asp{:- b_1, ..., b_n.} representing a rule with falsity in the head such that if all of \asp{b_1} to \asp{b_n} are true in an interpretation then there is a contradiction and therefore the interpretation is not an answer--set. Choice rules are of the form \asp{ \{ a_1, ..., a_n \} :- b_1, ..., b_n. }, meaning that any atom in \asp{a_1, ..., a_n} can arbitrarily be picked for inclusion in an answer--set if \asp{b_1, ..., b_n.} is true. Aggregates are present in the body of rules and are of the form \asp{ l \{ b_1; ...; b_n \} u } where \asp{l} and \asp{u} are positive integers meaning that at least \asp{l} and at most \asp{u} elements of \asp{b_1, ..., b_n} must be true for the aggregate to be true. Omitting \asp{l} or \asp{u} removes the respective constraint. Finally, conditional literals can be contained in the body of a rule and are of the form \asp{b_1 : b_2, ..., b_n}. They follow the semantics of material implication; conditional literals are true if the head is true or the body is false. Note that there are no conditional literals in a strict sense, however an aggregate of the form \asp{ l \{ b_1 : b_2, ..., b_n \} u } means that \asp{b_1} is counted as being true when restricted to the domain of \asp{b_2, ..., b_n}. Without variables this simply means that \asp{b_1} is counted as true when \asp{b_2, ..., b_n} is true and hence operates similarly to conditional literals.

\section{Approach}
\label{secASPApproach}

In this section we describe the overall approach we take to reasoning about multi--level governance institutions using answer--set programming. The general idea is to take a formal representation of a multi--level governance institutions. Then, to produce a corresponding representation of that formal description as a set of answer--set programming rules. Moreover, to also provide a corresponding set of answer--set programming rules which capture multi--level governance institution's formal semantics.

The ASP representation for multi--level governance institutions comprises facts representing the signature of each institution's formal representation (events and fluents) and their initial states. Furthermore, the ASP representation comprises rules, which represent each individual institution's rules described by their functions (event generation, consequence and fluent derivation).

Each ASP rule, representing an institution rule, comprises an antecedent and a consequent. The antecedent corresponds to the parameters the functions take. For the event generation and state consequence functions, the corresponding ASP rules' antecedents comprise the occurrence of events and the state conditions. For the fluent derivation function, expressing constitutive rules of the form ``fluent A derives (counts--as) fluent B in context C'' the corresponding ASP rules' antecedents comprise conditions on the state modelling the fluent A and the context C. The consequence of a rule corresponds to the \textit{effect} of the function's returned value on a multi--level governance institution model. This is an event caused to occur according to $\pazocal{G}$, the initiation and termination of fluents according to $\pazocal{C}$, and non--inertial fluents holding in a state according to a fluent derivation function $\pazocal{D}$. 

Multi--level governance institution semantics is represented in ASP as more general rules. For example, stating that if an inertial fluent is initiated then it holds until it is terminated. Finally, composite traces are mapped to a corresponding ASP representation as sets of facts, each stating that an event has been observed at a particular point in time.

The mapping to an ASP program makes use of the same common predicates used previously in work extending InstAL to settings with multiple institutions \cite{Li2014,Li2013b,Li2013}. In turn these bear similarities to constructs used in the Event Calculus \cite{Kowalski1986}. To give context for what follows, the predicates are summarised in their non--ground form:

\begin{itemize}
\item \asp{holdsat(F, In, I)} denotes that the fluent \asp{F} holds in the institution \asp{In} at time \asp{I}.
\item \asp{observed(E, In, I)} denotes that the event \asp{E} is observed by the institution \asp{In} at time \asp{I} corresponding to the exogenous event that has occurred in the synchronised trace for the institution.
\item \asp{occurred(E, In, I)} denotes that the event \asp{E} occurs in the institution \asp{In} at time \asp{I}.
\item \asp{initiated(F, In, I)} denotes that the fluent \asp{F} is initiated in the institution \asp{In} at time \asp{I}.
\item \asp{terminated(F, In, I)} denotes that the fluent \asp{F} is terminated in the institution \asp{In} at time \asp{I}.
\item \asp{instant(I)} denotes \asp{I} is a time instant.
\item \asp{start(I)} denotes \asp{I} is the first time instant.
\item \asp{final(I)} denotes \asp{I} is the last time instant.
\item \asp{next(I, J)} denotes \asp{J} is a time instant that is strictly after \asp{I} such that there is no time instant between \asp{I} and {J}.
\end{itemize}

The aforementioned predicates are used in both antecedents and consequents of rules. Such as, stating \textit{conditional} on particular fluents (not) holding in a state and an event occurring, then an event \textit{occurs}. This means events in function parameters correspond to \asp{occurred/3} whilst state conditions correspond to sets containing positive and negative \asp{holdsat/3} predicates. An empty state condition (the empty set) is always true and replaced with the special atom $\asp{\#true}$.

In more detail, the translation for representing a multi--level governance institution, comprising a set of ASP rules, is split into two parts containing the translation of the individual institutions and the translation of the links between them. For example in the Charter of Fundamental Rights, \verb;cfr;, \verb;exConsent; is an exogenous event, \verb;consent; is an institutional event and \verb;consentedDataProcessing(ada,isp); is an inertial fluent:

\begin{lstlisting}[mathescape]
evtype(exConsent,cfr,ex). 
evtype(consent,cfr,in).
ifluent(consentedDataProcessing(ada,isp),cfr).
\end{lstlisting}

An event generation function is translated to rules. Each rule contains an \verb;occurred/3; atom in the head representing the event that is caused to occur. Each rule's body comprises an \verb;occurred/3; atom representing the \textit{causal event}, and positive and negative \verb;holdsat/3; atoms representing the rule's condition on the state. For example, the following rule states that non--consensual data processing occurs if Ada's personal data has been stored, but she has not consented and non--consensual data processing is empowered to occur:

\begin{lstlisting}[mathescape]
occurred(nonConsensualDataProcessing(ada),cfr,I) :- 
   occurred(storeData(isp,ada,personal),cfr,I),
   holdsat(pow(cfr,nonConsensualDataProcessing(ada)),cfr,I),
   not holdsat(consentedDataProcessing(ada,isp),cfr,I), instant(I).
\end{lstlisting}

A consequence function is translated to ASP rules, using \verb;initiated/3; and \\ \verb;terminated/3; atoms in the head for the initiation and termination of fluents. Each fluent initiation and termination rule's body comprises an \verb;occurred/3; atom representing the event causing a fluent to be initiated/terminated, and positive and negative \verb;holdsat/3; atoms representing the context in which the fluent initiation/termination is conditional on. For example, in the Charter of Fundamental Rights institution the fluent stating that Ada has consented to data processing is initiated if she consents. In the Data Retention Directive institution the obligation to oblige metadata is stored is initiated (i.e. imposed) when Ada uses electronic communications:

\begin{lstlisting}[mathescape]
initiated(consentedDataProcessing(ada,isp),cfr,I) :-
   occurred(consent(ada,isp),cfr,I), instant(I).
initiated(obl(obl(storeData(isp,ada,metadata),now),now),drd,I) :-
   occurred(useElectronicCommunication(ada,isp),drd,I), instant(I).
\end{lstlisting}

The fluent derivation function is represented as ASP rules with \verb;holdsat/3; atoms in the head and body. For example, in the \verb;cfr; institution `unfair data processing' is derived from an obligation to process data non--consensually:

\begin{lstlisting}[mathescape]
holdsat(obl(nonConsensualDataProcessing(ada),now),cfr,I) :-
   holdsat(unfairdataprocessing(ada),cfr,I), instant(I).
\end{lstlisting}

The links between institutions are also represented as rules. Generally, a link from a lower--level institution institution L to a higher--level institution H which governs L comprises rules with occurred and holdsat atoms in the head. The occurred rules state a norm dischargement/violation event occurs in the institution H when it occurs in the institution L. Likewise, a normative fluent holds in H when it holds in L. All of these rules are produced such that only the dischargement/violation events occurring in L and consequently in H are about normative fluents L imposes. Likewise, rules state only normative fluents hold in H when they hold in L for those normative fluents that L itself imposes. Thus, if L receives norm dischargement/violation events or normative fluents from another institution, these do not get passed up to H from L.

To give an example, the following rule states that when the prohibition on deleting data before 12 months holds in the UK's Data Retention Regulations, \verb;drr;, then it also holds in the EU's Data Retention Directive for checking compliance.

\begin{lstlisting}[moredelim={[is][]{@@}{@@}}]
holdsat(pro(deleteData(isp, ada, metadata),@@time@@(m12)),drd,I) :-
   holdsat(pro(deleteData(isp, ada, metadata),@@time@@(m12)),drr,I),
   instant(I).
\end{lstlisting}

Abstraction according to deontological counts--as is also represented as rules, where the head is a holdsat atom representing the abstract normative fluent conditional on some concrete normative fluents holding.

Finally, the semantics are represented as more general rules. For example, the following rules state that an inertial fluent holds in a state if it is initiated or if it held in the previous state and was not terminated (capturing the common--sense law of inertia):

\begin{lstlisting}[mathescape]
holdsat(P,In,J) :- holdsat(P,In,I),
    not terminated(P,In,I), next(I,J), ifluent(P, In).
holdsat(P,In,J) :- initiated(P,In,I), next(I,J), ifluent(P, In).
\end{lstlisting}

\section{Computational Framework}
\label{secMLGCompMapping}

In this section we detail the computational framework in full. We provide translations from multi--level governance institution's representation and semantics in the formal framework to an ASP program. The transformation provides an ASP program which is sound and complete with respect to the formal framework, that is, the transformation produces a \textit{corresponding} ASP program. By executing the ASP program we can determine compliance of lower--level institutions with higher--level institutions in a multi--level governance institution.

\subsection{Multi--level Governance Institution ASP Representation}
\label{secAPPTransMLIRep}
We begin with a convenient translation between a state condition as it appears in an event generation or consequence function and its representation in ASP as a set of positive and negative \verb;holdsat/3; atoms. That is, we translate a state condition $\textit{Exp}$ for an institution $\pazocal{I}$ accordingly:

\begin{definition}{\textbf{Expression Translation}}\label{defASPExprTrans} Let literals be denoted $x_i$ for $i \in \mathbb{N}$. The translation of an expression $\textit{Exp}$ for an individual institution $\pazocal{I}$ with the label $\textit{In}$ is defined as:
\begin{align*}
\textit{EX}(\textit{Exp, } \pazocal{I}) =
\begin{cases}
\hfill \asp{\#true} \hfill & \text{ if } \textit{Exp = } \emptyset \\
\hfill \asp{holdsat(}\textit{p,In}\asp{ , I})  \hfill & \text{ if } \textit{Exp = p} \\
\hfill \asp{not\;} \asp{holdsat(}\textit{p,In}\asp{ , I})  \hfill & \text{ if } \textit{Exp =} \neg \textit{p} \\
            \hfill \textit{EX}(x_0, \pazocal{I}), ..., \textit{EX}(x_n, \pazocal{I}) \hfill & \text{ if } \textit{Exp =} \{ x_0, ..., x_n \} \\
\end{cases}
\end{align*}
\end{definition}

Individual institutions are translated according to a set of rules for translating each individual institution in a multi--level governance institution from their formal representation to a corresponding ASP representation. Formally and subsequently described in more detail:
\begin{definition}{\textbf{Multi--level Governance Institution ASP Translation}}\label{defASPMLTrans} \allowdisplaybreaks[1] Let $\pazocal{ML} = \langle \pazocal{T}, R \rangle$ be a multi--level governance institution s.t. $\pazocal{T} = \langle \pazocal{I}^{1}, ..., \pazocal{I}^{n} \rangle$. The program $\Pi^{\textit{insts}}$ is the multi--level governance institutions program obtained for $\pazocal{ML}$ according to the following translation:
$\forall i \in [1, n]
(\pazocal{I}^{i} = \langle \pazocal{E}^{i}, \pazocal{F}^{i}, \pazocal{C}^{i}, \pazocal{G}^{i}, \pazocal{D}^{i}, \Delta^{i} \rangle) :$
\begin{longtable}[l]
{%
     >{\collectcell\myalign}%
      m{\textwidth}%
     <{\endcollectcell}%
}
\endfirsthead
\endhead
{
\pazocal{I}^{i} \; & \Leftrightarrow && \asp{inst(}\textit{In}^i\asp{).} && \tag{D\ref{defASPMLTrans}.1}\label{eqASPMLTrans1} \\
e \in \pazocal{E}_{\textit{obs}}^{i} \; & \Leftrightarrow && \; \asp{evtype(e,} \textit{In}^i \asp{,ex).} && \tag{D\ref{defASPMLTrans}.2}\label{eqASPMLTrans2} \\[1ex]
e \in \pazocal{E}_{\textit{inst}}^{i} \; & \Leftrightarrow && \asp{evtype(e,} \textit{In}^i \asp{,inst).} && \tag{D\ref{defASPMLTrans}.3}\label{eqASPMLTrans3} \\[1ex]
f \in \pazocal{F}^{i}_{\textit{inert}} \; & \Leftrightarrow && \asp{ifluent(f, }\textit{In}^i\asp{).} && \tag{D\ref{defASPMLTrans}.4}\label{eqASPMLTrans4} \\[1ex]
f \in \pazocal{F}^{i}_{\textit{ninert}} \; & \Leftrightarrow && \asp{nifluent(f, }\textit{In}^i\asp{).} && \tag{D\ref{defASPMLTrans}.5}\label{eqASPMLTrans5}  \\[1ex]
& && \asp{initiated(} p\asp{, } \textit{In}^i \asp{, I) :- } && \\
\exists X \in \pazocal{X}^{i}, e \in \pazocal{E}^{i}, p \in \pazocal{C}^{i\uparrow}(X,e) \; & \Leftrightarrow &&  \; \;\asp{occurred(} e\asp{, } \textit{In}^i \asp{,} I\asp{),} && \\
& && \; \; \asp{not}\; \asp{holdsat(}p, \textit{In}^{i}, \asp{I)}, && \tag{D\ref{defASPMLTrans}.6}\label{eqASPMLTrans6} \\
& && \; \; \textit{EX(X,} \pazocal{I}^i), \asp{instant(I).} && \\[1ex]
& && \asp{terminated(} p\asp{, }\textit{In}^i \asp{, I) :- } && \\
\exists X \in \pazocal{X}^{i}, e \in \pazocal{E}^{i}, p \in \pazocal{C}^{i\downarrow}(X,e) \; & \Leftrightarrow && \; \; \asp{occurred(} e\asp{,} \textit{In}^i\asp{, I),} \textit{EX(X,} \pazocal{I}^i), && \tag{D\ref{defASPMLTrans}.7}\label{eqASPMLTrans7} \\
& && \; \; \asp{holdsat(}f\asp{,} \textit{In}^{i}\asp{, I),} \asp{instant(I).} && \\[1ex]
& && \asp{occurred(} e^{\prime}\asp{, }\textit{In}^i \asp{, I) :- } && \\
\exists X \in \pazocal{X}^{i}, e \in \pazocal{E}^{i}, e^{\prime} \in \pazocal{G}^{i}(X,e) \; & \Leftrightarrow && \; \; \asp{occurred(} e\asp{, }\textit{In}^i\asp{, I),} && \tag{D\ref{defASPMLTrans}.8}\label{eqASPMLTrans9} \\
& && \; \; \asp{holdsat(pow(}e^{\prime}\asp{),} \textit{In}^{i}\asp{, I),} && \\
& && \; \; \textit{EX(X,} \pazocal{I}^i), \\
& && \; \; \asp{instant(I).} && \\[1ex]
\exists X \in \pazocal{X}^{i}, f \in \pazocal{F}^{i}, f^{\prime} \in \pazocal{D}^{i}(X, f) \; & \Leftrightarrow && \asp{holdsat(} f^{\prime}\asp{, }\textit{In}^i \asp{, I) :- } && \tag{D\ref{defASPMLTrans}.9}\label{eqASPMLTrans12} \\
& && \; \; \asp{holdsat(} f\asp{, }\textit{In}^i\asp{, I),} \; \textit{EX(X,} \pazocal{I}^i), && \\
& && \; \; \asp{instant(I).} && \\[1ex]
f \in \Delta^{i} \; & \Leftrightarrow && \asp{holdsat(} f\asp{, } \textit{In}^i, \asp{I)} \asp{:-} \asp{start(I)}. && \tag{D\ref{defASPMLTrans}.10}\label{eqASPMLTrans10} \\[1ex]
\exists \langle h, i \rangle \in R, n \in (\pazocal{F}^{h}_{\textit{anorm}} \cup \pazocal{F}^{h}_{\textit{cnorm}}) \cap \pazocal{F}^{i}_{\textit{ninert}} \; & \Leftrightarrow &&  \asp{holdsat(} n \asp{, }\textit{In}^i \asp{, I) :- } && \tag{D\ref{defASPMLTrans}.11}\label{eqASPMLTrans13}\\
& && \; \; \asp{holdsat(} n\asp{, }\textit{In}^h\asp{, I).}  && \\[1ex]
}
\end{longtable}
\begin{longtable}[l]
{%
     >{\collectcell\myalign}%
      m{0.9\textwidth}%
     <{\endcollectcell}%
}
\endfirsthead
\endhead
{
\exists \langle h, i \rangle \in R, e \in \pazocal{E}^{h}_{\textit{norm}} \cap \pazocal{E}^{i}_{\textit{norm}} \; & \Leftrightarrow && \; \asp{occurred(} e \asp{, }\textit{In}^i \asp{, I) :- } && \tag{D\ref{defASPMLTrans}.12}\label{eqASPMLTrans14} \\
& && \; \; \asp{occurred(} e\asp{, }\textit{In}^h\asp{, I).} && \\[1ex]
\forall f, f^{\prime} \in \pazocal{F}_{\textit{cnorm}} \cup \pazocal{F}_{\textit{anorm}} : f \equiv f^{\prime} \; & \Leftrightarrow && \asp{holdsat(} \textit{f}\asp{, } \textit{In}^i, \asp{I) :-} && \tag{D\ref{defASPMLTrans}.13}\label{eqASPMLTrans11} \\
& && \; \; \asp{holdsat(} \textit{f}^{\prime}\asp{, } \textit{In}^i, \asp{I), } \\
& && \; \; \asp{instant(I).}  && \\
}
\end{longtable}
\end{definition}

The translation is described as follows for an individual institution. The first translation \ref{eqASPMLTrans1} ensures the unique label of the institution is a fact. Translations \ref{eqASPMLTrans2}~to~\ref{eqASPMLTrans5} add facts corresponding to the signature of the institution for representing exogenous events, institutional events, inertial fluents and non--inertial fluents. Translations \ref{eqASPMLTrans6}~and~\ref{eqASPMLTrans7} add ASP rules corresponding to the consequence function, ensuring fluents are initiated and terminated if a condition on a state holds and an event occurs. Translation \ref{eqASPMLTrans9} adds rules for producing events according to the event generation function conditional on a state and another event occurring. Translation \ref{eqASPMLTrans12} adds rules for ensuring non--inertial fluents hold if they hold according to the fluent derivation function. The initial set of inertial fluents is added to the institution's initial state with transformation \ref{eqASPMLTrans10}. If the institution governs a lower--level institution, then any normative fluents which hold in the lower--level institution are `passed up' (if they are non--inertial fluents in the higher--level institution) \ref{eqASPMLTrans13}. Likewise for norm dischargement and violation events \ref{eqASPMLTrans14}. Finally, transformation \ref{eqASPMLTrans11} ensures any normative fluent that is equivalent to another normative fluent that holds in a state, also holds in that state. This concludes the transformations providing specific rules for a multi--level governance institution describing when events occur and fluents are initiated and terminated.

\subsection{Deontological Counts--as ASP Representation}
\label{secAPPDCARepr}

The formal semantics for deontological counts--as, which abstracts concrete norms, is represented as a set of ASP rules for each individual institution. The formal semantics for deontological counts--as is a \textit{general} function defined for any state describing which relatively concrete normative fluents if they hold cause another, more abstract normative fluent to hold. This is based on concepts the concrete fluents prescribe count--as more abstract concepts in a particular state. This is according to the event generation function, fluent derivation function, and counts--as between normative fluents according to deontological counts--as itself. The semantics for deontological counts--as are defined in a general way based on whether concrete concepts \textit{could} cause a more abstract concept to occur/hold in a state. This general definition is based on inspecting the input and output of the functions representing counts--as rules. In other words, the formal semantics are defined by inspecting the left and right hand side of ``A counts--as B in a context C'' rules for whether the context C holds in a state and therefore deducing whether A \textit{could} cause B to occur/hold. 

However, such a general definition is not possible in ASP. This is because counts--as rules are represented as ASP rules. It is not possible to inspect the left and right hand--side of ASP rules and determine whether, for a counts--as rule ``A counts--as B in a context C'', if A \textit{could} cause B. We address the problem of having no foreseeable way to represent the formal deontological counts--as semantics as general ASP rules. That is by translating from individual institutions' formal representation to \textit{specific} rules that provide the same effects as deontological counts--as in a \textit{flattened} form.

Each of these specific rules for deontological counts--as causes a specific abstract normative fluent to hold in the head. They are conditional on specific concrete normative fluents holding in the body depending on the context. Deontological counts--as is context--sensitive, an obligation to store personal data counts--as an obligation to process data non--consensually only if a user has not consented. This means it is only necessary for certain concrete normative fluents to hold in specific contexts for abstraction to take place, obliging personal data is only abstracted if a user has not consented. In different contexts the concrete normative fluents required for abstraction to take place are potentially different or no abstraction takes place. If a user has consented then an obligation to store personal data is not abstracted to non--consensual data processing. So, the body states in which contexts the concrete normative fluents necessarily hold for abstraction to take place and that at least one concrete normative fluent must be eligible for abstraction in order for abstraction to take place.

To give an example for abstracting obliging storing personal data to obliging non--consensual data processing in the Charter of Fundamental Rights interpretation of the Data Retention Directive:

\begin{lstlisting}[mathescape]
holdsat(obl(obl(nonConsensualDataProcessing(ada),now),now), 
        cfr, I) :-
   holdsat(obl(obl(storeData(CommServProv0,Agent0,personal),now),
               now), cfr, I):
   not holdsat(consentedDataProcessing(ada,isp),cfr,I);
   1{holdsat(obl(obl(storeData(isp,ada,personal),now),now),cfr,I):
     not holdsat(consentedDataProcessing(ada,isp),cfr,I)}, 
   holdsat(pow(cfr,nonConsensualDataProcessing(ada)),cfr,I),
   instant(I).
\end{lstlisting}

The head atom is the more abstract normative fluent obliging an obligation for data to be stored non--consensually. In the body, \verb;holdsat(obl(obl(;\\ \verb;storeData(CommServProv0,Agent0,personal),now),now),cfr,I):;\\ \verb;not holdsat(consentedDataProcessing(ada,isp),cfr,I);, represents that \textit{in the context} that Ada has not consented to data processing the concrete obligation to oblige storing personal data must hold in order to be abstracted to obliging an obligation to store data non--consensually. At least one context in which abstraction can take place must hold in order for abstraction to actually take place according to \\ \verb;1{holdsat(obl(obl(storeData; \\ \verb;(isp,ada,personal),now),now),cfr,I):; \\ \verb;not holdsat(consentedDataProcessing(ada,isp),cfr,I)};. In other words, if Ada has consented to data processing, then no abstraction can take place.

Such specific rules to represent deontological counts--as are formally defined as and subsequently described in detail:

\begin{definition}{\textbf{Deontological Counts--as Translation}}\label{defASPDCATrans} Let $\pazocal{ML} = \langle \pazocal{T}, \pazocal{B} \rangle$ be a multi--level governance institution s.t. $\pazocal{T} = \langle \pazocal{I}^{1}, ..., \pazocal{I}^{n} \rangle$. The program $\Pi^{\textit{abstr}}$ is the deontic abstraction program obtained for $\pazocal{ML}$ according to the following translation:
$\forall i \in [1, n]
(\pazocal{I}^{i} = \langle \pazocal{E}^{i}, \pazocal{F}^{i}, \pazocal{C}^{i}, \pazocal{G}^{i}, \pazocal{D}^{i}, \Delta^{i} \rangle) :$
\begin{longtable}[l]
{%
     >{\collectcell\myalign}%
      m{0.985\textwidth}%
     <{\endcollectcell}%
}
\endfirsthead
\endhead
{
\textit{obl}(b, d) \in \pazocal{F}^{i}_{\textit{anorm}}, b \in \pazocal{E}^{i} \; & && \asp{holdsat(obl(}b, d\asp{)}, \textit{In}^i \asp{,I) :- } && \\
a \in \pazocal{E}^{i}, X \in \pazocal{X}, b  \in \pazocal{G}^{i}(X, a) \; & \Leftrightarrow && \; \; \textit{CN}\asp{,} C \asp{,} && \tag{D\ref{defASPDCATrans}.1}\label{eqASPDCATrans1} \\
\textit{CN} = \{ \textit{EX(obl(a,d), } \pazocal{I}^{i}) \asp{:} \textit{EX(X, } \pazocal{I}^{i})\asp{;} \} & && \; \; \asp{holdsat(pow(}b\asp{),} \textit{In}^{i}\asp{, I),} && \\
C = \asp{1\{} \textit{CN} \asp{\}} & && \; \; \asp{instant(I).} && \\[1ex]
\textit{pro}(b, d) \in \pazocal{F}^{i}_{\textit{anorm}}, b \in \pazocal{E}^{i} & && \asp{holdsat(pro(}b, d\asp{)}, \textit{In}^i \asp{,I) :- } && \\
\textit{CN} = \{ \textit{EX(pro(a,d),} \pazocal{I}^{i}) \asp{:} \textit{EX(X, } \pazocal{I}^{i})\asp{;} \mid & \Leftrightarrow && \; \; \textit{CN}\asp{,} \textit{C} \asp{,} && \tag{D\ref{defASPDCATrans}.2}\label{eqASPDCATrans2} \\
a \in \pazocal{F}^{i}, X \in \pazocal{X}, b  \in \pazocal{G}^{i}(X, a) \} & && \; \; \asp{holdsat(pow(}b\asp{),} \textit{In}^{i}\asp{, I),} && \\
\textit{CN} \neq \emptyset, \; \; C = \asp{1\{} \textit{CN} \asp{\}} & && \; \; \asp{instant(I).} \\[1ex]
\textit{obl}(b, d) \in \pazocal{F}^{i}_{\textit{anorm}}, b \in \pazocal{E}^{i} & && \asp{holdsat(obl(}b, d\asp{)}, \textit{In}^i \asp{,I) :- } && \\
a \in \pazocal{F}^{i}, X \in \pazocal{X}, b  \in \pazocal{D}^{i}(X, a) \; & \Leftrightarrow && \; \; \textit{CN}\asp{,} C \asp{,} && \tag{D\ref{defASPDCATrans}.3}\label{eqASPDCATrans5} \\
\textit{CN} = \{ \textit{EX(obl(a,d), } \pazocal{I}^{i}) \asp{:} \textit{EX(X, } \pazocal{I}^{i})\asp{;} \} \; &  && \; \; C = \asp{1\{} \textit{CN} \asp{\}} && \\[1ex]
\textit{pro}(b, d) \in \pazocal{F}^{i}_{\textit{anorm}}, b \in \pazocal{E}^{i} & && \asp{holdsat(pro(}b, d\asp{)}, \textit{In}^i \asp{,I) :- } && \\
\textit{CN} = \{ \textit{EX(pro(a,d),} \pazocal{I}^{i}) \asp{:} \textit{EX(X, } \pazocal{I}^{i})\asp{;} \mid & \Leftrightarrow && \; \; \textit{CN}\asp{,} \textit{C} \asp{,} && \tag{D\ref{defASPDCATrans}.4}\label{eqASPDCATrans6} \\
a \in \pazocal{F}^{i}, X \in \pazocal{X}, b  \in \pazocal{D}^{i}(X, a) \}, & && \; \; \asp{instant(I).} && \\
\textit{CN} \neq \emptyset, \; \; C = \asp{1\{} \textit{CN} \asp{\}} \\[1ex]
\textit{obl}(b, d), b \in \pazocal{F}_{\textit{anorm}}^{i}, & &&  \\[1ex]
r \in \Pi^{\textit{abstr}}(\textit{H}(r) = \asp{holdsat(}b\asp{,} \textit{In}^{i}, \asp{I)}), & &&  && \\
\textit{CN} = \{ \textit{EX(obl(a,d), } \pazocal{I}^{i}) \asp{:} c\asp{;} \mid &  && \asp{holdsat(obl(}b, d\asp{)}, \textit{In}^i \asp{,I) :- } && \tag{D\ref{defASPDCATrans}.5}\label{eqASPDCATrans3} \\
\asp{holdsat(}a, \textit{In}^i \asp{,I) : } c\asp{;} \in \textit{B}(r) \} & \Leftrightarrow && \; \; \textit{CN}, C  \asp{,} \textit{POWS}, && \\
\textit{POWS} = \textit{EX}(\{ \textit{pow(e)} \mid & && \; \; \asp{instant(I).} && \\
\asp{holdsat(pow(}e\asp{),} \textit{In}^{i}\asp{, I)} \in \textit{B}(r) \}, \pazocal{I}^{i}), & && && \\
\textit{CN} \neq \emptyset, \; \; C = \asp{1\{}\textit{CN}\asp{\}} & && && \\[1ex]}
\end{longtable}
\begin{longtable}[l]
{%
     >{\collectcell\myalign}%
      m{0.985\textwidth}%
     <{\endcollectcell}%
}
\endfirsthead
\endhead
{
\textit{pro}(b, d), b \in \pazocal{F}^{i}_{\textit{anorm}}, & &&  && \\
R = \{ r \in \Pi^{\textit{abstr}} : \textit{H}(r) = \asp{holdsat(}b\asp{,} \textit{In}^{i}, \asp{I)} \}, &  && \asp{holdsat(pro(}b, d\asp{)}, \textit{In}^i \asp{,I) :- } && \\
\textit{CN} = \{ \textit{EX(pro(a,d), } \pazocal{I}^{i}) \asp{:} c\asp{;} \mid r \in R, & \Leftrightarrow && \; \; \textit{CN}, C \asp{,} \textit{POWS}, && \tag{D\ref{defASPDCATrans}.6}\label{eqASPDCATrans4} \\
\asp{holdsat(}a, \textit{In}^i \asp{,I) : } c\asp{;} \in B(r) \} & && \; \; \asp{instant(I).}  && \\
\textit{POWS} = \textit{EX}(\{ \textit{pow(e)} \mid & && && \\
\asp{holdsat(pow(}e\asp{),} \textit{In}^{i}\asp{, I)} \in \textit{B}(r) \}, \pazocal{I}^{i}), & && && \\
\textit{CN} \neq \emptyset, \; \;  C = \asp{1\{}\textit{CN}\asp{\} } \in B(r) & && &&
}
\end{longtable}
\end{definition}

In more detail, the first two translations \ref{eqASPDCATrans1} and \ref{eqASPDCATrans2} produce ASP rules for abstracting first--order norms. Abstraction is based on rules of the type ``A counts--as B in context C'' according to the event generation $\pazocal{G}^{i}$ function. The second two translations \ref{eqASPDCATrans5} and \ref{eqASPDCATrans6} are, like the first, for producing first--order norm abstraction but this time based on the fluent derivation function  $\pazocal{D}^{i}$. For all four translations \ref{eqASPDCATrans1} to \ref{eqASPDCATrans6} the resulting rules are of the type ``obliging/prohibiting A(s) counts--as obliging/prohibiting B(s) in context(s) C(s)''. These new abstracting counts--as rules act as base cases for producing further abstracting counts--as rules for higher--order norms by translations \ref{eqASPDCATrans3} and \ref{eqASPDCATrans4}. That is, translations \ref{eqASPDCATrans3} and \ref{eqASPDCATrans4} provide rules which abstract concrete higher--order normative fluents to abstract higher--order normative fluents based on whether the prescribed lower--order normative fluents count--as abstract lower--order normative fluents according to deontological counts--as itself.

To give further examples, the rules for abstracting first--order obligations are provided by the translation according to \ref{eqASPDCATrans1} and \ref{eqASPDCATrans5}. The translation is conditional on there being an event/fluent $a$ counting--as a more abstract event/fluent $b$ in a context $X$. This counts--as relation is according to the event generation and fluent dependency functions. The result is a rule that states if a concrete obligation fluent about an event/fluent $a$ holds in a state and that state entails the context $X$ (in which $a$ counts--as $b$) then a more abstract obligation fluent prescribing $b$ to occur/hold also holds. In the Charter of Fundamental Rights two rules cause an abstract obligation to store personal data to hold if either metadata or content data are obliged to be stored in any context (denoted with $\asp{\#true}$):
\begin{lstlisting}[mathescape]
holdsat(obl(storeData(commServ,user,personal),now), cfr, I) :-
   holdsat(obl(storeData(commServ,user,content),now), cfr, I): 
   #true;
   1{ holdsat(obl(storeData(commServ,user,content),now), cfr, I): 
      #true; }, 
   holdsat(pow(storeData(commServ,user,personal)), cfr, I),
   instant(I).
holdsat(obl(storeData(commServ,user,personal),now), cfr, I) :-
   holdsat(obl(storeData(commServ,user,metadata),now), cfr, I): 
   #true;
   1{ holdsat(obl(storeData(commServ,user,metadata),now), cfr, I):   
      #true; }, 
   holdsat(pow(storeData(commServ,user,personal)), cfr, I),
   instant(I).
\end{lstlisting}

Rules for abstracting first--order prohibitions are provided by the translations according to \ref{eqASPDCATrans2} and \ref{eqASPDCATrans6}. The translation is conditional on there being events/fluents $a$ which in different contexts $X$ count--as an abstract event/fluent $b$. This counts--as relation is according to the event generation and fluent dependency functions. The result is a rule that specifies for events/fluents $a$(s) that count--as an event/fluent $b$ in contexts $X$ entailed by a state, if all the events/fluents $a$ are prohibited (before the same deadline) then a more abstract prohibition on $b$ also holds (with the same deadline). The rule is conditional on at least one context $X$ in which an $a$ counts--as a $b$ holds. The following is an example rule produced in the Charter of Fundamental Rights institution based on storing content data or storing metadata unconditionally counting--as storing personal data:

\begin{lstlisting}[mathescape]
holdsat(pro(storeData(commServ,user,personal),never), cfr, I) :-
   holdsat(pro(storeData(commServ,user,content),never), cfr, I): 
   #true;
   holdsat(pro(storeData(commServ,user,metadata),never), cfr, I): 
   #true;
   1{ holdsat(pro(storeData(commServ,user,content),never), cfr, I):
   #true;
   holdsat(pro(storeData(commServ,user,metadata),never), cfr, I): 
   #true; }, 
   holdsat(pow(storeData(commServ,user,personal)), cfr, I),
   instant(I).
\end{lstlisting}

Higher--order obligation abstraction rules are provided by the translation according to \ref{eqASPDCATrans3}. In more detail, the translation is conditional on there being a rule that states concrete normative fluents $a$(s) count--as an abstract normative fluent $b$ in contexts $X$. The result is a rule that is similar, albeit abstracting higher--order obligations. The new rule states that obliging $a$(s) before the same deadline counts--as an abstract obligation $b$ before the same deadline in contexts $X$, so long as one of the contexts for an $a$ to count--as a $b$ holds. To give an example for the Charter of Fundamental Rights as before, albeit this time a rule that abstracts obligations on prohibitions to store content and metadata to an obligation to prohibit storing personal data:

\begin{lstlisting}[mathescape]
holdsat(obl(pro(storeData(commServ,user,personal),never), now), 
        cfr, I) :-
   holdsat(obl(pro(storeData(commServ,user,content),never), now), 
           cfr, I): #true;
   holdsat(obl(pro(storeData(commServ,user,metadata),never), now), 
           cfr, I): #true;
   1{holdsat(obl(pro(storeData(commServ,user,content),never), now),
              cfr, I): #true;
   holdsat(obl(pro(storeData(commServ,user,metadata),never), now), 
           cfr, I): #true;}, 
   holdsat(pow(storeData(commServ,user,personal)), cfr, I), 
   instant(I).
\end{lstlisting}

Higher--order prohibition abstraction rules are provided by the translation according to \ref{eqASPDCATrans4}. In more detail,
the translation is conditional on there being rules expressing that concrete normative fluents $a$s count--as an abstract normative fluent $b$ in contexts $X$. The result is a rule that states prohibiting normative fluents $a$s before the same deadline in the contexts $X$ counts--as prohibiting normative fluent $b$ before the same deadline, so long as one of the contexts for an $a$ to count--as a $b$ holds. To give an example for the Charter of Fundamental rights, a rule causes an abstract prohibition on prohibiting storing personal data to hold if prohibiting prohibitions to store content and metadata hold (in all contexts, since content and metadata count--as personal data unconditional on a state).

\begin{lstlisting}[mathescape]
holdsat(pro(pro(storeData(commServ,user,personal),never), never), 
        cfr, I) :-
   holdsat(pro(pro(storeData(commServ,user,content),never), never), 
           cfr, I): #true;
   holdsat(pro(pro(storeData(commServ,user,metadata),never), 
           never), cfr, I): #true;
   1{holdsat(pro(pro(storeData(commServ,user,content),never), 
   never), cfr, I): #true;
   holdsat(pro(pro(storeData(commServ,user,metadata),never), 
           never), cfr, I): #true;}, 
       holdsat(pow(storeData(commServ,user,personal)), cfr, I),
       instant(I).
\end{lstlisting}

This concludes our presentation of the ASP representation for deontological counts--as, as a set of specific rules which \textit{flatten} the general formal deontological counts--as semantics.

\subsection{Multi--level Governance Semantics ASP Representation}
\label{secAPPSem}

Now we present the final component of the computational framework, the general semantics of multi--level governance represented as ASP rules. These semantics comprise two ASP programs, $\Pi^{\textit{trace}(n)}$ representing a composite event trace and $\Pi^{\textit{base}(n)}$ containing \textit{general rules} to produce a multi--level governance model for a trace of composite events represented in ASP. We begin with the translation from the formal representation of a composite trace, to a corresponding ASP representation, comprising facts representing an event that is observed at each point in time:

\begin{definition}{\textbf{Composite Trace Translation}}\label{defASPCompTraceTrans} Let $\textit{ctr} = \langle \textit{e}_0, ..., e_k \rangle$ be a composite trace. The program $\Pi^{\textit{trace}(k)}$ is the ASP trace program for $\textit{ctr}$ comprising the following set of rules:
\begin{equation*}
\{ \asp{compObserved(\textit{e}_{\textit{i}}, \textit{i}).} \mid i \in [n] \} \tag{D\ref{defASPCompTraceTrans}.1}\label{eqASPCompTraceTrans}
\end{equation*}
\end{definition}

We now present the ASP program $\Pi^{\textit{base}(k)}$ comprising general rules for producing sequences of states transitioned between by sets of events for a composite trace, that is, a multi--level governance institution model. Formally and subsequently described:

\begin{definition}{\textbf{Multi--level Governance Semantics ASP Translation}}\label{defASPMLSemTrans} The program $\Pi^{\textit{base}(k)}$ is the base multi--level governance ASP program:
\allowdisplaybreaks[1]
\begin{subequations}
\begin{align*}
& \begin{aligned}
\asp{holdsat(P,In,J):-} & \asp{holdsat(P,In,I),not\; terminated(P,In,I),
    next(I,J),} \\ &
    \asp{ifluent(P, In).}
\end{aligned} \tag{D\ref{defASPMLSemTrans}.1}\label{eqASPSemTrans1} \\
&\begin{aligned}
\asp{holdsat(P,In,J):-} & \asp{initiated(P,In,I),
	next(I,J),}
    \asp{ifluent(P, In).}
\end{aligned} \tag{D\ref{defASPMLSemTrans}.2}\label{eqASPSemTrans2} \\
&\begin{aligned}
\asp{occurred(now, In, I) :- instant(I), inst(In).}
\end{aligned} \tag{D\ref{defASPMLSemTrans}.3}\label{eqASPSemTrans3} \\
& \begin{aligned}
\asp{occurred(E,In,I):- evtype(E,In,ex),observed(E,In,I).}
\end{aligned} \tag{D\ref{defASPMLSemTrans}.4}\label{eqASPSemTrans4} \\
& \begin{aligned}
\asp{occurred(null,In,I) :- not\; evtype(E,In,ex), observed(E,In,I).}
\end{aligned} \tag{D\ref{defASPMLSemTrans}.5}\label{eqASPSemTrans5} \\
& \begin{aligned}
\asp{\{compObserved(E, J)\}:-} & \asp{evtype(E,In,ex),instant(J),} \\ & 
\asp{not\; final(J), inst(In).}
\end{aligned} \tag{D\ref{defASPMLSemTrans}.6}\label{eqASPSemTrans6} \\
& \begin{aligned}
\asp{ :- } & \asp{compObserved(E,J),compObserved(F,J),instant(J), evtype(E,InX,ex),} \\ & \asp{evtype(F,InY,ex), E!=F,inst(InX;InY).}
\end{aligned} \tag{D\ref{defASPMLSemTrans}.7}\label{eqASPSemTrans7} \\
& \begin{aligned}
& \asp{obs(I):- compObserved(E,I),evtype(E,In,ex).}
\end{aligned} \tag{D\ref{defASPMLSemTrans}.8}\label{eqASPSemTrans8} \\
& \begin{aligned}
\asp{ :- not\; obs(I), not\; final(I), instant(I).}
\end{aligned} \tag{D\ref{defASPMLSemTrans}.9}\label{eqASPSemTrans9} \\
&\begin{aligned}
\asp{observed(E,In,I) :- compObserved(E,I), inst(In).}
\end{aligned} \tag{D\ref{defASPMLSemTrans}.10}\label{eqASPSemTrans10} \\
& \begin{aligned}
\asp{occurred(disch(obl(A, D)),In,I) :-} & \asp{holdsat(obl(A, D),In,I),} \\
& \asp{1\{occurred(A,In,I),} \\ &
\asp{holdsat(A, In, I)\}.}
\end{aligned} \tag{D\ref{defASPMLSemTrans}.11}\label{eqASPSemTrans11} \\
& \begin{aligned} 
\asp{occurred(viol(obl(A, D)),In,I)} :- &
   \asp{holdsat(obl(A, D),In,I),} \\ &
   \asp{not\; occurred(disch(obl(A, D)), In, I), } \\
   & \asp{1\{occurred(D,In,I), holdsat(D, In, I)\}.}
\end{aligned} \tag{D\ref{defASPMLSemTrans}.12}\label{eqASPSemTrans12} \\
& \begin{aligned}
\asp{occurred(disch(pro(A, D)),In,I) :-} & \asp{
   holdsat(pro(A, D),In,I),} \\ &
   \asp{1\{occurred(D,In,I),} \\ & 
   \asp{holdsat(D, In, I)\}.}
\end{aligned} \tag{D\ref{defASPMLSemTrans}.13}\label{eqASPSemTrans13} \\
&\begin{aligned}
\asp{occurred(viol(pro(A, D)),In,I) :-} &
   \asp{holdsat(pro(A, D)),In,I),} \\ & 
   \asp{1\{occurred(A,In,I), holdsat(A, In, I)\}}, \\ & 
   \asp{not\; occurred(disch(pro(A, D)), In, I).}
\end{aligned} \tag{D\ref{defASPMLSemTrans}.14}\label{eqASPSemTrans14} \\
&\begin{aligned}
\asp{terminated(N,In,I) :-} &
   \asp{1\{occurred(viol(N),In,I),} \\ & 
   \asp{occurred(disch(N),In,I)\}}, \\ &
   \asp{ifluent(N,In)}, \asp{holdsat(N,In,I).}
\end{aligned} \tag{D\ref{defASPMLSemTrans}.15}\label{eqASPSemTrans15} \\
& \begin{aligned}
& \asp{final(\textit{k}).}\;
 \asp{start(0).} \;
 \asp{instant(0..T) :- final(T).} \; \\
 & \asp{next(T, T+1) :-} \asp{instant(T).}
\end{aligned} \tag{D\ref{defASPMLSemTrans}.16}\label{eqASPSemTrans16}
\end{align*}
\end{subequations}
\end{definition}

The common--sense law of inertia is captured with the first rule, \ref{eqASPSemTrans1}. This states an inertial fluent held in the previous state and it has not yet been terminated, then it continues to hold. Initiated fluents hold in the state after they are initiated due to rule \ref{eqASPSemTrans2}. The event of `\textit{now}' occurs in every time instant according to \ref{eqASPSemTrans3}. Collectively, \ref{eqASPSemTrans4} and \ref{eqASPSemTrans5} ensure there is a \textit{synchronised} trace for each institution for a given composite trace. If an event is observed and an institution recognises it as an exogenous event, then that event occurs in the institution according to \ref{eqASPSemTrans4}. If there is no event recognised by an institution that is observed, then the null event occurs in that institution \ref{eqASPSemTrans5}. Rule \ref{eqASPSemTrans6} gives an arbitrary choice to include an observed event in the composite trace, which, together with rules \ref{eqASPSemTrans7} to \ref{eqASPSemTrans10} ensures there is a complete composite trace of length \textit{k} and that no two observable events occur at the same time. Rules \ref{eqASPSemTrans11} to \ref{eqASPSemTrans14} ensure relevant compliance events occur when normative fluents that hold are discharged/violated. Rule \ref{eqASPSemTrans15} ensures discharged and violated normative fluents are terminated. Finally, facts and rules in \ref{eqASPSemTrans16} ensure the model and composite trace is of length \textit{k}, that there are facts representing time instants between zero and \textit{k}, and that the time instants are strictly ordered according to $\asp{next/2}$.

This concludes our presentation of the ASP programs which comprise the computational framework, thus, allowing multi--level governance institutions and their semantics to be executed and consequently compliance checks performed.

\section{Executed Case Study}
\label{secASPCompiler}

We have written the case study in the high--level computational framework specification language. By compiling from the specification language to an AnsProlog representation we are able to assess compliance in our case study's multi--level governance institution. This is by executing the resulting AnsProlog program together with a trace of events.

The case study is instantiated for a domain comprising four types. Firstly the agents acting in the system (\verb;ada; and \verb;charles;). Secondly, we specify various types of role, since we need to distinguish between the agents/organisations and their social status. The case study differentiates between citizens and law enforcement officials as well as Internet Service Providers (ISPs) thus we have the roles \verb;lawEnforcement; and \verb;isp;. Thirdly, we distinguish between different data types (\verb;content;, \verb;metadata; and \verb;personal;).

The case study is run against an observable event trace. We chose an observable event trace which shows the framework's context--sensitivity to abstract norm reasoning. This is by testing the use of electronic communications and ISP's fulfilment of metadata storage obligations in different social contexts. Namely, the context that an agent, Ada, has not consented and the context that she has. The trace is given below:

\begin{lstlisting}[mathescape]
observed(exUseElectronicCommunication(ada, isp), 0).
observed(exStoreData(isp, ada, metadata), 1).
observed(exRequestData(charles, isp, ada), 2).
observed(exConsent(ada, isp), 3).
observed(exUseElectronicCommunication(ada, isp), 4).
observed(exStoreData(isp, ada, metadata), 5).
\end{lstlisting}

First the agent Ada uses electronic communications provided by the service provider ISP. Then, the service provider, ISP, stores Ada's communications metadata. An agent, Charles, requests data from ISP concerning Ada. Ada consents to her data being stored (after it is stored). Ada uses ISP's electronic communications again. Finally, ISP stores Ada's metadata again. In the first half of this observable event trace Ada's data is stored without her consent, in the latter half Ada's data is stored after she has given consent.

\begin{figure}[!t]
	\centering
    \begin{longtable}{@{}l@{}}

\resizebox{1\textwidth}{!}{

\begin{tikzpicture}
[start chain=trace going right,
start chain=state0 going down,
start chain=state1 going down,
start chain=state2 going down,
start chain=state3 going down,
node distance=1cm and 5.2cm
]
{{ [continue chain=trace]
\node[circle,draw,on chain=trace](i0) {$S_{0}$};
\draw[color=white](i0)+(180:5.6cm) --node[above]{}(i0);
\draw(i0)+(-3,0)node[rotate=90,anchor=south]{};
}
{ [continue chain=state0 going below]
\node [on chain=state0,below=of i0,rectangle,draw,inner frame sep=0pt] (s0) {
\begin{minipage}{5cm}\raggedright\everypar={\hangindent=1em\hangafter=1}
\text{1)} \textbf{+} is(\allowbreak{}charles, lawEnforcement): \textbf{drr}, \textbf{drr} \\
\text{2)} \textbf{+} pro(\allowbreak{}dataUnprotected(\allowbreak{}ada, personal), never): \textbf{cfr}\\
\text{3)} \textbf{+} pro(\allowbreak{}privacyDisrespected, never): \textbf{cfr}\\
\text{4)} \textbf{+} pro(\allowbreak{}uncontrolByIndepAuth, never): \textbf{cfr} \\
\text{5)} \textbf{+} pro(\allowbreak{}unfairDataProcessing(\allowbreak{}ada), never): \textbf{cfr}\\
\text{6)} \textbf{-- +} obl(\allowbreak{}pro(\allowbreak{}storeData(\allowbreak{}isp, ada, content), never), now): cfr, \textbf{drd}\\
\text{7)} \textbf{+} pro(\allowbreak{}storeData(\allowbreak{}isp, ada, content), never): cfr, drd, \textbf{drr}\\
\end{minipage}
};
} 
\draw (i0) -- (s0);

}

{{ [continue chain=trace]
\node[circle,draw,on chain=trace](i1) {$S_{1}$};
\draw[-latex,thin](i0) -- 
node[above]{\begin{tabular}{>{\centering}m{6cm}}
\\
 useElectronicCommunication(\allowbreak{}ada, isp): \textbf{cfr}, \textbf{drd}, \textbf{drr} \\
\textbf{\color{white}{\hlg{disch(\allowbreak{}obl(\allowbreak{}pro(\allowbreak{}storeData(\allowbreak{}isp, ada, content), never), now))}}}: cfr, \textbf{drd} \\
\end{tabular}}
(i1);
}
{ [continue chain=state1 going below]
\node [on chain=state1,below=of i1,rectangle,draw,inner frame sep=0pt] (s1) {
\begin{minipage}{5cm}\raggedright\everypar={\hangindent=1em\hangafter=1}
\text{1)} is(\allowbreak{}charles, lawEnforcement): \textbf{drd},   \textbf{drr} \\
\text{2)} \textbf{--} pro(\allowbreak{}dataUnprotected(\allowbreak{}ada, personal), never): \textbf{cfr} \\
\text{3)} \textbf{--} pro(\allowbreak{}privacyDisrespected, never): \textbf{cfr} \\
\text{4)} \textbf{--} pro(\allowbreak{}uncontrolByIndepAuth, never): \textbf{cfr} \\
\text{5)} \textbf{--} pro(\allowbreak{}unfairDataProcessing(\allowbreak{}ada), never): \textbf{cfr} \\
\text{6)} pro(\allowbreak{}storeData(\allowbreak{}isp, ada, content), never): cfr, drd, \textbf{drr} \\
\text{7)} \textbf{-- +} obl(\allowbreak{}storeData(\allowbreak{}isp, ada, metadata), now): cfr, drd, \textbf{drr} \\
\text{8)} \textbf{-- +} obl(\allowbreak{}obl(\allowbreak{}storeData(\allowbreak{}isp, ada, metadata), now), now): cfr, \textbf{drd}  \\
\text{9)} \textbf{-- +} \textbf{\textcolor{blue}{obl(\allowbreak{}obl(\allowbreak{}storeData(\allowbreak{}isp, ada, personal), now), now)}} : \textbf{cfr(from 8)} \\
\text{10)} \textbf{-- +} \textbf{\textcolor{blue}{obl(\allowbreak{}obl(\allowbreak{}nonConsensualDataProcessing(\allowbreak{}ada), now), now)}}: \textbf{cfr (from 9)} \\
\text{11)} \textbf{-- +} \textbf{\textcolor{blue}{obl(\allowbreak{}dataProcessed, now)}} : \textbf{cfr (from 9)} \\
\text{12)} \textbf{-- +} \textbf{\textcolor{blue}{obl(\allowbreak{}dataUnprotected(\allowbreak{}ada, personal), now)}}: \textbf{cfr (from 9)} \\
\text{13)} \textbf{-- +} \textbf{\textcolor{blue}{obl(\allowbreak{}nonConsensualDataProcessing(\allowbreak{}ada), now)}} : \textbf{cfr (from 10)}\\
\text{14)} \textbf{-- +} \textbf{\textcolor{blue}{obl(\allowbreak{}privacyDisrespected, now)}} : \textbf{cfr (from 9)} \\
\text{15)} \textbf{-- +} \textbf{\textcolor{blue}{obl(\allowbreak{}unfairDataProcessing(\allowbreak{}ada), now)}} : \textbf{cfr (from 13)}\\
\text{16)} \textbf{+} \textbf{\textcolor{blue}{dataProcessed}} : \textbf{cfr (from 11)} \\
\text{17)} \textbf{+} \textbf{\textcolor{blue}{dataUnprotected(\allowbreak{}ada, personal)}} : \textbf{cfr (from 12)} \\
\text{18)} \textbf{+} \textbf{\textcolor{blue}{privacyDisrespected}} : \textbf{cfr (from 14)} \\
\text{19)} \textbf{+} \textbf{\textcolor{blue}{uncontrolByIndepAuth}} : \textbf{cfr (from 16)} \\
\text{20)} \textbf{+} \textbf{\textcolor{blue}{unfairDataProcessing(\allowbreak{}ada)} (from 15) } : \textbf{cfr} \\
\end{minipage}
};
} 
\draw (i1) -- (s1);

}

{{ [continue chain=trace]
\node[circle,draw,on chain=trace](i2) {$S_{2}$};
\draw[-latex,thin](i1) -- 
node[above]{\begin{tabular}{>{\centering}m{6cm}}
\\ storeData(\allowbreak{}isp, ada, metadata): \textbf{cfr}, \textbf{drd}, \textbf{drr} \\
 nonConsensualDataProcessing(\allowbreak{}ada): \textbf{cfr} \\
\textbf{\color{white}{\hlg{disch(\allowbreak{}obl(\allowbreak{}obl(\allowbreak{}storeData(\allowbreak{}isp, ada, metadata), now), now))}}}: cfr, \textbf{drd} \\
\textbf{\color{white}{\hlg{disch(\allowbreak{}obl(\allowbreak{}storeData(\allowbreak{}isp, ada, metadata), now))}}}: cfr, drd, \textbf{drr} \\
\textbf{\color{white}{\hlr{viol(\allowbreak{}pro(\allowbreak{}uncontrolByIndepAuth, never))}}}: \textbf{cfr} \\
\textbf{\color{white}{\hlr{viol(\allowbreak{}pro(\allowbreak{}privacyDisrespected, never))}}}: \textbf{cfr} \\
\textbf{\color{white}{\hlr{viol(\allowbreak{}pro(\allowbreak{}dataUnprotected(\allowbreak{}ada, personal), never))}}}: \textbf{cfr} \\
\textbf{\color{white}{\hlr{viol(\allowbreak{}pro(\allowbreak{}unfairDataProcessing(\allowbreak{}ada), never))}}}: \textbf{cfr}
\end{tabular}}
(i2);
}
{ [continue chain=state2 going below]
\node [on chain=state2,below=of i2,rectangle,draw,inner frame sep=0pt] (s2) {
\begin{minipage}{5cm}\raggedright\everypar={\hangindent=1em\hangafter=1}
\text{1)} is(\allowbreak{}charles, lawEnforcement): \textbf{drd}, \textbf{drr} \\
\text{2)} pro(\allowbreak{}storeData(\allowbreak{}isp, ada, content), never): cfr, drd, \textbf{drr} \\
\text{3)} pro(\allowbreak{}storeData(\allowbreak{}isp, charles, content), never): cfr, drd, \textbf{drr} \\
\text{4)} \textbf{+} pro(\allowbreak{}dataUnprotected(\allowbreak{}ada, personal), never): \textbf{cfr} \\
\text{5)} \textbf{+} pro(\allowbreak{}deleteData(\allowbreak{}isp, ada, metadata), time(\allowbreak{}m12)): drd, \textbf{drr} \\
\text{6)} \textbf{+} pro(\allowbreak{}privacyDisrespected, never): cfr\\
\text{7)} \textbf{+} pro(\allowbreak{}uncontrolByIndepAuth, never): \textbf{cfr} \\
\text{8)} \textbf{+} pro(\allowbreak{}unfairDataProcessing(\allowbreak{}ada), never): \textbf{cfr} \\
\text{9)} \textbf{+} pro(\allowbreak{}unfairDataProcessing(\allowbreak{}charles), never): \textbf{cfr} \\
\text{10)} \textbf{+} obl(\allowbreak{}deleteData(\allowbreak{}isp, ada, metadata), time(\allowbreak{}m13)): drd, \textbf{drr} \\
\text{11)} \textbf{+} pro(\allowbreak{}deleteData(\allowbreak{}isp, ada, metadata), time(\allowbreak{}m12)): drd, \textbf{drr} \\
\text{12)} \textbf{-- +} obl(\allowbreak{}ensure\_data\_retention\_period(\allowbreak{}ada, isp, metadata, m6, m24), now) : cfr, \textbf{drd} \\
\textbf{+} \textbf{\textcolor{blue}{ensure\_data\_retention\_period(\allowbreak{}ada, isp, metadata, m6, m24)}} : \textbf{drd from 10 and 11} \\
\end{minipage}
};
} 
\draw (i2) -- (s2);
}

{{ [continue chain=trace]
\node[circle,draw,on chain=trace](i3) {$S_{3}$};
\draw[-latex,thin](i2) -- 
node[above]{\begin{tabular}{>{\centering}m{6cm}}
\\ requestData(\allowbreak{}charles, isp, ada): \textbf{cfr}, \textbf{drd}, \textbf{drr} \\
\textbf{\color{white}{\hlg{disch(obl(ensure\_data\_retention\_period(ada, isp, metadata, m6, m24), now))}}}: cfr, \textbf{drd} \\
\end{tabular}}
(i3);
}
{ [continue chain=state3 going below]
\node [on chain=state3,below=of i3,rectangle,draw,inner frame sep=0pt] (s3) {
\begin{minipage}{5cm}\raggedright\everypar={\hangindent=1em\hangafter=1}
\text{1)} is(\allowbreak{}charles, lawEnforcement): \textbf{drd}, \textbf{drr} \\
\text{2)} obl(\allowbreak{}deleteData(\allowbreak{}isp, ada, metadata), time(\allowbreak{}m13)): drd, \textbf{drr}\\
\text{3)} pro(\allowbreak{}deleteData(\allowbreak{}isp, ada, metadata), time(\allowbreak{}m12)): drd, \textbf{drr} \\
\text{4)} pro(\allowbreak{}dataUnprotected(\allowbreak{}ada, personal), never): cfr\\
\text{5)} pro(\allowbreak{}privacyDisrespected, never): cfr \\
\text{6)} pro(\allowbreak{}storeData(\allowbreak{}isp, ada, content), never): cfr, drd, \textbf{drr} \\
\text{7)} pro(\allowbreak{}uncontrolByIndepAuth, never): cfr \\
\text{8)} pro(\allowbreak{}unfairDataProcessing(\allowbreak{}ada), never): cfr \\
\text{9)} \textbf{\textcolor{blue}{ensure\_data\_retention\_period(\allowbreak{}ada, isp, metadata, m6, m24)}} : \textbf{drd from 2 and 3} \\
\text{10)} \textbf{+} obl(\allowbreak{}provideData(\allowbreak{}isp, charles, ada), time(\allowbreak{}m1)): drd, \textbf{drr} \\
\text{11)} \textbf{+} \textbf{\textcolor{blue}{obl(\allowbreak{}provideData(\allowbreak{}isp, charles, ada), undue\_delay)}} : \textbf{drd from 10} \\
\text{12)} \textbf{-- +} obl(\allowbreak{}obl(\allowbreak{}provideData(\allowbreak{}isp, charles, ada), undue\_delay), now) : drd \\
\end{minipage}
};
} 
\draw (i3) -- (s3);

}

{ [continue chain=trace]
\node[on chain=trace] (i4) {};
\draw[-latex,dashed](i3) --
node[above]{}(i4);
}

\end{tikzpicture}
}
\end{longtable}
\caption[Multi--level governance case study execution]{Case study execution. The originating institutions for a fluent are in \textbf{bold}, `\textbf{+}' indicates an initiated fluent, `\textbf{--}' indicates a terminated fluent. Non--inertial fluents are in \textcolor{blue}{\textbf{bold}} denoting they are derived from other fluents according to the fluent derivation and deontological counts--as (norm abstraction) operations. Norm {\textbf{\color{white}\hlg{discharge}}} and {\textbf{\color{white}\hlr{violation}}} events are highlighted.}
\label{figExampleTrace}
\end{figure}

\begin{figure}[t!]
    \ContinuedFloat
    \captionsetup{list=off,format=cont}
    \centering
    \begin{longtable}{@{}l@{}}


\resizebox{1\textwidth}{!}{

\begin{tikzpicture}
[
start chain=trace going right,
start chain=state4 going down,
start chain=state5 going down,
start chain=state6 going down,
start chain=state7 going down,
node distance=1cm and 5.2cm
]
{{ [continue chain=trace]
\node[circle,draw,on chain=trace](i4) {$S_{4}$};
\draw[-latex,dashed](i4)+(180:5.2cm) --node[above]{\begin{tabular}{>{\centering}m{6cm}}\\ consent(\allowbreak{}ada, isp): \textbf{cfr} \\
\textbf{\color{white}{\hlg{disch(\allowbreak{}obl(\allowbreak{}obl(\allowbreak{}provideData(\allowbreak{}isp, charles, ada), undue\_delay), now))}}}: cfr, \textbf{drd}
\end{tabular}}(i4);
}
{ [continue chain=state4 going below]
\node [on chain=state4,below=of i4,rectangle,draw,inner frame sep=0pt] (s4) {
\begin{minipage}{5cm}\raggedright\everypar={\hangindent=1em\hangafter=1}
\text{1)} is(\allowbreak{}charles, lawEnforcement): \textbf{drd}, \textbf{drr} \\
\text{2)} obl(\allowbreak{}deleteData(\allowbreak{}isp, ada, metadata), time(\allowbreak{}m13)): drd, \textbf{drr} \\
\text{3)} pro(\allowbreak{}deleteData(\allowbreak{}isp, ada, metadata), time(\allowbreak{}m12)): drd, \textbf{drr} \\
\text{4)} obl(\allowbreak{}provideData(\allowbreak{}isp, charles, ada), time(\allowbreak{}m1)): drd, \textbf{drr}  \\
\text{5)} pro(\allowbreak{}dataUnprotected(\allowbreak{}ada, personal), never): cfr \\
\text{6)} pro(\allowbreak{}privacyDisrespected, never): cfr\\
\text{7)} pro(\allowbreak{}storeData(\allowbreak{}isp, ada, content), never): cfr, drd, \textbf{drr}\\
\text{8)} pro(\allowbreak{}uncontrolByIndepAuth, never): cfr\\
\text{9)} pro(\allowbreak{}unfairDataProcessing(\allowbreak{}ada), never): cfr \\
\text{10)} \textbf{\textcolor{blue}{ensure\_data\_retention\_period(\allowbreak{}ada, isp, metadata, m6, m24)}} : \textbf{drd from 2 and 3} \\
\text{11)} \textbf{+} consentedDataProcessing(\allowbreak{}ada, isp): \textbf{cfr} \\
\end{minipage}
};
} 
\draw (i4) -- (s4);

}

{{ [continue chain=trace]
\node[circle,draw,on chain=trace](i5) {$S_{5}$};
\draw[-latex,thin](i4) -- 
node[above]{\begin{tabular}{>{\centering}m{6cm}}
\\ useElectronicCommunication(\allowbreak{}ada, isp): \textbf{cfr}, \textbf{drd}, \textbf{drr} \\
\end{tabular}}
(i5);
}
{ [continue chain=state5 going below]
\node [on chain=state5,below=of i5,rectangle,draw,inner frame sep=0pt] (s5) {
\begin{minipage}{5cm}\raggedright\everypar={\hangindent=1em\hangafter=1}
\text{1)} is(\allowbreak{}charles, lawEnforcement): \textbf{drd}, \textbf{drr} \\
\text{2)} pro(\allowbreak{}dataUnprotected(\allowbreak{}ada, personal), never): cfr \\
\text{3)} pro(\allowbreak{}privacyDisrespected, never): \textbf{cfr} \\
\text{4)} pro(\allowbreak{}uncontrolByIndepAuth, never): \textbf{cfr} \\
\text{5)} consentedDataProcessing(\allowbreak{}ada, isp): \textbf{cfr} \\\
\text{6)} obl(\allowbreak{}provideData(\allowbreak{}isp, charles, ada), time(\allowbreak{}m1)): drd, \textbf{drr} \\
\text{7)} obl(\allowbreak{}provideData(\allowbreak{}isp, charles, ada), undue\_delay): \textbf{drd} \\
\text{8)} pro(\allowbreak{}dataUnprotected(\allowbreak{}charles, personal), never): \textbf{cfr} \\
\text{9)} pro(\allowbreak{}unfairDataProcessing(\allowbreak{}ada), never): cfr \\
\text{10)} pro(\allowbreak{}storeData(\allowbreak{}isp, charles, content), never): cfr, drd, \textbf{drr} \\
\text{11)} \textbf{+} obl(\allowbreak{}deleteData(\allowbreak{}isp, ada, metadata), time(\allowbreak{}m13)): drd, \textbf{drr} \\
\text{12)} pro(\allowbreak{}deleteData(\allowbreak{}isp, ada, metadata), time(\allowbreak{}m12)): drd, \textbf{drr} \\
\text{13)} \textbf{-- +} obl(\allowbreak{}storeData(\allowbreak{}isp, ada, metadata), now): cfr, drd, \textbf{drr} \\
\text{14)} \textbf{-- +} obl(\allowbreak{}obl(\allowbreak{}storeData(\allowbreak{}isp, ada, metadata), now), now): cfr, \textbf{drd} \\
\text{15)} \textbf{\textcolor{blue}{ensure\_data\_retention\_period(\allowbreak{}ada, isp, metadata, m6, m24)}} : \textbf{drd from 11 and 12} \\
\text{16)} \textbf{-- +} \textbf{\textcolor{blue}{obl(\allowbreak{}obl(\allowbreak{}storeData(\allowbreak{}isp, ada, personal), now), now)}} : \textbf{cfr from 14} \\
\text{17)} \textbf{-- +} \textbf{\textcolor{blue}{obl(\allowbreak{}privacyDisrespected, now)}}: \textbf{cfr from 16} \\
\text{18)} \textbf{-- +} \textbf{\textcolor{blue}{obl(\allowbreak{}dataProcessed, now)}}: \textbf{cfr from 16} \\
\text{19)} \textbf{-- +} \textbf{\textcolor{blue}{obl(\allowbreak{}dataUnprotected(\allowbreak{}ada, personal), now)}}: \textbf{cfr from 16} \\
\text{20)} \textbf{+} \textbf{\textcolor{blue}{dataUnprotected(\allowbreak{}ada, personal)}} : \textbf{cfr from 18} \\
\text{21)} \textbf{+} \textbf{\textcolor{blue}{privacyDisrespected}} : \textbf{cfr from 17} \\
\text{22)} \textbf{+} \textbf{\textcolor{blue}{uncontrolByIndepAuth}} : \textbf{cfr from 18} \\
\end{minipage}
};
} 
\draw (i5) -- (s5);

}

{{ [continue chain=trace]
\node[circle,draw,on chain=trace](i6) {$S_{6}$};
\draw[-latex,thin](i5) -- 
node[above]{\begin{tabular}{>{\centering}m{6cm}}
\\
 storeData(\allowbreak{}isp, ada, metadata): \textbf{cfr}, \textbf{drd}, \textbf{drr} \\
\textbf{\color{white}{\hlg{disch(\allowbreak{}obl(\allowbreak{}obl(\allowbreak{}storeData(\allowbreak{}isp, ada, metadata), now), now))}}}: cfr, \textbf{drd} \\
\textbf{\color{white}{\hlg{disch(\allowbreak{}obl(\allowbreak{}storeData(\allowbreak{}isp, ada, metadata), now))}}}: cfr, drd, \textbf{drr} \\
\textbf{\color{white}{\hlg{disch(\allowbreak{}obl(\allowbreak{}obl(\allowbreak{}storeData(\allowbreak{}isp, ada, metadata), now), now))}}}: cfr, \textbf{drd} \\
\textbf{\color{white}{\hlg{disch(\allowbreak{}obl(\allowbreak{}storeData(\allowbreak{}isp, ada, metadata), now))}}}: cfr, drd, \textbf{drr} \\
\textbf{\color{white}{\hlr{viol(\allowbreak{}pro(\allowbreak{}uncontrolByIndepAuth, never))}}}: \textbf{cfr} \\
\textbf{\color{white}{\hlr{viol(\allowbreak{}pro(\allowbreak{}privacyDisrespected, never))}}}: \textbf{cfr} \\
\textbf{\color{white}{\hlr{viol(\allowbreak{}pro(\allowbreak{}dataUnprotected(\allowbreak{}ada, personal), never))}}}: \textbf{cfr}
\end{tabular}}
(i6);
}
{ [continue chain=state6 going below]
\node [on chain=state6,below=of i6,rectangle,draw,inner frame sep=0pt] (s6) {
\begin{minipage}{5cm}\raggedright\everypar={\hangindent=1em\hangafter=1}
\text{1)} is(\allowbreak{}charles, lawEnforcement): \textbf{drd}, \textbf{drr} \\
\text{2)} obl(\allowbreak{}provideData(\allowbreak{}isp, charles, ada), time(\allowbreak{}m1)): drd, \textbf{drr} \\
\text{3)} obl(\allowbreak{}provideData(\allowbreak{}isp, charles, ada), undue\_delay): \textbf{drd} \\
\text{4)} pro(\allowbreak{}dataUnprotected(\allowbreak{}ada, personal), never): \textbf{cfr} \\
\text{5)} pro(\allowbreak{}privacyDisrespected, never): cfr \\
\text{6)} pro(\allowbreak{}storeData(\allowbreak{}isp, ada, content), never): cfr, drd, \textbf{drr} \\
\text{7)} pro(\allowbreak{}uncontrolByIndepAuth, never): \textbf{cfr} \\
\text{8)} pro(\allowbreak{}unfairDataProcessing(\allowbreak{}ada), never): \textbf{cfr} \\
\text{9)} consentedDataProcessing(\allowbreak{}ada, isp): cfr\\
\text{10)} obl(\allowbreak{}deleteData(\allowbreak{}isp, ada, metadata), time(\allowbreak{}m13)): drd, \textbf{drr} \\
\text{11)} pro(\allowbreak{}deleteData(\allowbreak{}isp, ada, metadata), time(\allowbreak{}m12)): drd, \textbf{drr} \\
\text{12)} \textbf{\textcolor{blue}{ensure\_data\_retention\_period(\allowbreak{}ada, isp, metadata, m6, m24)}}: \textbf{drd from 10 and 11} \\
\text{13)} \textbf{-- +} obl(\allowbreak{}ensure\_data\_retention\_period(\allowbreak{}ada, isp, metadata, m6, m24), now): cfr, \textbf{drd} \\
\end{minipage}
};
} 
\draw (i6) -- (s6);

}

{{ [continue chain=trace]
\node[circle,draw,on chain=trace](i7) {$S_{7}$};
\draw[-latex,thin](i6) -- 
node[above]{\begin{tabular}{>{\centering}m{6cm}}
\\
\textbf{\color{white}{\hlg{disch(obl(ensure\_data\_retention\_period(ada, isp, metadata, m6, m24), now))}}}: cfr \textbf{drd} \\
\end{tabular}}
(i7);
}
}


{ [continue chain=trace]
\node[on chain=trace] (i8) {};
\draw[color=white,-latex,dashed](i7) -- (i8);
}

\end{tikzpicture}
}\\
\end{longtable}
    \caption{}
\end{figure}

The resulting multi--level governance institution model is depicted in Figure~\ref{figExampleTrace}, for brevity edited to just containing those fluents that are relevant to the discharge and violation of norms. The model is described subsequently.

We first look at the interaction between the UK--DRR and the EU--DRD which governs the UK--DRR. Accordingly:
\begin{itemize}
\item State $S_0$ -- Contains fluents stating the agent \verb;charles; is playing the role of \\ \verb;lawEnforcement; in the UK--DRR and EU--DRD. The EU--DRD obliges that it is prohibited for \verb;isp; to store the \verb;content; of \verb;ada;'s data. The UK--DRR does indeed prohibit \verb;isp; from storing the content of ada's communications data. Thus, the obligation to prohibit storing content data in the EU--DRD is immediately discharged as denoted by the discharge event occurring in the transition to the next state. The transition to the next state also includes the event of \verb;ada; using electronic communications provided by \verb;isp;, due to the occurrence of the \textit{exogenous} event in the timeline program stating the same.
\item State $S_1$ -- Includes new fluents. Firstly, the EU--DRD imposes an obligation on the UK--DRR to oblige \verb;isp; to store \verb;ada;'s communications' metadata. Secondly, the UK--DRR imposes an obligation for \verb;isp; to store \verb;ada;'s communications' metadata. The UK--DRR's first--order obligation to store metadata discharges the EU--DRD's second--order obligation to oblige an obligation to store metadata. The transition to state $S_2$ includes the event of \verb;ada;'s communications' metadata being stored by \verb;isp;, driven by the corresponding exogenous event, and consequently the discharge of the obligation from the UK--DRR for \verb;isp; to store Ada's communications metadata.
\item State $S_2$ -- The EU--DRD, which governs the UK--DRR, obliges data retention to be ensured for between 6 and 24 months. In comparison, the UK--DRR, both obliges that Ada's communications metadata is deleted before 13 months and prohibits deletion before 12 months. Hence, the UK--DRR requires that metadata is stored for between 12 and 13 months, which is abstractly interpreted in the EU--DRD as ensuring data is retained between 6 and 24 months \\ (\verb;ensure_data_retention_period(ada, isp, metadata, m6, m24;) discharging the EU--DRD's obligation for data to be stored within this timeframe. In the transition to state $S_3$ \verb;charles; requests Ada's data from \verb;isp;.
\item State $S_3$ -- Since \verb;charles; is playing the role of \verb;lawEnforcement;, this causes the EU--DRD to oblige that \verb;isp; is obliged to provide the data before any \verb;undue_delay;. Meanwhile, the UK--DRR obliges \verb;isp; to provide \verb;charles; with the data within one month (\verb;m1;). According to the EU--DRD anything occurring before one month \textit{counts--as} the event of \verb;undue_delay;. Thus the EU--DRD interprets the obligation to provide data within one month as the abstract obligation to provide data before any undue delay. This causes discharge of the obligation to oblige data is provided before any undue delay.
\item States $S_4$ to $S_6$ follow largely the same pattern. In the transition to the next state $S_4$ \verb;ada; consents to her data being stored, which has no affect on the UK--DRR or the EU--DRD. Transitioning to state $S_5$ \verb;ada; uses electronic communications, then \verb;ada;'s data is stored, causing the same obligations and prohibitions to be imposed by the UK--DRR and EU--DRD as when these events occurred previously.
\end{itemize}

In conclusion, for this trace of events the UK--DRR is compliant with the EU--DRD. All of the EU--DRD's normative fluents it imposes are discharged and none are violated. In comparison, the EU--DRD is \textit{non--compliant} with the EU--CFR as we will see:

\begin{itemize}
\item State $S_0$ -- the EU--CFR prohibits the EU--DRD's regulations from being uncontrolled by an independent authority. What this means is that data retention should be within the EU jurisdiction. Likewise, the EU--CFR also prohibits data from being unprotected (i.e. stored without anonymisation), privacy from being disrespected (i.e. personal data being stored) and data being processed unfairly (i.e. personal data being stored without an agent's consent).
\item State $S_1$ -- a number of the EU--CFR's prohibitions are violated:
\begin{itemize}
\item \textbf{Violation of the CFR's prohibition on regulations not being controlled by an independent authority} (meaning, compliance with the EU--CFR's data protection rights must be observable by an independent authority, such as by ensuring data is retained within the EU). The EU--DRD obliges the UK--DRR to oblige \verb;ada;'s communications' metadata is stored. According to the EU--CFR obliging storing data (of any type) counts--as data being processed, hence an obligation to oblige storing metadata is abstracted to an obligation to process data, which is abstracted further to processing data. The EU--CFR views processing data without a prohibition on it being stored outside of the EU counting--as regulations not being controlled by an independent authority. Hence, the prohibition on regulations being uncontrolled by an independent authority is violated.
\item \textbf{Violation of the CFR's prohibition on unfair data processing}. The EU--CFR interprets storing metadata as storing personal data, thus it determines that there is an abstract obligation to oblige personal data is stored. In the EU--CFR, storing personal data in the context that an agent has not consented counts--as non--consensual data processing (\verb;nonConsensualDataProcessing(ada);). Thus the EU--CFR determines that there is an obligation to oblige non--consensual data processing of \verb;ada;'s data. According the EU--CFR an obligation to store data non--consensually counts--as unfair data processing, hence an obligation to oblige non--consensual data processing is abstracted to an obligation to process data unfairly. An obligation to process data unfairly in turn, counts--as unfair data processing (i.e. from the perspective of the EU--CFR it does not matter if data is actually processed unfairly or just obliged, both are unfair data processing). This causes the EU--CFR's prohibition on processing data unfairly to be violated.
\item \textbf{Violation of the CFR's prohibition on disrespecting privacy}. The obligation to oblige storing metadata imposed by the EU--DRD is abstracted to an obligation to oblige storing personal data. In the EU--CFR obliging storing personal data counts--as the non--inertial fluent for privacy to be disrespected. Hence, obliging an obligation to store personal data is further abstracted to obliging privacy is disrespected which also counts--as simply disrespecting privacy. Hence the EU--CFR's prohibition on disrespecting privacy is violated.
\item \textbf{Violation of the CFR's prohibition on data being unprotected}. The  obliges an obligation for Ada's metadata to be stored (according to the  an obligation to oblige personal data to be stored) even in the context that it is not anonymised. The EU--CFR views an obligation to oblige storing personal data as being the same thing as processing data, which in the context that the data is not anonymised is abstractly the same thing as data being unprotected. Hence, the EU--CFR's prohibition on data being unprotected is violated.
\end{itemize}
Each violated prohibition in the EU--CFR is initiated in the next state. 
\item States $S_2$ and $S_3$ contain nothing of interest from the perspective of the EU--CFR. In the transition to state $S_4$ Ada consents to her personal data being stored.
\item State $S_4$ contains a fluent stating Ada has consented to her personal data being stored.
\item State $S_5$ also contains prohibitions in the EU--CFR which are violated by the EU--DRD, as in state $S_1$, with one difference:
\begin{itemize}
\item \textbf{The CFR's prohibition on data being processed unfairly is \textit{not} violated}. The EU--DRD, from the perspective of the EU--CFR, obliges an obligation to store personal data. However, since Ada has consented the obligation to oblige personal data being stored is not abstracted to an obligation to oblige non--consensual data processing and not subsequently abstracted to `unfair data processing'. Hence, in state $S_5$, unlike in state $S_2$ the EU--CFR's prohibition on unfair data processing is not violated since the context is different (Ada has consented to her data being stored). Meanwhile, the rest of the EU--CFR's prohibitions are violated (for the second time).
\end{itemize}
\end{itemize}

From this case study we can see the UK--DRR is compliant with the EU--DRD (i.e. the UK's legislation correctly implements the directive). On the other hand, the EU's data retention directive is non--compliant with the EU--CFR. In particular, the EU--DRD was found to be non--compliant in a particular social context with particular prohibitions issued by the EU--CFR. In different contexts the same prohibitions might not be violated. As we saw in the context that Ada had consented to her personal data being processed, the directive did not the violate the prohibition on unfair data processing. This is because the directive's normative fluents were not interpreted by the charter as more abstractly counting--as unfair data processing. Hence, whether there is compliance depends on the context which determines the abstract meaning of normative fluents. A full compliance check examines all possible social contexts by checking all possible traces of events that result in unique answer--sets (or in our formal framework, models).

\section{Computational Framework Soundness and Completeness}
\label{secMLGCompTheorems}

We now demonstrate that the computational framework provides an executable implementation of the formal framework. This is with theorems stating the computational framework is sound and complete with respect to the formal framework (proofs are provided in the appendices). We begin by packaging, for convenience, the aforementioned ASP programs into a single ASP program $\Pi^{\pazocal{ML}(k)}$.

\begin{definition}{\textbf{Multi--level Governance ASP Program}} Let $\pazocal{ML} = \langle \pazocal{T}, R \rangle$  be a multi--level governance institution. Let $\textit{ctr}$ be a composite trace for $\pazocal{ML}$ of length $n$. Let $\Pi^{\textit{insts}}$ and $\Pi^{\textit{abstr}}$ be the institutions and deontic abstraction programs obtained for $\pazocal{ML}$. Let, $\Pi^{\textit{trace}(n)}$ be the trace program obtained for $\textit{ctr}$ and let $\Pi^{\textit{base}(n)}$ be a multi--level governance base program. A multi--level governance institution ASP program for $\pazocal{ML}$ and a composite trace $\textit{ctr}$ is:
\begin{equation*}
\Pi^{\pazocal{ML}(n)} = \Pi^{\textit{base}(n)} \cup \Pi^{\textit{trace}(n)} \cup \Pi^{\textit{abstr}} \cup \Pi^{\textit{insts}}
\end{equation*}
\end{definition}

The first property we present gives the set of of events that are \textit{maximally} in the set of events according to the event generation operation.

\begin{lemma}\label{theoremGRMaximality} If $\textit{GR}^{i}$ is the event generation operation for an institution $\pazocal{I}^{i}$ with respect to a tuple of events $E$ and a tuple of events $S$, and $E^{\prime} = \textit{GR}^{i}(S^{i}, E^{i})$ w.r.t. $S$ and $E$ then
\begin{subequations}
$\forall e^{\prime} \in E^{\prime}$:
\begin{align*}
& e^{\prime} = \textit{now} & \textbf{or} \tag{T\ref{theoremGRMaximality}.1}\label{eqGR1} \\
&e^{\prime} \in E^{i} & \textbf{or} \tag{T\ref{theoremGRMaximality}.2}\label{eqGR2} \\
& \exists X, e : X \in \pazocal{X} \wedge e \in E \wedge S^{i} \models X \wedge e^{\prime} \in \pazocal{G}(X, e) \wedge S^{i} \models \textit{pow}(e^{\prime}) & \textbf{or} \tag{T\ref{theoremGRMaximality}.3}\label{eqGR4} \\
& e^{\prime} = \textit{disch}(\textit{obl}(a, d)) \text{s.t.} \exists \textit{obl}(a, d) : S^{i} \models \textit{obl}(a, d) \wedge (a \in E^{\prime} \vee S^{i} \models a) & \textbf{or} \tag{T\ref{theoremGRMaximality}.4}\label{eqGR5} \\
& e^{\prime} = \textit{viol(obl(a, d))} \text{s.t.} \exists \textit{obl}(a, d) : S^{i} \models \textit{obl}(a, d) \wedge (d \in E^{\prime} \vee S^{i} \models d) \wedge & \\ & \; \; \; \; \; \; \; \; \; \; \; \; \; \; \; \; \; \; \; \; \; \; \; \; \; \; \; \; \; \; \; \; \; \; \; \; \; \; \; \; \; \; \; \; \; \; \; \; \; \textit{disch}(\textit{obl}(a, d)) \not \in E^{\prime} & \textbf{or} \tag{T\ref{theoremGRMaximality}.5}\label{eqGR6} \\
& e^{\prime} = \textit{disch}(\textit{pro}(a, d)) \text{s.t.} \exists \textit{pro}(a, d) : S^{i} \models \textit{pro}(a, d) \wedge (d \in E^{\prime} \vee S^{i} \models d) & \textbf{or} \tag{T\ref{theoremGRMaximality}.6}\label{eqGR7} \\
& e^{\prime} = \textit{viol}(\textit{pro}(a, d)) \text{s.t.} \exists \textit{pro}(a, d) : S^{i} \models \textit{pro}(a, d) \wedge (a \in E^{\prime} \vee S^{i} \models a) \wedge & \\ & \; \; \; \; \; \; \; \; \; \; \; \; \; \; \; \; \; \; \; \; \; \; \; \; \; \; \; \; \; \; \; \; \; \; \; \; \; \; \; \; \; \; \; \; \; \; \; \; \; \; \textit{disch}(\textit{pro}(a, d)) \not \in E^{\prime} & \textbf{or} \tag{T\ref{theoremGRMaximality}.7}\label{eqGR8} \\
& \exists \langle h, i \rangle : \langle h, i \rangle \in R, e \in \pazocal{E}^{h}_{\textit{norm}} \cap \pazocal{E}^{i}_{\textit{norm}}  \tag{T\ref{theoremGRMaximality}.8}\label{eqGR9}
\end{align*}
\end{subequations}
\end{lemma}
\begin{proof}
Immediately follows from Definition~\ref{defGR}.
\end{proof}

The second property we present gives the set of fluents that are maximally in the closure of deontological counts--as.

\begin{lemma}\label{theoremDCAMaximality} If $\textit{DC}^{\omega}$ is the closure of the deontological counts--as function for an institution $\pazocal{I} = \langle \pazocal{E}, \pazocal{F}, \pazocal{C}, \pazocal{G}, \pazocal{D}, \Delta \rangle$, a state $S \in \Sigma$ for $\pazocal{I}$ and $S^{\prime} = \textit{DC}^{\omega}(S)$ then
\begin{subequations}
$\forall f^{\prime} \in S^{\prime}:$
\begin{alignat*}{3}
f^{\prime} \in S && \quad \textbf{or} & \quad \tag{T\ref{theoremDCAMaximality}.1}\label{eqDCCl3} \\
\exists \langle N^{\prime}, f^{\prime} \rangle \in \textit{DC}(S^{\prime}) : N^{\prime} \subseteq S^{\prime} && \quad  & \quad \tag{T\ref{theoremDCAMaximality}.2}\label{eqDCCl4}
\end{alignat*}
\end{subequations}
\end{lemma}
\begin{proof}Immediately follows from Definition~\ref{defDC}.
\end{proof}

We now give the soundness property for the deontic abstraction representation in ASP with respect to the formal definition of deontological counts--as. In doing so, we demonstrate that we have provided a transformation that flattens the deontological counts--as function described in the formal framework to an executable set of ASP rules. The property states that a state in the answer--set for a multi--level governance answer--set program is equivalent to the same state in the formal model for the formal framework with the deontic counts--as function $\textit{DC}^{i \omega}$ applied.

\begin{lemma}\label{theoremAbstraction} Let $\pazocal{ML} = \langle \pazocal{T}, R \rangle$ be a multi--level governance institution s.t. $\pazocal{T} = \langle \pazocal{I}^{1}, ..., \pazocal{I}^{n} \rangle$, and $\textit{ctr}$ be a composite trace of length $k$. Let $\forall i \in [1,n]$ $\textit{In}^{i}$ be a unique label for $\pazocal{I}^{i}$. Let $\Pi^{\pazocal{ML}(k)}$ be the multi--level governance ASP program for $\pazocal{ML}$ and $\textit{ctr}$. Let $M_{P}$ be an answer--set for the program $P^{*} = \textit{ground}(\Pi^{\pazocal{ML}(k)})$. Given a set $S^{i}_{j}$ such that
\begin{align*}
& \forall i \in [1, m], \forall j \in [k] : M_P \models \asp{holdsat(}f, \textit{In}^{i}, j) \Rightarrow f \in S^{i}_{j} 
\end{align*}
then $S^{i}_{j} = \textit{DC}^{i \omega}(S^{i}_{j})$.
\end{lemma}
\begin{proof} See the appendices.
\end{proof}

The next property we are interested in is soundness for the translation to an ASP program as a whole. Specifically, the property states any answer--set for a multi--level governance ASP program for a given trace of events corresponds to a multi--level governance institution model in the formal framework for the same trace of events.

\begin{theorem}{(Soundness)}\label{theorSound} Given a multi--level governance institution $\pazocal{ML} = \langle \pazocal{T}, R \rangle$ s.t. $\pazocal{T} = \langle \pazocal{I}^{1}, ..., \pazocal{I}^{n} \rangle$. Let $\textit{ctr} = \langle e_0, ..., e_k \rangle$ be a composite trace for $\pazocal{ML}$. Let $\Pi^{\pazocal{ML}(k)}$ be the multi--level governance ASP program for $\pazocal{ML}$ and $\textit{ctr}$. Let $\forall i \in [1, n] : \textit{str}^{i} = \langle \textit{se}^{i}_{0}, ..., \textit{se}^{i}_{k} \rangle$ be a synchronised trace for $\pazocal{I}^{i}$ w.r.t. $\textit{ctr}$. Let $M_{P}$ be an answer--set for the program $P^{*} = \textit{ground}(\Pi^{\pazocal{ML}(k)})$. Then $M = \langle M^{1}, ..., M^{n} \rangle$ with $\forall i \in [n] : M^{i} = \langle S^{i}, E^{i} \rangle, S^{i} = \langle S^{i}_{0}, ..., S^{i}_{k+1} \rangle, E^{i} = \langle E^{i}_0, ..., E^{i}_k \rangle$ such that:
\begin{align*}
& \forall h \in [1, n], \forall j \in [k], : M_P \models \asp{holdsat(}f, \textit{In}^{h}, j) \Rightarrow f \in S^{h}_{j} \tag{T\ref{theorSound}.1}\label{eqThSo1}  \\
& \forall h \in [1, n], \forall j \in [k], \forall e \neq \asp{null} : M_P \models \asp{occurred(}e, \textit{In}^{h}, j) \Rightarrow e \in E^{h}_{j} \tag{T\ref{theorSound}.2}\label{eqThSo3} \\
& \forall h \in [1, n], \forall j \in [k] : M_P \models \asp{occurred(null}, \textit{In}^{h}, j) \Rightarrow e_{\textit{null}} \in E^{h}_{j} \tag{T\ref{theorSound}.3}\label{eqThSo4}
\end{align*}
is the model of $\pazocal{ML}$ w.r.t. $\textit{ctr}$.
\end{theorem}
\begin{proof} See the appendices.
\end{proof}

The next property we are interested in is completeness. This states that for any model of a multi--level governance institution in the formal framework, for a trace of events, the multi--level governance ASP program produces a corresponding answer--set for the same trace of events.

\begin{theorem}{(Completeness)}\label{theorCompl} Given a multi--level governance institution $\pazocal{ML} = \langle \pazocal{T}, R \rangle$ s.t. $\pazocal{T} = \langle \pazocal{I}^{1}, ..., \pazocal{I}^{n} \rangle$. Let  $\textit{ctr} = \langle e_0, ..., e_k \rangle$ be a composite trace for $\pazocal{ML}$. Let $\forall i \in [1,n] : \textit{str}^{i} = \langle \textit{str}^{i}_{0}, ..., \textit{str}^{i}_{k} \rangle$ be a synchronised trace for $\pazocal{I}^{i}$ w.r.t. $\textit{ctr}$. Let $M = \langle M^{1}, ..., M^{n} \rangle$ be the multi--level governance institution model $\pazocal{ML}$ w.r.t. \textit{ctr} where $\forall i \in [1,n] : M^{i} = \langle S^{i}, E^{i} \rangle, S^{i} = \langle S^{i}_0, ..., S^{i}_{k+1} \rangle, E^{i} = \langle E^{i}_0, ..., E^{i}_{k} \rangle$. Let $\Pi^{\pazocal{ML}(k)}$ be the multi--level structure ASP program for $\pazocal{ML}$ and a composite trace $\textit{ctr}$. Let $M_{P}$ be the set of atoms:
\begin{align*}
\forall i \in [1, n], \forall j \in [k + 1] : S^{i}_{j} \models f \Rightarrow M_P \models \asp{holdsat(}f, \textit{In}^{i}, j) & \tag{T\ref{theorCompl}.1}\label{eqThCo1} \\
\forall i \in [1, n], \forall j \in [k] : e \in E^{i}_{j} \Rightarrow M_P \models \asp{occurred(}e, \textit{In}^{i}, j) & \tag{T\ref{theorCompl}.2}\label{eqThCo2} \\
\forall i \in [1,n], \forall j \in [1, k] : f \in (S_{j} \backslash S_{j-1}) \cap \pazocal{F}^{i}_{\textit{inert}} \Rightarrow M_{P} \models \asp{initiated}(f, \textit{In}^{i}, j - 1) & \tag{T\ref{theorCompl}.3}\label{eqThCo3} \\
\forall i \in [1,n], \forall j \in [1, k] : f \in (S^{i}_{j} \backslash S^{i}_{j+1}) \cap \pazocal{F}^{i}_{\textit{inert}} \Rightarrow M_{P} \models \asp{terminated}(f, \textit{In}^{i}, j+1) & \tag{T\ref{theorCompl}.4}\label{eqThCo4} \\
\forall i \in [1, n], \forall j \in [k] : e = \textit{ctr}_j \Rightarrow M_P \models \asp{observed(}e, \textit{In}^{i}, j), & \\
M_P \models \asp{compObserved(}e, j), & \\
M_P \models \asp{obs(}j) \tag{T\ref{theorCompl}.5}\label{eqThCo5} \\
\forall i \in [1, n], \forall j \in [k] : e = \textit{str}^{i}_j \neq e_{\textit{null}} \Rightarrow M_P \models \asp{occurred(}e, \textit{In}^{i}, j) & \tag{T\ref{theorCompl}.6}\label{eqThCo6} \\
\forall i \in [1, n], \forall j \in [k] : e_{\textit{null}} = \textit{str}^{i}_j \Rightarrow M_P \models \asp{occurred(null}, \textit{In}^{i}, j) & \tag{T\ref{theorCompl}.7}\label{eqThCo7} \\
\forall i \in [1, n], \forall e \in \pazocal{E}^{i}_{\textit{obs}} : M_P \models \asp{evtype(e}, \textit{In}^{i}\asp{, ex)} & \tag{T\ref{theorCompl}.8}\label{eqThCo8} \\
\forall i \in [1, n], \forall e \in \pazocal{E}^{i}_{\textit{inst}} : M_P \models \asp{evtype(e}, \textit{In}^{i}\asp{, inst)} & \tag{T\ref{theorCompl}.9}\label{eqThCo9} \\
\forall i \in [1, n], \forall f \in \pazocal{F}^{i}_{\textit{inert}} : M_P \models \asp{ifluent(f}, \textit{In}^{i}\asp{)} & \tag{T\ref{theorCompl}.10}\label{eqThCo10} \\
\forall i \in [1, n], \forall f \in \pazocal{F}^{i}_{\textit{ninert}} : M_P \models \asp{nifluent(f}, \textit{In}^{i}\asp{)} & \tag{T\ref{theorCompl}.11}\label{eqThCo11} \\
\forall i \in [1, n] : M_P \models \asp{inst(}\textit{In}^{i}\asp{)} & \tag{T\ref{theorCompl}.12}\label{eqThCo12} \\
\forall i \in [k] : M_P \models \asp{instant(}i\asp{)} & \tag{T\ref{theorCompl}.13}\label{eqThCo13} \\
M_P \models \asp{start(}0\asp{)} & \tag{T\ref{theorCompl}.14}\label{eqThCo14} \\
\forall i, j \in [k] : j = i + 1 \Rightarrow M_P \models \asp{next(}i, j\asp{)} & \tag{T\ref{theorCompl}.15}\label{eqThCo15} \\
M_P \models \asp{final(}k\asp{)} & \tag{T\ref{theorCompl}.16}\label{eqThCo16} \\
\end{align*}
Then, $M_{P}$ is an answer set of $P^{*} = \textit{ground}(\Pi^{\pazocal{ML}(k)})$.
\end{theorem}
\begin{proof} See the appendices.
\end{proof}

This concludes the demonstration of the soundness and completeness of the formal and computational frameworks, with respect to each other.

\section{Related Work}
\label{secMLGCompRelatedWork}

In this chapter we presented a computational framework for reasoning about multi--level governance. That is, we focussed on the practical side. Whilst there are many practical institutional reasoning frameworks, we find none which contribute automated reasoning for determining compliance in multi--level governance. We discuss the related work in more detail as follows.

There have been many different approaches proposed to reason about institutions, normative systems and organisations which we split into three broad types. Firstly, those proposing a high--level institution specification language (e.g.  \cite{Lopez2003,Lopez2006,DInverno2012}) for institution designers to precisely specify an institution's software implementation. Secondly, those proposing or studying formal logics of norms and other institutional rules (e.g. \cite{Boella2003d,Dignum2002,Makinson2003,Grossi2005a,Grossi2006a,Torre1998}). Thirdly, those contributing frameworks for formally representing and reasoning about institutions and normative systems, with an aim for practical implementations using an algorithmic or logic--programming based approach (e.g. \cite{Cliffe2007,Cliffe2006,Jianga,Li2014,Governatori2005a,Governatori2007}). Our work most closely relates to the latter practical frameworks, which we discuss in more detail.

The most closely related framework, on which we build, is the Institutional Action Language, InstAL, first proposed by Cliffe et al. \cite{Cliffe2006,Cliffe2007}. Li et al. have made developments on InstAL for detecting conflicts between norms \cite{Li2013}, in particular in interacting institutions \cite{Li2013b} and cooperating institutions \cite{Li2014}. In the work of Li et al. institutions are linked with special \textit{bridge} institutions such that events occurring in one can cause events to occur in another and likewise for fluents being initiated or terminated. Such bridge institutions have a similar role to our links between different levelled institutions. The difference is that bridge also operate as a king of institution, evolving from one state to the next such that the bridges (links) between institutions can change over time. In contrast, we use institutional links which remain static over time. Further developments on InstAL were realised by Pieters et al. \cite{Pieters2013,Pieters2014} for reasoning about institutions as a means to police and enforce security policies. In their work, Pieters et al. \cite{Pieters2013,Pieters2014} extend InstAL with rules for non--inertial fluents which (in our own words) state ``when context C holds then so does fluent B''. These bear similarity to our fluent derivation rules of the form ``fluent A derives fluent B in context C''. But, in our case we view fluent derivation rules as firstly ascribing a special meaning to a concrete fluent `A' (hence they have a different form) and secondly serving as a basis for abstracting normative fluents. Our work is also related to our previous work which was loosely based on InstAL \cite{King2015a,King2015b} for reasoning about multi--tier institutions and higher--order norms. 

The main differences between these developments and this chapter is we have extended InstAL for representation and reasoning about multi--level governance. In more detail, there are differences in reasoning about permissive societies (where anything not prohibited is permitted), instantaneous and indefinite norms, bridged versus linked institutions, non--inertial fluent rules versus fluent derivation rules, and our main focus in this chapter: combining higher--order normative reasoning and norm abstraction. We summarise all of these differences in Table~\ref{tabInstALComp}.

\def\tick{\tikz\fill[scale=0.35](0,.35) -- (.25,0) -- (1,.7) -- (.25,.15) -- cycle;}
\begin{table}
\centering
\newcolumntype{C}{ >{\centering\arraybackslash} m{10mm} }
\newcolumntype{D}{ >{\centering\arraybackslash} m{12mm} }
\newcolumntype{E}{ >{\centering\arraybackslash} m{15mm} }
{\renewcommand{\arraystretch}{1.25}
\begin{tabular}{ l C E E D E }
    & InstAL \cite{Cliffe2006,Cliffe2007} & Li et al. \cite{Li2013b,Li2013,Li2014} & Pieters et al. \cite{Pieters2013,Pieters2014} & King et al. \cite{King2015a,King2015b} & This chapter \\ \hline
Individual Institutions & \tick & \tick & \tick & \tick & \tick \\ \hline
Empowerment & \tick & \tick & \tick & & \tick \\ \hline
\textbf{B}ridged vs. & & & & & \\
\textbf{L}inked Institutions & & \textbf{B} & & & \textbf{L} \\ \hline
\textbf{N}on--\textbf{I}nertial vs. & & & & & \\
Fluent \textbf{D}erivation rules & & & \textbf{NI} & & \textbf{D} \\ \hline
Permissive Society & & & & \tick & \tick \\ \hline
Instantaneous and & & & & & \\
Indefinite Norms & & & & & \tick \\ \hline
\textbf{Higher--order} & & & & & \\ 
\textbf{Normative Reasoning} & & & & \tick & \tick \\ \hline
\textbf{Norm Abstraction} & & & & & \tick \\ \hline
\end{tabular}} \quad
\caption[Multi--level governance reasoning comparison with InstAL]{Comparison between closely related developments on InstAL.}
\label{tabInstALComp}
\end{table}

Also relevant are practical approaches for reasoning about normative social commitments (e.g. contracts, promises) \cite{Gunay2012,Yolum2004}, using the Event Calculus \cite{Kowalski1986}. These bear similarity to our own approach since we use EC--like constructs (initiation and termination of fluents driven by events) and reasoning. Commitments have been  formalised with `lifecycle' elements not present in our notion of norms, such as the creation and deletion of the commitment/rule (e.g. through an utterance) which in turn imposes obligations in particular circumstances. Higher--order commitments are grammatical in these approaches but do not coincide with our notion of higher--order norms. In our case, a higher--order norm represents a statement such as `the \textit{outcome} of your rules should not be an obligation to do X in context C', where X can be a concrete or abstract notion. On the other hand, nested commitments represent `you have promised to me that you will not make a \textit{commitment} towards person X to do Y in context C', here the nesting simply represents that a rule of a specific form should not be created, regardless of the rule's effects and the abstract meaning of those effects. Consequently, nested commitments, come from a fundamentally different perspective and are not aligned with our formalisation of regulations which govern other regulations nor do they capture abstraction.

Another practical institutional reasoning approach is temporal defeasible deontic logic. Defeasible logic is a non--monotonic logic designed to be implemented in Prolog \cite{Antoniou2001,Nute1987}. There are three rules types in defeasible logic, \textit{strict} rules ($\rightarrow$) whose conclusion is true so long as the antecedent is true, \textit{defeasible} rules ($\Rightarrow$) whose conclusion is true unless the rule is rebutted by another rule, and \textit{defeating} rules ($\rightsquigarrow$) whose conclusion is never true but if the antecedent is true rebuts other rules with a contradictory conclusion. Defeasible logic comprises a proof procedure where rule conclusions are tested for whether they are true by first asserting them as an argument, then finding all counter--arguments which rebut them by applying defeating rules, and then recursively counter--attacking all rebuttals with further arguments, terminating by constraints on non--repeatability of arguments (e.g. \cite{Prakken1996a}). Defeasible \textit{temporal deontic logics} proposed by Governatori et al. \cite{Governatori2005a,Governatori2007} extend defeasible logic with rule types and proof procedures for obligations and temporalised outcomes. In these proposals various legal concepts are formalised, including constitutive rules and norms. But as far as we know there have been no developments on these approaches towards norms governing norms and/or norm abstraction, such as for reasoning about compliance in multi--level governance.

\section{Discussion}
\label{secMLGCompDiscussion}

In this chapter, we answered the question ``\textit{how can institutional design compliance in multi--level governance be computationally verified?}'' with a computational framework. In doing so, we provided a practical way to reason with the semantics proposed in the previous chapter. We assessed our proposal along two fronts. Firstly, with a comprehensive case study based on three--levelled governance in EU law where abstraction and context--sensitivity are important in determining non--compliance. Secondly, by proving that the practical implementation in Answer--Set Programming, the \textit{computational framework}, is indeed \textit{sound} and \textit{complete} with respect to the formal framework showing the two correspond. We used the fact that the formal framework corresponds to the computational framework to implement the proposal by extending the InstAL compiler, thereby offering users a high--level language in order to specify multi--level governance and the means to automatically detect (non--)compliance. To summarise, this chapter provides both a practical way to determine compliance in multi--level governance and also a way to assess the previous chapter's proposed semantics with a real--world case study and its execution.

The main weakness of this chapter's contributions is that it is heavily dependent on Answer--Set Programming. In one sense this is not a problem, since the previous chapter proposed a semantics which `stand on their own'. In another sense, this chapter aims to provide practical reasoning, whilst certain limitations of answer--set programming have practical implications. Specifically, due to answer--set programming's limitations, this chapter was unable to provide general rules for all of the semantics proposed in the previous section. In particular, the deontological abstraction semantics, characterised as a function in the previous section, are represented as a flattened function in the form of ASP rules in this section. What this means is that any time an institution design is changed and we wish to check its compliance, then the high--level description of an institution needs to be newly compiled in its entirety to an Answer--Set Program. If deontological abstraction's semantics had a general set of ASP rules, then only the changed rules in the changed institution would need a new corresponding ASP representation. To address this weakness, future work should investigate different representation results. Possible ways to achieve this include a corresponding representation in Prolog (e.g. as for defeasible logic \cite{Antoniou2001}) or an embedding in a higher--order logic theorem prover where the higher--order logic could act as a meta--language in which to represent the semantics from the previous section (e.g. as in \cite{Benzmuller2014}).
\chapter{Explanatory Rectifications for Non--compliant Institutions}
\label{chapter_5}

\epigraph[0pt]{Plurality is never to be posited without necessity.}{William of Ockham}

\epigraph[0pt]{Whenever possible, substitute constructions out of known entities for inferences to unknown entities.}{Bertrand Russell}

\blfootnote{\color{tck-grey}This chapter is based on the following paper:\\
\textbf{King, T. C.}, Li, T., Vos, M. De, Jonker, C. M., Padget, J., \& Riemsdijk, M. B. Van. (2016). Revising Institutions Governed by Institutions for Compliant Regulations. Coordination, Organizations, Institutions, and Normes in Agent Systems XI: COIN 2015 International Workshops, COIN@ AAMAS, Istanbul, Turkey, May 4, 2015, COIN@ IJCAI, Buenos Aires, Argentina, July 26, 2015, Revised Selected Papers., 9628, 191 – 208. \cite{King2015b}}


\newpage

This chapter makes the following contributions:

\begin{itemize}
\item An application of Inductive Logic Programming (ILP) in ASP, demonstrated with an implementation, to provide a mechanism that \textit{successfully} revises rules of a non--compliant institution to ensure it is compliant.
\item A revision mechanism that ensures institution revisions are \textit{minimal} in their regulatory effects through adding, deleting and modifying rules, thereby keeping as closely as possible to the designer's original intentions.
\end{itemize}

In chapters \ref{chapter_3} and \ref{chapter_4} we presented a formal and computational framework for determining compliance in multi--level governance. The contributions can be used by a judiciary to determine punishments for non--compliance or by a legislature to avoid enacting non--compliant legislation. However, at this point in the dissertation it is unclear how a legislature should address non--compliance by changing an institutional design. In this chapter we return to the issue of rule change, but this time asking the question, how can we mechanically find ways in which legislation can be modified to ensure it is compliant?

On the one hand, an institution designer wishes to avoid punishment for non--compliant institution designs. On the other hand, an institution designer designed the non--compliant institution with the aim of meeting particular societal outcomes. At the same time, it is important not to just rectify non--compliance but also find the simplest explanation for non--compliant institution designs in order to support a legislator in avoiding making the same mistakes. 

It is an arduous task to determine non--compliance, but even more arduous to rectify it in a way that keeps as closely as possible to the legislature's original intentions. Firstly because different rule modifications, namely addition, deletion and changes to the antecedent, must be tried to determine if they result in compliance in various contexts and therefore find a possible explanation for non--compliance. Secondly, they must be compared to alternative revisions that result in compliance to see if one makes fewer changes to the rules than another, in order to find the \textit{simplest} and most general explanation for non--compliance. Thirdly, because different revisions must also be optimised for keeping as closely as possible to the designer's original intentions by still producing the same legal effects (e.g. obligations and prohibitions) that do not result in the institution design being non--compliant. Clearly, this is a mechanical task ripe for computation. 

The idea is to test different changes to an institution design in order to rectify non--compliance and therefore find possible explanations for non--compliance. Then, to take the \textit{simplest} explanation, which minimally changes an institution's outcomes and rules, and apply it as a remedy. In this chapter we address the burden on the institution designer in revising to be compliant with a computational mechanism which also minimise the changes to the outcomes of the institution design.

This chapter builds on the previous chapters' formal and computational frameworks. However, this chapter deals with a simplified version of the formal and computational framework. We discuss this simpler framework, its application to a new case study and the approach we take in \ref{secReviApproach}. Then, we present the computational mechanism for successfully and minimally revising an institution for compliance in section~\ref{secReviForCompliance}. We finish this chapter by comparing to other mechanisms for revising institutions and sets of rules in \ref{secRevisionsRelatedWork} and conclude with discussion on limitations and wider implications in section~\ref{secReviDisc}.

\section{Approach}
\label{secReviApproach}

In this chapter, we look at revising rules for compliance. As we saw in the previous chapter, individual institutions which are governed by other institutions (in that case, within multi--level governance), can be represented as ASP rules. Building on that idea, the approach we take to revising an institution to be compliant is to revise its representation as ASP rules, using ASP itself to try various rule modifications. 

However, as we discussed at the end of the preceding chapter, there is a limitation of the approach taken due to two contributing factors. Firstly, because whenever a constitutive rule is changed in an institution, then the abstract meaning of concrete normative fluents, which is derived from the constitutive rules, also changes. Secondly, because normative fluent abstraction is represented in ASP as specific ASP rules (rather than a general semantics as in our formal framework). Taking these two factors together, this means whenever a constitutive rule is changed in ASP then potentially other ASP rules must also be changed representing the abstraction of a normative fluent. Consequently, if we use the framework presented in the preceding section which includes ASP rules for abstracting normative fluents, then the process of changing rules involves firstly, changing the rules, secondly determining all other normative fluent abstraction rules which need to be changed or added. This is a complex process and consequently we look at a simpler problem where we change rules in institutions governed by institutions, where each governance level \textit{operates at the same level of abstraction}.

To address this simpler problem, we first re--introduce a simpler version of our framework. In this version of the framework, we define \textit{multi--tier institutions}, first presented in \cite{King2015a}, which defines a tiering of institutions, each institutional tier governing the tier below, where norms are not abstracted. We overview the simpler framework in subsection~\ref{secReviMultiTierInsts}, then introduce the formal representation in subsection~\ref{secReviRepr}, then we provide the operational semantics in ASP in subsection~\ref{secReviComp}. Finally, we describe the main technique we use, Inductive Logic Programming (ILP) theory revision, in subsection~\ref{secReviILP}.

\subsection{Multi--tier Institutions}
\label{secReviMultiTierInsts}

\begin{figure}[t!]
\centering
\def\State at (#1,#2,#3,#4){
\coordinate (Pos) at (#1+0.085,#2+0.1);

\draw (#1,#2) [black][rounded corners=1pt] ++(0,0) rectangle +(0.16,0.11);
\node [above right] at (#1,#2) {$S_{#4}^{\textit{#3}}$};
}

\def\InstEvo at (#1,#2,#3,#4){

\coordinate (S0) at (#1+-0.185,#2+0.6);
\State at (#1+-0.185,#2+0.6,#3,0)
\draw (S0) [->, very thick] +(0.16,0.05) -- +(0.485,0.05);
\node [above] at (#1+0.15,#2+0.65) {Events$_{0}^{\textit{#3}}$};

\coordinate (S1) at (#1+0.3,#2+0.6);
\State at (#1+0.3,#2+0.6,#3,1)
\draw (S1) [->, very thick] +(0.16,0.05) -- +(0.485,0.05);
\node  [above] at (#1+0.635,#2+0.65) {Events$_{1}^{\textit{#3}}$};

\coordinate (S2) at (#1+0.785,#2+0.6);
\State at (#1+0.785,#2+0.6,#3,2);
\draw (S2) [->, very thick] +(0.16,0.05) -- +(0.485,0.05);
\draw (#1,#2) [fill][white] +(1.05,0.6) rectangle +(1.15,0.7);
\node [above, font=\bf] at (#1+1.1,#2+0.6) {$...$};

\State at (#1+1.27,#2+0.6,#3,k+1);

\node [above right,font=\bf] at (#1+-0.25, #2+0.8) {#4};
\draw (#1,#2)[black][thick,rounded corners=3pt] +(-0.25,0.575) rectangle +(1.5,0.78);
}

\def\ObsEvent at (#1,#2,#3,#4){
\coordinate (Pos) at (#1+0.085,#2+0.1);
\draw 		[thick, ->] (Pos) arc (180:-55:0.05);

\draw (#1,#2) [black][rounded corners=1pt] ++(0,0) rectangle +(0.16,0.1);
\node [above right] at (#1,#2) {$S_{#4}^{\textit{#3}}$};
}

\def\ObsTrace at (#1,#2){

\coordinate (S0) at (#1+-0.185,#2+0.6);

\node [above] at (#1+0.15,#2+0.85) (ObEv0) {Obs. Event$_{0}$};
\node  [above] at (#1+0.635,#2+0.85) (ObEv1) {Obs. Event$_{1}$};
\node  [above] at (#1+1.2,#2+0.85) (ObEvk) {Obs. Event$_{k}$};
\draw [->, very thick] (ObEv0.east) -- (ObEv1.west);
\draw [->, very thick] (ObEv1.east) -- (ObEvk.west);

\draw (#1,#2) [fill][white] +(0.85,0.8) rectangle +(0.95,1);
\node [above, font=\bf] at (#1+0.9,#2+0.85) {$...$};

}
\begin{tikzpicture}[xscale=5.5,yscale=5.5]
\InstEvo at (0,1,n,Nth-tier Institution)
\InstEvo at (0,0.6,2,Second-tier Institution)
\InstEvo at (0,0.2,1,First-tier Institution)

\ObsTrace at (0,-0.4)

\fill (0.675,1.522) circle (0.13mm);
\fill (0.675,1.472) circle (0.13mm);
\fill (0.675,1.422) circle (0.13mm);

\draw [line width=1.5mm,
preaction={-triangle 90,line width=0.8mm,draw,shorten >=-1mm}] (0.675,1.0186) -- (0.675,1.1436);
\node [above right, text width = 3cm] at (0.75,1.0386) {Link};

\draw [line width=1.5mm,
preaction={-triangle 90,line width=0.8mm,draw,shorten >=-1mm}] (0.675,0.609) -- (0.675,0.734);
\node [above right, text width = 3cm] at (0.75,0.5782) {Input for all Institutions};

\node [above right,font=\bf, text width=3cm] at (1.5,1.6) {Nth-order Norms};
\node [above right,font=\bf, text width=3cm] at (1.5,1.2) {Second-order Norms};
\node [above right,font=\bf, text width=3cm] at (1.5,0.8) {First-order Norms};
\end{tikzpicture}
\vspace{-5cm}
\caption[Multi--tier institution schema]{Schematic view of a multi--tier institution model}
\label{figModularOverview}
\end{figure}
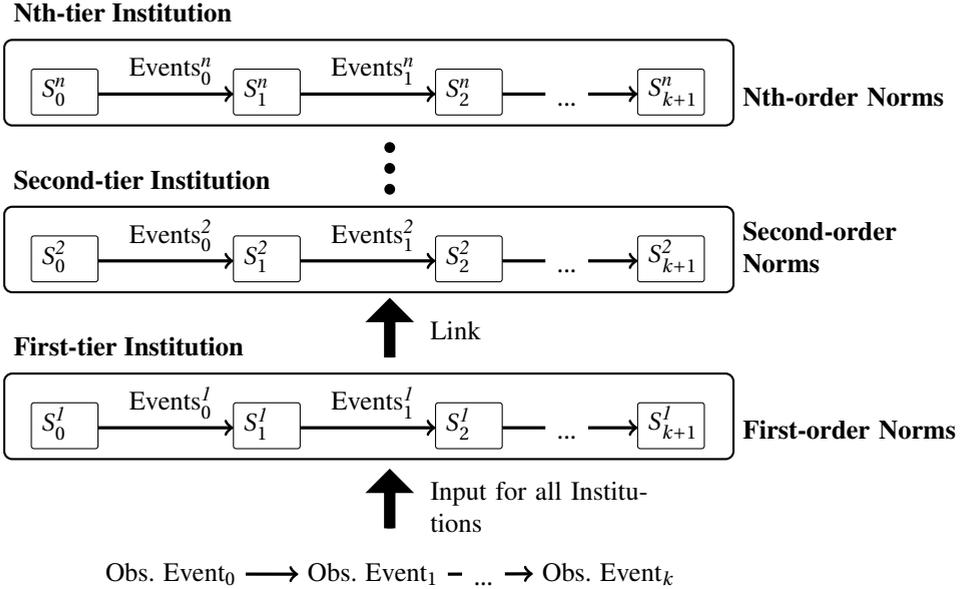

We begin by providing an overview of individual and multi--tier institutions, schematically depicted in figure~\ref{figModularOverview}. To recap the main concepts, an individual legal institution acts as a mechanism to guide the behaviour of the system it governs. Institutions define a set of constitutive and regulative rules which respectively establish an institutional description and prescription of reality (see Searle's counts--as relation \cite{Searle2005}). Constitutive rules \textit{describe} the system governed through creating institutional facts that can represent events caused by other events (e.g. entering a location which is private counts--as entering a private location), or they can represent changes to the institutional state (e.g. entering a private location causes an agent to be at a private location). Regulative rules \textit{prescribe} what properties should hold/events should occur in a system by creating obligations and prohibitions in states (e.g. when requested an agent is obliged to share their location). An institution's regulative rules regulate over a social interpretation of reality constructed from brute facts by constitutive rules.

Conceptually, a multi--tier institution extends the notion of an individual institution governing an MAS to institutions governing institutions in a tiered structure. Each institutional tier governs the tier below. The first--tier imposes norms on what occurs and holds in an MAS (first--order norms), the second--tier norms on the norms imposed by the first (norms about first--order norms, i.e. second--order norms), and so on. When a lower--tier imposes a normative fluent which is non--compliant with a higher--tier, events denoting a higher--order norm violation occur.

The operational semantics of a multi--tier institution, first presented in \cite{King2015a}, allow such non--compliance to be determined by checking a multi--tier institution model for norm violations. Depicted in Figure~\ref{figModularOverview}, the model describes how each $i$th--tier institution evolves over time, as an event--state sequence, in response to the evolution of the tier below. The first--tier evolves in response to a trace of observable events that \textit{could} occur in an MAS (i.e. produced for a pre--runtime check). Each tier above the first evolves in response to  the event--state sequence of the institution they govern (i.e. the tier below). States contain domain fluents describing the MAS and normative fluents prescribing the events that should occur and fluents that should hold in the tier below (including other normative fluents). Each state transition is caused by events occurring in the institution from the previous state, which are in turn driven by the events and states from the tier below. If a normative fluent in a state is violated by an event or fluent in the tier below (including another normative fluent) a norm violation event occurs in the transition to the next state. Thus, model--checking can be used to compliance--monitor one institution with another by checking for higher--order norm violation events.

Differing from previous chapters \ref{chapter_3} and \ref{chapter_4} on multi--level governance institutions, in this chapter \textit{multi--tier} institutions restrict each $i$th tier in regulating the tier below $i$th--order norms. Moreover, each institutional state does not include the abstraction of any concrete normative fluents. 

\subsection{Formal Representation}
\label{secReviRepr}

In this section we provide the formal representation of institutions and multi--tier institutions from \cite{King2015a} as follows. We start with the representation for normative fluents, which oblige/prohibit an event occurring or another fluent holding (the aim) before an event occurs or fluent holds (the deadline). The language of normative fluents is over a set of propositions denoting fluents and events describing the system being governed. If the set of propositions includes only non--normative events and fluents, then only first--order normative fluents can be expressed. If, however, the set of propositions contains first--order normative fluents, then second--order normative fluents can be expressed and so on. Such higher--order normative fluents are categorised as: obliging/prohibiting a normative fluent holds before an event or non--normative fluent holds, obliging/prohibiting an event or non--normative fluent before a normative fluent holds, and obliging/prohibiting a normative fluent to hold before another normative fluent holds.

\begin{definition}{\textbf{Normative Fluents}} Let $P$ be a set of propositions denoting fluents and events, $a$ be the norm's aim, $d$ the deadline and $a, d \in P$. The set of normative fluents $\pazocal{N} \vert _{P}$ is the set of all norms $n$ expressed as:
\begin{alignat*}{3}
& \textit{n} & \; \; \; ::= & \; \; \; \textit{obl}(\textit{a}, \textit{d}) \; \mid \textit{pro}(\textit{a}, \textit{d})
\end{alignat*}
\end{definition}

Institutions in this chapter share many common characteristics of institutions presented earlier. Formally, an institution is defined as and described subsequently:

\begin{definition}{\textbf{Individual Institution}}\label{def:inst}  An institution is a tuple $\pazocal{I} \vert _{P} = \langle \pazocal{E}, \pazocal{F}, \pazocal{C}, \pazocal{G}, \Delta \rangle$, restricted to the set of propositions $P$, given $\pazocal{E}_{\textit{obs}}, \pazocal{E}_{\textit{instact}},\pazocal{F}_{\textit{dom}} \subseteq P$, $\pazocal{I} \vert _{P}$ is defined as:
\begin{itemize}
\item $\pazocal{F}_{\textit{norm}} \subseteq \pazocal{N} \vert _{P}$ is a set of normative fluents.
\item $\pazocal{F} = \pazocal{F}_{\textit{dom}} \cup \pazocal{F}_{\textit{norm}}$ is a set of fluents.
\item$\begin{aligned}\pazocal{E}_{\textit{norm}} = \{ \textit{disch}(n), \textit{viol}(n) \mid n \in \pazocal{F}_{\textit{norm}} \} \end{aligned}$
\item $\pazocal{E}_{\textit{inst}} = \pazocal{E}_{\textit{instact}} \cup \pazocal{E}_{\textit{norm}}$ where $\pazocal{E}_{\textit{instact}}$ and $\pazocal{E}_{\textit{norm}}$ are disjoint.
\item $\pazocal{E} = \pazocal{E}_{\textit{obs}} \cup \pazocal{E}_{\textit{inst}}$.
\item $\pazocal{C} : \pazocal{X} \times \pazocal{E} \rightarrow 2^{\pazocal{F}} \times 2^{\pazocal{F}}$ is a state consequence function.
\item $\pazocal{G} : \pazocal{X} \times \pazocal{E} \rightarrow 2^{\pazocal{E}_{\textit{instact}}}$ is an event generation function.
\item $\Delta \subseteq \pazocal{F}$ is the initial institutional state.
\end{itemize}
\end{definition}

In more detail, an institutional specification gives the \textit{signature} of events ($\pazocal{E}$) that can occur and fluents ($\pazocal{F}$) that can hold in the institution. The signature is specified from a set of propositions $P$ to which the institution $\pazocal{I}\vert_{P}$ is restricted to (just $\pazocal{I}$ is used if $P$ is unimportant). The events $\pazocal{E}$ is the set of observable events ($\pazocal{E}_{\textit{obs}}$), and the set of institutional events ($\pazocal{E}_{\textit{inst}}$). The set of institutional events ($\pazocal{E}_{\textit{inst}}$) consists of events signifying something unrelated to a norm has happened ($\pazocal{E}_{\textit{instact}}$), or a norm has been discharged/violated ($\pazocal{E}_{\textit{norm}}$). The set of an institution's fluents ($\pazocal{F}$), describe the state of a domain ($\pazocal{F}_{\textit{dom}}$) such as that being governed (e.g. an agent is at a location), and the normative fluents ($\pazocal{F}_{\textit{norm}}$) that can hold in the institution. Each institution starts at an initial state ($\Delta$). An institution evolves from one state to the next in response to observable events, where, an institutional state is a set of fluents that are true at that point in time, $\Sigma = 2^{\pazocal{F}}$ denoting the set of all possible institutional states. Describing the events and state changes that occur are an \textit{event generation} function and \textit{state consequence} function. Both functions' arguments are a condition on a state, describing the things that must and must not hold in a state to cause the events/state change, and a set of events. A condition on a state is described with state formulae, $\pazocal{X} = 2^{\pazocal{F} \cup \neg \pazocal{F}}$ denoting the set of all state formulae, where $\neg \pazocal{F} = \{ \neg f \mid f \in \pazocal{F}\}$ is the weak negation of all fluents denoting they do not hold. The event generation function ($\pazocal{G} : \pazocal{X} \times \pazocal{E} \rightarrow 2^{\pazocal{E}_{\textit{inst}}}$) states when, conditional on a state, one event counts--as another. The consequence function ($\pazocal{C} : \pazocal{X} \times \pazocal{E} \rightarrow 2^{\pazocal{F}} \times 2^{\pazocal{F}}$) provides the fluents that are initiated and terminated by events from one state to the next. As previously, we use $\pazocal{C}(X, e) = \langle \pazocal{C}^{\uparrow}(X, e), \pazocal{C}^{\downarrow}(X, e) \rangle$ to denote the consequence function's result, where $\pazocal{C}^{\uparrow}(X, e)$ is the set of fluents initiated and $\pazocal{C}^{\downarrow}(X, e)$ is the set of fluents terminated by the event $e$ when the state entails the state condition $X$.

The approach we take to representing a multi--tier institution is to restrict each $i$th--tier institution such that it can contain events and fluents from all the tiers below for monitoring, but can only govern the tier directly below by imposing $i$thorder norms. This restriction is defined by, starting with a set of propositions $P$ which describe the MAS' events and fluents, the first--tier $\pazocal{I}^{1} \vert_{P^1}$ imposes normative fluents over the events and fluents of the MAS (i.e. first--order normative fluents) such that $P^1 = P$. Then, each $i$th--tier above the first $\pazocal{I}^{i} \vert_{P^i}$ can contain the normative fluents and events (dischargement and violation) from the tiers below for monitoring ($P^i = P^{i-1} \cup \pazocal{N} \vert_{P^{i-1}} \cup \pazocal{E}^{i-1}_{\textit{norm}}$), but each $i$th--tier is restricted in only initiating and terminating normative fluents over these (i.e. $i$thorder norms). This means, an institution can potentially also impose norms about the dischargement and violation of norms in the tier below. We leave this to the discretion of the designer, since in some cases it can make sense, for example obliging a norm is violated before an obligation to pay a fine is imposed. For monitoring, we will later define using ASP rules how each tier is linked, such that each $i$th--tier contains the normative events and normative fluents produced/imposed by the tier below. Formally, a multi--tier institution is defined as:

\begin{definition}{\textbf{Multi--tier Institution}} Let $P$ be a set of propositions denoting the domain, a multi--tier institution is a tuple $\pazocal{T} = \langle \pazocal{I}^{1} \vert _{P^{1}}, ..., \pazocal{I}^{n} \vert _{P^{n}} \rangle$ where:
\begin{itemize}
\item Each element of $\pazocal{T}$ is an individual institution s.t. \\ $\forall i \in [n] : \pazocal{I}^i \vert_{P^{i}} = \langle \pazocal{E}^i, \pazocal{F}^i, \pazocal{C}^i, \pazocal{G}^i, \Delta^{i} \rangle, \pazocal{X}^{i} = 2^{\pazocal{F}^{i} \cup \neg \pazocal{F}^{i}},$ \\ $\Sigma^{i} = 2^{\pazocal{F}^{i}}$
\item $P^{1} = P$ and $\forall i \in [2, n]$, ${P}^{i} = P^{i-1} \cup \pazocal{N} \vert_{P^{i-1}} \cup \pazocal{E}^{i-1}_{\textit{norm}}$ -- each $i$th--tier can contain the events and fluents that can be defined in the tier below and normative fluents over these.
\item $\forall i \in [2, n], \forall S \in \pazocal{X}^i, \forall e \in \pazocal{E}^{i} : \pazocal{C}^{i\uparrow}(S, e) \cap \pazocal{N}\vert_{P^{i-1}} = \emptyset, \pazocal{C}^{i\downarrow}(S, e) \cap \pazocal{N}\vert_{P^{i-1}} = \emptyset$ -- the $i$th--tier can only initiate and terminate $i$thorder norms.
\end{itemize}
\end{definition}

This concludes the representation for individual and multi--tier institutions.

\subsection{Case Study}
\label{secReviCaseStudy}

Now, we introduce our case study and demonstrate the multi--tier institution representation by formalising it. Our case study is in the context of a system for crowdsourcing audio data from users using specialised cellphone apps, called a soundsensing system \cite{Lu2009}. Our case study is for a tier--1 soundsensing institution (SS) which governs users contributing audio data using cellphone apps, and for the tier--2 governmental institution (GI) which regulates the regulations of the soundsensing system away from placing unacceptable limits on user's rights. The case study is formalised in table~\ref{tabFormalTier1Example} and table~\ref{tabFormalTier2Example}. We follow the convention that upper--case symbols stand for variables. For brevity we leave out the set of events and fluents for each institution. Both institutions consist of rules describing the domain (e.g. an agent entering a new location causes the agent to be at that location) and consider the location `street b' to be private and the agent `Bertrand' to be a child (see the initial states). 

The soundsensing institution is described as follows and formalised in Table~\ref{tabFormalTier1Example}:

\textbf{Soundsensing Tier--1 Institution}
\begin{itemize}
\item Entering a new location causes an agent to leave the previous location (\ref{exGenSSI1}).
\item In general when a norm is violated a generic norm violation event occurs (\ref{exGenSSI2}, \ref{exGenSSI3} and \ref{exGenSSI4}).
\item Due to a designer error, leaving a location also causes a generic norm violation event (\ref{exGenSSI5}).
\item A user entering a location which is private counts--as entering a private location (\ref{exGenSSI6}).
\item Likewise, leaving a location which is private counts--as leaving a private location (\ref{exGenSSI7}).
\item Entering a new location causes a user to be at that location (\ref{exConISSI1}).
\item Likewise, leaving a location means the user is no longer at their previous location (\ref{exConISST1}).
\item When a user turns 14 years of age, they are no longer considered a child (\ref{exConISST2}) as defined in real--world regulations pertaining to the retention of children's data\cite{COPPA1998}. 
\item Users are obliged to provide their location on request to give the collected data location context (\ref{exConISSI2}).
\item If a user violates a norm they are obliged to pay a fine (\ref{exConISSI3}).
\item Users are forbidden from turning their microphone off to ensure data is collected continuously (\ref{exSSIS}).
\end{itemize}

In turn, the soundsensing institution is governed by a tier--2 governmental institution designed to meet different aims (e.g. maintaining agents' rights). It is formalised in table~\ref{tabFormalTier2Example} and described as follows:

\textbf{Governmental Tier--2 Institution} 
\begin{itemize}
\item Entering a new location causes an agent to leave the previous location (\ref{exGenGI1}).
\item When a first--order norm is violated a generic norm violation event occurs (\ref{exGenGI2} and \ref{exGenGI3}).
\item Entering a location causes an agent to be at that location (\ref{exConIG1}).
\item Likewise, leaving a location causes an agent to no longer be at that location (\ref{exConTG1}).
\item If someone turns 14 years of age then they are no longer a child (\ref{exConTG2}), again, based on real--world regulations pertaining to the retention of children's data\cite{COPPA1998}.
\item It is obliged that fines are only imposed on users after they violate a norm. In the initial state (\ref{exGIIS}) there is an obligation for a norm violation to occur before an obligation to pay a fine is imposed. Naturally, the obligation for a norm violation to occur first does not hold once a fine is imposed. If there is an obligation to pay a fine and that obligation is discharged (i.e. the fine is paid), then the obligation for a norm to be violated before a fine is imposed is reinstated (\ref{exConIG2}).
\item When a user is in an area that forbids audio recording (i.e. a private area), it is forbidden to forbid them from turning their microphone off (\ref{exConIG3}).
\item It is forbidden to oblige children (users under the age of 14) to share their location (\ref{exGIIS}). Similar regulations can be found in the United States Government's Child Privacy and Protection Act \cite{COPPA1998}.
\end{itemize}

This concludes the formalisation of our case study. In the next section we will introduce the computational framework we use in this chapter and present the results of using the computational framework to check the compliance of the soundsensing institution with the governmental institution.

\begin{table}[h!]
\begin{spacing}{0.1}
\begin{longtable}[H]
{%
     >{\collectcell\myalign}%
      m{0.05\textwidth}%
     <{\endcollectcell}%
     >{\collectcell\myalign}%
      m{0.2\textwidth}%
     <{\endcollectcell}%
}
\caption[Soundsensing Institution formalisation]{\textbf{Soundsensing Institution Formalisation}}
\label{tabFormalTier1Example}
\endfirsthead
\endhead
{
\pazocal{G}^{\textit{SS}}(\{ \textit{at}(\textit{Loc0}, \textit{Ag0}) \}, \textit{enter}(\textit{Loc1}, \textit{Ag0})) \ni \textit{leave}(\textit{Loc0}, \textit{Ag0})
} &
{
\tag{SSIG.1}\label{exGenSSI1}
} \\
{
\pazocal{G}^{\textit{SS}}(&\emptyset, \textit{viol(obl(share\_location(Ag0),} \textit{leave(Ag0, Loc0)))}) \ni \\
& \textit{norm\_violation(Ag0)}
} &
{
\tag{SSIG.2}\label{exGenSSI2}
} \\
{
\pazocal{G}^{\textit{SS}}(&\emptyset, \textit{viol(obl(share\_location(Ag0),} \textit{leave(Ag0, Loc0)))}) \ni \\ 
& \textit{norm\_violation(Ag0)}
} &
{
\tag{SSIG.3}\label{exGenSSI3}
} \\
{
\pazocal{G}^{\textit{SS}}(&\emptyset, \textit{viol(pro(microphone\_off(Ag0),} \textit{leave\_soundsensing(Ag0)))}) \ni \\ &
\textit{norm\_violation(Ag0)} 
} &
{
\tag{SSIG.4}\label{exGenSSI4}
}
\end{longtable}
\end{spacing}
\end{table}

\begin{table}[h!]
\begin{spacing}{0.1}
\begin{longtable}[H]
{%
     >{\collectcell\myalign}%
      m{0.05\textwidth}%
     <{\endcollectcell}%
     >{\collectcell\myalign}%
      m{0.2\textwidth}%
     <{\endcollectcell}%
}
\endfirsthead
\endhead
{
\pazocal{G}^{\textit{SS}}(& \emptyset, \textit{enter(Ag0, Loc0)}) \ni \textit{norm\_violation(Ag0)}
} &
{
\tag{SSIG.5}\label{exGenSSI5}
} \\
{
\pazocal{G}^{\textit{SS}}( & \{ \textit{private(Loc0)} \}, \textit{enter(Ag0, Loc0)}) \ni \textit{enter\_private(Ag0)}
} &
{
\tag{SSIG.6}\label{exGenSSI6}
} \\
{
\pazocal{G}^{\textit{SS}}(& \{ \textit{private(Loc0)} \}, \textit{leave(Ag0, Loc0)}) \ni \textit{leave\_private(Ag0)}
} &
{
\tag{SSIG.7}\label{exGenSSI7}
} \\
{
\pazocal{C}^{SS\uparrow}(\emptyset, \textit{enter(Loc0, Ag0)}) \ni \textit{at(Ag0, Loc0)} 
} &
{
\tag{SSICI.1}\label{exConISSI1}
} \\
{
\pazocal{C}^{SS\uparrow}(&\emptyset, \textit{request\_location(Ag0)}) \ni \\ & \textit{obl(share\_location(Ag0), leave(Ag0, Loc0))}
} &
{
\tag{SSICI.2}\label{exConISSI2}
} \\
{
\pazocal{C}^{SS\uparrow}(&\emptyset, \textit{norm\_violation(Ag0)}) \ni \\ & \textit{obl(pay\_fine(Ag0), leave\_soundsensing(Ag0))}
} &
{
\tag{SSICI.3}\label{exConISSI3}
} \\
{
\pazocal{C}^{SS\downarrow}(& \emptyset, \textit{leave(Loc0, Ag0)}) \ni \textit{at}(Loc0, Ag0)
} &
{
\tag{SSICT.1}\label{exConISST1}
} \\
{
\pazocal{C}^{SS\downarrow}(&\{ \textit{child(Ag0)} \}, \textit{birthday(Ag0, 14)}) \ni \textit{child(Ag0)}
} &
{
\tag{SSICT.2}\label{exConISST2}
} \\
{
\Delta^{\textit{SS}} = \{ & \textit{private(street\_b)}, \textit{at(ada, street\_b)}, \\ & \textit{at(bertrand, street\_c)}, \textit{child(bertrand)}, \\ &
\textit{pro(microphone\_off(Ag0),}\textit{leave\_soundsensing(Ag0))} \}
} &
{
\tag{SSIIS}\label{exSSIS}
}
\end{longtable}
\end{spacing}
\end{table}

\begin{table}[h!]
\begin{spacing}{0.1}
\begin{longtable}[T]
{%
     >{\collectcell\myalign}%
      m{0.05\textwidth}%
     <{\endcollectcell}%
     >{\collectcell\myalign}%
      m{0.2\textwidth}%
     <{\endcollectcell}%
}
\caption[Governmental Institution formalisation]{\textbf{Governmental Institution Formalisation}}
\label{tabFormalTier2Example}
\endfirsthead
\endhead
{
\pazocal{G}^{\textit{GI}}(\{ \textit{at}(\textit{Loc0}, \textit{Ag0}) \}, \textit{enter}(\textit{Loc1}, \textit{Ag0})) \ni \textit{leave}(\textit{Loc0}, \textit{Ag0})
} &
{
\tag{GGI.1}\label{exGenGI1}
} \\
{
\pazocal{G}^{\textit{GI}}(&\emptyset, \textit{viol(obl(share\_location(Ag0)}, \textit{leave(Ag0, Loc0)))}) \ni \\ &
\textit{norm\_violation(Ag0)}
} &
{
\tag{GGI.2}\label{exGenGI2}
} \\
{
\pazocal{G}^{\textit{GI}}(& \emptyset, \textit{viol(pro(microphone\_off(Ag0),} \textit{leave\_soundsensing(Ag0)))}) \ni \\ & \textit{norm\_violation(Ag0)}
} &
{
\tag{GGI.3}\label{exGenGI3}
} \\
{
\pazocal{C}^{GI\uparrow}(\emptyset, \textit{enter(Loc0, Ag0)}) \ni \textit{at(Ag0, Loc0)}
} &
{
\tag{CIGI.1}\label{exConIG1}
} \\
{
\pazocal{C}^{GI\uparrow}(&\emptyset, \textit{disch(obl(pay\_fine(Ag0),} \textit{leave\_soundsensing(Ag0)))}) \ni \\ & \textit{obl(norm\_violation(Ag0),} \\ &
\; \; \; \; \; \; \; \; \; \; \textit{obl(pay\_fine(Ag0),}\textit{leave\_soundsensing(Ag0)))}
} &
{
\tag{CIGI.2}\label{exConIG2}
} \\
{
\pazocal{C}^{GI\uparrow}( & \{ \textit{private(Loc0)} \}, \textit{enter(Loc0)}) \ni \\ & \textit{pro(pro(microphone\_off(Ag0),} \textit{leave\_soundsensing(Ag0)),} \\ & 
\; \; \; \; \; \; \; \; \; \; \textit{leave(Loc0))}
} &
{
\tag{CIGI.3}\label{exConIG3}
} \\
{
\pazocal{C}^{GI\downarrow}(\emptyset, \textit{leave(Loc0, Ag0)}) \ni \textit{at}(Loc0, Ag0)
} &
{
\tag{CTGI.1}\label{exConTG1}
} \\
{
\pazocal{C}^{GI\downarrow}(\{ \textit{child(Ag0)} \}, \textit{birthday(Ag0, 14)}) \ni \textit{child(Ag0)}
} &
{
\tag{CTGI.2}\label{exConTG2}
} \\
{
\Delta^{\textit{GI}} = \{ & \textit{obl(norm\_violation(Ag0),} \\ &
\; \; \; \; \; \; \; \; \; \; \textit{obl(pay\_fine(Ag0),} \textit{leave\_soundsensing(Ag0))}, \\ &
\textit{pro(obl(share\_location(bertrand),}  \textit{leave(Ag0, Loc0)),} \\ & 
\; \; \; \; \; \; \; \; \; \; \textit{birthday(bertrand, 14))} \} \cup \Delta^{\textit{SS}}
} &
{
\tag{GIIS}\label{exGIIS}
}
\end{longtable}
\end{spacing}
\end{table}

\clearpage

\subsection{Multi--Tier Institution Operationalisation in ASP}
\label{secReviComp}

The formal representation for multi--tier institutions is complemented by a corresponding ASP operationalisation, much like in chapter~\ref{chapter_4}, for automatic compliance checking of lower--tier institutions with higher--tier institutions. The ASP program, comprises rules using the familiar $\asp{initiated}(\textit{p, In, }\asp{I)}$, $\asp{terminated}(\textit{p, In, }\asp{I)}$, $\asp{occurred}(\textit{e, In, }\asp{I)}$, \\$\asp{holdsat}(\textit{f, In, }\asp{I)}$, $\textit{start}(I)$ and $\textit{instant}(I)$ literals outlined previously, as well as the set of state formula literals $\textit{EX(X, In, I)}$.

The ASP program comprises several components which we refer to later in this chapter. Firstly, programs presented in the preceding chapter:
\begin{itemize}
\item A program representing a trace of observable events used as input for producing multi--tier models, the timeline program  $\Pi^{\textit{trace}(k)}$ given previously in definition~\ref{defASPCompTraceTrans}.
\item an implementation of the operational semantics, the reasoning program $\Pi^{\textit{base}(k)}$ given previously in definition~\ref{defASPMLSemTrans}.
\end{itemize}

Secondly, two new types of ASP program:

\begin{itemize}
\item Programs $\Pi^{\pazocal{I}^{i}}$ for each individual institution $\pazocal{I}^{i}$ in the multi--tier institution $\pazocal{M}$ according to the translation given in the subsequent definition~\ref{defASPMTIndInstTrans}.
\item A program $\Pi^{\textit{mtreas}}$, given in subsequent definition~\ref{defASPMTProgram}. It represents the filtering functions linking the individual institutions in a multi--tier institution, thereby ensuring normative fluents and norm compliance events are passed from lower--tiers to higher--tiers.
\end{itemize}

The new types of ASP program are defined as follows. First, the programs $\Pi^{\pazocal{I}^{i}}$ for representing each individual institution in a multi--tier institution in ASP. Defined and described subsequently as:

\begin{definition}\label{defASPMTIndInstTrans}{\textbf{Multi--tier Institution ASP Translation}} Let $\pazocal{T} = \langle \pazocal{I}^{1} \vert _{P^{1}}, ..., \pazocal{I}^{n} \vert _{P^{n}} \rangle$ be a multi--tier institution. The programs $\Pi^{\pazocal{I}^{i}}$ are defined for each individual institution $\pazocal{I}^{i}$ according to the following translation: $\forall i \in [1, n], \; 
(\pazocal{I}^{i} = \langle \pazocal{E}^{i}, \pazocal{F}^{i}, \pazocal{C}^{i}, \pazocal{G}^{i}, \Delta^{i} \rangle)$:
\begin{spacing}{0.1}
\begin{longtable}[T]
{%
     >{\collectcell\myalign}%
      m{0.9\textwidth}%
     <{\endcollectcell}%
}
{
\pazocal{I}^{i} & \Leftrightarrow &&
\asp{tier(}\textit{In}\asp{, }\textit{i}\asp{).}\asp{inst(}\textit{In}\asp{).} \in \Pi^{\pazocal{I}^{i}} && \tag{IT.1}\label{eqIT1} \\
e \in \pazocal{E}_{\textit{obs}}^{i} & \Leftrightarrow && \asp{evtype(e,} \textit{In} \asp{,ex).} \in \Pi^{\pazocal{I}^{i}} && \tag{IT.2}\label{eqIT2} \\
e \in \pazocal{E}_{\textit{instact}}^{i} & \Leftrightarrow && \asp{evtype(e,} \textit{In} \asp{,in).} \in \Pi^{\pazocal{I}^{i}} && \tag{IT.3}\label{eqIT3} \\
f \in \pazocal{F}^{i} & \Leftrightarrow && \asp{ifluent(f, }\textit{In}\asp{).} \in \Pi^{\pazocal{I}^{i}} && \tag{IT.4}\label{eqIT4} \\
\pazocal{C}^{i\uparrow}(X,e) = P & \Leftrightarrow &&
\forall p \in P : \asp{initiated(} p\asp{, } \textit{In} \asp{, I) :- } && \\ 
& && \; \; \; \; \; \; \; \; \; \; \; \; \; \; \; \; \; \; \; \; \; \; \; \asp{occurred(} e\asp{, } \textit{In} \asp{, I),} && \tag{IT.5}\label{eqIT5} \\ 
& && \; \; \; \; \; \; \; \; \; \; \; \; \; \; \; \; \; \; \; \; \; \; \; \textit{EX(X, \textit{In}, I)} \asp{.} \in \Pi^{\pazocal{I}^{i}} &&} \\
{\pazocal{C}^{i\downarrow}(X,e) = P & \Leftrightarrow &&
\forall p \in P : \asp{terminated(} p\asp{, }\textit{In} \asp{, I) :- } && \\ 
& && \; \; \; \; \; \; \; \; \; \; \; \; \; \; \; \; \; \; \; \; \; \; \; \asp{occurred(} e\asp{,} In\asp{,} I \asp{),} && \tag{IT.6}\label{eqIT6} \\ 
& && \; \; \; \; \; \; \; \; \; \; \; \; \; \; \; \; \; \; \; \; \; \; \; \textit{EX(X, \textit{In}, I)} \asp{.} \in \Pi^{\pazocal{I}^{i}} && \\
\pazocal{G}^{i}(X,e) = E & \Leftrightarrow && \forall e^{\prime} \in E : \asp{occurred(} e^{\prime}\asp{, }\textit{In} \asp{, I) :- } && \\
& && \; \; \; \; \; \; \; \; \; \; \; \; \; \; \; \; \; \; \; \; \; \; \; \asp{occurred(} e\asp{, }\textit{In}\asp{, I),} && \tag{IT.7}\label{eqIT7} \\ 
& && \; \; \; \; \; \; \; \; \; \; \; \; \; \; \; \; \; \; \; \; \; \; \; \textit{EX(X, \textit{In}, I)} \asp{.} \in \Pi^{\pazocal{I}^{i}} && \\
f \in \Delta^{i} & \Leftrightarrow && \asp{holdsat(} f\asp{, } \textit{In}, \asp{I) :- start(I).} \in \Pi^{\pazocal{I}^{i}} &&
\tag{IT.8}\label{eqIT8}
}
\end{longtable}
\end{spacing}
\end{definition}

In more detail, the tier each institution operates at is declared as a fact (\ref{eqIT1}). Events and fluents are declared as facts as usual (\ref{eqIT2} to \ref{eqIT4}). Fluents are initiated (\ref{eqIT5}) and terminated (\ref{eqIT6}) as usual according to the state consequence function. Likewise, events occur as usual according to the event generation function (\ref{eqIT7}). Finally, the initial state is represented with \asp{holdsat/3} atoms (\ref{eqIT8}).

The links between tiers, for `passing up' normative fluents and norm compliance events from lower--tiers to higher--tiers for compliance checking are defined as the following ASP program:

\begin{definition}\label{defASPMTProgram}{\textbf{Multi--tier Links ASP Program}} The program $\Pi^{\textit{mtreas}}$ is the following ASP program:
\allowdisplaybreaks[1]
\begin{subequations}
\begin{align*}
\begin{aligned}
\asp{occurred(disch(obl(A, D)), HIn, I) :-} & \asp{occurred(disch(obl(A, D)), LIn, I),} \\
& \asp{tier(LIn, I_1),} \asp{tier(HIn, I_2),} \\
& \asp{I_2 == I_1 + 1.}
\end{aligned} \tag{MTS.1}\label{exMTS1} \\
\begin{aligned}
\asp{occurred(disch(pro(A, D)), HIn, I) :-} & \asp{occurred(disch(pro(A, D)), LIn, I),} \\
& \asp{tier(LIn, I_1),} \asp{tier(HIn, I_2),} \\
& \asp{I_2 == I_1 + 1.}
\end{aligned} \tag{MTS.2}\label{exMTS2} \\
\begin{aligned}
\asp{occurred(viol(obl(A, D)), HIn, I) :-} & \asp{occurred(viol(obl(A, D)), LIn, I),} \\
& \asp{tier(LIn, I_1),} \asp{tier(HIn, I_2),} \\
& \asp{I_2 == I_1 + 1.}
\end{aligned} \tag{MTS.3}\label{exMTS3} \\
\begin{aligned}
\asp{occurred(viol(pro(A, D)), HIn, I) :-} & \asp{occurred(viol(pro(A, D)), LIn, I),} \\
& \asp{tier(LIn, I_1),} \asp{tier(HIn, I_2),} \\
& \asp{I_2 == I_1 + 1.}
\end{aligned} \tag{MTS.4}\label{exMTS4} \\
\begin{aligned}
\asp{holdsat(obl(A, D), HIn, I) :-} & 
\asp{holdsat(obl(A, D), LIn, I),} \\
& \asp{tier(LIn, I_1),} \asp{tier(HIn, I_2),} \\
& \asp{I_2 == I_1 + 1, start(I).}
\end{aligned} \tag{MTS.5}\label{exMTS5} \\
\begin{aligned}
\asp{holdsat(pro(A, D), HIn, I) :-} & 
\asp{holdsat(pro(A, D), LIn, I),} \\
& \asp{tier(LIn, I_1),} \asp{tier(HIn, I_2),} \\
& \asp{I_2 == I_1 + 1, start(I).}
\end{aligned} \tag{MTS.6}\label{exMTS6} \\
\begin{aligned}
\asp{initiated(obl(A, D), HIn, I) :-} & 
\asp{initiated(obl(A, D), LIn, I),} \\
& \asp{tier(LIn, I_1),} \asp{tier(HIn, I_2),} \\
& \asp{I_2 == I_1 + 1, start(I).}
\end{aligned} \tag{MTS.7}\label{exMTS7} \\
\begin{aligned}
\asp{initiated(pro(A, D), HIn, I) :-} & 
\asp{initiated(pro(A, D), LIn, I),} \\
& \asp{tier(LIn, I_1),} \asp{tier(HIn, I_2),} \\
& \asp{I_2 == I_1 + 1, start(I).}
\end{aligned} \tag{MTS.8}\label{exMTS8} \\
\begin{aligned}
\asp{terminated(obl(A, D), HIn, I) :-} & 
\asp{terminated(obl(A, D), LIn, I),} \\
& \asp{tier(LIn, I_1),} \asp{tier(HIn, I_2),} \\
& \asp{I_2 == I_1 + 1, start(I).}
\end{aligned} \tag{MTS.9}\label{exMTS9} \\
\begin{aligned}
\asp{terminated(obl(A, D), HIn, I) :-} & 
\asp{terminated(obl(A, D), LIn, I),} \\
& \asp{tier(LIn, I_1),} \asp{tier(HIn, I_2),} \\
& \asp{I_2 == I_1 + 1, start(I).}
\end{aligned} \tag{MTS.10}\label{exMTS10}
\end{align*}
\end{subequations}
\end{definition}

This concludes our short introduction of multi--tier institution operationalisation in ASP.

\subsection{Executed Case Study}
	
\begin{figure*}[t!]
\begin{center}
\input{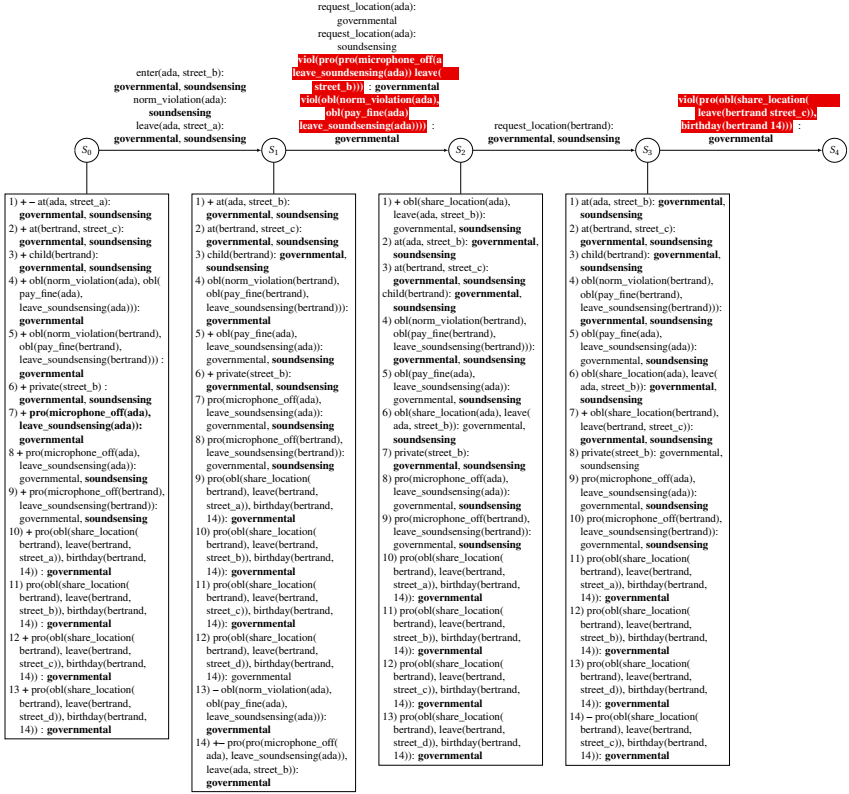}
\end{center}
\caption[Example trace as input for finding non--compliance explanations]{Results for a trace, the originating institutions for an event or a fluent are in \textbf{bold}, newly initiated fluents are denoted with a \textbf{+} and terminated fluents which will cease to hold in the next state are denoted with \textbf{--}. Violated second--order norms in the second--tier governmental institution are \textbf{\color{white}{\hlr{highlighted}}}.}
\label{figReviExampleTrace}
\end{figure*}

In this section we execute our case study, comprising the soundsensing institution which is governed by a governmental institution, formalised as a multi--tier institution. That is by executing the multi--tier institutions' corresponding ASP programs, $\Pi^{\pazocal{I}^{\textit{SS}}}$, $\Pi^{\pazocal{I}^{\textit{GI}}}$, together with the program linking them $\Pi^{\textit{mtreas}}$, the base semantics program $\Pi^{\textit{base(k)}}$ and an observable trace program $\Pi^{\textit{trace(k)}}$, in the last two cases for a trace of observable events of length $k$. In our case, the trace program comprises an event sequence with three elements, represented as the following ASP facts:

\begin{lstlisting}[mathescape]
compObserved(ex_enter(ada, street_b), 0).
compObserved(ex_request_location(ada), 1).
compObserved(ex_request_location(bertrand), 2).
\end{lstlisting}

In this short trace, first Ada enters a new location, \asp{street\_b}, then Ada's location is requested and finally Bertrand's location is requested. For this short trace, three cases of non--compliance of the soundsensing institution with the governmental institution are discovered, as depicted in figure~\ref{figReviExampleTrace}.

Firstly, when Ada enters a new location due to an erroneous rule in the soundsensing institution (a designer error), a generic norm violation event occurs in the soundsensing institution (the first state transition). Due to the generic norm violation event occurring, the soundsensing institution imposes an obligation for Ada to pay a fine in state $S_1$. However, the governmental institution only recognises actual norm violation events as causing a generic norm violation event and obliges an actual norm is violated before an obligation to pay a fine is imposed. Consequently, the governmental institution's higher--order norm for there to be a norm violation before an obligation to pay a fine is imposed is violated in the transition from $S_1$ to $S_2$.

Secondly, in the soundsensing institution agents are unconditionally prohibited from turning their microphone off, however the governmental institution prohibits such a prohibition when an agent enters a private 
location.  When Ada enters a private location ($\asp{street\_b}$) there is still a prohibition on her from turning her microphone off. Consequently, the prohibition on prohibiting Ada from turning her microphone off when she is in a private location is violated during the transition from state $S_1$ to $S_2$.

Thirdly, when an agent is requested to provide their location the soundsensing institution obliges them to do so, but this is forbidden by the governmental institution if the agent is a child, such as Bertrand. In state $S_3$ the soundsensing institution obliges Bertrand, a child, to share his location. Consequently, in the final state transition the governmental institution's prohibition on obliging Bertrand to share his location is violated. This concludes our presentation of non--compliance for the case study in this chapter. In the final part of this section we will discuss the approach we take to automatically resolving non--compliance, by using Inductive Logic Programming.

\subsection{Inductive Logic Programming: A brief overview}
\label{secReviILP}

We view the problem of revising lower--tier institutions to be compliant with higher--tier institutions as a theory revision (TR) problem which can be solved using Inductive Logic Programming (ILP). ILP \cite{Muggleton95} is a machine learning technique concerned with the induction of logic theories that generalise (positive and negative) examples with respect to a prior background knowledge. In non--trivial problems it is crucial to define the search space accurately. This is done by a {\em language bias}, that can be expressed using the notion of mode declarations \cite{Muggleton95}, describing the structure of the elements in the target theory. In the case presented here, we want to find the ASP rules that contain certain elements in the head and body. So we will have head and body mode declarations.

An {\em ILP theory revision task\/} is a tuple $\langle P, B, M \rangle$  where $P$ is a set of conjunctions of literals,  called {\em properties},  $B$ is a normal program, called the {\em background theory}, $M$ is a set of mode declarations describing the form that rules in the revised theory can take and $s(M)$ is the set of rules adhering to $M$. A theory $H$, called a {\em hypothesis}, is an inductive solution for the task $\langle P, B, M \rangle$, if (i)~$H \subseteq s(M)$,  and (ii)~$P$ is true in all the answer sets of $B \cup H$.

Our approach to revising a lower--tier institution in a multi--tier institution to be compliant is based on the introduction of new rules, and deleting and revising existing ones. As discussed in \cite{Corapi2009}, non--monotonic inductive logic programming can be used to revise an existing theory. The key concept is that of {\em minimal revision}. In general, a TR system is biased towards  the computation of theories that are similar to a given revisable theory. The difference between two programs $T$ and $T'$ is denoted as $c(T,T')$.

The theory $T'$, called a {\em revised theory}, is a {\em TR solution} for the task $\langle P, B, T, M \rangle$ with distance $c(T, T')$, iff
\begin{inparaenum}[(i)]
\item \label{lab:cond1}$T' \subseteq s(M)$,
\item \label{lab:cond2}$P$ is true in all the answer sets of $B \cup T'$,
\item if a theory $S$ exists that satisfies conditions (\ref{lab:cond1}) and (\ref{lab:cond2}) then $c(T, S) \geq c(T, T')$, (i.e. minimal revision).
\end{inparaenum}

\section{Revising Institutions For Compliance}
\label{secReviForCompliance}

In this section we give the details of this chapter's main contribution: a system for revising a lower--tier institution to be compliant with a higher--tier institution in a multi--tier institution. In particular, we are interested in revising the institution which the system user (an institutional designer) has the power to effect change. We call this institution to be revised a \textit{mutable} institution. We are interested in revising a mutable institution to meet two properties:

\begin{itemize}
\item \textbf{Success} meaning that a formerly non--compliant institution for an event--trace is compliant for the same event trace after being revised. This means when normative fluents are obliged to be imposed they are, and conversely any prohibited normative fluents are not imposed.
\item \textbf{Minimality} is a requirement for any revision to minimise the change in \textit{consequences} of the new institution compared to the old one. That is, following changes to the institution the institution's states are as close as possible to the states prior to the change(s) for a trace of events. To give an example, the soundsensing institution prohibits agents to turn their cellphone microphone off, whilst the governmental institution prohibits such a prohibition in areas deemed `private'. In this case, an institution revision can be successful by removing the soundsensing institution's prohibition altogether, but only successful \textit{and} minimal by removing the prohibition in just those cases where an agent is at a private location.
\end{itemize}

We instantiate the problem of revising a mutable institution as an ILP theory revision task in Section~\ref{secReviILP}. Then, we take a computational approach to solving the ILP theory revision task by performing \textit{inductive} search in ASP \cite{Corapi2010}. Inductive search is achieved by transforming the mutable institution represented in ASP to an ASP representation encoding the space of ILP theory revisions and enabling different revisions to be tried. We describe our computational approach using ASP in Section~\ref{secReviASP}, and the implementation and revision results for our case study in Section~\ref{secRevImpl}.

\subsection{Revising Institutions to be Compliant is an ILP Theory Revision Task Instance}

In this section, we define the revision for compliance task as an ILP revision task according to the revision for compliance requirements outlined previously. We begin by formally defining the search space of possible revisions with \textit{mode declarations}. Mode declarations define the literals that can appear in the head and body of rules. In the case of revising a mutable institution in a multi--tier institution, the mode declarations describe the valid rules for: generating non--normative institutional events, initiating and terminating domain fluents, and given the mutable institution is the $i$th--tier, initiating and terminating $i$thorder normative fluents (i.e. restricted to only initiating/terminating a normative fluent $f$ if it is not in the language of norms $\pazocal{N}\vert_{\textit{Pr}^{i-1}}$ of the tier $i-1$ below).

\begin{definition}\textbf{Mode Declarations.} Let $\pazocal{I}^{i} = \langle \pazocal{E}^{i}, \pazocal{F}^{i}, \pazocal{G}^{i}, \pazocal{C}^{i}, \Delta^{i} \rangle$ be a mutable institution for which $\textit{In}$ is a unique label. The mode declarations for $\pazocal{I}^{i}$ are a pair $M = \langle M^{h}, M^{b} \rangle$ where $M^{h}$ is the set of head mode declarations and  $M^{b}$ the set of body mode declarations, defined as:
\begin{equation*}
\begin{array}{l c l}
M^{h} & = & \{ \asp{initiated(}f, \textit{In}, \asp{I)}, \asp{terminated(}f, \textit{In}, \asp{I)} : f \in \pazocal{F} \; \backslash \; \pazocal{N} \vert _{\textit{Pr}^{i-1}} \} \cup \\ & &
\{ \asp{occurred(}e, \textit{In}, \asp{I)} : e \in \pazocal{E}^{i}_{\textit{instact}} \} \\
M^{b} & = & \{ \asp{holdsat(}f \asp{,} \textit{In}, \asp{I)}, \asp{\neg holdsat(}f \asp{,} \textit{In}, \asp{I)} : f \in \pazocal{F}^{i} \} \cup \\ & &
\{ \asp{occurred(}e, \textit{In}, \asp{I)} : e \in \pazocal{E}^{i} \}
\end{array}
\end{equation*}
\end{definition}

The set of \textit{compatible rules} with the head and body mode declarations are also required to contain one and only one event in the body. The compatible rules are defined as:

\begin{definition}\textbf{Compatible Rules.} Let $M = \langle M^{h}, M^{b} \rangle$ be the mode declarations for a mutable institution $\pazocal{I}^{i} = \langle \pazocal{E}^{i}, \pazocal{F}^{i}, \pazocal{G}^{i}, \pazocal{C}^{i}, \Delta^{i} \rangle$. An ASP rule $\asp{l_0 \; :- \; l_1, ..., l_n.}$ where $n \in \mathbb{N}$ is compatible with $M$ iff $l_0 \in M^{h}$, $\forall i \in [1, n] : l_i \in M^{b}$ and $|\{ l_1, ..., l_n \} \cap \{ \asp{occurred(}e, \textit{In}, \asp{I)} : e \in \pazocal{E}^{i} \}| = 1$. The set of all compatible rules with $M$ is $s(M)$.
\end{definition}

Having described the search space of revisions, a theory revision task \emph{TR} needs to be instantiated with the properties $P$ that a solution must meet. These properties are typically positive examples (formulae that are true following a revision) and negative examples (formulae that are false following a revision). In our case we are only interested in supplying negative examples, stating that non--compliance is eradicated in a solution to $\textit{TR}$. The negative examples in $P$ are represented as ASP integrity constraints requiring a revised mutable institution is compliant with all higher--order norms it can violate~-- including those it does not violate before revision~-- ensuring revision does not cause further non--compliance.

\begin{definition}\textbf{Compliance Properties}. Let $\pazocal{I}^{i}$ be a mutable institution and \\$\pazocal{I}^{i+1} = \langle \pazocal{E}^{i + 1}, \pazocal{F}^{i + 1}, \pazocal{C}^{i + 1}, \pazocal{G}^{i + 1}, \Delta^{i + 1} \rangle$ be the institution with unique name $\textit{In}^{i+1}$ governing $\pazocal{I}^{i}$ where $i \in \mathbb{N}$. The compliance properties for $\pazocal{I}^{i}$ is the set of constraints:
\begin{align*}
P = \{ \asp{:- \; occurred(viol(} n), \textit{In}^{i+1} \asp{, I), instant(I).} : n \in \pazocal{F}^{i+1}_{\textit{norm}} \}
\end{align*}
\end{definition}

We can now instantiate an ILP theory revision task, as a \textit{compliance} theory revision task in a multi--tier institution according to the previous definitions:

\begin{definition}\textbf{Compliance Theory Revision Task}. Let $\pazocal{I}^{i}$ be a mutable institution in the multi--tier institution $\pazocal{M}$. An ILP theory revision task $\textit{TR} = \langle P, B, T, M \rangle$ is a compliance theory revision task for $\pazocal{I}^{i}$ iff:
\begin{inparaenum}[(a)]
\item $P$ is a set of compliance properties for $\pazocal{I}^{i}$,
\item $B$ is the normal program comprising
\begin{inparaenum}[(i)]
\item a multi--tier reasoning program $\Pi^{\textit{mtreas}}$,
\item the timeline program $\Pi^{\textit{trace}(k)}$ and
\item the institution representation program $\Pi^{\pazocal{I}^{j}}$ for each institution $\pazocal{I}^{j}$ in $\pazocal{M}$ apart from the mutable institution $\pazocal{I}^{i}$,
\end{inparaenum}
\item $T$ is the institution representation program $\Pi^{\pazocal{I}^{i}}$ for the mutable institution $\pazocal{I}^{i}$, and (iv) $M$ is the set of mode declarations for $\pazocal{I}^{i}$.
\end{inparaenum}
\end{definition}

As outlined previously, we require solutions to theory revision to minimise the revision cost in order to remain as close to an institution designer's original intentions as possible. More precisely, the requirement is that the changed, mutable, institution's model for a composite trace contains as many similarities between states compared to before the changes were made (i.e. minimising the changes to consequences). We derive the cost of revision from the changes in consequences rather than the number of rule changes~-- as used in \cite{Li2014}~-- since due to non--monotonicity, as the changes in consequences between two versions of a mutable institution increases, the number of rule changes does \textit{not} necessarily monotonically increase. The changes in consequences are the number of added and deleted fluents in the answer set for $B \cup T$ compared to the answer--set $B \cup T^{\prime}$ for some revised institution $T^{\prime}$ (i.e. the symmetric set difference between the answer--sets for $B \cup T$ and for $B \cup T^{\prime}$).

\begin{definition}\textbf{Theory Revision Cost} Let $\textit{TR} = \langle P, B, T, M \rangle$ be a compliance theory revision task for a mutable institution $\pazocal{I}$ with unique label $\textit{In}$, $T^{\prime}$ be a solution to $\textit{TR}$, $\textit{ans}$ be the answer--set for $B \cup T$ and $\textit{ans}^{\prime}$ be the answer--set for $B \cup T^{\prime}$ and $\oplus$ be the set symmetric difference operation. The cost $c(T, T^{\prime})$ is defined as:
\begin{align*}
c(T, T^{\prime}) = \left\vert\left\lbrace
\begin{array}{ll}
	f = \asp{holdsat}(p, \textit{In}, i) : i \in \mathbb{N}, \; \; f \in \textit{ans} \oplus \textit{ans}^{\prime}
\end{array}
\right\rbrace\right\vert
\end{align*}
\end{definition}

Thus, a solution to an ILP theory revision task, instantiated as revision for compliance, $\textit{TR} = \langle P, B, T, M \rangle$ is a search problem. The search space consists of all possible revised versions of a mutable institution theory (normal program) and a solution $T^{\prime}$ to the theory revision task is one which  ensures the mutable institution is now compliant for a particular trace (meets the properties in $P$) and minimal such that there is no revised theory $S$ which is `more minimal' ($c(T, S) \geq c(T, T')$). This requires searching all possible solutions and comparing them. In order to perform this search we need a way to try different changes to $T$ and find optimal solutions. In the next subsection we show how we can use ASP to perform this search.

\subsection{Solving ILP Institution Revision in ASP}
\label{secReviASP}

Based on \cite{Li2014} we use inductive search in ASP to solve an ILP theory revision task $\textit{TR} = \langle P, B, T, M \rangle$ instantiated as institutional revision for compliance. The approach we take is to transform the theory to be revised $T$ (a mutable institution) into an ASP program where different changes to the theory can be tried (body literal and rule addition and deletion) that fit into the space of possibilities $s(M)$. This is by redefining certain predicates in rules as \textit{undefined}, rather than being a necessary element of a rule. Such undefined predicates are called \textit{abducibles}. 

We call this program, with previously fully--defined predicates now defined as abducibles, the \textit{revision} program $\Pi^{\textit{rev}}$. The background theory $B$ remains unchanged and provides both the unchangeable parts of the multi--tier institution and multi--tier reasoning. The background theory allows the effects of different revisions to be determined. The properties to be met, $P$, constrain any revisions found by the ASP program $\Pi^{\textit{rev}}$ to result in a compliant institution. The cost measure between a revisable $T$ and revised theory $T^{\prime}$, $c(T, T^{\prime})$ is encoded as an ASP optimisation statement. Computing the answer--sets for these components as a single ASP program explores the search space, with each answer--set representing an outcome (revised theory) that meets the properties $P$ and with those that minimise the difference (changes in consequences) ranked highest and presented to the user for selection. The advantage of this approach is that the representation and reasoning for the non--revisable portions of the multi--tier institution are encoded as the same ASP programs for the computational and revision framework requiring no re--implementation.

\begin{table}[t!]
\begin{tabular}{p{0.45\textwidth} p{0.45\textwidth}}
\textbf{Rules Describing Institution Changes} & \textbf{Explanation} \\ \hline
$\begin{aligned}
& l_0 \asp{:-} \; l_1, ..., l_n, \asp{rev(}\textit{In, i,} \asp{details(rDel)}) \asp{.} \\
& \asp{\{ rev(}\textit{In, i,} \asp{details(rDel))} \}.
\end{aligned}$ &
\vspace{-0.45cm} \textit{Rule deletion:} Existing rules are extended with an abducible  $\asp{rev(}\textit{In, i,} \asp{details(rDel)})$, which when included in an answer--set has the effect of deleting the rule with index $i$. \\
\hline
$\begin{aligned}
& l_0 \asp{:-} \; l_1, ..., l_{j-1}, \asp{try(}i, j, B^{+}_{-}(l_j), l_j\asp{)}, \\ & 
\; \; \; \; \; \; \; \; \; l_{j+1}, ..., l_n \asp{.} \\ &
\asp{try(}i, j, B^{+}_{-}(l_j), l_j \asp{)} \asp{ :- } \; l_j, \\ & 
\; \; \; \; \; \; \; \; \asp{not} \; \asp{rev(}\textit{In}, i, \asp{details(bDel}, j))\asp{.} \\ &
\asp{try(}i, j, B^{+}_{-}(l_j), l_j\asp{)} \asp{ :- } \; \\ & 
\; \; \; \; \; \; \; \; \asp{rev(}\textit{In}, i, \asp{details(bDel}, j))\asp{.} \\ &
\asp{\{ \asp{rev(}\textit{In}, i, \asp{details(bDel}, j)) \}}.
\end{aligned}$ &
\vspace{-1.8cm} \textit{Body literal deletion:} Each body literal $l_j$ of an existing rule is replaced with the literal $\asp{try/4}$ for trying to delete the body literal $l_j$. When the abducible $\asp{rev(}\textit{In}, i, \asp{details(bDel}, j))$ is included in an answer--set the effect is to make the try literal true and thus effectively delete the literal $l_j$, otherwise the try literal is only true when $l_j$ is true (effectively keeping $l_j$).
\\ \hline
$\begin{aligned} 
&l_0 \asp{:-} \; \asp{rev}(\textit{In, i,} \asp{details(rAdd)}), l_1, ..., l_n \asp{.} \\ &
\asp{\{ rev}(\textit{In, i,} \asp{details(rAdd))\}.}
\end{aligned}$  & \vspace{-0.45cm} \textit{Rule addition}: Including the abducible $\asp{rev}(\textit{In, i,} \asp{details(rAdd)})$ has the effect of including the rule with index $i$ in the program.
\\ \hline
$\begin{aligned}
& l_0  \asp{ :- } l_1, l_2, ..., l_n, \\ & \; \; \; \; \; \; \; \;\asp{extension(}i, l_0, l_{n+1}, B^{+}_{-}(l_{n+1})) \asp{.}\\ &
\asp{extension(}i, l_0, l_{n+1}, B^{+}_{-}(l_{n+1})) \asp{ :- } \\ &
\; \; \; \asp{not} \; \; \asp{rev}(\textit{In}, i, \asp{details(bAdd, } \; \\ & 
\; \; \; \; \; \; \; \; \; \; \; \; \; \; \; \; \; B^{+}_{-}(l_{n+1}), l_{n+1})) \asp{.} \\ &
\asp{extension(}i, l_0, l_{n+1} , B^{+}_{-}(l_{n+1})) \asp{ :- } \\ &
\; \; \; \asp{rev}(\textit{In}, i, \asp{details(bAdd, } \; \\ &
\; \; \; \; \; \; \; \; \; B^{+}_{-}(l_{n+1}), l_1), l_1 \asp{.} \\ &
\asp{\{rev}(\textit{In}, i, \asp{details(bAdd, } \\ &
\; \; \; \; \; \; \; \; \; B^{+}_{-}(l_{n+1}), l_{n+1}))\asp{\}}.
\end{aligned}$ & 
\vspace{-2.5cm} \textit{Body literal addition:} Existing rules are appended with $\asp{extension/4}$ predicates for each body mode literal a rule can be extended with. Including the abducible $\asp{rev}(\textit{In}, i, \asp{details(bAdd, } \; \textit{pos}, l_1)$ in an answer--set has the effect of extending the rule with index $i$ with the body literal $l_1$ (constraining the rule). That is, adding the revision predicate to an answer set makes the extension predicate true only when the literal with the specified variable bindings are true, effectively adding a constraint/body--literal to the rule. Otherwise, the extension predicate is always true (no constraint is tried for addition).
\end{tabular}
\caption[Abducible predicates explanation]{Explanation of how abducible revision predicates can (re--)define institutional rules for finding revisions of the institution $\textit{In}$}
\label{tabRevisableInst}
\end{table}

In order to go from a revisable theory $T$ representing a mutable institution to a revision program $\Pi^{\textit{rev}}$, we need to alter $T$ in some way such that adding new rules and changing existing rules can be tried by the new program with each answer--set corresponding to different revised theories. The approach we take, as in \cite{Corapi2010,Li2014}, is to introduce \textit{abducible} predicates which represent the different revision operations and are selected by the program for inclusion in answer--sets. If an abducible is selected for answer--set inclusion then the effect is to  perform the revision operation the abducible represents. The abducibles have the form $\asp{rev}(\textit{In}\asp{, i, details(...))}$ conveying to the user the revision operation described in $\asp{details(...)}$ (e.g. a rule deletion operation) is carried out on a rule with label $i$ in institution $\textit{In}$. To give a simple example the rule $\asp{l_0 \; :- \; l_1.}$ cannot be selected for deletion by an ASP program, but we can modify it to become $\asp{l_0 \; :- l_1, not \; rev(\textit{In}, i, details(rDel)).}$ meaning if the abducible $rev(\textit{In}, i, \asp{details(rDel)})$ is included by the program in an answer--set the effect is to delete the rule $i$ by ensuring the body is never true. The selection of revision tuples for inclusion in an answer--set is encoded in the ASP revision program using the ASP \textit{choice} construct of the form $\asp{\{ \textit{rev}(\textit{In}, i, details(...)) \}}$.

Each type of revision operation (rule and body literal addition and deletion) requires a different abducible and set of rules in the ASP revision program $\Pi^{\textit{rev}}$. In Table~\ref{tabRevisableInst} we describe the details of the different rules for trying revisions and the transformation from a revisable theory $T$ to a revision program $\Pi^{\textit{rev}}$ using $\textit{In}$ to represent an institution's name, $i$ to represent a rule identifier (e.g. an integer) and $B^{+}_{-}(l)$ to represent whether a literal $l$ is positive or negative.

Finally, the cost $c(T, T^{\prime})$ between two theories is encoded as an ASP optimisation constraint causing the ASP program to only present answer--sets that are minimal in the changes to consequences between $T \cup B$ and $T^{\prime} \cup B$, which we also extend with a secondary preference for revisions that \textit{generalise} the institution (deleting body literals and rules) rather than specialising (adding new body literals and rules). The optimisation statement is given below where $X\asp{@}n$ represents the priority $n$ of minimising the numerical value $X$, $\asp{difference/1}$ measures the difference between the states in the answer--set for the institution before and after revision (in terms of added and removed fluents for each state), $\asp{rAdd/1}$ counts the rule additions, $\asp{bAdd/1}$ the body additions, $\asp{bDel/1}$  the body deletions and $\asp{rDel/1}$ the rule deletions.

\begin{lstlisting}[mathescape]
#minimize {D@5: difference(D); RA@4: rAdd(RA); BA@3: bAdd(BA);
           BD@2: bDel(BD); RD@1: rDel(RD)}.
\end{lstlisting}

\subsection{Implementation and Results}
\label{secRevImpl}

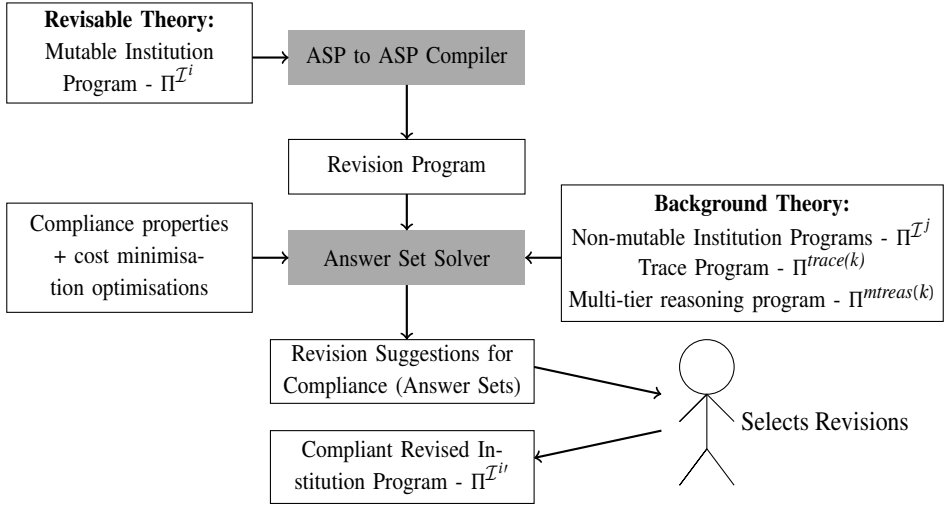
\begin{figure*}[t]
\begin{center}
\begin{tikzpicture}[xscale=0.6, yscale=0.6, every node/.style={xscale=0.9}]
\draw  (-7.4,4.4) rectangle (-2,2.2) node[pos=.5, text width=3.5cm, align=center] {\small \textbf{Revisable Theory: } \\ Mutable Institution Program - $\Pi^{\pazocal{I}^{i}}$};
\fill  [fill={rgb:black,1;white,2}] (-1.2,3.8) rectangle (4,2.6) node[pos=.5, text width=5cm, align=center] {\small ASP to ASP Compiler};

\draw  (-1.2,1.4) rectangle (4,0.2) node[pos=.5, text width=5cm, align=center] {\small Revision Program};

\draw  (-7.4,0) rectangle (-2,-2.4) node[pos=.5, text width=3.5cm, align=center] {\small Compliance properties + cost minimisation optimisations};
\fill  [fill={rgb:black,1;white,2}] (-1.2,-0.6) rectangle (4,-1.8) node[pos=.5, text width=5cm, align=center] {\small Answer Set Solver};
\draw  (4.8,0.4) rectangle (13.2,-2.6) node[pos=.5, text width=10cm, align=center] {\small \textbf{Background Theory:} \\ Non-mutable Institution Programs - $\Pi^{\pazocal{I}^{j}}$ \\ Trace Program - $\Pi^{\textit{trace(k)}}$ \\ Multi-tier reasoning program - $\Pi^{\textit{mtreas}(k)}$};

\draw  (-1.6,-3) rectangle (4.2,-4.4) node[pos=.5, text width=5cm, align=center] {\small Revision Suggestions for Compliance (Answer Sets)};
\draw  (-1.6,-5) rectangle (4.2,-6.6) node[pos=.5, text width=4cm, align=center] {\small Compliant Revised Institution Program - $\Pi^{\pazocal{I}^{i\prime}}$};

\draw  (8,-3.6) ellipse (0.6 and 0.6);
\draw (8,-4.2) coordinate (v2) {} -- (8,-5.4) coordinate (v1) {};
\draw (v1) -- (7.4,-6.2) -- (v1) -- (8.6,-6.2);
\draw (v2) -- (7.4,-4.8) -- (v2) -- (8.6,-4.8);
\node at (10.6,-4.8) {Selects Revisions};
\draw [->, thick] (-2,3.2) -- (-1.2,3.2);
\draw [->, thick] (1.4,2.6) -- (1.4,1.4);
\draw [->, thick] (1.4,0.2) -- (1.4,-0.6);
\draw [->, thick] (-2,-1.2) -- (-1.2,-1.2);
\draw [->, thick] (4.8,-1.2) -- (4,-1.2);
\draw [->, thick] (1.4,-1.8) -- (1.4,-3);
\draw [->, thick] (4.2,-3.6) -- (7,-4.2);
\draw [->, thick] (7,-5) -- (4.2,-5.6);
\end{tikzpicture}
\end{center}
\caption[Compiler for finding non--compliance explanations overview]{Overview of using the implemented compiler and the multi--tier institution framework to resolve non--compliance.}
\label{figDataflow}
\end{figure*}

A prototype system for revising a lower--tier institution to be compliant with a higher--tier is implemented according to the description in the preceding sections.\footnote{The prototype, multi--tier reasoning in ASP and the examples used in this chapter can be found at \url{https://sourceforge.net/projects/multitierinstitutionlearning/files/}.} The implementation is a compiler written in Java which, as depicted in Figure~\ref{figDataflow}, takes as input the mutable institution the institution designer has the power to effect change represented in ASP (the mutable institution program $\Pi^{\pazocal{I}^{i}}$) and outputs a revision program $\Pi^{\textit{rev}}$. The revision program is then put together with compliance properties to be met by revisions, revision cost minimisation optimisations and the background theory to remain unchanged (the non--mutable institutions, the timeline program and the multi--tier institution reasoning). An answer--set solver applied to the composition of these programs then produces minimal revision suggestions for compliance (answer sets). The suggestions are passed to a user who selects and applies a set of revisions, resulting in a compliant institution represented as an ASP program.

In addition to the system presented in this chapter, the ASP compiler also addresses an apparent lack of re--usability of institutions (e.g. using the same institution for different sets of agents) due to their propositional nature. Rather than taking just propositional institutions as input, the compiler also takes \textit{first--order} institution theories containing variables in the head and body of rules, together with bindings and monadic predicates to denote types. To give an example, $\asp{agent(ada)}$ denotes $\asp{ada}$ is of type agent and $\asp{agent(X)}$ denotes the variable $\asp{X}$ is any ground term of type agent. Thus, a designer does not need to write a new propositional institution for the case where a new agent, Charles, joins the institutionalised society with all the norms and domain fluents that are about Charles. Instead, a fact $\asp{agent(charles)}$ can be added stating Charles is of type agent. In turn, the compiler takes these more re--usable first--order institution theories as input and outputs a first--order institution revision program that tries different variable bindings between head and body literals' variables of the same type.

For our case study, we have used our prototype compiler to produce a revision program for sub--sets of the compliance problem. That is, dividing the program up into smaller parts and resolving one case of non--compliance at a time for tractability, and testing all revision suggestions together at the end to confirm they are consistent. Some of the minimal and successful revisions found are given below (we keep to those we find most intuitive).

The first change suggested addresses the issue of non--compliance due to an obligation to pay a fine being imposed by the tier--1 soundsensing institution when an agent enters a new area. Non--compliance occurs, because an agent entering a new area triggers a norm violation event in the first tier institution regardless of whether a norm has been violated, whilst the second tier obliges that a norm is genuinely violated before a fine is imposed. The revision suggestion is to delete the rule in the first--tier soundsensing institution causing a norm violation event to occur when an agent enters an area:

\begin{lstlisting}[mathescape]
$\text{\sout{occurred(norm\_violation(Agent0), soundsensing, I) :- }}$
      $\text{\sout{agent(Agent0), instant(I),}}$
      $\text{\sout{occurred(enter(Agent0, Location0), soundsensing, I),}}$
      $\text{\sout{location(Location0).}}$
\end{lstlisting}

The second issue is that children (people under the age of 14) are obliged to share their location when requested, but this is prohibited by the tier--2 governmental institution. The following suggestion is one of several minimal changes found to ensure the non--compliant obligation is not imposed on children. An additional constraint is placed that an agent, \verb;Agent2;, is not a child and the variable \verb;Agent2; is bound to the variable \verb;Agent0; denoting the agent who would be obliged to share their location. This means that the obligation can not be imposed on a child. The new variable \verb;Agent2; is introduced since the implementation relies on using unique variables for all literals and then systematically trying different optional bindings between the variables (or no bindings). The new rule is:

\begin{lstlisting}[mathescape]
initiated(obl(share_location(Agent0), 
              leave(Agent0, Location0)), soundsensing, I) :-
   occurred(request_location(Agent1), soundsensing, I),
   holdsat(at(Agent1, Location0), soundsensing, I),
   $\textcolor{white} {\textbf{not holdsat(child(Agent2), soundsensing, I)}}$, Agent0 = Agent1, 
   $\textcolor{white} {\textbf{Agent0 = Agent2}}$,	agent(Agent0), agent(Agent1), $\textcolor{white} {\textbf{agent(Agent2)}}$,
   location(Location0), instant(I).
\end{lstlisting}

Finally, the tier--2 governmental institution prohibits a prohibition on an agent to turn their microphone off when they are in a private area. Yet, the tier--1 soundsensing institution always prohibits turning a microphone off until the agent leaves the system (the prohibition exists in the initial state). The revisions found are not to delete the rule initiating a prohibition in the tier--1 soundsensing institution's initial state, but instead, to terminate the prohibition when an agent enters a private area and then initiate it again when they leave. Although the revision adds two rules, it is minimal in the outcome of the tier--1 soundsensing institution since there is still a prohibition on turning the microphone off in all other cases where it is allowed by the tier--2 governmental institution:

\begin{lstlisting}[mathescape]
$\textcolor{white} {\textbf{terminated(pro(microphone\_off(Agent0), leave\_soundsensing(Agent0)),}}$
   $\textcolor{white} {\textbf{soundsensing, I) :- }}$ 
   $\textcolor{white} {\textbf{occurred(enter\_private(Agent2), soundsensing, I), }}$ 
   $\textcolor{white}{\textbf{agent(Agent2), agent(Agent0), Agent0=Agent2, instant(I).}}$
$\textcolor{white} {\textbf{initiated(pro(microphone\_off(Agent0), leave\_soundsensing(Agent0)),}}$
   $\textcolor{white} {\textbf{soundsensing, I) :-}}$ 
   $\textcolor{white} {\textbf{occurred(leave\_private(Agent2), soundsensing, I), agent(Agent2),}}$ 
   $\textcolor{white}{\textbf{agent(Agent0), Agent0=Agent2, instant(I).}}$
\end{lstlisting}

\section{Related Work}
\label{secRevisionsRelatedWork}

There has been much work on norm change in normative systems, however, as far as we are aware we are the first to propose a way to revise institutions to be \textit{compliant} with other institutions in a multi--tier institution.

The most closely related work is by Li et al. \cite{Li2013b,Li2014} who also uses inductive search in ASP to resolve an ILP \textit{theory revision} task. Unlike us, their focus is on resolving norm conflicts between multiple institutions governing a group of agents (e.g. when an agent is prohibited to perform an action by one institution and obliged by another) and later the general case of debugging ASP programs \cite{Li2015}. In comparison, we focus on revising \textit{non--compliance} between lower--tier and higher--tier institutions in a \textit{multi--tier} institution. Our proposal is based on Li et al. and extended to revising an $i$th--tier institution by adding new rules or modifying/deleting pre--existing rules to impose $i$th--order norms. We also extend the work to revising with minimal changes in the \textit{consequences} of a revised institution (as opposed to changed rules), finally we look at the creation and deletion of existing rules which in our case study provides more minimal changes in the consequences compared to rule modification.

Vasconcelos et al. \cite{Vasconcelos2007} have proposed a technique for revising conflicting norms based on first--order unification. Their proposal provides a fine--grained way to revise obligation/permission/prohibition predicates' terms. For example, an obligation to be in an area that overlaps with a prohibited area is revised by changing the obliged/prohibited areas for an agent to be in. In contrast to our work, their focus is on modifying the obligation/permission/prohibition predicates and not with adding/removing/modifying \textit{rules} to meet a particular property (compliance between institutions in our case).

Governatori and Rotolo \cite{Governatori2010} propose a way to use a defeasible logic to modify legal systems by introducing new norms which derogate, abrogate and annul norms using defeasible rules. Central to their proposal is the idea of a legal system being \textit{versioned} and having two timelines: the versioning timeline and the timeline of the legal system's evolution (i.e. which norms are imposed and when). We only consider the latter timeline, the evolution of an institution (in our case during pre--runtime model checking) and focus on diagnosing \textit{causes} of non--compliance between institutions rather than assuming it is known what the new information (rules) is.

Finally, on the more conceptual and theoretical side, Boella et al.\cite{Boella2009a} look at how to classify different systems of norm change by investigating a set of rational norm change postulates. Specifically, they look at normative system change to incorporate new \textit{conditional} norms in input/output logics and they investigate the set of consistent postulates for different input/output logics. Again, this work also presupposes which conditional norms should be added to the normative system/institution, thus any system meeting these postulates is quite different from our proposal.

\section{Discussion}
\label{secReviDisc}

In this chapter we contributed an implemented automated system for revising a lower--tier institution's regulations to be compliant with the regulations of a higher--tier institution it is governed by. The proposal addressed a problem created by pervasive legal artefacts in the social world, where on the one hand institutions are used to govern other institutions in a vertical governance structure we call multi--tier institutions, creating the potential for non--compliant regulations. On the other hand, revising institutions' regulations to be compliant is non--trivial due to their inherent complexity. 

Our proposal takes our previous formal and computational framework \cite{King2015a} for determining the compliance of institutions in multi--tier institutions. Then, viewing the problem of revising an institution to be compliant as an instance of an ILP (Inductive Logic Programming) theory revision task, we use inductive search in ASP based on \cite{Li2014} to solve the ILP theory revision task for compliance. Inductive search in ASP is performed by translating, using an implemented compiler, from an ASP representation of an institution that needs to be revised to be complaint where revisions cannot be tried and searched for, to an ASP representation where revisions can be tried and thus revisions for compliance determined. Then, our system goes about finding revisions that are successful in resolving non--compliance and minimal in the changes to the institution's consequences thus keeping the changed institution as close as possible to the institution designer's original intentions.

The system for revising institutions, for tractability, considers a fragment of the search space of revisions: modifying and deleting existing rules and extending a single mutable institution with a limited number of rules. The successful and minimal revisions that do exist (if any) within the space explored are guaranteed to be found. However, there may be more minimal revisions that result in a well--formed institution (according to our  representation of institutions) outside of this space, but this space is bigger and takes longer to explore. 

We consider this a problem that is important to address. Firstly with formal analysis of the complexity of the full problem. Secondly, by studying the applicability of various heuristics to the full search problem (e.g. genetic algorithms) that cannot guarantee a minimal solution is found (i.e. in the case of genetic algorithms instead converging on local optima) but can help resolve tractability issues. As yet, it is unclear which heuristics are appropriate and how they can be incorporated into ILP revision as abducible search in ASP, presenting an interesting challenge for future work.
\chapter{Formalising Institutional Enactment Validity}
\label{chapter_6}

\epigraph[0pt]{One of the major problems encountered in time travel is not that of becoming your own father or mother. There is no problem in becoming your own father or mother that a broad--minded and well--adjusted family can't cope with. [...] The major problem is simply one of grammar.}{Douglas Adams, The Restaurant at the End of the Universe}

\blfootnote{\color{tck-grey}This chapter is based on the following paper:\\
\textbf{King, T. C.}, Dignum, V., \& Jonker, C. M. (2016). When Do Rule Changes Count--as Legal Rule Changes? In Proceedings of the 22nd European Conference on Artificial Intelligence (ECAI 2016). Frontiers in Artificial Intelligence and Applications. Vol 285. (pp. 3 – 11). IOS Press. http://doi.org/10.3233/978--1--61499--672--9--3 \cite{King2016}}

\newpage

Previously, Chapter~\ref{chapter_3}, Chapter~\ref{chapter_4} and Chapter~\ref{chapter_5} covered soft constraints for institutional design governance and how to comply with these soft constraints. In this chapter we look at \textit{hard institutional constraints} for institutional enactment governance, which make the social actions of enacting institutional changes possible. This chapter makes the following contributions:

\begin{itemize}
\item A novel representation of institutions for which changes can be enacted according to secondary legal rules regulating rule change. In particular, secondary legal rules represented as counts--as rule that make rule change possible and are conditional on an extended notion of social context that includes conditions on past and present states, institution versions and hypothetical effects of rule change.
\item A novel semantics for reasoning about rule changes ascribed by rule--modifying counts--as rules.
\end{itemize}

This chapter looks at institutional governance from the \textit{can} perspective, in terms of which rule changes can be enacted in which contexts and by whom. Specifically, we look at the kinds of rules found in administrative law, which describes the procedures that must be followed for rule change to take place and specifies what \textit{types} of rule change are possible. These are hard institutional constraints on rule enactment, in the sense that they make rule changes (im)possible.

Whilst constitutive counts--as rules ascribe a social reality and how it changes over time, they themselves are also subject to evolving over time when they are modified, such as by a legislative body. Rule--modifying counts--as rules regulate changes to counts--as rules themselves, ``A counts--as \textit{modifying a rule} in context C'', describing the legislative procedure and the possible modification types. Rule--modifying counts--as rules are also modifiable according to the rule modification regulations. In (\cite{Governatori2005b,Governatori2010}) a defeasible logic formalises legal rule change over time but not rule change regulation. In (\cite{Boella2004}) counts--as rules that regulate rule modifications are formalised, but not in a temporal setting. However, thus far there has been no attention paid to formalising rule change regulated by counts--as rules in a temporal setting.

This raises the question, in a temporal setting ``how can we formalise practical reasoning to determine when institutional rule changes count--as legally valid rule change enactments? should we define when rule changes count--as legal and valid rule changes?'' To give some examples:

\begin{itemize}
\item Based on (\cite[Art. 71]{RepublicOfItaly1947})  parliament voting to enact a bill counts--as enacting it within one month. Rule change affects institutional states (e.g. enacting a bill obliging fences to be painted white, or changing what counts--as parliament); rule changes are conditional on the institutional state (e.g. whether a body constituting parliament has voted).
\item The UK government voted to \textit{retroactively} require UK residents in a business partnership abroad to pay tax (\cite[Sec. 58]{FinanceAct2008}), criminalising people in the past. Criminalising retroactive modifications are not possible according to the European Convention of Human Rights (\cite[Art. 7]{ECHR1953}). Rule change affects institutional states (e.g. criminalising people in the past); rule change is conditional on how it would change institutional states (e.g. if the change would criminalise people in the past then the change is not possible).
\item A monarch or parliament can enact and repeal laws. The monarch enacts a law obliging all fences are painted white. The parliament retroactively repeals the power for the monarch to enact laws, reversing the fence--painting law enactment. Retroactive rule change affects past rule--modifying counts--as rules; past rule modifications can be unravelled due to a retroactive modification.
\end{itemize}

An interdependency exists between the counts--as rules that construct a social reality and rule--modifying counts--as rules. Changing counts--as rules affects the past/present/future institutional context and can change the modifications which happened in the past up until the present; rule modifications are conditional on the past/present/future institutional context and the hypothetical rule change effects. Whether a rule change counts--as a \textit{legal} rule change requires assessing the context in which the change takes place -- comprising both the present state of affairs and the potential rule change effects  affecting whether a rule change is legal in the first place.

In this chapter we address the question ``how can we formally define when legally valid institutional change enactments occur?''. The question is addressed by contributing the first formalisation of rule change ascribed by counts--as rules accounting for the aforementioned cases in a temporal setting. We introduce the notion of temporal rule--modifying counts--as rules, ``A counts--as modifying a rule in the past/present/future in context C'', which ascribe rule changes. To account for institutions dually evolving from one social reality to the next built by counts--as rules, as well as from on counts--as rule set to another, we adopt social reality (state) and institution version timelines. We enhance context from referring to a built social reality, to also a past social reality in different institution versions and the hypothetical effects of rule change. Rule modifications can potentially affect past modifications by changing the social context or contradicting previous modifications, we posit that the most recent modifications are prioritised and can override past modifications.

The rest of this chapter is organised as follows. We begin by comparing what follows with the InstAL framework  which we use as a basis for our novel formalisation, and the formalisation used in the previous chapters in \ref{secoverview}. Then we present the formal representation \ref{secReprFramework}. The novel formal semantics are given in \ref{secSemanticFramework}. We apply our framework to real case studies and examples used to test difficult edge cases in \ref{secCaseStudies}. Finally, we compare with closely related work in \ref{secRelatedWork} and conclude with wider implications and directions for future work in \ref{secConclusions}.

\section{Comparison with InstAL}
\label{secoverview}

In this chapter we extend institutions first formalised in the InstAL (Institution Action Language) framework (\cite{Cliffe2006,Cliffe2007}) where rules cannot be modified and rule change cannot be regulated, to be modifiable and modification--regulating. In InstAL an institution is a signature comprising events, fluents and regulatory rules. An event is either institutional providing a social interpretation (e.g. a person paying tax) of other institutional events and observable events or is observable and exogenous to the institution corresponding to brute facts (e.g. the event we call a person paying tax). Fluents  describe/prescribe the domain (e.g. a person is a business partner who is obliged to pay tax). We view all rules as ``A \textit{counts--as} B in context C'', stating an event (the `A') in a social context (`C') causes further institutional events to occur or different fluents to hold in the next state (the `B's'). An InstAL institution contains rules regulating its evolution from one social reality (state) to the next, but the rules remain static.
\looseness=-1

We extend institutions comprising counts--as rules to counts--as rules which are modifiable and ascriptively regulate rule changes. Rule  modifications activate and deactivate rules in the past/present/future, analogous to enacting regulatory changes. Rule modifications are regulated with counts--as rules, ``an event A counts--as a \textit{rule--modifying} event B in context C'', which ascribes the actions counting--as past/present/future rule modifications in a particular context.

Importantly, unlike in InstAL, institutions with modifiable rules do not just evolve from one state to the next according to their own counts--as rules. Rather, institutions evolve along two timelines, from one version comprising rules to another version comprising different rules. Each version evolves over time from state to state according to its rules. When a rule change occurs according to rule--modifying counts--as rules in one version, the institution evolves to the next version potentially comprising different (active) counts--as rules at different times. For example, if in one institution version on Wednesday a rule is \textit{retroactively} added on the preceding Monday then that version's past does not change. Instead, the institution evolves to a new version where the rule is active from the Monday potentially causing the version to evolve differently from then onwards. Institutions evolve from version to version where one versions can evolve from one state to another differently according to different rules.

Contexts in counts--as rules are also extended. Contexts can be conditional on the present state but also past institution versions and states. This allows testing potential retroactive rule modification effects. For example, a retroactive rule change is ascribed by a government voting for the change in the context that it does not criminalise people in the past \textit{compared to the previous version's past}. To summarise we extend institutions to evolve along rule version and state timelines according to counts--as rules conditional on contexts comprising past versions and states, and potential rule change effects.

In comparison to Chapter~\ref{chapter_3}, Chapter~\ref{chapter_4} and Chapter~\ref{chapter_5} we are only interested in institutions ascribing non--deontic social facts. For example, social actions such as voting in parliament or enacting a rule change, and states of affairs such as someone being a tax payer. In other words, we are interested in the hard institutional constraints that make social actions possible according to counts--as rules. We are not interested in obligations and prohibitions which prescribe the ideal social actions and states of affairs. Consequently, unlike in the previous chapters we provide no specific representation for deontic positions nor a semantics to determine when they are violated. Moreover, we do not consider whether the effects of regulatory rules (i.e. the obligations and prohibitions they impose) are good or bad and hence we do not capture higher--order norms. We use counts--as rules, but unlike in Chapter~\ref{chapter_3} and Chapter~\ref{chapter_4} we do not capture links between concrete and abstract regulations according to counts--as rules. Rather, we capture the notion of governance in terms of counts--as rules making institutional change enactments regular and \textit{possible}, hence deontic elements are not necessary.

\section{Representation}
\label{secReprFramework}

We begin with representing institutions which regulate their own temporal rule modifications.

\begin{definition}{\textbf{Institution}}\label{def:inst} An institution is a tuple $\pazocal{I} = \langle \pazocal{E}, \pazocal{F}, \pazocal{C}, \pazocal{G}, \Delta \rangle$. Institutions are distinguished with a superscript (e.g. $\pazocal{I}^{\textit{uk}} = \langle \pazocal{E}^{\textit{uk}}, \pazocal{F}^{\textit{uk}}, \pazocal{C}^{\textit{uk}}, \pazocal{G}^{\textit{uk}}, \Delta^{\textit{uk}} \rangle$). $\Sigma = 2^{\pazocal{F}}$ denotes all states for $\pazocal{I}$.
\end{definition}
Where:
\begin{enumerate}[leftmargin=*]
\item $\pazocal{E} = \pazocal{E}_{\textit{obs}} \cup \pazocal{E}_{\textit{inst}} \cup \pazocal{E}_{\textit{mod}}$ is a finite set of \textit{events} comprising:
\begin{itemize}[leftmargin=*]
\item Observable events $\pazocal{E}_{\textit{obs}}$ and institutional events $\pazocal{E}_{\textit{inst}}$.
\item Rule modification events $\pazocal{E}_{\textit{mod}} = \{ \textit{mod(op, id, t)} \mid \textit{op} \in \{ \textit{act}, \textit{deact} \}, \textit{id} \in \mathbb{ID}, t \in \mathbb{N} \}$ -- a rule with the identifier $\textit{id}$ (the identifier set being $\mathbb{ID}$) is activated/deactivated ($\textit{op}$) at a time $t$.
\end{itemize}
\item $\pazocal{F} = \pazocal{F}_{\textit{dom}} \cup \pazocal{F}_{\textit{ract}}$ is a finite set of \textit{fluents} describing the:
\begin{itemize}[leftmargin=*]
\item Domain $\pazocal{F}_{\textit{dom}}$.
\item Active rules $\pazocal{F}_{\textit{ract}} = \{ \textit{active(id)} \mid \textit{id} \in \mathbb{ID} \}$ identified as $\textit{id}$.
\end{itemize}
\item $\pazocal{X}$ is the set of all \textit{contexts} $\varphi$ expressible in the following grammar for fluents $f \in \pazocal{F}$:
\begin{alignat*}{3}
& \varphi & \; ::= & \; \; \; \top \; \mid f \; \mid \neg \varphi \mid \; \varphi \wedge \varphi \mid \; \varphi \vee \varphi \mid \; \varphi \rightarrow \varphi \mid  \; \textit{P} \mid \; \\
& & & \; \; \; \textit{PrS}(\varphi) \mid \; \textit{PaS}(\varphi) \mid \; \textit{PrV}(\phi) \mid \; \textit{PaV}(\phi) \\
& \phi & \; ::= & \; \; \; \varphi \; \mid \textit{NS}(\varphi)
\end{alignat*}
Each expression's informal meaning is the usual for propositional logic symbols. The operators bear truth in the following cases: \begin{inparaenum} \item $\textit{P}$ if the context is retroactive (i.e. the state in which $\textit{P}$ operates on is at a time before the version to which it belongs becomes the current version), and \item if $\varphi$ is true in: the previous state ($\textit{PrS}(\varphi)$), all past states ($\textit{PaS}(\varphi)$), the same state in the previous version ($\textit{PrV}(\varphi)$), the same state in all past versions ($\textit{PaV}(\varphi)$), and the next state ($\textit{NS}(\varphi)$).\end{inparaenum} The next state operator is restricted to past versions, meaning rules are never conditional on the actual future.
\item $\pazocal{G} : \pazocal{X} \times 2^{\pazocal{E}} \rightarrow 2^{\pazocal{E}_{\textit{inst}}}$ -- is the  \textit{event generation function} where $\pazocal{G}(X, E)$ is an event set caused by the events that occur ($E$) when the context $X$ holds.
\item $\pazocal{C} : \pazocal{X} \times \pazocal{E} \rightarrow 2^{\pazocal{F}_{\textit{dom}}} \times 2^{\pazocal{F}_{\textit{dom}}}$ is the \textit{state consequence function} where for a context $X \in \pazocal{X}$ and an event $e \in \pazocal{E}$ the consequence function's result is notated $\pazocal{C}(X, e) = \langle \pazocal{C}^{\uparrow}(X, e), \pazocal{C}^{\downarrow}(X, e) \rangle$ s.t. the initiated fluent set is $\pazocal{C}^{\uparrow}(X, e)$ and the terminated fluent set is $\pazocal{C}^{\downarrow}(X, e)$
\item $\Delta \subseteq \pazocal{F}$ is the \textit{initial institution state}
\end{enumerate}

For example, the following rule states that if Ada is found guilty  ($\textit{g(ada)}$) then she becomes a criminal ($\textit{crim(ada)}$). That is, the fluent $\textit{crim(ada)}$ is initiated by the event of being found guilty according to the consequence function ($\pazocal{C}^{\uparrow}$).
\begin{align*}
\pazocal{C}^{\uparrow}(\top, \textit{g(ada)}) \ni \textit{crim(ada)}
\end{align*}
A government rule change ($\textit{gmod(act, id, t)})$) that does not retroactively criminalise people counts--as a legal rule change. The condition is in all past retroactive states someone is not a criminal ($\textit{crim(ada)}$) if in the previous version (prior to rule change) they were not.
\begin{align*}
\pazocal{G}(&\textit{PaS}(\textit{P} \rightarrow \textit{PrV}(\neg \textit{crim(ada)}) \rightarrow \neg \textit{crim(ada)})), \\ & \{ \textit{gmod(act, id, t)} \}) \ni \textit{act(id, t)}
\end{align*}
In order to reason about modifying specific institutional rules, we tie rule identifiers to the institutional rules they represent. Specifically we map the inputs and single outputs of $\pazocal{G}$ and $\pazocal{C}$ to identifiers (i.e. not the whole set of events or initiated/terminated fluents).

\begin{definition}{\textbf{Rule Identifier Function}} A rule identifier function for an event generation function $\pazocal{G} : \pazocal{X} \times 2^{\pazocal{E}} \rightarrow 2^{\pazocal{E}_{\textit{inst}}}$ is $\textit{rid}^{\pazocal{G}} : \pazocal{X} \times 2^{\pazocal{E}} \times \pazocal{E}_{\textit{inst}} \rightarrow \mathbb{ID}$. The rule identifier functions for a consequence function $\pazocal{C} : \pazocal{X} \times \pazocal{E} \rightarrow 2^{\pazocal{F}_{\textit{dom}}} \times 2^{\pazocal{F}_{\textit{dom}}}$ are $\textit{rid}^{\pazocal{C}^{\uparrow}} : \pazocal{X} \times \pazocal{E} \times \pazocal{F}_{\textit{dom}} \rightarrow \mathbb{ID}$ and $\textit{rid}^{\pazocal{C}^{\downarrow}} : \pazocal{X} \times \pazocal{E} \times \pazocal{F}_{\textit{dom}} \rightarrow \mathbb{ID}$.
\end{definition}

So, the previous rule criminalising Ada has the ID $\textit{crim0} = \textit{rid}^{\uparrow}(\top, \textit{g(ada)}, \textit{crim(ada))}$. Examples/case studies omit this function.

\section{Semantics}
\label{secSemanticFramework}

This section defines institution semantics, following InstAL's method using just sets and functions, with the following considerations.

Observable events cause an institution rule version to transition from state to state by generating transitioning events according to the event generation function $\pazocal{G}$ and initiating and terminating fluents according to the consequence function $\pazocal{C}$. An institution transitions from one version of rules to another when rule modifying events are generated by the event generation function $\pazocal{G}$.

An institutional interpretation represents this dual evolution as a tuple  $M = \langle R, V \rangle$ where:
\begin{inparaenum}
\item $V = \langle V_0, ..., V_j \rangle$ is a tuple of versions each comprising a state and event set sequence up to length $k$ with typical element $V_v = \langle S_v, E_v \rangle$. The state sequence for $v$ is $S_v = \langle S_{v:0}, ..., S_{v:k+1} \rangle$ with typical element $S_{v:i} \in \Sigma$ and the event set sequence (the events transitioning between states) is $E_v = \langle E_{v:0}, ..., E_{v:k} \rangle$ with typical element $E_{v:i} \subseteq \pazocal{E}$. States denoted $S_{v:i}$ and event sets $E_{v:t}$ are denoted with the version $v$ to which they belong and their time instant $i$.
\item $R : [0,k] \rightarrow [0,j]$ is a function stating which institution version is the \textit{current} version for a given time.
\end{inparaenum}

$R$ also represents when rule change events occurring in a version can change that version's rules. Rule modification events only change version rules if the institution has not already evolved to a later version. For example, if on Monday a rule is added, then the institution evolves to a new \textit{current} version where that rule is actually added on Monday. When the version evolves, previous versions become \textit{obsolete} from then onwards (e.g. Monday) meaning their rules are not changeable. If $R(i) \leq v$ then an event occurring in version $v$ at time $i$ can modify rules in $v$ since the version is not yet obsolete.

The semantics are defined with respect to the interpretation $M = \langle R, V \rangle$, an institution $\pazocal{I} = \langle \pazocal{E}, \pazocal{F}, \pazocal{C}, \pazocal{G}, \Delta \rangle$, the set of all institutional interpretations $\mathbb{I}$, and an observable event set trace $\textit{et} = \langle O_0, ..., O_k \rangle$ with typical element $O_i \subseteq \pazocal{E}_{\textit{obs}}$.

\subsection{Institutional Change}

Counts--as rules, causing institution state and version change, are conditional on a context being \textit{modelled} by the state in an \textit{interpretation}.

\begin{definition}\label{defModCont}{\textbf{Modelling Context}} For all $X \in \pazocal{X}$ and $f \in \pazocal{F}$, context models $\langle M, S_{v:t} \rangle \models X$ is defined for $\top$, $\vee$ and $\rightarrow$ w.r.t. $\neg$ and $\wedge$ as usual and for the other symbols as:
\begin{subequations}
\begin{align*}
& \langle M, S_{v:t} \rangle \models f & \Leftrightarrow & \; \; \; f \in S_{v:t} & \tag{\ref{defModCont}.1}\label{eqMC1} \\
& \langle M, S_{v:t} \rangle \models \neg \psi & \Leftrightarrow & \; \; \;  \langle M, S_{v:t} \rangle \not \models \psi & \tag{\ref{defModCont}.2}\label{eqMC2} \\
& \langle M, S_{v:t} \rangle \models \psi \wedge \phi & \Leftrightarrow & \; \; \;  \langle M, S_{v:t} \rangle \models \psi \; \textbf{ and } \\
& & & \; \; \; \langle M, S_{v:t} \rangle \models \phi & \tag{\ref{defModCont}.3}\label{eqMC3} \\ & 
\langle M, S_{v:t} \rangle \models \textit{P} & \Leftrightarrow & \; \; \; R(t) < v \;   
\tag{\ref{defModCont}.4}\label{eqMC4} \\
& \langle M, S_{v:t} \rangle \models \textit{PrS}(\psi) & \Leftrightarrow & \; \; \; \langle M, S_{v:t-1} \rangle \models \psi \tag{\ref{defModCont}.6}\label{eqMC6} \\
& \langle M, S_{v:t} \rangle \models \textit{PaS}(\psi) & \Leftrightarrow & \; \; \; \forall t^{\prime} \in [0,t-1] : \langle M, S_{v:t-1} \rangle \models \psi \tag{\ref{defModCont}.7}\label{eqMC7} \\
& \langle M, S_{v:t} \rangle \models \textit{PrV}(\psi) & \Leftrightarrow & \; \; \; \langle M, S_{v-1:t} \rangle \models \psi \tag{\ref{defModCont}.8}\label{eqMC8} \\
& \langle M, S_{v:t} \rangle \models \textit{PaV}(\psi) & \Leftrightarrow & \; \; \; \forall v^{\prime} \in [0, v-1] : \langle M, S_{v^{\prime}-1:t} \rangle \models \psi \tag{\ref{defModCont}.9}\label{eqMC9} \\
& \langle M, S_{v:t} \rangle \models \textit{NS}(\psi) & \Leftrightarrow & \; \; \; \langle M, S_{v:t+1} \rangle \models \psi \tag{\ref{defModCont}.10}\label{eqMC10}
\end{align*}
\end{subequations}
\end{definition}

Semantics are as usual for modelling a fluent (\ref{eqMC1}), weak negation (\ref{eqMC2}) and conjunction (\ref{eqMC3}). A state is retroactive if at that time the version is not the current version but it will be in the future (\ref{eqMC4}) -- for example, if on a Wednesday the institution evolves to a new version, then anything occurring on the Monday is retroactive to the new version (i.e. occurring in the version's past). States model formula as expected for a previous state (\ref{eqMC6}), all previous states (\ref{eqMC7}), the previous version (\ref{eqMC8}), all past versions (\ref{eqMC9}) and the next state (\ref{eqMC10}).

An event `B' occurs when transitioning to a new state in a version according to a rule -- ``A counts--as B in context C'' ($\pazocal{G}$) -- if an event `A' occurs, the context `C' is modelled by the state and the counts--as rule itself is active in the version's state. Events occurring in response  to observable events $E$ are formalised as an event generation operation.

\begin{definition}{\textbf{Event Generation Operation}}\label{defGR} The event generation operation $\textit{GR} : \Sigma \times 2^{\pazocal{E}} \times \mathbb{I} \rightarrow 2^{\pazocal{E}}$ is defined such that $\textit{GR}(S_{v:t}, E, M) = E^{\prime}$ iff $E^{\prime}$ only satisfies the following conditions:
\begin{subequations}
\allowdisplaybreaks[1]
\begin{align*}
& E \subseteq E^{\prime} & \tag{\ref{defGR}.1}\label{eqGR1} \\
& \exists X \in \pazocal{X}, e \subseteq E, e^{\prime} \in \pazocal{G}(X, e) \cap \pazocal{E}_{\textit{inst}} : \textit{id} = \textit{rid}^{\pazocal{G}}(X, e, e^{\prime}), \\ &
\langle M, S_{v:t} \rangle \models X \wedge \textit{active(id)} \Rightarrow e^{\prime} \in E^{\prime} & \tag{\ref{defGR}.2}\label{eqGR2} \\
& \exists X \in \pazocal{X}, e \subseteq E, e^{\prime} \in \pazocal{G}(X, e) \cap \pazocal{E}_{\textit{mod}} : \textit{id} = \textit{rid}^{\pazocal{G}}(X, e, e^{\prime}),
\\ & 
\langle M, S_{v:t} \rangle \models X \wedge \textit{active(id)}, R(t) \neq v \Rightarrow e^{\prime} \in E^{\prime} &  & \tag{\ref{defGR}.3}\label{eqGR3} \\
& \exists X \in \pazocal{X}, e \subseteq E, e^{\prime} \in \pazocal{G}(X, e) \cap \pazocal{E}_{\textit{mod}} : \textit{id} = \textit{rid}^{\pazocal{G}}(X, e, e^{\prime}), \\ &
\langle M, S_{v:t} \rangle \models X \wedge \textit{active(id)}, R(t) = v \Rightarrow (e^{\prime} \in E^{\prime} \; \textbf{or} \; e^{\prime} \not \in E^{\prime}) &  & \tag{\ref{defGR}.4}\label{eqGR4}
\end{align*}
\end{subequations}
Any fixed point reached after iterative applications of $\textit{GR}$ is denoted as $\textit{GR}^{\omega}(S_{v:t}, E, M)$.
\end{definition}

Events that have occurred still occur (\ref{eqGR1}). If an active rule states an event $e$ causes an event $e^{\prime}$ in a context modelled by the state, then $e$ \textit{can} cause $e^{\prime}$ to occur depending on $e^{\prime}$'s type. Specifically, whether $e^{\prime}$ is a type that could cause an inconsistency (e.g. removing rules that ascribe rule modifications, for more on the paradox of rule change see \cite{Suber1990}). An event $e^{\prime}$ always occurs if it is a non--rule--modifying institutional event (\ref{eqGR2}) or occurs when the version is obsolete and it cannot modify rules (\ref{eqGR3}). Rule modifying events in non--obsolete versions \textit{can} cause rule changes and a potential paradox. So they \textit{optionally} occur in a non--obsolete version where they can cause rule change and/or a paradox (\ref{eqGR4}). Hence, $\textit{GR}$ is \textit{multi--valued}.

Iterating the event generation operation until a \textit{fixed point} is reached obtains all events which occur. At least one fixed point is guaranteed.

\begin{lemma}\label{thGRFP} For any set of events $E \subseteq \pazocal{E}$, interpretation $M$ and state $S_{v:t} \in \Sigma$ there exists a fixed point $\textit{GR}^{\omega}(S_{v:t}, E, M)$.
\end{lemma}
\begin{proof} $\textit{GR}$ always has a monotonically increasing value (w.r.t. set inclusion) and a finite domain.
\end{proof}

An institution version transitions between states, driven by event occurrences, according to a state transition operation.

\begin{definition}{\textbf{State Transition Operation}}\label{defTR} The state transition operation $\textit{TR} : \Sigma \times 2^{\pazocal{E}} \times \mathbb{I}  \rightarrow 2^{\pazocal{E}}$ is defined for a state $S_{v:t}$, a set of events $E_{v:t}$ and an interpretation $M$ as:
\begin{subequations}
\allowdisplaybreaks[1]\\
$\textit{TR}(S_{v:t}, E_{v:t}, M) =$
\begin{align*}
\{ f \mid \; & f \in S_{v:t} \cap \textit{TERM}(S_{v:t}, E_{v:t}, M) \; \; \textbf{or} \tag{\ref{defTR}.1}\label{eqTR1} \\ &
f \in \textit{INIT}(S_{v:t}, E_{v:t}, M)
\} \tag{\ref{defTR}.2}\label{eqTR2}
\end{align*}
where:\\
$\textit{INIT}(S_{v:t}, E_{v:t}, M) =$
\begin{align*}
\{ f \mid & \exists e \in E_{v:t}, X \in \pazocal{X} :
\textit{id} = \textit{rid}^{\pazocal{C}^{\uparrow}}(X, e, f), \\ &
f \in \pazocal{C}^{\uparrow}(X, e) \cap \pazocal{F}_{\textit{dom}}, \langle M, S_{v:t} \rangle \models X \wedge \textit{active(id)} \; \; \; \; \textbf{or} & \tag{\ref{defTR}.3}\label{eqTR3} \\ &
\exists t^{\prime} \in [0,k], \nexists  t^{\prime \prime} \in [t^{\prime},k] : \textit{id} = \textit{rid}^{\pazocal{C}^{\uparrow}}(X, e, f), \\ &
R(t^{\prime}) \leq v, R(t^{\prime\prime}) \leq v,
\textit{mod(act, id, t)} \in E_{v:t^{\prime}}, \\ &\textit{mod(deact, id, t)} \in E_{v:t^{\prime \prime}}, f = \textit{active(id)}\} \tag{\ref{defTR}.4}\label{eqTR4}
\end{align*}
$\textit{TERM}(S_{v:t}, E_{v:t}, M) =$
\begin{align*}
\{ f \mid & \exists e \in E_{v:t}, X \in \pazocal{X} : \textit{id} = \textit{rid}^{\pazocal{C}^{\downarrow}}(X, e, f), \\ &
f \in \pazocal{C}^{\downarrow}(X, e) \cap \pazocal{F}_{\textit{dom}}, \langle M, S_{v:t} \rangle \models X \wedge \textit{active(id)} \; \; \textbf{or} & \tag{\ref{defTR}.5}\label{eqTR5} \\ &
\exists t^{\prime} \in [0,k], \nexists  t^{\prime \prime} \in [t^{\prime},k] :  \textit{id} = \textit{rid}^{\pazocal{C}^{\downarrow}}(X, e, f) \\ &
R(t^{\prime}) \leq v, R(t^{\prime\prime}) \leq v,\textit{mod(deact, id, t)} \in E_{v:t^{\prime}}, \\ & 
\textit{mod(act, id, t)} \in E_{v:t^{\prime \prime}},
f = \textit{active(id)}\}  \tag{\ref{defTR}.6}\label{eqTR6}
\end{align*}
\end{subequations}
\end{definition}

Transitioning from one state to the next follows common--sense inertia -- a fluent holds in a new state if it held in the previous state and was not terminated (\ref{eqTR1}) or it was initiated in the previous state (\ref{eqTR2}). A domain fluent is initiated/terminated if an event causes it to be according to a rule defined by the state consequence function $\pazocal{C}$ that is active in the current state with a condition (context) that is modelled in the state (\ref{eqTR3} for initiation and \ref{eqTR5} for termination). A fluent denoting an active rule is initiated/terminated in a state if a rule activating/deactivating event occurs at a time when the version is not obsolete and no contradictory deactivation/activation event occurs at a later time when the version is not obsolete (\ref{eqTR4} for activating rules and \ref{eqTR6} for deactivating rules). The most recent modifications in a version take precedent if they occur when the version is a non--obsolete version and simultaneous contradictory rule modifications are cancelled.

\subsection{Models}

Now we define when an interpretation is an institutional model for an observable event set trace. An institutional interpretation is, broadly speaking, an institutional model for an observable event set trace iff: \begin{inparaenum} \item each version evolves according to the event generation and state transition operations, and \item the institution evolves from one version to another when rules are modified. \end{inparaenum} However, the event generation operation is multi--valued since rule modifications are \textit{optional}. Thus, there are potentially multiple candidate event sets for transitioning between states and therefore multiple interpretations to select as models. 

We want to maximise the rule modification events that are not self--contradicting (e.g. not applying modifications that retroactively remove a rule making retroactive rule removal possible). Interpretations are prioritised, denoted as $<$, based on maximising rule modifications. An interpretation has higher priority over another if at the earliest time in the earliest version in which the interpretation differ it contains a superset of rule modifying events compared to the `same' set for the lower priority interpretation.

\begin{definition}{\textbf{Prioritised Interpretation}}\label{defPI} Let $M^{0} = \langle R^{0}, V^{0} \rangle \in \mathbb{I}$ and $M^{1} = \langle R^{1}, V^{1} \rangle \in \mathbb{I}$ be two interpretations for institution $\pazocal{I}$ where: $V^{0} = \langle V^{0}_0, ..., V^{0}_i \rangle$ with typical element $V^{0}_v = \langle E^{0}_v, S^{0}_v \rangle$ s.t. $E^{0}_v = \langle E^{0}_{v:0}, ..., E^{0}_{v:k} \rangle$, and $V^{1} = \langle V^{1}_0, ..., V^{1}_j \rangle$ with typical element $V^{1}_v = \langle E^{1}_v, S^{1}_v \rangle$ s.t. $E^{1}_v = \langle E^{1}_{v:0}, ..., E^{1}_{v:k} \rangle$. The ordering $<$ is a relation between interpretations $M^{0}$ and $M^{1}$ such that:
\begin{subequations}
\allowdisplaybreaks[1]
\begin{align*}
M^{0} < M^{1} & \; \; \Leftrightarrow & & \exists t \in [0,k], \nexists t^{\prime} \in [0,t\textit{-}1] : \\ & & & 
v = R^{0}(t), E^{0}_{v:t} \cap \pazocal{E}_{\textit{mod}} \supset E^{1}_{v:t} \cap \pazocal{E}_{\textit{mod}} \\ & & &
v^{\prime} = R^{0}(t^{\prime}), E^{0}_{v^{\prime}:t^{\prime}} \neq E^{1}_{v^{\prime}:t^{\prime}}
\end{align*}
\end{subequations}
\end{definition}

We operationally characterise a model by constructing a `correct' interpretation. That is, constructing versions comprising correct state transitions and generated events. We could construct each institution version by starting at an initial state and proceeding from one state to the next according to the event generation and state transition operations. However, this would require knowing which rule modification events happen in each version's past, present and future.

To give an example for an observable event set trace $\langle O_0, ..., O_k \rangle$. An institution starts at an initial state only comprising an active rule enabling a government to make retroactive modifications ($\Delta = S_{0:0} = \{ \textit{active(gov0)} \}$). First, a fence is observably built ($O_0 = \{ \textit{fb} \}$, occurring during the first state transition $\textit{fb} \in E_{0:0} = \textit{GR}^{\omega}(S_{0:0}, O_0, M)$). But, there is no active rule that causes the next state to be different ($S_{0:1} = \textit{TR}(S_{0:0}, E_{0:0}) = S_{0:0} = \{ \textit{active(gov0)} \}$). Then, the government votes to retroactively activate a rule in state zero, stating building a fence initiates an obligation to paint it. Consequently, the second state which has already been determined, $S_{0:1}$, seems wrong since it lacks the fence painting rule and its effects. In fact, the institution should transition to a new rule version $V_{1}$. This new version should start at the same initial state $S_{1:0} = \Delta$. But, crucially, transition to the next state ($S_{1:0} = \textit{TR}(S_{1:0}, E_{1:0})$) with the knowledge that in the future of the new version the fence painting rule will be retroactively added at state zero ($S_{1:0}$) and become active in the second state ($S_{1:1}$). State transitions are defined with respect to an interpretation comprising past/present/future rule modification events which might be unknown when each state and transitioning event set is constructed.

We define an interpretation successor operation which addresses the problem of constructing a `correct' interpretation without the knowledge of each version's past/present/future. The successor operation takes as input a preceding interpretation which supplies versions comprising a past/present/future on which each version in the new succeeding interpretation can be constructed according to $\textit{TR}$ and $\textit{GR}$. That is, a new interpretation is produced using the version timelines of the previous interpretation, taking into account past/present/future rule modifications from the preceding interpretation's version timelines.

A succeeding interpretation might not be the same as the previous interpretation, since the previous interpretation might have been built without knowledge of its own past/present/future. That is, the new interpretation might differ in its temporal evolution (comparable version timelines in each interpretation being different). Consequently, the succeeding interpretation might have new, previously unknown, rule modification events that also need to be accounted for and thus another succeeding interpretation must be produced. 

The idea is to iteratively apply the institution successor operation until a succeeding interpretation is produced that is the same as the previous interpretation. That is, until the operation reaches a fixed point, which is guaranteed according to lemma~\ref{thFixedPoint} we give later on. Intuitively, the fixed point characterises an interpretation that is built taking into account its own past/present/future modifications in each version (since it was built with respect to an identical preceding interpretation). Formally, the successor interpretation operation is:

\begin{definition}\textbf{Successor Interpretation Operation}\label{defSIO} \;Let $\textit{et} = \langle O_0, ..., O_k \rangle$ be an observable event trace for $\pazocal{I}$ of length $k$. Let $M^{\prime} = \langle R^{\prime}, V^{\prime} \rangle \in \mathbb{I}$ be an interpretation such that $V^{\prime} = \langle V^{\prime}_0, ..., V^{\prime}_{j^{\prime}} \rangle$ is a tuple of institution versions. The interpretation successor operation $\textit{SUCC} : \mathbb{I} \times \textit{ET} \rightarrow \mathbb{I}$ is defined for the interpretation $M$ w.r.t. $\pazocal{I}$ and $\textit{et}$ such that $\textit{SUCC}(M, \textit{et}) = M^{\prime}$ iff $M^{\prime}$ satisfies the following conditions:
\begin{subequations}
\allowdisplaybreaks[1]
\begin{align*}
& \forall v \in [0,j^{\prime}] : S^{\prime}_{v:0} =  \Delta \tag{\ref{defSIO}.1}\label{eqSIO1} \\
& \forall v \in [0, j^{\prime}], t \in [0,k] : E^{\prime}_{v:t} = \textit{GR}^{\omega}(S^{\prime}_{v:t}, O_t, M) \tag{\ref{defSIO}.2}\label{eqSIO2} \\ &
\forall v \in [0, j^{\prime}], t \in [0,k] : S^{\prime}_{v:t+1} = \textit{TR}(S^{\prime}_{v:t}, E^{\prime}_{v:t}, M) \tag{\ref{defSIO}.3}\label{eqSIO3} \\
& R^{\prime}(t) = \left\{\begin{array}{lr}
        0, & t = 0, E^{\prime}_{0:t} \cap \pazocal{E}_{\textit{mod}} = \emptyset\\
        1, & t = 0, E^{\prime}_{0:t} \cap \pazocal{E}_{\textit{mod}} \neq \emptyset\\
        R^{\prime}(t\textit{-}1), & t > 0, E^{\prime}_{R(t\textit{-}1):t} \cap \pazocal{E}^{\prime}_{\textit{mod}} = \emptyset \\
        R^{\prime}(t\textit{-}1) \textit{+} 1, & t > 0, E^{\prime}_{R(t\textit{-}1):t} \cap \pazocal{E}^{\prime}_{\textit{mod}} \neq \emptyset
        \end{array}\right.
  \tag{\ref{defSIO}.4}\label{eqSIO4} \\
& \text{Given that } V^{\prime} = \langle V^{\prime}_0, ..., V^{\prime}_{j^{\prime}} \rangle, R^{\prime}(k) = j^{\prime} \tag{\ref{defSIO}.5}\label{eqSIO5}
\end{align*}
\end{subequations}
\end{definition}

Every institution version starts at the same initial state (\ref{eqSIO1}). Each state transition (an event set) in a version is produced by the event generation operation applied to the previous state and the observable events occurring at that time (\ref{eqSIO2}). The next state in a version is the state produced by the state transition operation applied to the previous state and the transitioning events occurring in that version \textit{with respect to} the preceding institutional interpretation (\ref{eqSIO3}). That is, transitioning from one state to the next takes into account the rule modification events occurring in the past/present/future of the same version in the preceding interpretation. Rule modifications in the latest version cause the current version to evolve/increment to the next version. If no rule modification takes place the version remains the same or the zeroeth version for the zeroeth time instant (\ref{eqSIO4}). If a rule modification does take place in the latest version, then the current version at that time incremented by one, or is the first version for the zeroeth time point (\ref{eqSIO4}). The version sequence only goes up until the current version at the last time instant (\ref{eqSIO5}).

At least one fixed point for the successor interpretation operation, starting at any initial interpretation, is always guaranteed. A fixed point is denoted as $\textit{SUCC}^{\omega}(M, et)$. To see why, the general idea is that there always exists a series of successive interpretations that monotonically increase which versions and states they agree on.

The following lemma is used to prove that there always exists a series of such interpretations and therefore that there always exists a fixed point. Informally, the lemma is conditional on there being two successors $M^{\prime}$ and $M^{\prime \prime}$ to any interpretation that agree with each other up until a particular time ($h$) in a version ($j$). The consequence is that the second interpretation $M^{\prime \prime}$ has the same events at time $h$ and state transition at time $h+1$ in version $j$ as if the event and state transitions were produced with respect to $M^{\prime \prime}$'s \textit{own} past/present/future timeline.

\begin{lemma}\label{thNextSucc}
If $\pazocal{I}$ is an institution, $M$ an interpretation and $\textit{et}$ an observable event trace of length $k$ for $\pazocal{I}$ and there exists interpretations $M^{\prime} = \textit{SUCC}(\textit{M, et})$ and $M^{\prime \prime} = \textit{SUCC}(M^{\prime}, \textit{et})$ where $\exists h \in [0, k], j \in [0, v^{\prime}], \forall i \in [0,k]:$
\begin{align*}
& \langle V^{\prime}_0, ..., V^{\prime}_{j-1} \rangle = \langle V^{\prime \prime}_0, ..., V^{\prime \prime}_{j-1} \rangle \tag{A\ref{thNextSucc}.1}\label{eqNextSucc1} \\
& \langle S^{\prime}_{j:0}, ..., S^{\prime}_{j:h} \rangle = \langle S^{\prime \prime}_{j:0}, ..., S^{\prime \prime}_{j:h} \rangle \tag{A\ref{thNextSucc}.2}\label{eqNextSucc2} \\
& \begin{aligned} 
& \textit{mod(op, id, h)} \in E^{\prime}_{v:i}, \\ & R^{\prime}(i) \leq j
\end{aligned} \Leftrightarrow 
\begin{aligned} & \textit{mod(op, id, h)} \in E^{\prime \prime}_{j:i}, \\ &
 R^{\prime \prime}(i) \leq j 
\end{aligned} \tag{A\ref{thNextSucc}.3}\label{eqNextSucc3}
\end{align*}
then $E^{\prime \prime}_{j:h} = \textit{GR}^{\omega}(S^{\prime \prime}_{j:h}, O_h, M^{\prime \prime})$ and $S^{\prime \prime}_{j:h+1} = \textit{TR}(S^{\prime \prime}_{j:h}, E^{\prime \prime}_{j:h}, M^{\prime \prime})$
\end{lemma}

The previous lemma's assumptions can always be met starting from \textit{any} interpretation $M$. Firstly, since in the worst case, from any interpretation we can obtain a successor starting at the institution's initial state -- so both successors agree at least on the initial state. Secondly, by making the non--deterministic choice in the event generation operation to select the same rule modifications for both the successor and the successor to the successor (in the worst case, no rule modifications). We can continue to incrementally produce successive interpretations that monotonically increase the time point they agreed upon. Note that, this may mean backtracking by changing preceding interpretations (e.g. selecting no rule modifications).

\begin{lemma}\label{thFixedPoint} There exists a fixed point for the interpretation successor operation denoted $\textit{SUCC}^{\omega}(M, et)$ for any $M$ and $\textit{et}$.
\end{lemma}

In fact, there can be multiple fixed points, as exemplified:

\begin{example}\label{exFixedPoints} An institution $\pazocal{I}$ contains a legislative rule with the id $\textit{leg0} \in \mathbb{ID}$ stating that an agent, Ada, voting to activate a rule ($\textit{vote}_{a}(\textit{act, id, t}) \in \pazocal{E}_{\textit{obs}}$) \textit{counts--as} activating the rule: $\pazocal{G}(\top, \{ \textit{vote}_{a}(\textit{act, id, t}) \}) \ni \textit{mod(act, id, t)}$. In the initial state the legislative rule is active $\Delta = \{ \textit{active(leg0)} \}$. In an observable event trace $\textit{et} = \langle O_0 \rangle$ Ada votes to activate another rule with the id $\textit{leg1} \in \mathbb{ID}$ in the initial state $O_0 = \{ \textit{vote}_{a}\textit{(act, leg1, 0}) \}$.
\end{example}

From an initial empty interpretation $M$ we have the following successors and interpretations for example~\ref{exFixedPoints}(differences are in \textbf{bold}):
\begingroup\makeatletter\def\f@size{7}\check@mathfonts
\def\maketag@@@#1{\hbox{\m@th\large\normalfont#1}}
\begin{align*}
& M^{2} = \textit{SUCC(M, et)} = \textit{SUCC}^{\omega}\textit{(M, et)} \text{ s.t. } V^{2} = \langle V^{2}_{0} \rangle, R^{2}(0) = 0, R^{2}(1) = 0, \\ & 
S^{2}_{0:0} = \{ \textit{active(leg0)} \}, S^{2}_{0:1} = \{ \textit{active(leg0)} \}, 
E^{2}_{0:0} = \{ \textit{vote}_{a}\textit{(act, leg1, 0}) \} \\ &
M^{1} = \textit{SUCC(M, et)} = \textit{SUCC}^{\omega}\textit{(M, et)} \text{ s.t. } V^{1} = \langle V^{1}_{0}, V^{1}_{1} \rangle, R^{1}(0) = 1, R^{1}(1) = 1, \\ &
S^{1}_{0:0} = \{ \textit{active(leg0)} \}, S^{1}_{0:1} = \{ \textit{active(leg0)} \}, \\ & 
E^{1}_{0:0} = \{ \textit{vote}_{a}\textit{(act, leg1, 0}),\textbf{\textit{mod}(\textit{act, leg1 , 0})}  \} \\ &
S^{1}_{1:0} = \{ \textit{active(leg0)} \}, S^{1}_{1:1} = \{ \textit{active(leg0)} \}, E^{1}_{1:0} = \{ \textit{vote}_{a}\textit{(act, leg1, 0}) \} \\ &
M^{0} = \textit{SUCC(M, et)} = \textit{SUCC}^{\omega}\textit{(M, et)} \text{ s.t. } V^{0} = \langle V^{0}_{0}, V^{0}_{1} \rangle, R^{0}(0) = 1, R^{0}(1) = 1, \\ & 
S^{0}_{0:0} = \{ \textit{active(leg0)} \}, S^{0}_{0:1} = \{ \textit{active(leg0)} \}, \\ & 
E^{0}_{0:0} = \{ \textit{vote}_{a}\textit{(act, leg1, 0}), \textbf{\textit{mod}(\textit{act, leg1 , 0})}  \} \\ &
S^{0}_{1:0} = \{ \textit{active(leg0)} \}, S^{0}_{1:1} = \{ \textit{active(leg0)}, \textbf{\textit{active(leg1)}} \}, \\ &
E^{0}_{1:0} = \{ \textit{vote}_{a}\textit{(act, leg1, 0}), \textbf{\textit{mod}(\textit{act, leg1 , 0})} \}
\end{align*}
\endgroup
Each fixed point has different rule modifications. $M^{2}$ does not add the rule $\textit{leg1}$. $M^{1}$ contains an attempt to add the rule in the version zero but not in version one. Finally, $M^{0}$ adds the rule in the version zero and version one, version one being the current version when the rule is added meaning the rule addition is successful. In fact, the following prioritisation holds $M^{0} < M^{1} < M^{2}$ meaning that $M^{0}$ maximises successful rule modifications.

Models are interpretations which maximise successful rule modifications. Thus we characterise models by combining the successor interpretation fixed point and interpretation prioritisation. Given an empty interpretation we find a fixed point successor interpretation for a given event set trace (\ref{eqM1}). The fixed point is a model if there is no greater prioritised successor fixed point interpretation (\ref{eqM2}).

\begin{definition}{\textbf{Models}}\label{defM}
Let $M = \langle R, V \rangle$ be an empty interpretation such that $V = \langle V_0 \rangle$, $V_0 = \langle E_0, S_0 \rangle$, $E_0 = \langle \rangle$ and $S_0 = \langle \rangle$. The interpretation $M^{\prime} = \langle R^{\prime}, V^{\prime} \rangle$ is a model for $\pazocal{I}$ w.r.t. an observable event set trace $\textit{et} = \langle O_0, ..., O_k \rangle$ iff:
\begin{subequations}
\begin{align*}
& M^{\prime} = \textit{SUCC}^{\omega}(M, \textit{et}) \; \; \; \textbf{and} & \tag{\ref{defM}.1}\label{eqM1} \\
& \text{There does not exist an } M^{\prime \prime} < M^{\prime} \text{ meeting \ref{eqM1}}. & \tag{\ref{defM}.2}\label{eqM2}
\end{align*}
\end{subequations}
\end{definition}

From lemma~\ref{thFixedPoint} and definition~\ref{defM} we have the following property.

\begin{lemma}\label{thModelExistence} There exists at least one model for any institution $\pazocal{I}$ w.r.t. an observable event set trace $\textit{et}$. \end{lemma}

These semantics operationalise answering ``when does a rule change count--as a legal rule change?''. Generally, a physical or institutional actions counts--as the social action of a legal rule change \textit{if and only if} the rule change is ascribed by secondary legal counts--as rules conditional on a social context, which can include hypothetical rule change effects, that holds before and after the rule change takes place. Models always contain `legal' rule modifications, defined as fixed point interpretations which maximise rule modifications. So, `legal' rule--changes occur in at least one model whilst illegal rule changes do not occur at all (the non--deterministic choice for a rule modification to occur in \ref{eqGR4}) and the institution continues to operate `as usual', meeting our desiderata.

\section{Case Studies}
\label{secCaseStudies}

Now we apply the framework to concrete case studies. For brevity we use variables to denote: all rule identifiers ($\textit{id} \in \mathbb{ID}$), all rule change operations ($\textit{op} \in \{ \textit{act, deact} \}$), and all time instants ($t \in \mathbb{N}$). The first case concerns a simple rule change procedure.

\begin{case}\label{exSimple}
An institution $\pazocal{I}^{\textit{sgov}}$ describes a \textbf{s}imple \textbf{gov}ernment comprising two agents, Ada and Bertrand. Both Ada and Bertrand voting to activate or deactivate a rule in the context that neither are criminals ($\textit{crim(ada)}, \textit{crim(bert)} \in \pazocal{F}^{\textit{sgov}}_{\textit{dom}}$) \textit{counts--as} activating/deactivating the rule. The rule modifying counts--as rules are identified with $\textit{leg0} \in \mathbb{ID}$ and formalised as $\pazocal{G}^{\textit{sgov}}(\neg \textit{crim(ada)} \wedge \neg \textit{crim(bert)}, \{ \textit{vote}_{a}(\textit{op, id, t}), \textit{vote}_{b}(\textit{op, id, t}) \}) \ni \textit{mod(act, id, t)}$. At time point one Ada and Bertrand vote to add a rule with id $\textit{crim0}$, $O_1 = \{ \textit{vote}_{a}(\textit{act, crim0, 1}), \textit{vote}_{b}(\textit{act, crim0, 1}) \}$. The rule identified as $\textit{crim0}$ states that if Ada or Bertrand are found guilty of a crime ($\textit{g(ada)}, \textit{g(bert)} \in \pazocal{E}^{\textit{sgov}}_{\textit{obs}}$) then they become criminals, formally -- $\pazocal{C}^{\uparrow}(\top, \textit{g(ada)}) \ni \textit{crim(ada)}$ and $\pazocal{C}^{\textit{sgov}\uparrow}(\top, \textit{g(bert)}) \ni \textit{crim(bert)}$. Next, Bertrand is found guilty of a crime $O_2 = \{ \textit{g(bert)} \}$. Finally, Bertrand and Ada vote to deactivate the criminalising rule, $O_3 = \{ \textit{vote}_{a}(\textit{act, crim0, 3}), \textit{vote}_{b}(\textit{act, crim0, 3}) \}$.
\end{case}

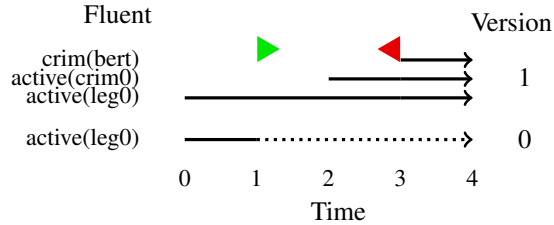
\begin{figure}[h]
\centering
\begin{tikzpicture}[xscale=0.95]

\draw [->, very thick] (-3.5,-2.45) -- (-2.5,-2.45);
\draw [very thick] (-4.5,-2.7) -- (-3.5,-2.7);
\draw [->, very thick] (-3.5,-2.7) -- (-2.5,-2.7);
\draw [very thick] (-6.5,-2.95) -- (-3.5,-2.95);
\draw [->, very thick] (-3.5,-2.95) -- (-2.5,-2.95);

\draw [very thick] (-6.5,-3.5) -- (-5.5,-3.5);
\draw [->, very thick, dotted] (-5.5,-3.5) -- (-2.5,-3.5);

\node[below, text width = 1.3cm] at (-5.9,-3.78) {\small 0};
\node[below, text width = 1.3cm] at (-4.9,-3.78) {\small 1};
\node[below, text width = 1.3cm] at (-3.9,-3.78) {\small 2};
\node[below, text width = 1.3cm] at (-2.9,-3.78) {\small 3};
\node[below, text width = 1.3cm] at (-1.9,-3.78) {\small 4};

\node[below right] at (-8.0318,-1.6) {Fluent};

\node[left] at (-7,-2.45) {\small crim(bert)};
\node[left] at (-7,-2.7) {\small active(crim0)};
\node[left] at (-7,-2.95) {\small active(leg0)};

\node[left] at (-7,-3.5) {\small active(leg0)};

\node[below right, text width=1.3cm] at (-2.6578,-1.7) {Version};

\node[right, text width=1.3cm] at (-2,-2.6724) {1};
\node[right, text width=1.3cm] at (-2,-3.5) {0};

\node[below right] at (-4.88,-4.2) {Time};

\node [below right,regular polygon,
        regular polygon sides=3,
        fill=black!10!green,
        inner sep =0.2em,
        minimum size=1em,
        shape border rotate = -90] at (-5.5,-2.2) {};

\node [below left,regular polygon,
        regular polygon sides=3,
        fill= black!10!red,
        inner sep =0.2em,
        minimum size=1em,
        shape border rotate = 90] at (-3.5,-2.2) {};

\end{tikzpicture}

\caption[Institutional enactment governance case~\ref{exSimple}]{Model for case~\ref{exSimple} with two institution versions.}
\label{figExampleSimple}
\end{figure}

For clarity, models are represented graphically. The model for case~\ref{exSimple} is shown in Figure~\ref{figExampleSimple}. Lines represent when domain and active rule fluents hold. We distinguish between whether a fluent holds in a state $S_{v:t}$: \begin{inparaenum} \item retroactively in the version's past and \textit{not} in the previous version, \tikz[baseline=-0.5ex, very thick, color=blue, dashed]{ \draw [-] (0,0) -- (5ex,0); } (i.e. $R(t) < v$ and $\langle M, S_{v-1:t} \rangle \not \models f$), \item when the version is the \textit{current version}, \tikz[baseline=-0.5ex, very thick]{ \draw [-] (0,0) -- (5ex,0); }) (i.e. $R(t) = v$), and \item when the version is \textit{obsolete}, \tikz[baseline=-0.5ex, dotted, color=red, very thick]{ \draw [-] (0,0) -- (5ex,0); } (i.e. $R(t) > v$). \end{inparaenum} Time instants are marked if they have successful or non--successful rule modification events in versions where modifications can have an effect (i.e. non--obsolete versions): \begin{inparaenum} \item \tikz{ \node [below right,regular polygon,
        regular polygon sides=3,
        fill=black!30!green,
        inner sep =0.2em,
        minimum size=1em,
        shape border rotate = -90] at (-2.75,0.75) {};} denoting that all the rule modification events occurring in the previous version occur again (i.e. $E_{v:t} \cap \pazocal{E}_{\textit{mod}} = E_{v-1:t} \cap \pazocal{E}_{\textit{mod}}$). Meaning, the conditions (contexts) for the rule modifying events to be ascribed are consistent with the version and therefore with applying the rule modifications (the non--deterministic choice to include a rule modification in $E_{v:t}$ according to \ref{eqGR4} is always made) \item \tikz{
\node [below left,regular polygon,
        regular polygon sides=3,
        fill=black!10!red,
        inner sep =0.2em,
        minimum size=1em,
        shape border rotate = 90] at (-3.5,0.75) {};} denoting that at least one rule modification event which occurred in the previous version does not occur again (i.e. $E_{v:t} \cap \pazocal{E}_{\textit{mod}} \neq E_{v-1:t} \cap \pazocal{E}_{\textit{mod}}$). Meaning, the conditions (contexts) for rule modifying events to be ascribed are inconsistent with the version they occur in and therefore with applying the rule modifications (a non--deterministic choice according to \ref{eqGR4} to \textit{not} include a rule modification is made when building $E_{v:t}$). \end{inparaenum}
        
Figure~\ref{figExampleSimple} shows case~\ref{exSimple}'s model. Throughout version zero the legislative rule ($\textit{leg0}$) is active, stating Ada and Bertrand voting to add a rule counts--as adding a rule. When at time instant one Ada and Bertrand vote to add a new rule ($\textit{crim0}$), stating people found guilty become criminals, the model succeeds to version one where the new rule is successfully added. At time instant three Bertrand becomes a criminal. When they vote again to modify a rule it is unsuccessful, since rule change is conditional on neither being criminals. Adding a criminalising rule altered the built social reality in version one's future, changing what could be ascribed as a legal rule modification.

The next case presents an institution $\pazocal{I}^{\textit{uk}}$ representing the UK's legislation rules. The cases are based on past changes to a court decision on UK tax laws (\cite{PadmoreIRC1987}), and past changes to tax laws (\cite{FinanceAct2008}). The UK government can unconditionally enact any rules effective at any time. Observable events where the government activates/deactivates a rule (\textit{gmod(op, id, t)}) count--as modifying the rule (\textit{mod(op, id, t)}. Legislative rules identified as $\textit{leg0} \in \mathbb{ID}$ cause rule activations $\pazocal{G}^{\textit{uk}}(\top, \{ \textit{gmod(act, id, t)} \}) \ni \textit{mod(act, id, t)}$ and legislative rules identified as $\textit{leg1} \in \mathbb{ID}$ cause rule deactivations $\pazocal{G}^{\textit{uk}}( \top, \{ \textit{gmod(deact, id, t)} \}) \ni \textit{mod(deact, id, t)}$. A model $M^{\textit{uk}} = \langle R^{\textit{uk}}, V^{\textit{uk}} \rangle$ is produced for an observable event trace $\textit{et} = \langle O_0, O_1, O_2, O_3, O_4 \rangle$ for $\pazocal{I}^{\textit{uk}}$. The model comprises four versions $V^{\textit{uk}} = \langle V^{\textit{uk}}_0, V^{\textit{uk}}_1, V^{\textit{uk}}_2, V^{\textit{uk}}_3 \rangle$. We begin the case:

\begin{case}\label{exUK1} A rule states that any UK resident (e.g. person \textit{a} resides in the UK --\textit{r(a, uk)}) in a business partnership in the UK (\textit{p(a, uk)}) or elsewhere such as Jersey (\textit{p(a, jers)}) in the first tax year month is obliged to pay tax (\textit{oblt}). We have for all locations $L \in \{ \textit{uk, jers} \}$ a tax rule $\pazocal{C}^{\textit{uk}\uparrow}(\textit{r(a, uk)} \wedge \textit{p(a, L)}, \textit{mon1}) \ni \textit{oblt}$ identified as $\textit{tax0} \in \mathbb{ID}$. Initially the legislative rules \textit{leg0} and \textit{leg1}, and the tax rule $\textit{tax0}$ are active ($\Delta^{\textit{uk}} = \{ \textit{active(leg0)}, \textit{active(leg1)}, \textit{active(tax0)} \}$). At time point one it is the first tax year month ($O_1 = \{ \textit{mon1} \}$). Following a court challenge (\cite{PadmoreIRC1987}) the government retroactively replaces the tax rule with id \textit{tax0} with a new rule with id \textit{tax1} ($O_2 = \{ \textit{gmod(deact, tax0, 0)}, \textit{gmod(act, tax1, 0)} \}$). The new rule, \textit{tax1}, states that only people in a UK business partnership are obliged to pay tax -- $\pazocal{C}^{\textit{uk}\uparrow}(\textit{r(a, uk)} \wedge \textit{p(a, uk)}, \textit{mon1}) \ni \textit{oblt}$. \end{case}

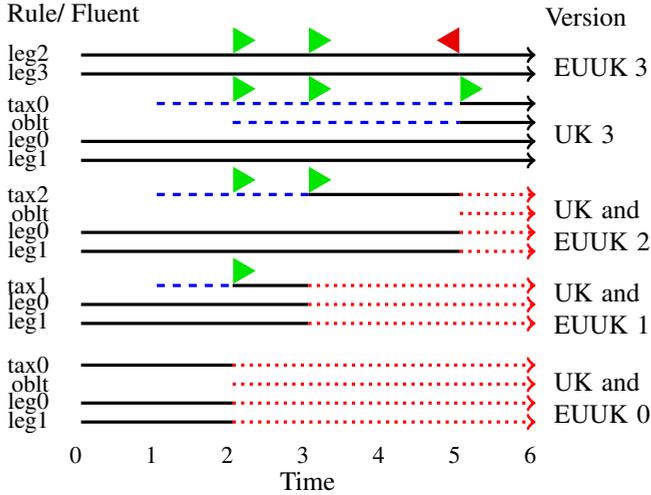
\begin{figure}[h]
\centering
\begin{tikzpicture}

\draw [very thick] (-6.5,1) -- (-1.5,1);
\draw [very thick] (-6.5,0.75) -- (-1.5,0.75);
\draw [->, very thick] (-1.5,1) -- (-0.5,1);
\draw [->, very thick] (-1.5,0.75) -- (-0.5,0.75);
\node[below right] at (-7.6,1.8) {Rule/ Fluent};
\node[left] at (-6.8,1) {\small leg2};
\node[left] at (-6.8,0.75) {\small leg3};
\node[below right, text width=1.3cm] at (-0.5,1.75) {Version};
\node[right] at (-0.38,0.84) {EUUK 3};
\node [below left,regular polygon,
        regular polygon sides=3,
        fill=black!10!red,
        inner sep =0.2em,
        minimum size=1em,
        shape border rotate = 90] at (-1.5,1.3) {};
\node [below right,regular polygon,
        regular polygon sides=3,
        fill=black!10!green,
        inner sep =0.2em,
        minimum size=1em,
        shape border rotate = -90] at (-3.5,1.3) {};
\node [below right,regular polygon,
        regular polygon sides=3,
        fill=black!10!green,
        inner sep =0.2em,
        minimum size=1em,
        shape border rotate = -90] at (-4.5,1.3) {};

\draw [very thick, dashed, color=blue] (-5.5,0.36) -- (-1.5,0.36);
\draw [->, very thick] (-1.5,0.36) -- (-0.5,0.36);
\draw [very thick, dashed, color=blue] (-4.5,0.11) -- (-1.5,0.11);
\draw [->, very thick] (-1.5,0.11) -- (-0.5,0.11);
\draw [->, very thick] (-6.5,-0.14) -- (-0.5,-0.14);
\draw [->, very thick] (-6.5,-0.39) -- (-0.5,-0.39);

\draw [very thick, dashed, color=blue] (-5.5,-0.84) -- (-3.5,-0.84);
\draw [very thick] (-3.5,-0.84) -- (-1.5,-0.84);
\draw [->, very thick, dotted, color=red] (-1.5,-0.84) -- (-0.5,-0.84);
\draw [->, very thick, dotted, color=red] (-1.5,-1.09) -- (-0.5,-1.09);
\draw [very thick] (-6.5,-1.34) -- (-1.5,-1.34);
\draw [->, very thick, dotted, color=red] (-1.5,-1.34) -- (-0.5,-1.34);
\draw [very thick] (-6.5,-1.59) -- (-1.5,-1.59);
\draw [->, very thick, dotted, color=red] (-1.5,-1.59) -- (-0.5,-1.59);

\draw [very thick, dashed, color=blue] (-5.5,-2.04) -- (-4.5,-2.04);
\draw [very thick] (-4.5,-2.04) -- (-3.5,-2.04);
\draw [->, very thick, dotted, color=red] (-3.5,-2.04) -- (-0.5,-2.04);
\draw [very thick] (-6.5,-2.29) -- (-3.5,-2.29);
\draw [->, very thick, dotted, color=red] (-3.5,-2.29) -- (-0.5,-2.29);
\draw [very thick] (-6.5,-2.54) -- (-3.5,-2.54);
\draw [->, very thick, dotted, color=red] (-3.5,-2.54) -- (-0.5,-2.54);

\draw [very thick] (-6.5,-3.09) -- (-4.5,-3.09);
\draw [->, very thick, dotted, color=red] (-4.5,-3.09) -- (-0.5,-3.09);
\draw [->, very thick, dotted, color=red] (-4.5,-3.34) -- (-0.5,-3.34);
\draw [very thick] (-6.5,-3.59) -- (-4.5,-3.59);
\draw [->, very thick, dotted, color=red] (-4.5,-3.59) -- (-0.5,-3.59);
\draw [very thick] (-6.5,-3.84) -- (-4.5,-3.84);
\draw [->, very thick, dotted, color=red] (-4.5,-3.84) -- (-0.5,-3.84);

\node[below, text width = 1.3cm] at (-6,-4.04) {\small 0};
\node[below, text width = 1.3cm] at (-5,-4.04) {\small 1};
\node[below, text width = 1.3cm] at (-4,-4.04) {\small 2};
\node[below, text width = 1.3cm] at (-3,-4.04) {\small 3};
\node[below, text width = 1.3cm] at (-2,-4.04) {\small 4};
\node[below, text width = 1.3cm] at (-1,-4.04) {\small 5};
\node[below, text width = 1.3cm] at (0,-4.04) {\small 6};

\node[left] at (-6.8,0.36) {\small tax0};
\node[left] at (-6.8,0.11) {\small oblt};
\node[left] at (-6.8,-0.14) {\small leg0};
\node[left] at (-6.8,-0.39) {\small leg1};

\node[left] at (-6.8,-0.84) {\small tax2};
\node[left] at (-6.8,-1.09) {\small oblt};
\node[left] at (-6.8,-1.34) {\small leg0};
\node[left] at (-6.8,-1.59) {\small leg1};

\node[left] at (-6.8,-2.04) {\small tax1};
\node[left] at (-6.8,-2.29) {\small leg0};
\node[left] at (-6.8,-2.54) {\small leg1};

\node[left] at (-6.8,-3.09) {\small tax0};
\node[left] at (-6.8,-3.34) {\small oblt};
\node[left] at (-6.8,-3.59) {\small leg0};
\node[left] at (-6.8,-3.84) {\small leg1};

\node[right, text width=1.3cm] at (-0.38,-0.05) {UK 3};
\node[right, text width=1.5cm] at (-0.38,-1.25) {UK and EUUK 2};
\node[right, text width=1.5cm] at (-0.38,-2.34) {UK and EUUK 1};
\node[right, text width=1.5cm] at (-0.38,-3.52) {UK and EUUK 0};

\node[below right] at (-4,-4.38) {Time};

\node [below right,regular polygon,
        regular polygon sides=3,
        fill=black!10!green,
        inner sep =0.2em,
        minimum size=1em,
        shape border rotate = -90] at (-1.5,0.66) {};

\node [below right,regular polygon,
        regular polygon sides=3,
        fill=black!10!green,
        inner sep =0.2em,
        minimum size=1em,
        shape border rotate = -90] at (-3.5,0.66) {};

\node [below right,regular polygon,
        regular polygon sides=3,
        fill=black!10!green,
        inner sep =0.2em,
        minimum size=1em,
        shape border rotate = -90] at (-4.5,0.66) {};

\node [below right,regular polygon,
        regular polygon sides=3,
        fill=black!10!green,
        inner sep =0.2em,
        minimum size=1em,
        shape border rotate = -90] at (-3.5,-0.54) {};

\node [below right,regular polygon,
        regular polygon sides=3,
        fill=black!10!green,
        inner sep =0.2em,
        minimum size=1em,
        shape border rotate = -90] at (-4.5,-0.54) {};
        
\node [below right,regular polygon,
        regular polygon sides=3,
        fill=black!10!green,
        inner sep =0.2em,
        minimum size=1em,
        shape border rotate = -90] at (-4.5,-1.74) {};
        
\end{tikzpicture}
\caption[Institutional enactment governance case~\ref{exUK1}]{Model for case~\ref{exUK1} with four institution versions for the institution $\pazocal{I}^{uk}$ (denoted UK) and a model for case~\ref{exEUUK} with four versions for the institution $\pazocal{I}^{euuk}$ (denote EUUK). Institutions $\pazocal{I}^{uk}$ and $\pazocal{I}^{euuk}$ have identical versions 0 to 2. Not shown, in all states person `a' is in a Jersey--based business partnership (\textit{p(a,jersey)}) and is a UK resident (\textit{r(a,uk)}).}
\label{figUKExampleAndEUUKExample}
\end{figure}

In Figure~\ref{figUKExampleAndEUUKExample} version zero ($V^{\textit{uk}}_0$) obliges person `a' to pay tax in state two. In state two ($S^{\textit{uk}}_{0:2}$) the current institution version changes when the rule obliging UK residents to pay tax (\textit{tax0}) is replaced  with the rule obliging UK business partners to pay tax (\textit{tax0}) ($\textit{act(t1,0)}, \textit{deact(t0,0)} \in  E^{\textit{uk}}_{0:2}$) to version one ($V^{\textit{uk}}_1$ s.t. $R^{\textit{uk}}(2) = 1$). Due to this change, the version one ($V^{\textit{uk}}_1$), does not oblige tax to be paid in its third state ($S^{\textit{uk}}_{1:2}$) since person \textit{a} resides in the UK but is in a Jersey--based business partnership.

\addtocounter{case}{-1}
\begin{case}[continued]
The government partially reverses the tax change at time point three. This is by retroactively replacing the rule obliging people in a UK business partnership to pay tax (\textit{tax1}) with a new rule  identified as \textit{tax2} \\ ($O_3 = \{ \textit{gmod(deact, tax1, 0)}, \textit{gmod(act, tax2, 0)} \}$). The new rule obliges UK \textit{residents} in a business partnership to pay tax if it does not criminalise them retroactively (i.e. in a retroactive state an obligation to pay tax is initiated conditional on the obligation holding in the next state of the previous version). For all locations $L \in \{ \textit{uk, jersey} \}$ the rule is $\pazocal{C}^{\textit{uk}\uparrow}((\textit{r(a, uk)} \wedge \textit{p(a, L)}) \rightarrow (\textit{P}) \rightarrow \textit{PrV}(\textit{NS(} \textit{oblt}))), \textit{mon1}) \ni \textit{oblt}$. Next, it is the first tax year month again ($O_4 = \{ \textit{mon1} \}$).
\end{case}

In Figure~\ref{figUKExampleAndEUUKExample} version two ($V^{\textit{uk}}_2$), like version zero, does not oblige `a' to pay tax in the past. But, it does oblige them to pay tax after the second time the first tax year month occurs ($\textit{mon1} \in E^{\textit{uk}}_{2:4}$).

\addtocounter{case}{-1}
\begin{case}[continued] The UK government decides to reverse the previous judgements going back to the original rule set ($O_5 = \{ \textit{gmod(deact, tax2, 0)}, \textit{gmod(act, tax0, 0)} \}$).
\end{case}

In Figure~\ref{figUKExampleAndEUUKExample}, version three ($V^{\textit{uk}}_3$) reverts to the original legislation. Thus we have the same situation as if the legislation in version zero had not been modified. That is, an obligation to pay tax after the first occurrence of the first tax year month ($\textit{mon1} \in E^{\textit{uk}}_{3:1}$).

The next case is a variation on the previous describing an institution $\pazocal{I}^{\textit{euuk}}$, incorporating EU human rights law.

\begin{case}\label{exEUUK} The European Convention on Human Rights \cite[Art. 7]{ECHR1953} (ECHR) blocks retroactive legislative modifications that \textit{criminalise} formerly innocent people. The institution $\pazocal{I}^{\textit{euuk}}$ contains the same rules as $\pazocal{I}^{\textit{uk}}$ with the same identifiers minus the legislative rules $\textit{leg0}$ and $\textit{leg1}$. Instead, legislative rules state that observable rule modifications \textit{count--as} rule modifications \textit{conditional} on the changes not retroactively criminalising people. In all states where rules are being applied retroactively, if there is not an obligation to pay tax in the previous version then there must not be an obligation to pay tax in the current version. We have rules with the identifier \\$\textit{l2}$: $\pazocal{G}^{\textit{euuk}}(\textit{PaS}(\textit{P} \rightarrow \textit{PrV}(\neg \textit{oblt}) \rightarrow \neg \textit{oblt})), \{ \textit{gmod(act, id, t)} \}) \ni \textit{act(id, t)}$, and rules with the identifier $\textit{l3}$: $\pazocal{G}^{\textit{euuk}}(\textit{PaS}(\textit{P} \rightarrow \textit{PrV}(\neg \textit{oblt}) \rightarrow \neg \textit{oblt})), \{ \textit{gmod(deact, id, t)} \}) \ni \textit{deact(id, t)}$. Initially, person `a' is in a Jersey based business partnership (\textit{p(a,jersey)}) and is a UK resident (\textit{r(a,uk)}), and the first tax rule and the legislative rules conditional on being non retroactively criminalising are active such that $\Delta^{\textit{euuk}} = \{ \textit{p(a,uk)}, \textit{r(a,uk)}, \textit{tax0}, \textit{l2}, \textit{l3} \}$. The same events occur as in case~\ref{exUK1}, $\textit{et} = \langle \emptyset, \{ \textit{mon1} \},$ $\{ \textit{gmod(deact, t0, 0)},$ $\textit{gmod(act, t1, 0)} \},$ $\{ \textit{gmod(deact, t1, 0)},$ $\textit{gmod(act, t2, 0)} \},$ $\{ \textit{mon1} \},$ $\{ \textit{gmod(deact, t2, 0)},$ $\textit{gmod(act, t0, 0)} \} \rangle$.
\end{case}

Figure~\ref{figUKExampleAndEUUKExample} shows a model $M^{\textit{euuk}}$ for $\pazocal{I}^{\textit{euuk}}$. The first three versions are identical to our previous case~\ref{exUK1} (where the UK's legislature was not constrained by EU rules blocking retroactively criminalising modifications), since the first two rule modifications do not criminalise people retroactively. Unlike in our previous case~\ref{exUK1}, the version two contains no tax rules. The reason being that tax rule two -- ``obliging uk residents in a business partnership to pay tax but on the condition that if it is retroactive then those people were obliged to pay tax in the previous version'', is deactivated since its deactivation does not criminalise retroactively. On the other hand, tax rule zero -- ``any UK resident in a business partnership in the first tax year month is obliged to pay tax'' ($\pazocal{C}^{\textit{uk}\uparrow}(\textit{r(a, uk)} \wedge \textit{p(a, L)}, \textit{mon1}) \ni \textit{oblt}$) is not reactivated, even though it was reactivated in our previous case~\ref{exUK1}. Its reactivation would retroactively criminalise people if activated in version three, meaning its activation does not occur since legislative rule -- $\textit{l2}$: $\pazocal{G}^{\textit{euuk}}(\textit{PaS}(\textit{P} \rightarrow \textit{PrV}(\neg \textit{oblt}) \rightarrow \neg \textit{oblt})), \{ \textit{gmod(act, id, t)} \}) \ni \textit{act(id, t)}$ -- has a condition that is not met.

The next cases look at modifying legislative rules themselves.

\begin{case}\label{exML1} An institution $\pazocal{I}^{\textit{p}}$ describes a parliament that can retroactively modify rules through a majority vote $\textit{pvote(act, id, t)} \in \pazocal{E}_{\textit{obs}}$. The legislative rules are identified the id $\textit{parl0} \in \mathbb{ID}$ for activating rules $\pazocal{G}^{\textit{p}}(\top, \{ \textit{pvote(act, id, t)} \}) \ni \textit{mod(act, id, t)}$ and with the id $\textit{parl1} \in \mathbb{ID}$ for deactivating rules $\pazocal{G}^{\textit{p}}(\top, \{ \textit{pvote(deact, id, t)} \}) \ni \textit{mod(deact, id, t)}$. In the initial state all rules are active such that $\textit{active(id)} \in \Delta$. In an observable event set trace $\textit{tr} = \langle O_0, O_1 \rangle$ at time point one the parliament votes to retroactively remove the rule which ascribes retroactive modifications ($O_1 = \{ \textit{pvote(deact, parl1, 0)} \}$.
\end{case}

Depicted in Figure~\ref{figMLExample1} a single model $M^{\textit{p}} = \langle R^{\textit{p}}, V^{\textit{p}} \rangle$ comprises two institution versions $V^{\textit{p}} = \langle V^{\textit{p}}_0, V^{\textit{p}}_1 \rangle$. An event occurs in version zero at time instant one, where the parliament votes to retroactively modify a rule and the corresponding rule modification event occurs ($E^{\textit{p}}_{0:1} = \{ \textit{pvote(deact, parl1, 0)}, \textit{mod(deact, parl1, 0)} \}$). Consequently the institution transitions to version one ($R^{\textit{p}}(1) = 1$). Importantly, in version one, the same rule modifying event \textit{does not occur}. The reason being, if the modification event did occur then the rule $\textit{parl1}$ ascribing the modification event -- $\pazocal{G}^{\textit{p}}(\top, \{ \textit{pvote(deact, id, t)} \}) \ni \textit{mod(deact, id, t)}$ -- would be inactive in version one state one $S^{\textit{p}}_{1:1}$, and the deactivation could not occur in the first place (contradiction). This exemplifies how the formalism always guarantees a model, paradoxical rule modifications do not occur if they make the rule modifying event impossible in the first place.

\begin{figure}[h]
\centering
\begin{tikzpicture}[xscale=0.95]
\draw [very thick] (-6.5,-2.25) -- (-5.5,-2.25);
\draw [->, very thick] (-5.5,-2.25) -- (-4.5,-2.25);
\draw [very thick] (-6.5,-2.5) -- (-5.5,-2.5);
\draw [->, very thick] (-5.5,-2.5) -- (-4.5,-2.5);

\draw [very thick] (-6.5,-3.05) -- (-5.5,-3.05);
\draw [->, very thick, dotted, color=red] (-5.5,-3.05) -- (-4.5,-3.05);
\draw [very thick] (-6.5,-3.3) -- (-5.5,-3.3);
\draw [->, very thick, dotted, color=red] (-5.5,-3.3) -- (-4.5,-3.3);

\node[below, text width = 1.3cm] at (-6,-3.66) {\small 0};
\node[below, text width = 1.3cm] at (-5,-3.66) {\small 1};
\node[below, text width = 1.3cm] at (-4,-3.66) {\small 2};

\node[below, text width = 1.3cm] at (-5.25,-4.12) {\small Time};

\node[below right] at (-7.95,-1.4) {Fluent};

\node[left] at (-7,-2.25) {\small active(parl0)};
\node[left] at (-7,-2.5) {\small active(parl1)};

\node[left] at (-7,-3.05) {\small active(parl0)};
\node[left] at (-7,-3.3) {\small active(parl1)};

\node[below right, text width=1.3cm] at (-4.55,-1.46) {Version};

\node[right, text width=1.3cm] at (-4,-2.37) {1};
\node[right, text width=1.3cm] at (-4,-3.17) {0};
     
\node [below left,regular polygon,
        regular polygon sides=3,
        fill=black!10!red,
        inner sep =0.2em,
        minimum size=1em,
        shape border rotate = 90] at (-5.5,-1.93) {};

\end{tikzpicture}
\caption[Institutional enactment governance case~\ref{exML1}]{Model for case~\ref{exML1}}
\label{figMLExample1}
\end{figure}

The next case extends the previous case~\ref{exML1}:

\begin{case}\label{exML3} This case describes an institution $\pazocal{I}^{\textit{mp}}$ where a monarch and a parliament can retroactively modify rules, including all the rules from the previous case's institution $\pazocal{I}^{\textit{p}}$. Additionally, a rule identified as $\textit{fence0} \in \mathbb{ID}$ states that if a fence is built $\textit{fb} \in \pazocal{E}^{\textit{mp}}_{\textit{obs}}$ it is obliged the fence is painted white $\textit{oblpf} \in \pazocal{F}^{\textit{mp}}_{\textit{inst}}$ -- $\pazocal{C}^{\textit{mp}\uparrow}(\top, \textit{fb}) \ni \textit{oblpf}$. A rule identified as \textit{mon0} states the monarch issuing a rule change decree $\textit{mdecree(act, id, t)} \in \pazocal{E}^{\textit{mp}}_{\textit{obs}}$  to activate a rule counts--as activating the rule -- $\pazocal{G}^{\textit{mp}}( \top, \{ \textit{mdecree(act, id, t)} \}) \ni \textit{mod(act, id, t)}$. A rule identified as \textit{mon1} state the monarch issuing a decree to deactivate a rule counts--as deactivating the rule $\pazocal{G}^{\textit{mp}}(\top, \{ \textit{mdecree(deact, id, t)} \}) \ni \textit{mod(deact, id, t)}$. All legislative rules are initially active, but the fence painting rule is not (s.t. $\textit{active(fence0)} \not \in \Delta^{\textit{mp}}$). At time point one the parliament votes for the fence--painting obligation rule to be activated, ($O_1 = \{ \textit{pvote(act, fence0, 1)} \}$), a fence is built ($O_2 = \{ \textit{fb} \}$), the monarch issues by decree the fence--building rule to be retroactively deactivated at the time it was activated, cancelling its activation ($O_3 = \{ \textit{mdecree(deact, fence0, 1)} \}$). Finally, the parliament votes to retroactively dis--enable the monarch from deactivating rules ($O_4 = \{ \textit{pvote(deact, mon1, 0)} \}$).
\end{case}

\begin{figure}[h]
\centering
\begin{tikzpicture}[xscale=0.95]

\draw [very thick] (-6.5,0.25) -- (-5.5,0.25);
\draw [->, very thick] (-5.5,0.25) -- (-1.5,0.25);
\draw [very thick] (-6.5,0) -- (-5.5,0);
\draw [->, very thick] (-5.5,0) -- (-1.5,0);
\draw [very thick] (-6.5,-0.25) -- (-5.5,-0.25);
\draw [->, very thick] (-5.5,-0.25) -- (-1.5,-0.25);
\draw [very thick] (-6.5,-0.5) -- (-6.5,-0.5);
\draw [->, very thick] (-6.5,-0.5) -- (-5.5,-0.5);
\draw [very thick, dashed, color=blue] (-4.5,-0.75) -- (-2.5,-0.75);
\draw [->, very thick] (-2.5,-0.75) -- (-1.5,-0.75);
\draw [very thick, dashed, color=blue] (-3.5,-1) -- (-2.25,-1);
\draw [->, very thick] (-2.5,-1) -- (-1.5,-1);

\draw [very thick] (-6.5,-1.55) -- (-2.5,-1.55);
\draw [->, very thick, dotted, color=red] (-2.5,-1.55) -- (-1.5,-1.55);
\draw [very thick] (-6.5,-1.8) -- (-2.5,-1.8);
\draw [->, very thick, dotted, color=red] (-2.5,-1.8) -- (-1.5,-1.8);
\draw [very thick] (-6.5,-2.05) -- (-2.5,-2.05);
\draw [->, very thick, dotted, color=red] (-2.5,-2.05) -- (-1.5,-2.05);
\draw [very thick] (-6.5,-2.3) -- (-2.5,-2.3);
\draw [->, very thick, dotted, color=red] (-2.5,-2.3) -- (-1.5,-2.3);

\draw [very thick] (-6.5,-2.91) -- (-3.5,-2.91);
\draw [->, very thick, dotted, color=red] (-3.5,-2.91) -- (-1.5,-2.91);
\draw [very thick] (-6.5,-3.16) -- (-3.5,-3.16);
\draw [->, very thick, dotted, color=red] (-3.5,-3.16) -- (-1.5,-3.16);
\draw [very thick] (-6.5,-3.41) -- (-3.5,-3.41);
\draw [->, very thick, dotted, color=red] (-3.5,-3.41) -- (-1.5,-3.41);
\draw [very thick] (-6.5,-3.66) -- (-3.5,-3.66);
\draw [->, very thick, dotted, color=red] (-3.5,-3.66) -- (-1.5,-3.66);
\draw [very thick] (-4.5,-3.91) -- (-3.5,-3.91);
\draw [->, very thick, dotted, color=red] (-3.5,-3.91) -- (-1.5,-3.91);
\draw [very thick] (-3.5,-4.16) -- (-3.5,-4.16);
\draw [->, very thick, dotted, color=red] (-3.5,-4.16) -- (-1.5,-4.16);

\draw [very thick] (-6.5,-4.71) -- (-5.5,-4.71);
\draw [->, very thick, dotted, color=red] (-5.5,-4.71) -- (-1.5,-4.71);
\draw [very thick] (-6.5,-4.96) -- (-5.5,-4.96);
\draw [->, very thick, dotted, color=red] (-5.5,-4.96) -- (-1.5,-4.96);
\draw [very thick] (-6.5,-5.21) -- (-5.5,-5.21);
\draw [->, very thick, dotted, color=red] (-5.5,-5.21) -- (-1.5,-5.21);
\draw [very thick] (-6.5,-5.46) -- (-5.5,-5.46);
\draw [->, very thick, dotted, color=red] (-5.5,-5.46) -- (-1.5,-5.46);

\node[below, text width = 1.3cm] at (-5.85,-5.73) {\small 0};
\node[below, text width = 1.3cm] at (-4.85,-5.73) {\small 1};
\node[below, text width = 1.3cm] at (-3.85,-5.73) {\small 2};
\node[below, text width = 1.3cm] at (-2.85,-5.73) {\small 3};
\node[below, text width = 1.3cm] at (-1.85,-5.73) {\small 4};
\node[below, text width = 1.3cm] at (-0.85,-5.73) {\small 5};

\node[below right] at (-7.9,1.2) {Fluent};

\node[left] at (-7,0.25) {\small active(parl0)};
\node[left] at (-7,0) {\small active(parl1)};
\node[left] at (-7,-0.25) {\small active(mon0)};
\node[left] at (-7,-0.5) {\small active(mon1)};
\node[left] at (-7,-0.75) {\small active(fence0)};
\node[left] at (-7,-1) {\small oblf};

\node[left] at (-7,-1.55) {\small active(parl0)};
\node[left] at (-7,-1.8) {\small active(parl1)};
\node[left] at (-7,-2.05) {\small active(mon0)};
\node[left] at (-7,-2.3) {\small active(mon1)};

\node[left] at (-7,-2.91) {\small active(parl0)};
\node[left] at (-7,-3.16) {\small active(parl1)};
\node[left] at (-7,-3.41) {\small active(mon0)};
\node[left] at (-7,-3.66) {\small active(mon1)};
\node[left] at (-7,-3.91) {\small active(fence0)};
\node[left] at (-7,-4.16) {\small oblf};

\node[left] at (-7,-4.71) {\small active(parl0)};
\node[left] at (-7,-4.96) {\small active(parl1)};
\node[left] at (-7,-5.21) {\small active(mon0)};
\node[left] at (-7,-5.46) {\small active(mon1)};

\node[below right, text width=1.3cm] at (-1.75,1.2) {Version};

\node[right, text width=1.3cm] at (-1,-0.2) {3};
\node[right, text width=1.3cm] at (-1,-2) {2};
\node[right, text width=1.3cm] at (-1,-3.4) {1};
\node[right, text width=1.3cm] at (-1,-5) {0};

\node[below right] at (-3.91,-6.19) {Time};

\node [below right,regular polygon,
        regular polygon sides=3,
        fill=black!10!green,
        inner sep =0.2em,
        minimum size=1em,
        shape border rotate = -90] at (-2.75,0.59) {};

\node [below left,regular polygon,
        regular polygon sides=3,
        fill=black!10!red,
        inner sep =0.2em,
        minimum size=1em,
        shape border rotate = 90] at (-3.5,0.59) {};

\node [below right,regular polygon,
        regular polygon sides=3,
        fill=black!10!green,
        inner sep =0.2em,
        minimum size=1em,
        shape border rotate = -90] at (-5.5,0.59) {};

\node [below right,regular polygon,
        regular polygon sides=3,
        fill=black!10!green,
        inner sep =0.2em,
        minimum size=1em,
        shape border rotate = -90] at (-3.5,-1.21) {};

\node [below right,regular polygon,
        regular polygon sides=3,
        fill=black!10!green,
        inner sep =0.2em,
        minimum size=1em,
        shape border rotate = -90] at (-5.5,-1.21) {};

\node [below right,regular polygon,
        regular polygon sides=3,
        fill=black!10!green,
        inner sep =0.2em,
        minimum size=1em,
        shape border rotate = -90] at (-5.5,-2.57) {};

\end{tikzpicture}
\caption[Institutional enactment governance case~\ref{exML3}]{Model for case~\ref{exML3} with four institution versions.}
\label{figMLExample3}
\end{figure}
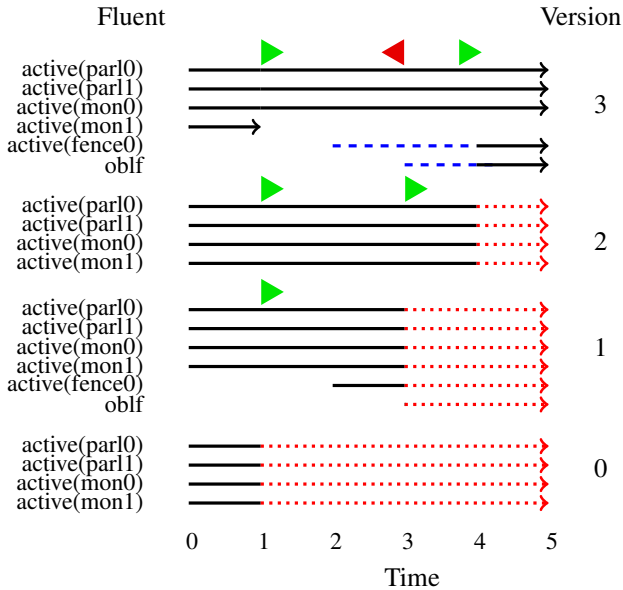

Depicted in Figure~\ref{figMLExample3} the model $M^{\textit{mp}} = \langle R^{\textit{mp}}, V^{\textit{mp}} \rangle$ comprises four versions $V^{\textit{mp}} = \langle V^{\textit{mp}}_{0}, V^{\textit{mp}}_{1}, V^{\textit{mp}}_{2}, V^{\textit{mp}}_{3} \rangle$. At version zero time instant zero the parliament votes to add the rule obliging built fences to be painted white, causing a rule modification event ($E^{\textit{mp}}_{0:1} = \{ \textit{pvote(act, fence0, 1)}, \textit{mod(act, fence0, 1)} \}$) and the institution to transition to the version one ($R^{\textit{mp}}(1) = 1$) where the same modification occurs \\($E^{\textit{mp}}_{1:1} = \{ \textit{pvote(act, fence0, 1)}, \textit{mod(act, fence0, 1)} \}$). In the version one time instant two building a fence ($\textit{fb} \in E^{\textit{mp}}_{1:2}$) causes an obligation to paint the fence $\textit{oblpf} \in S^{\textit{mp}}_{1:3}$. At time instant three the monarch retroactively deactivates the fence painting rule \\($\textit{mdecree(deact, fence0, 1)} \in E^{\textit{mp}}_{1:3}$) causing the institution to transition to the version two ($R^{\textit{mp}}(3) = 2$) where the modification takes effect ($\textit{mdecree(deact, fence0, 1)} \in E^{\textit{mp}}$). Consequently, the fence painting obligation rule is deactivated and its effects (an obligation) no longer hold. When the parliament retroactively removes the ability for the monarch to deactivate rules the institution transitions to the final version three ($R^{\textit{mp}}(4) = 3$) where the parliament's retroactive rule removal takes effect \\($\textit{pvote(deact, mon1, 0)}, \textit{mod(deact, mon1, 0)} \in E^{\textit{mp}}_{3:4}$) causing the monarch's modifications to be unravelled (note that at the final version's third time instant the monarch's rule modification is unsuccessful even though it was successful in the previous version). Consequently, the fence painting obligation rule and its effects (an obligation) is reinstated by retroactively removing the ability to deactivate the fence painting rule.

\subsection{Related Work}
\label{secRelatedWork}

In this chapter we contributed a framework for reasoning about the legality of rule changes. Key to our contribution is the idea of rule--modifying constitutive rules, which express things like ``A counts--as modifying a rule in the past/present/future in context C''. The main idea is to contribute a formal framework for reasoning about the legality of rule change ascribed by constitutive rules. In particular, accounting for the fact that changing rules in the past/present/future affects which rule changes are possible in the first place. Moreover, reasoning about conditions on rule change that are dependent on the rule change effects, such as not criminalising people in the past that were previously innocent (non--retroactively criminalising). Broadly speaking, there are three competing formalisations which look at the legality of rule change. We compare these as follows.

In (\cite{Governatori2005b, Governatori2010}) a defeasible logic is proposed for temporal rule modification operations. Operations include, in (\cite{Governatori2010}), complete rule removal (annulment) and removing immediate rule effects (abrogation). Meta--rules are used to introduce rule changes, which bear similarity to our rule--modifying counts--as rules. However, the meta--rules are only conditional on the social context at a single point in time. For comparison, we formalise richer conditions in rule--modifying constitutive rules. Specifically, rule change being conditional on the social reality in previous institution versions with previous rule--sets, previous institutional states and hypothetical rule changes required to capture a number of important examples we address (such as rule change being non--retroactively criminalising). The focus of these papers is on rule change operations found in the legal domain, rather than the relation between ascribing a social reality using constitutive rules and ascribing rule modifications with constitutive rules conditional on the social reality.

In (\cite{Lopez2003,Lopez2006}) electronic institutions are specified in the Z specification language where legislation norms restrict legislative actions. The conditions for legislation norms are less expressive than our proposal and the authors do not consider the interdependency between changing rules in the past/present/future and the built social reality. 

On the other hand, in \cite{Boella2004} rule modifications ascribed by counts--as rules are formalised where there is such a potential interdependency. The focus in \cite{Boella2004} is on playing games in a setting where agents can act on the environment and also the rules which govern their actions. Most significantly, the main difference is that in \cite{Boella2004} a static setting is formalised where institutions do not evolve over time from state to state or one set of rules to another. In our framework we explicitly look at these aspects not examined in \cite{Boella2004}.

Slightly further afield, the MOISE+ organisational framework has been used to construct organisations that act to enact organisational changes, in this case applied to a football team case study \cite{Hubner2004}. The focus is on the structural organisational elements required to re--organise an organisation. For example, the organisational roles required to assess when a re--organisation is required and what direction the organisational design should take. In contrast to this chapter, the focus is not on constraints making the social actions of norm change possible (or in their case, organisational change). Nor is the meta--level, the change of the rules making rule change possible. Instead, the claim is that they propose an organisational architecture required to make organisational change effective, and their claim is supported with an empirical simulation using a Markov Decision Process to select appropriate organisational changes.

More generally, other work looks at the problem of normative system change. This body of research includes: \begin{inparaenum} \item norm change postulates (\cite{Boella2004}), \item detecting and/or resolving norm inconsistencies (\cite{Jiang2015,Jianga,Kollingbaum2007,Corapi2011,Li2014,Vasconcelos2008}) and \item temporal norm updates (\cite{Alechina2013,Knobbout2014})\end{inparaenum}. However, these frameworks do not look at rule change legality ascribed by constitutive rules over time.

\section{Discussion}
\label{secConclusions}

This chapter answers the question ``how should we define when rule changes count--as legal and valid rule changes?'' with a novel formal framework. Our framework supports reasoning about institutional rule change over time, where rule changes are ascriptively regulated by counts--as rules. We presented a novel semantics defining how an institution evolves from one social reality to the next and from one version of rules to another. Under the proposed semantics counts--as rules define the past/present/future social reality. In turn, rule modifications change counts--as rules in the past/present/future and therefore the constructed social reality. Rule modifications are in turn conditional on the built social reality and their potential effects. To summarise the general answer to the question ``how should we define when rule changes count--as legal and valid rule changes?'', according to our formalisation a legal rule change occurs \begin{inparaenum} \item if the rule change is ascribed by counts--as rules, conditional on a context taking into account the potential changes to the context the rule modification \textit{would} make \item taking into account past/present/future rule modifications effect on counts--as rules and thus the context in which rule changes are conditional on. \end{inparaenum}

In chapters \ref{chapter_3}, \ref{chapter_4} and \ref{chapter_5} our Searlian notion of an institution is a social language that ascribes institutional facts, such as obligations and various social notions such as murder. If an institutional fact is ascribed by a brute fact that is an agentive observable event (i.e. an agent \textit{doing} something) then we say the agent that performed the action had the \textit{legal power} in the Hohfeldian sense to cause the institutional fact to occur or hold. That is, in previous chapters \ref{chapter_3}, \ref{chapter_4} and \ref{chapter_5} we were dealing with the legal power for agents to bring about institutional facts, including obligations when we are dealing with norms. In comparison, in this chapter it is the norms and other counts-as rules themselves that are ascribed from brute facts, or brought about by agents with sufficient legal power. Thus, unlike in previous chapters, this chapter goes beyond merely the notion of legal power to ascribe institutional facts in a limited sense. This chapter, when it talks about legality, is really talking about the legal power to ascribe norms and other counts-as rules themselves, which in turn give agents legal power to bring about further institutional facts and rules.

In terms of future work, there are two important formal and application--orientated avenues. From a formal perspective, we considered rule changes conditional on their potential effects in the past/present/future, but not whether affecting the past/present/future in general. Yet, some administrative laws explicitly block retroactive modifications (e.g. \cite[Art. 1 Sec. 9 Cl. 3]{USConstitution} ``No Bill of Attainder or ex post facto Law shall be passed''). In our framework a weak notion of preventing retroactive modifications can be expressed as a lack of retroactive rule--modifying counts--as rules. A stronger notion cannot be expressed \textit{blocking} rules that allow modifications in the past from being introduced. A natural representation is higher--priority rule modifying counts--as rules (e.g. from a higher authority), but requires a defeasible logic left for further investigation.

In principle, the framework's fixed--point institution model characterisation can be implemented in any adequately expressive language. Future work should investigate if  there is a suitable representation in declarative languages, well suited to rule--based reasoning. One candidate is Answer--Set Programming (ASP) \cite{Gebser2011,Gelfond1988} as used in the InstAL framework \cite{Cliffe2006}. The main obstacles are finding an adequate representation of the institutions presented in this chapter (i.e. rules that can be modified) and optimising for answer--sets that contain a superset of rule modifications over other answer--sets (thus corresponding to our notion of a model).
\chapter{Application}
\label{chapter_7}

\epigraph[0pt]{If you find that you're spending almost all your time on theory, start turning some attention to practical things; it will improve your theories. If you find that you're spending almost all your time on practice, start turning some attention to theoretical things; it will improve your practice.}{Donald Knuth}

\blfootnote{\color{tck-grey}This chapter is loosely based on the following papers:\\
\textbf{King, T. C.}, Liu, Q., Polevoy, G., Weerdt, M. de, Dignum, V., Riemsdijk, M. B. van, \& Warnier, M. (2014). Request Driven Social Sensing (Demonstration). In A. Lomuscio, P. Scerri, A. Bazzan, \& M. Huhns (Eds.), Proceedings of the 2014 International Conference on Autonomous Agents and Multiagent Systems (AAMAS 2014) (pp. 1651 – 1652). Paris, France: International Foundation for Autonomous Agents and Multiagent Systems. \cite{King2014} \\
\textbf{King, T. C.}, Riemsdijk, M. B. Van, Dignum, V., \& Jonker, C. M. (2015). Supporting Request Acceptance with Use Policies. In Coordination, Organizations, Institutions, and Norms in Agent Systems X: COIN 2014 International Workshops, COIN@ AAMAS, Paris, France, May 6, 2014, COIN@ PRICAI, Gold Coast, QLD, Australia, December 4, 2014, Revised Selected Papers (pp. 114 – 131). Springer. \cite{King2014a}}

\newpage
\nocite{King2015c}
In this dissertation each formal contribution has been developed and assessed against case studies. In this chapter, we describe an application and implementation in the context of the SHINE project under which this research was conducted. The SHINE project aims to provide suitable mechanisms that help obtaining detailed environmental data through user participation. The idea is to get users to participate in providing data by contributing the sensors they own together with any actions the user needs to take such as taking a photograph or ensuring a microphone is turned on. In return for donating their time, taking useful actions and contributing their resources, users are able to receive the valuable information provided by the participation of other users. In this scenario, institutions in the form of contracts require users to contribute their cellphone sensors, via cellphone apps, in order to collect and aggregate spatio--temporal weather data. Not all users will wish to accept these contracts if they are considered to be `bad' in terms of how, when and for whom their cellphone sensors are used. Moreover, a contract might take away liberties the user wishes to maintain or not uphold highly--valued rights. Hence, we apply a version of the multi--level governance compliance checking framework (chapter~\ref{chapter_4} and \ref{chapter_5}) in order to automatically check whether these contracts are `good' from the perspective of the user's wishes over how, when and for whom their cellphone sensors are used.

The rest of this chapter begins by describing the application domain in section~\ref{secChap7Domain}. Then we provide an example in section~\ref{secChap7Example}. An architectural overview is given in section~\ref{secChap7Overview} of how the reasoning framework is used in the context of crowdsensing and describe the implementation. We finish with discussion on benefits and limitations of the application in section~\ref{secChap7Concl}.

\section{Contractually Crowdsensing Rain Data}
\label{secChap7Domain}

The SHINE (Sensing Heterogeneous Information Network Environment) project brings the perspective that there are many existing unused sensors owned by various stakeholders. At the same time, detailed environmental data is required to support governments and citizens in making decisions affected by environmental conditions. Such decisions include determining where to go to avoid flooded areas or deciding how city water infrastructure can be improved based on its current effectiveness. One possibility to provide detailed data is to buy sensors, deploy them and then collect data for a single purpose. But this is costly. SHINE aims to overcome a costly single--minded buy--deploy--sense cycle by re--purposing the existing, unused, sensors owned by independent stakeholders.

SHINE views the sensors as being heterogeneous and the stakeholders who own the sensors as being autonomous. Sensor heterogeneity means that the sensors' stakeholders own operate differently. For example, even if we only consider cellphone microphone sensors, one stakeholder might own a cellphone which runs a different operating system and therefore needs to be configured differently (e.g. ensuring data transfer is enabled or recording a video with a cellphone camera) to another stakeholder's cellphone. One way to address the heterogeneity of sensors is to ask stakeholders to ensure their sensors are configured correctly. Stakeholder autonomy means that we cannot force a stakeholder to donate the use of their sensors or to configure them in the correct way.

In SHINE stakeholder autonomy is addressed by contractually obliging stakeholders through bilateral agreement to ensure that they donate and configure their sensors to provide detailed data (e.g. rainfall locations). Specifically, in the context of this dissertation, SHINE aims to enact contracts between stakeholders and sensor networks to \textit{crowdsource} the use of the cellphone sensors stakeholders already own, also known as \textit{crowdsensing} \cite{Ganti2011}. Crowdsensing involves crowdsourcing existing sensing devices as a means to cost--effectively acquire detailed data in a variety of domains. For example, crowdsourcing people's cellphone battery sensors in order to obtain detailed temporal--spatial urban air temperatures \cite{Overeem2013}, using microwave links in cellular networks to obtain detailed rainfall data \cite{Overeem2013a} or using the cellphones in an area to form the wireless network to aggregate and transport the data \cite{Liu2013}. The SHINE project uses contractual crowdsensing to acquire people's cellphone sensors as a means to cost--effectively acquire detailed data in a variety of domains. 

We look at crowdsensing people's cellphones to obtain rain data with an implementation in a simulated prototype. In our case, rain data is obtained using a hypothetical cellphone app which uses the cellphone's audio sensors or communications links between devices to determine rainfall. The hypothetical owner of a cellphone is a human user and as such they have autonomy to join the crowdsensing system, ensure the cellphone sensors are on (i.e. the microphone, GPS and communications links) and that cellphone apps are able to transmit mobile data. The crowdsensing system binds users with contracts (a type of legal institution \cite{Ruiter1997}) for use of their cellphone devices and sensors by imposing various regulations on those users and their cellphones. For example, ``a cellphone is obliged to collect rain data when at a specific location and the user is prohibited from turning the sensor off until the data is collected''. In return, a user might be promised access to the rain data obtained by the crowdsensing system. By requesting a large number of users to agree to such contracts and enough users agreeing to those contracts detailed data can be gathered through the crowdsourcing of sensors.

However, if there are many contracts offered for the use of a user's resource, it is unmanageable for a user to assess the details of each and every one. On the one hand, if a user agrees to a contract without reading the fine print, they are liable to engage in a `bad' (e.g. unfair) contract. For example, a contract which requires the owner to keep audio sensors and data communications on even if the cellphone battery is depleted (i.e. below 10\%). On the other hand, if contracts are rejected by default the user might miss out on the opportunity to obtain useful information (rain data). In the second case, the system is likely to not meet its aims if too many users reject contracts on the basis of not having the time to assess them. Consequently, feasibility demands the automation of contract acceptance and rejection. Yet, the automation must respect an owner's desire to maintain autonomy over how, when and for whom their cellphone contributes data. Unfortunately, existing crowdsensing systems (e.g. \cite{Dutta2009,Hull2006}) do not let users control how their resources are used. In other words, the user needs a way to specify a policy stating regulations a contract should and should not impose for using the user's resource. We call such a policy a \textit{use policy}. Conceptually, fitting with our earlier theoretical work, a contract is a first--level institution governing users in an MAS (the crowdsensing system), whilst the use policy specified by users is a second--level institution governing offered contracts. Hence, a contract sets the space of compliant behaviour required to ensure the collection of crowdsensed data. In comparison, a user's use policy effectively states how, when and for whom their cellphone may be used, or what obligations and prohibitions a contract may impose.

Taking a formalised use policy and contract offered to a user as input, software should automatically accept or reject the contract. Acceptance or rejection should be based on whether the contract places obligations and prohibitions on a user and their device that the user does not like in certain contexts (e.g. the context that the battery is depleted). We use a version of our computational framework from chapter~\ref{chapter_4} for checking compliance of lower--level institutions. We implemented an automated reasoner for contract acceptance/rejection based on compliance in a simulated crowdsensing prototype.

\subsection{Example}
\label{secChap7Example}

In the implemented contract reasoner contracts and use policies are specified in a high--level language. In principle, there are many possible contracts and use policies that can be specified. In order to exemplify the application we will present one contract and one use policy adapted from \cite{King2014a}. For simplicity of explanation, all institutions operate at the same level of abstraction and thus the norm abstraction reasoning from chapters~\ref{chapter_4} and \ref{chapter_5} is not employed. In reality, the crowdsensing system owners can specify their own contracts to offer to users and users can specify their own use policies on the basis of which contracts are accepted or rejected.

The following contract and use policy is formalised using the institution syntax from chapter~\ref{chapter_4}. The contract is a first--level institution $\pazocal{I}^{c}$ offered by the owner of the crowdsensing system, Ada, to a cellphone stakeholder called Bertrand for collecting data. The contract abstracts away from how the data is collected and sent, since this is dependent on the sensor. Data collection may or may not require user input, for example. The contract is formalised in table~\ref{tabChap7ContrForm} and comprises the following rules:

\begin{itemize}
\item Ada is interested in collecting rain data for a specific location, a square, and therefore the contract obliges Bertrand's cellphone to gather rainfall data when he enters the square before leaving the square (rule~\ref{exContr1}).
\item Once Bertrand's cellphone has collected rain data he is obliged to send it within one minute (rule~\ref{exContr2}).
\item If Bertrand enters the square then the contract states that he is forbidden to turn the cellphone's sensors off until the data is collected (rule~\ref{exContr3}).
\end{itemize}

Bertrand has specified a use policy, the institution $\pazocal{I}^{\textit{up}}$, formalised in table~\ref{tabChap7UsePolForm}. It defines the following second--order  norms:

\begin{itemize}
\item Bertrand wishes to be paid for the use of his resource. Every time Bertrand discharges an obligation to send data, it is obliged that the counter--party, Ada, is obliged to pay Bertrand within two minutes where this obligation must be imposed \textit{immediately} (rule~\ref{exUsePol1}).
\item When Bertrand's cellphone battery is depleted (i.e. drops below 10\%), he wishes to conserve it for more important tasks other than collecting rainfall data. Hence, when the battery becomes depleted it is forbidden to oblige Bertrand to keep the cellphone sensors on, until the battery is replenished sufficiently (rule~\ref{exUsePol2}).
\item Bertrand wishes to maintain the liberty to move freely, hence it is always forbidden to oblige Bertrand to collect data before leaving his location (initial state fluent~\ref{exUsePol3}).
\end{itemize}

\setcounter{rowno}{0}
\begin{spacing}{0.1}
\begin{longtable}[l]
{%
     >{\collectcell\myalign}%
      m{0.05\textwidth}%
     <{\endcollectcell}%
     >{\collectcell\myalign}%
      m{0.15\textwidth}%
     <{\endcollectcell}%
}
\caption[Crowdsensing contract institution formalisation]{A Contract for Crowdsensing Cellphone Sensors to Gather Rain Data}
\label{tabChap7ContrForm}
\endfirsthead
\endhead
{
& \pazocal{C}^{\textit{c}}(\emptyset, \textit{enter(bertrand, square)}) \ni \\ & 
\textit{obl(collectData(bertrand, rain), leave(bertrand, square))}
} &
{
\tag{Contr.1}\label{exContr1}
} \\
{
& \pazocal{C}^{\textit{c}}(\emptyset, \textit{collectData(bertrand, rain)}) \ni
\textit{obl(sendData(bertrand, rain), m1)}
} &
{
\tag{Contr.2}\label{exContr2}
} \\
{
& \pazocal{C}^{\textit{c}}(\emptyset, \textit{enter(bertrand, square)}) \ni \\ & 
\textit{pro(turnSensorsOff(bertrand), collectData(bertrand, rain))}
} &
{
\tag{Contr.3}\label{exContr3}
}
\end{longtable}
\end{spacing}

\setcounter{rowno}{0}
\begin{spacing}{0.1}
\begin{longtable}[l]
{%
     >{\collectcell\myalign}%
      m{0.05\textwidth}%
     <{\endcollectcell}%
     >{\collectcell\myalign}%
      m{0.125\textwidth}%
     <{\endcollectcell}%
}
\caption[Crowdsensing contract formalisation]{A contract for Crowdsensing Cellphone Sensors to Gather Rain Data}
\label{tabChap7UsePolForm}
\endfirsthead
\endhead
{
& \pazocal{C}^{\textit{up}}(\emptyset, \textit{disch(obl(sendData(bertrand, rain), leave(bertrand, square)))}) \ni \\ & 
\textit{obl(obl(pay(ada, bertrand), m5), now)}
} &
{
\tag{UsePol.1}\label{exUsePol1}
} \\
{
& \pazocal{C}^{\textit{up}}(\emptyset, \textit{batteryDepleted(bertrand)}) \ni \\ &
\textit{pro(pro(turnSensorsOff(bertrand), collectData(bertrand, rain))}, \\ &
\; \; \; \; \; \; \; \; \; \textit{batteryReplenished(bertrand))}
} &
{
\tag{UsePol.2}\label{exUsePol2}
} \\
{
& \Delta^{\textit{up}} \ni
\textit{pro(obl(collectData(bertrand, rain), leave(bertrand, square)), never)}
} &
{
\tag{UsePol.3}\label{exUsePol3}
}
\end{longtable}
\end{spacing}

Intuitively, we can see that Ada's contract violates Bertrand's use policy. Firstly, Bertrand demands payment for providing data, but Ada's contract does not offer payment. Secondly, Bertrand wishes to be able to turn his cellphone sensors off when the cellphone battery is depleted, but Ada's contract requires the sensors to be kept on when collecting data. Thirdly, Ada wants Bertrand to collect data before leaving his location, however, Betrand has stated that he wants the liberty of free movement. Hence, the automated reasoner in this case would reject Ada's contract on the basis of violating the terms of Bertrand's sensor use policy.

\section{System Overview}
\label{secChap7Overview}

\begin{figure}[t]
\centering
\begin{tikzpicture}
\draw [very thick, rounded corners] (-3.3,1.05) rectangle (2.55,-3);
\draw  [thick] (0.8,0.5) node [anchor=north west, text width=3cm] {Use \\Policy} rectangle (2.05,-0.5);
\draw  [thick] (-2.7,0.5) node [anchor=north west, text width=3cm] {Resource \\ Governance} rectangle (-0.7,-0.5);
\draw [<-, very thick, dashed] (-0.7,0) -- (0.8,0);
\draw [very thick, dashed] (-3.2,-1) -- (2.55,-1);
\draw [thick]  (-2.7,-1.5) node [anchor=north west, align=center, text width=4.5cm] {Data Collection and Aggregation} rectangle (2.05,-2.5);
\node [align=left, anchor=west] at (2.8,0) {Application Layer};
\node [align=left, anchor=west] at (2.8,-2) {Network Layer};
\draw [color=white, fill=white] (-2.85,1.2) node [anchor=west, align=left, text width=4.5cm, color=black] {Cellphone app} rectangle (-0.25,0.75);

\draw [very thick, rounded corners] (-3.3,-4.45) rectangle (2.55,-5.2);

\draw [color=white, fill=white] (-2.9,-4.3) node [anchor=west, align=left, text width=4.5cm, color=black] {Cellphone app} rectangle (-0.5,-4.6);
\draw [color=white, fill=white] (-3.45,-4.9) rectangle (2.7,-5.5);
\draw [thick, ->] (-2.25,2.4) node [anchor=north east, align=right] {\\Contract offer} -- (-2.25,1.5);
\draw [thick, ->] (-1.8,1.5) -- (-1.8,2.4) node [anchor=north west, align=left] {\\Accept/reject contract};

\draw [fill=white] (-3,2.6) ellipse (0.3 and 0.3) node {1};
\draw [fill=white] (-2.85,0.6) ellipse (0.3 and 0.3) node {2};
\draw [fill=white] (-2.1,-1.5) ellipse (0.3 and 0.3) node {3};
\draw [fill=white] (-2.1,-2.55) ellipse (0.3 and 0.3) node {4};
\draw [fill=white] (-1.95,-3.45) ellipse (0.3 and 0.3) node {5};
\draw [thick, <->] (-1.5,-3) node [anchor=north west, align=left] {\\ Send/receive data} -- (-1.5,-4.05);
\draw [->, thick] (-5.7,-1.9) node [anchor=south west, align=right] {Request Data} -- (-3.3,-1.9);
\end{tikzpicture}
\caption[Simulated application architecture]{A cellphone app is modelled in the simulated prototype comprising two layers, an application layer for the governance of the user's resource (the cellphone and its sensors) and a network layer for using the resource (collecting, aggregating and sending data).}
\label{figChap7Architecture}
\end{figure}
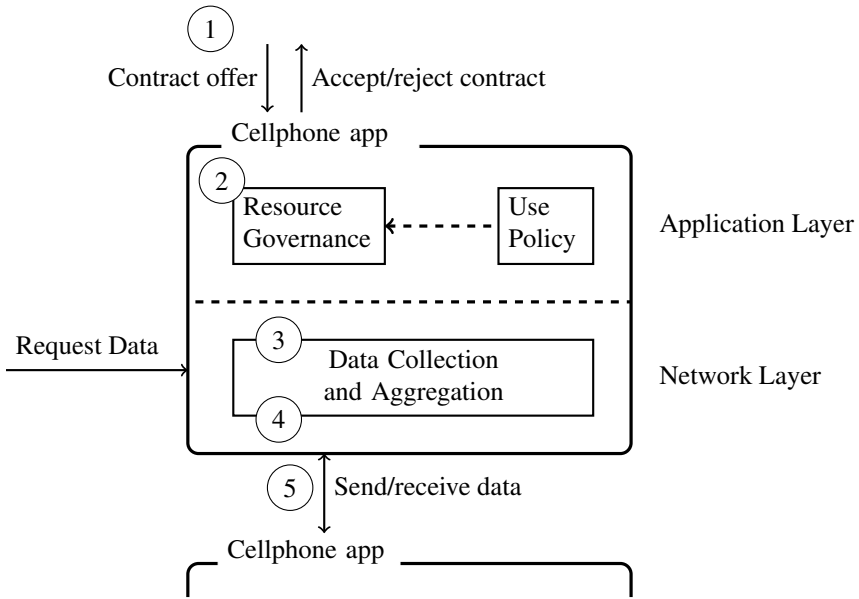

Reasoning about contracts and use policies, as exemplified previously, was implemented as a part of a simulated crowdsensing prototype. In the prototype a simulated cellphone app is instantiated for each simulated user. Each cellphone app combines reasoning for resource governance and data collection. The cellphone app's architecture is depicted in figure~\ref{figChap7Architecture} and described as follows in terms of interaction with the crowdsensing system as a whole and other user's individual cellphone apps:

\begin{enumerate}
\item Before a user's cellphone is crowdsourced in providing sensing information, a contract is offered to the cellphone to provide rainfall sensing data. The contract is ongoing in the sense that it is not for a single data point but rather multiple data points requested on--demand by the system. 
\item A resource governance component, the multi--level governance compliance reasoning, accepts or rejects the contract based on whether it is compliant with a user's use policy. Each user has a different use policy and different users may be engaged in different contracts, hence cellphones do not necessarily uniformly accept/reject a contract. If the contract is accepted, then the cellphone joins the crowdsensing system. 
\item A cellphone which has accepted a contract is requested for data. When data is requested, the cellphone forms a cluster with nearby users that have also accepted a contract for crowdsourcing rain data in the vicinity. The cluster collectively decides how many and which users collect data, based on the granularity of data required according to the algorithms described by the authors of~ \cite{Liu2013}. 
\item Once the data is collected, it is sent using one of two methods. If data delivery speed is unimportant, the cellphone communicates with other cellphones that have accepted a contract to form an ad--hoc network to transmit the data whilst conserving energy (compared to using a traditional cellular network). In this case, users' cellphones are also crowdsourced in helping to transport the data, where one cellphone receives data from other cellphones, (aggregates) packages it together and forwards it to the next cellphone in the ad--hoc network. If speed is more important, cellphones communicate using the cellular network. 
\item In either case, other users' devices that are contracted in providing data also receive rain data as a part of the contractual agreement.
\end{enumerate}

The prototype crowdsensing system is implemented in the NetLogo\footnote{\url{http://ccl.northwestern.edu/netlogo/}} simulation environment. In our prototype simulation, the population of use policies assigned to each user is configurable. Moreover, the contractual rules are also configurable. This is true both before and during runtime. Consequently, use policies can be edited dynamically to demonstrate their representation and how changing various second--order norms affects the acceptance and rejection of contracts. This supports policy makers (i.e. the owners of the crowdsensing system) in assessing the acceptability  of contracts and the level of users likely to join the crowdsensing system before run time. However, the main point of the system is as a \textit{proof of concept} rather than to conduct empirical research.

\begin{figure}[t]
\centering
\includegraphics[width=0.40\textwidth, trim=0cm 0cm 0cm 1cm, clip]{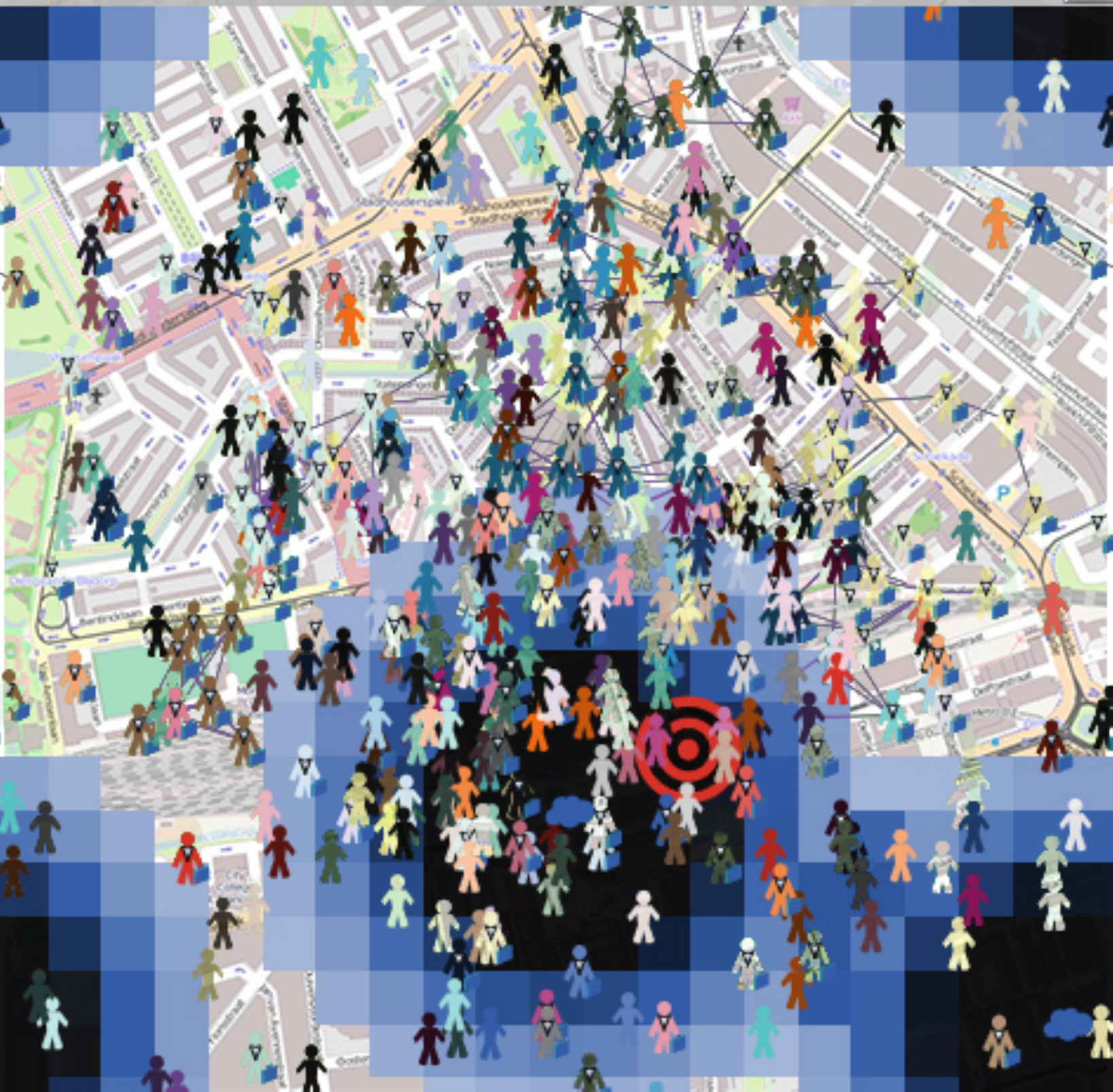}
\caption[Prototype simulated crowdsensing system visualisation]{Visualisation of data aggregation. Lines between users (not to scale) indicate data aggregation, adjacent users with the same colour indicate a cluster where only one in that cluster is providing data. Rainclouds are indicated with blue and black squares. The map is taken from \copyright OpenStreetMap.}
\label{figChapter7Screenshot}
\end{figure}

The visualisation of the prototype is shown in figure~\ref{figChapter7Screenshot}, where each cellphone owner that has accepted a contract and joined the crowdsensing system is depicted as a figure, a map represents the geographical area and rain clouds are represented with blocks in shades of black to light blue. Adjacent contracted sensing devices that have formed a cluster in order to select a single cellphone to collect data for a geographical area, share the same colour. Users that are crowdsourced into forming an ad--hoc network for delivering the data are linked with lines (a central station to which the data is eventually sent is denoted as a circle with alternating black and red inner circles).

\section{Discussion}
\label{secChap7Concl}

In this chapter we have described a proof of concept for the application of the compliance checking system, contributed by chapters~\ref{chapter_3} and \ref{chapter_4}, to the contracted crowdsourcing of cellphone sensors for detecting rain levels. In principle, the same system can be used to automatically govern contracts for crowdsourcing other types of data, such as temperature, crowd levels or photographic weather reports. However, although this chapter illustrates the proof of concept, it does not assess the usability of the PARAGon framework and hence its applicability is \textit{in principle} only. Future work should investigate whether users can adequately describe governance of socio--technical systems and the accuracy of the framework in correctly governing resources to meet user's expectations. Consequently, this chapter only argued the case for multi--level governance automation's applicability, in one case and a limited form, to SHINE--like systems.

To re--iterate, the main advantage of using the multi--level governance compliance checking components of the PARAGon framework is that it supports users in specifying how, when and for whom their devices are used. Users are able to state in which contexts specific obligations and prohibitions should (not) be placed on themselves and any counter--party. Moreover, a use policy, by containing temporal second--order norms, is able to declare temporal requirements for a contract to be accepted. For example, a user can specify that they should be paid in advance of being obliged to provide data, keep sensors turned on, etc. Moreover, the abstraction reasoning presented in chapter~\ref{chapter_3} and \ref{chapter_4} was not implemented for this prototype, but in principle would support a user in defining a use policy using abstract terminology they understand. Hence, the PARAGon framework supports users in specifying use policies and automatically accepting or rejecting contracts governed by such use policies.

The main limitation of this approach is that a single occurrence of non--compliance in a certain social context is aggregated to the contract being non--compliant as a whole. In reality, contracts are neither always accepted nor rejected outright due to a single flaw from the perspective of one party nor are contracts binarily good or bad. Rather, to an individual there are many levels of preference in how good or bad (compliant) a contract is. Consequently, contracts are often negotiated in order to reach a compromise between two parties. The reasoning implemented does not support negotiating in terms of preferences for contracts for the main reason that use policies do not elicit a preference relation between offered contracts. Consequently, the formalism for compliance checking in multi--level governance would need to be extended to reasoning about different levels of ideality in order to form a partial--ordering between contracts. One way would be to adopt the approach in \cite{Torre1999} where contrary--to--duty norms are used to represent different levels of ideality (e.g. you ought not murder, if you murder you ought to do so gently \cite{Forrester1984}). This approach is in a strictly non--temporal setting and so would need extending to reasoning about temporal norms. Another approach is to combine a temporal logic with a preference logic as in \cite{Bienvenu2006}, describing preferences over temporal states of affairs (in our case, the obligations and prohibitions imposed by a contract) and do away with higher--order norms altogether. Adopting one of these approaches when combined with the norm abstraction reasoning presented earlier, would enable users to specify abstract temporal preferences over contracts as a use policy.
\chapter{Conclusions}
\label{chapter_8}

\epigraph[0pt]{Of course, errors are not good for a chess game, but errors are unavoidable and in any case, a game without errors, or as they say `flawless game' is colourless.}{Mikhail Tal}

\epigraph[0pt]{When you see a good move, look for a better one.}{Emanuel Lasker}

\newpage

This dissertation contributed formal accounts of institutions involved in the governing of governance, which collectively comprise the PARAGon framework. The formalisations focussed on two governance types involved in the governing of governance. Namely, institutional design governance and institutional enactment governance. These governance types are found in the social world, where institution designers are governed in how institutions \textit{should} be designed and when institutional changes \textit{can} be enacted. Concerning how institutions should be designed, this dissertation formalised compliance in multi--level governance. Moreover, this dissertation contributed reasoning for diagnosing and rectifying the underlying causes of non--compliant institution designs. Concerning when institutional changes can be enacted, this dissertation formalised secondary legal rules defining the space of possible institutional change enactments. The primary objective, namely formalisation, was complemented with two practical objectives. Firstly, reasoning about realistic institution designs. Secondly, a practical means to reason about governance, which means a computational implementation or an obvious way to build a logical model. Whilst this dissertation contributed prescriptive research, it is nevertheless important to interrogate the intuitions involved. To this end, PARAGon was assessed against a number of real world and imagined institution designs. Together, these contributions aimed to extend the knowledge in the area of institutional reasoning by formalising a new aspect of institutions, in a practical way, and assessing the formalisation intuitions by way of case study.

Central to this dissertation's results was a shift in perspective and a new view on the constitutive rules that comprise institutions. This dissertation's perspective shifted from institutions guiding agent behaviour typically focussed on in the field, to institutions guiding and governing institutional design and enactment. This dissertation's formalisations of institutions seen from this new perspective resulted in a new view of constitutive rules' roles in institutions. Whilst much of the literature viewed the role of constitutive rules as making abstract social actions and other institutional facts possible, this dissertation enlarged the responsibility of constitutive rules. For institutional design governance, this dissertation expanded constitutive rules' responsibility from defining abstract concepts to also indirectly defining the abstract meaning of concrete norms in multi--level governance. For institutional enactment governance, this dissertation expanded constitutive rules' responsibility from making social actions and facts possible to making social actions of institutional change enactment possible. These two new views of constitutive rules are central to this dissertation, as we will discuss further in this chapter.

The rest of this chapter continues by concluding the dissertation with a discussion of results in Section~\ref{secChap8DiscussResults}, we discuss how the results can be applied to the SHINE project under which this research was conducted in Section~\ref{secChap8ApplyingTheResults}, then we discuss future work in Section~\ref{secChap8FutureWork} and finally we reflect on the dissertation as a whole in Section~\ref{secChap8Reflection}.

\section{Discussion of Results}
\label{secChap8DiscussResults}

Whilst it might have appeared that much of governance already had strong logical foundations, in Chapter~\ref{chapter_1} and Chapter~\ref{chapter_2} we saw that governing governance was, until now, mostly formless in the literature. Hence, this dissertation aimed to provide precise definitions for institutional governing of governance in terms of institutional design and enactment governance. This lead to the following overarching research question:

\begin{quote}
How can institutional design and enactment governance be supported with formal reasoning?
\end{quote}

The overarching research question formed the basis for the sub--research questions presented in Chapter~\ref{chapter_1}, which are listed as follows:

\begin{itemize}
\item \textbf{Sub research question 1:} What is a suitable representation to specify institutional design and enactment governance?
\item \textbf{Sub research question 2:} How can we formalise compliance in multi--level governance?
\item \textbf{Sub research question 3:} How can institutional design compliance in multi--level governance be computationally verified?
\item \textbf{Sub research question 4:} How can non--compliant institution designs be explained in order to rectify non--compliance according to the institution designer's objectives?
\item \textbf{Sub research question 5:} How can we formally define when legally valid institutional change enactments occur?
\end{itemize}

In the remainder of this section we discuss what we did to answer these research questions.

\subsection*{What is a suitable representation to specify institutional design and enactment governance?}
\label{secConclResultsRepresentation}

In order to answer this question, we wanted a \textit{natural} way to capture the representation for \textit{temporal} institutions operating in the following two governance types identified in Chapter~\ref{chapter_1}:

\begin{itemize}
\item \textbf{Institutional design governance:} multi--level governance where abstract temporal regulations contained within higher--level institutions govern the regulatory effects of lower--level institutions comprising concrete temporal regulations.
\item \textbf{Institutional enactment governance:} the legality of institutional rule change according to secondary legal rules in a temporal setting.
\end{itemize}

The question was answered for both of these governance types by contributing a high--level representation for institutional design governance in Chapter~\ref{chapter_3} and in a limited form in Chapter~\ref{chapter_5}, and for institutional enactment governance in Chapter~\ref{chapter_6}. The notion of institution used in this dissertation was founded on Searle's counts--as rules \cite{Searle2005} of the form ``A counts--as B in context C''. In our case, counts--as rules were used to represent an institution's temporal dynamics and played a central role for both institutional design and enactment governance. In order to provide a uniform representation for counts--as rules, we based our institution representations on those first proposed by the InstAL framework \cite{Cliffe2006,Cliffe2007}.

A representation specific to institutional design governance was contributed in Chapter~\ref{chapter_3} for multi--level governance, where counts--as rules both define which concrete concepts constitute abstract concepts and modal norms. The requirement for a natural representation specific to multi--level governance was met in three ways. Firstly, by offering the option to define explicit or implicit abstract norms governing concrete normative effects. Secondly, by adopting modal norms, which as we discussed in Chapter~\ref{chapter_3}, provide a natural generalisation to higher--order norms for the explicit governance of concrete normative effects. Thirdly, by \textit{not} requiring any explicitly represented links between concrete and abstract regulations and instead pushing the burden of determining relations between concrete and abstract regulations to a semantics. The temporal requirement was addressed by adopting temporal, immediate and indefinite norms. We showed how institutions defined according to our representation for multi--level governance have an, arguable, natural correspondence to the written law in Chapter~\ref{chapter_3}. In some cases explicit, but controversial, rule formalisation choices were necessary in the case studies examined. Specifically, in order to collapse obligations/prohibitions to non normative social facts, which although we provided argumentation for in Chapter~\ref{chapter_3}, due to having a controversial meaning is a representation \textit{choice}.

A representation specific to institutional enactment governance was contributed in Chapter~\ref{chapter_6}. The central representation construct was counts--as rules used to represent Hart's \cite[p.96]{Hart1961} secondary legal rules. As we discussed in Chapter~\ref{chapter_2}, following the conceptualisation of Biagoli \cite{Biagioli1997}, secondary rules make the social action of changing rules possible in the same way counts--as rules make social actions in general possible. In our representation, we contributed \textit{rule--modifying counts--as rules}, which ascribe rule modifications in the past, present or future conditional on a social context. The requirement for a natural representation was addressed in two ways. Firstly, by defining a richer social context to represent rule--modifying counts--as rules for real legal case studies. For example, to represent a condition, such as found in EU law \cite[Art. 7]{ECHR1953}, on past rule modifications only being possible if they would not, hypothetically, cause an innocent person to become criminalised in the past. Secondly, by not assuming any special representation to define secondary rules which modify other secondary rules, since there is no evidence supporting a special representation being required in the legal case studies we looked at. A number of real--world case studies in Chapter~\ref{chapter_6} were looked at to show an, arguable, natural morphism between the written law's secondary rules and our formalisation of those rules in the representation we propose.

In summary, PARAGon contributed a representation which limits assumptions for the governance types looked at, by providing a representation which conforms to written or mental institution's representation and court interpretations. Where necessary we offered a representation choice rather than making strong representation assumptions. There may be other ways to answer this research question with different representations. However, our requirement was for a natural representation for temporal institutions and hence a suitable representation without requiring any unnecessary constructs was contributed. That is, as far as we are able to tell for the case studies we looked at in Chapter~\ref{chapter_3}, Chapter~\ref{chapter_4}, Chapter~\ref{chapter_5} and Chapter~\ref{chapter_6}, our representation requires no further simplification in order to be natural nor any further complication in order to represent real--world institutional design and enactment governance.

\subsection*{How can we formalise compliance in multi--level governance?}
\label{secConclResultsMultiLevelGovernance}

In order to answer this question, PARAGon contributed a semantics for operationalising multi--level governance compliance in Chapter~\ref{chapter_3}. The semantics defined when a lower--level governance institution imposing relatively concrete regulations has regulatory effects that are non--compliant with higher--level institutions' abstract regulations.

Generally, the semantics assesses collective regulatory effects of lower--level institutions in different social contexts for whether they violate abstract regulations in higher--level institutions. Our semantics adopts the advice of David Makinson \cite{Makinson1999}, ``no logic of norms without attention to a system of which they form part''. That is, norms are understood in terms of the other constitutive rules contained in the same system. Constitutive norms are used to re--interpret the concrete regulatory effects in lower--level institutions for whether they have abstract meanings in higher--level institutions. The abstract meaning of a lower--level institution's concrete normative effects (caused by its concrete norms), according to constitutive rules, determines if those concrete norms violate abstract obligations and prohibitions in higher--level institutions. 

The abstract meaning of a lower--level institution's concrete normative effects is heavily dependent on the social \textit{context} lower--level institutions' obligations and prohibitions are imposed in \textit{from the perspective of} higher--level institutions. To return to a previous example, an obligation imposed by the UK's Data Retention Regulations to store communications metadata is abstractly interpreted by the Charter of Fundamental Rights' as unfair data processing  \textit{in the context that} the person who the data concerns has not consented. In turn, unfair data processing is prohibited by the Charter of Fundamental Rights, hence the UK's Data Retention Regulations institution violates a prohibition on unfair data processing when it imposes an obligation to store communications metadata \textit{in the context that} the person who the data concerns has not consented. Consequently, not only is the abstract meaning of a lower--level institutions normative effects context sensitive, but so is the compliance of a lower--level institution's norms.

Since social contexts are important, we chose to take into account their dynamics. That ism we chose to interpret institutions operating in multi--level governance as state--event transition systems where an institution evolves from one state to another, meaning social contexts also evolves over time. Consequently, our notion of compliance rejects a naïve static comparison between different regulations which would focus on individual rules, in favour of focusing on the combination of regulatory effects and social contexts they are applied in.

We contributed one definitional answer to the aforementioned research question, but there may be many more. Alternative definitions may vary significantly due to representation choices. For example, in our representation we adopted modal norms for the simplicity of representing higher--order norms. An alternative norm form, evaluative norms as discussed in Chapter~\ref{chapter_2}, does not have explicit deontic modalities and hence in this case a notion of compliance in multi--level governance might be defined differently. Moreover, there may be regulations found in multi--level governance that are not captured by our formalism.

To summarise, we answered this research question by defining compliance in multi--level governance as follows. An institutional design is compliant in multi--level governance \textit{if and only if} its concrete regulatory effects in different social contexts have compliant abstract meanings in higher--level institutions according to those higher--level institutions' interpretive counts--as rules. This notion of compliance accounts for the different concrete and abstract ontological views between governance levels and the dynamic \textit{meaning} of lower--level regulatory effects from the perspective of higher--levels.

\subsection*{How can institutional design compliance in multi--level governance be computationally verified?}
\label{secConclResultsCompMultiLevelGovernance}

In order to answer this research question, PARAGon contributed a computational framework for determining compliance in multi--level governance, presented in Chapter~\ref{chapter_4}. The computational framework provided a mapping from the formal representation of institutions in a multi--level governance relationship and their semantics to a corresponding executable ASP program. Institutions were represented in the computational framework as rules in a similar fashion to their formal counterparts. The multi--level governance semantics were represented as general ASP rules to capture the temporal dynamics of institutions and \textit{specific} ASP rules to represent the semantics for the abstraction of norms. Effectively, the ASP rules representing norm abstraction semantics flattened the deontological counts--as function, which defines how concrete normative effects are interpreted more abstractly, presented previously in the formal component of the framework for multi--level governance compliance (Chapter~\ref{chapter_3}). Whist a non--flattened and more general form might have been preferable, we found it was not possible due to current limitations of ASP. Naturally, we contributed theorems and their proofs demonstrating that the computational framework is sound and complete with respect to the formal framework for multi--level governance compliance checking. 

The computational framework was complemented with an implemented compiler. The compiler built on the InstAL framework's \cite{Cliffe2006,Cliffe2007} implementation, which allows institutions in a multi--level governance relationship to be specified in a high--level non--procedural language and compliance to be automatically checked. As we saw, compliance in multi--level governance is highly dependent on social context. Consequently, keeping with the aim of practicality, this means through our implementation we can reveal to institution designers in which social contexts their regulations are non--compliant. If certain social contexts are unreachable in reality and therefore compliance for these contexts is irrelevant, a user can decide based on the presented contextual information whether to ignore these contexts. Since we used ASP, it is possible domain knowledge can be incorporated into the resulting ASP programs in order to constrain compliance checks to only reachable social contexts. Consequently, the practical benefits were realised with an implemented and corresponding computational framework, which provides automated and contextualised compliance checks to users.

We only contributed one answer to this research question. Other corresponding computational frameworks may be possible. For example, using a higher--order logic as a meta--language to provide an axiomatisation of our formal semantics and a higher--order logic theorem prover to act as the computational implementation. Such an approach has been used to embed other logics, such as modal logic \cite{Benzmuller2014}. Speculatively, this approach could yield greater benefits by potentially enabling a computational framework comprising general rules/axioms rather than a flattening of the deontological counts--as function, which requires re--compilation every time institutional rules are changed. An entirely different answer to the research question could be to not provide a computational implementation, but to instead investigate the computational aspects of the formal framework. For example, decidability and complexity. In effect, our implementation shows our formal semantics is decidable, since ASP is decidable for finite Herbrand universes \cite{Gelfond2008} and we proved a correspondence between the ASP implementation and the formal framework. However, an alternative approach could have been to contribute a decision procedure, which would have removed the need for elaborate and recursive transformations between the formal representation and ASP--based implementation. Hence, there may be alternative and perhaps more elegant solutions to this research question.

By contributing a computational framework, we were able to formalise and execute a substantial portion of two related real--world multi--level governance legal cases. Specifically, the first case where the EU Data Retention Directive was enacted in 2006 \cite{EUDRD} to coordinate EU member state legislation in requiring communications metadata was stored and where the UK's implementation, its Data Retention Regulations \cite{UK2009} were enacted and found to be compliantly implement the directive in 2009. Moreover, the second case where in 2014 the Data Retention Directive was found to be non--compliant by the European Court of Justice \cite{ECJDRD} with the European Charter of Fundamental Rights \cite{EU2000}. 

By executing the case study, using interpretations provided by court judgements where possible, we were able to evaluate both our formal and computational multi--level governance compliance checking frameworks. One caveat is that the results for executing a case study can be highly dependent on how institutions are represented. Hence, we opted for a formalisation that arguably has a clear correspondence to the written law and court judgements. Bearing this in mind, we were able to see that we gained the same results as the real--world judgements of the European Court of Justice, using what we argue is a natural corresponding representation. Consequently, we contributed a computational implementation that demonstrates the formal definitions of compliance in multi--level governance are arguably `correct' and provides practical benefits with automated compliance checking.

\subsection*{How can non--compliant institution designs be explained in order to rectify non--compliance according to the institution designer's objectives?}
\label{secConclResultsExplNonCompl}

In order to answer this research question PARAGon contributed a formal definition of explanations for non--compliant institutional designs and a computational mechanism to \textit{induce} those explanations in Chapter~\ref{chapter_5}. That is, by generating explanations, treating those explanations as hypotheses for rectifying non--compliance and then testing those hypotheses for support over a trace of events and therefore multiple social contexts. The contribution means non--compliance can be automatically rectified by adopting the explanatory rectifications for non--compliance as institutional design changes.

The search for explanatory non--compliance rectifications was formalised as an Inductive Logic Programming (ILP) problem. We addressed the inconvenient idea that an institution designer prefers explanatory rectifications for non--compliance that keep an institution design as close to their original intentions, in order to ensure the proposal is of practical benefit. The idea being that our formal notion of explanatory rectifications for non--compliance are those that minimise changes to an institution's constructed social realities. Thereby, explanations for non--compliance offer rectifications that keep as closely as possible to the designer's original intentions, assuming those intentions were correctly captured by the original institutional design. Moreover, we formalised the notion of explanations as also optimizing, to a secondary degree, institutional design change simplicity. This means explanations are preferred if they are more compact, in the sense of optimising fewer rule changes explaining greater instances of non--compliance. Overall we formalised explanations as comprising four types of rule change which can be combined: adding rules, removing rules, adding rule constraints and removing rule constraints. Hence, we formalised explanatory rectifications for non--compliance as changes to rules that adhere to an institution designer's intentions, and are the simplest and most compact explanatory hypotheses.

We addressed explaining non--compliance from a practical perspective. That is, by implementing the search for the hypotheses that provide possible explanations and rectifications for non--compliance, as an ILP problem in ASP using existing embedding techniques. The implementation alleviates an otherwise arduous task of an institution designer in trying different institutional re--designs for the one which rectifies non--compliance in the best way according to our own optimisation criteria (closeness to the original design and simplicity). The general idea of the implementation is to take an ASP program defining a non--compliant institution and compile that program into another ASP program where hypotheses for rectifying non--compliance can be generated and tested for support. For an institution designer, the implementation provides flexibility in specifying for which social contexts rectifying non--compliance should be found and hence guide the process of finding explanatory rectifications.

In order to see how our formalisation of explanatory non--compliance rectifications behaved, we formalised a case study based on a fictional institution for a socio--technical system which in turn was governed by real--world and fictional laws. The system discovered a number of simple explanations for why the socio--technical system institution was non--compliant. These included the discovery of general design faults, which might potentially explain further issues the designer does not desire, removal of problematic rules and changes to existing rule constraints, and explanations comprising rule additions to ensure regulatory effects before rectification would still exist after rectification if they do not cause non--compliance.

We only contributed one possible answer to the aforementioned question based on the idea of inducing hypotheses for non--compliance. There may be other possible answers, such as by defining what constitutes an explanation differently. For example, rather than defining explanation as hypotheses comprising institutional rule changes, one possibility is to combine argumentation and abduction to provide explanations as causal stories (e.g. as proposed for legal cases \cite{Bex2010}). In this approach, an explanation is a chain of inferences showing how abduced probable causes of a true proposition provide an explanation of that proposition. Hence, a narrative causal chain could instead be used in this way to provide explanations for non--compliance. We do not attempt to exhaust all the feasible ways to define explanations for non--compliance, but note formal argumentation could provide alternative answers.

Together, the formal notion of explanations for non--compliance and the implementation for finding such explanations complement each other. The formal aspect can be tested by the implementation with various case studies to determine if the formal notions give intuitive results. The practical aspect is supported by addressing the need to find those solutions for non--compliance which minimise design change and offer the simplest explanations for understandability. We tested both of these aspects using a case study combining fictional and real--world rules.

\subsection*{How can we formally define when legally valid institutional change enactments occur?}
\label{secConclResultsFormGovInstChangeEnact}

In order to answer this research question PARAGon contributed a semantics that defined when institutional change enactments are valid in Chapter~\ref{chapter_6}. In this context, by institutional change enactment we mean changes to an institution's rules over time. An institutional rule change is valid if it is ascribed by an institution's secondary rules. These secondary rules were represented as a special type of counts--as rule of the form ``A counts--as modifying a rule in the past/present/future in context C'', which ascribe modifications to counts--as rules including other secondary rules. The semantics defined when social actions of institutional counts--as rule changes are ascribed, including to other secondary rules, and how the institution transitions from one version to another when rule changes are validly enacted.

The formalisation gave meaning to secondary (counts--as) rules. The secondary counts--as rules we formalised include a condition on a richer social context not given any formalisation, as far as we are aware, previously. In particular, social contexts acting as conditions on past institutional states, past institution versions and \textit{hypothetical} rule change effects. These conditions were used to represent cases such as institutional rule changes occurring conditional on those changes not criminalising formerly innocent people in the past. In order to deal with such expressivity we proposed a semantics that defines a rule change as occurring if the necessary conditions are met and all of the effects of the rule change are consistent with those necessary conditions.

We acknowledged that in reality secondary rules can make it possible for past, present and/or future rule changes to be made including to other secondary rules. Consequently, our semantics dealt with difficult cases where changes to secondary  rules potentially cause other, previously established, rule changes at various points in time to be unravelled or potentially further rule changes to be made. Moreover, our semantics handled cases where complex interactions between rule changes would, in a classical setting, cause inconsistency. For example, the case where the only rule ascribing rule modifications ascribes its own removal. In our case, we took the perspective that if a rule change causes inconsistency with the conditions for its own ascription then the rule change is not legally ascribed (i.e. is invalid) and thus does not occur in the first place.

We tested our definition against a number of case studies based on EU and UK law, as well as edge cases where we would expect there to be issues with an incorrect definition. The real world case studies demonstrated that through a simple representation we were able to reason about difficult kinds of rule change governance. For example, hard constraints on there being no changes to criminalising laws in the past, requiring a form of hypothetical reasoning to classify rule change effects. From the edge cases we saw that complex cases where rule changes interact, which are conceivably difficult to reason about without a mechanical process, are arguably handled intuitively by our formal account.

We did not attempt to exhaustively answer the research question. Instead, we offered a single formalisation of when institutional rule changes occur according to secondary rules. Other formalisations may be possible and, indeed, they exist for rule modifications in non--temporal settings \cite{Boella2004} and rule modifications in temporal settings \cite{Governatori2005b,Governatori2010} but with less expressive secondary rules. Moreover, an equivalent definitional answer could be given to this research question using a different formalisation. It's possible our semantics could be defined on top of an existing logic, either at the meta--level or object level. For example, we could try to adopt or modify default logic \cite{Reiter1980} by viewing conditions on rule change being ascribed as defaults that must hold before and after rule change. In our case, we chose to contribute a formal definition with a semantics that stands on its own, requiring no understanding of pre--existing logics aside from some rudimentary notions of sets and functions.

To summarise, we contributed a formal definition to answer this research question. Informally, a rule change occurs \textit{if and only if} it is ascribed by secondary legal counts--as rules conditional on a social context, which can include hypothetical rule change effects, that holds before and after the rule change takes place. Our account was assessed against a number of case studies, including edge cases to assess the intuitiveness of results.

\section{Applying the Research to the SHINE Project}
\label{secChap8ApplyingTheResults}

The research conducted in this dissertation was supported by the SHINE project of TU Delft. The project aimed to develop techniques for acquiring and coordinating large numbers of heterogeneous data resources that already exist in the environment (e.g. cellphone sensors, radars and people) in order to collect environmental data. This dissertation contributes techniques for formalising governing governance, which is arguably suitable for governance and coordination of `SHINE--like' sensor systems.

An implemented application of the research was presented in Chapter~\ref{chapter_7} for a simulated prototype as a proof--of--concept. In this application automated compliance checking in multi--level governance was used to determine if contracts offered to individuals to crowdsource their cellphone sensors should be accepted on the basis of being compliant with use policies. These use policies would be, conceivably, specified by cellphone owners stating how, when and for whom their cellphone sensors may be used. In this setting, the research was applied to giving people control over how their sensors are contributed to crowdsourcing systems without requiring sensor owners to arduously read the fine--print of the offered contracts. 

The research in Chapter~\ref{chapter_5} was also applied to a fictional institution used to govern a crowdsourcing system as a whole, which in turn was governed by external laws partially inspired by real--world legislation. In this application we demonstrated an automated system to find explanations for why the crowdsourcing system institution design was non--compliant with external laws. 

The research on compliance in multi--level governance in Chapter~\ref{chapter_3} and Chapter~\ref{chapter_4} was applied to a real--world multi--level governance case study. In this case we showed how the PARAGon framework was able to represent and reason about the EU Data Retention Directive \cite{EUDRD} and its non--compliance \cite{ECJDRD} with the EU Charter of Fundamental Rights \cite{EU2000} due to requiring the storage of people's communications metadata without their consent, which took away the right to respect for private and family life. In this case, the relevance was showing how we can check the compliance of laws concerning the collection of data, in much the same way a SHINE institution, which aims to guide MAS participants in contributing data, might be found non--compliant with similar external human rights and liberties laws. Hence, we applied the framework to a simulated prototype and case studies concerning SHINE--systems and data collection.

Outside of these realised applications, the multi--level governance compliance reasoning presented in Chapter~\ref{chapter_3} and Chapter~\ref{chapter_4} offers a \textit{potential} way to contribute automated reasoning for an architecture applied to governing heterogeneous sociotechnical systems. SHINE is concerned with heterogeneous sensor systems, which comprise different types of sensors. Rather than attempting to apply a single institution to govern heterogeneous sensors, which due to different behaviours might be unsuitable, multi--level governance offers a way to treat heterogeneous groups of sensors (e.g. crowdsourced cellphone sensors, a group of weather radars, etc.) as separate homogeneous self--governing sensor sub--systems. The idea is that each sensor sub--system has its own institutional governance designed by the stakeholders of that sub--system (e.g. sensor owners). Then, a SHINE institution is designed to govern institutions governing each sensor sub--system towards over--arching aims with abstract regulations. Thereby, multi--level governance delegates the design of sensor sub--system institution designs, removes problems of a single `one--size--fits--all' institution design and coordinates separate sub--system institution designs towards overarching data collection objectives. In this scenario, this dissertation supports governance of sensor sub--system institution designs with automated multi--level governance compliance checking, but further investigation is needed to determine if automating such governance can work for heterogeneous sensor `systems of sub--systems' in practice.

The institutional enactment governance presented in Chapter~\ref{chapter_6} provides a \textit{potential} way for stakeholders in SHINE sensor systems to specify flexible and automated governance of sensor systems' institutional change enactment. This means, system stakeholders are able to use the formal reasoning to determine how institutional changes can be enacted according to an institution's secondary rules when changing sensor aims require changed regulations. Moreover, stakeholders have flexibility in defining the rule change process for a given governed sensor system. The main benefit being, owing to heterogeneity between different sensor systems, alternative rule change processes for each sensor sub--system can be defined and reasoned about. For example, one system's rule change process can be defined as a direct democratic vote with certain constraints (e.g. not removing the right to a democratic vote from participants in the past) and another as requiring elected technocrats to make a collective decision. Hence, although untested, our secondary rule formalisation potentially allows automated operationalisation for a diverse range of governance and regulatory change enactment styles to be defined for different sensor systems, in a way that enables those system's regulations to adapt, as deemed appropriate, to new aims and needs.

None of this is to say that this dissertation contributes an exhaustive application to the governance of SHINE--like sensor systems. However, what has been shown is that in at least one area, the governance of contracts for crowdsourcing, the framework was applied in a prototype. Moreover, the framework has been applied to case studies relevant to sensor systems such as those envisaged by the SHINE project. Finally, there are potential avenues for future applications to SHINE--like systems.

\section{Future Work}
\label{secChap8FutureWork}

Future work is split into analysis of the formal frameworks contributed by this dissertation, improvements to this dissertation's contributions and new research lines influenced by this dissertation's contributions.

\subsection{Analysis}

The formalisms developed in this dissertation have been shown to meet various properties, such as there existing a sound and complete logic-programming implementation. Such properties were provided in order to demonstrate the expectations, set out in the introductions of the relevant chapters, have been met. However, there still remains work to be done on general properties that give a better understanding of how institutions behave under the defined semantics. 

The formalism defining compliance in multi-level governance, in Chapter~\ref{chapter_3} could be analysed for properties relating to inconsistency, (non-)compliance and redundancy. Inconsistency can occur in two senses. The first type of inconsistency that should be investigated is, what we can call, non-deontic inconsistency where due to the operations defining the semantics being undefined in some cases (e.g. the event generation operation) there is consequently no model for those cases. The second type of inconsistency is what we call deontic inconsistency, where it is impossible to discharge all obligations/prohibitions, or worse that some obligations/prohibitions are necessarily violated no matter what event occurs. Properties that coincide with the dischargement and violation of abstract and higher-order norms should also be investigated. For example, if an abstract norm is violated then what concrete norms, if any, must also necessarily be violated and vice versa? Finally, an important aspect of an institution or normative system addressed in the literature is \textit{redundancy} \cite{Boella2008a,Dastani2015} -- when would removing a counts-as rule have no effect whatsoever on how the institution evolves over time (for all traces of observable events)? Whilst norm redundancy has already been formalised and investigated, it would be interesting to investigate given the novel semantics of the multi-level governance compliance formalisation if there are general properties that coincide with a rule being redundant. Investigating properties relating to inconsistency, (non-)compliance and redundancy for the formalisation of compliance in multi-level governance would give a better understanding of how the formalisation behaves, in particular given the novel abstraction semantics, and some useful properties for understanding how to improve an institution operating under those semantics.

The formalism defining the legality of rule change, in Chapter~\ref{chapter_6}, could be analysed concerning (il)legal rule change, temporal dynamics and rationality. To recap the informal notion, a rule change is legal ``\begin{inparaenum}\textit{\item if the rule change is ascribed by counts--as rules, conditional on a context taking into account the potential changes to the context the rule modification would make \item taking into account past/present/future rule modifications effect on counts--as rules and thus the context in which rule changes are conditional on.}\end{inparaenum}'' However, such informality still leaves future investigation of formal properties that coincide with (il)legal rule change. Moreover, in terms of temporal dynamics, it is important to analyse under which conditions changes occurring in the future can affect the past, including past states and also the legality of past rule changes. In particular, concerning both temporal dimensions of the formalism. On the one hand we showed future rule changes can make past changes in the \textit{same version} become (il)legal and therefore potentially no longer occur. On the other hand we did not ascertain under which fragments of the language future rule changes are guaranteed to not affect the legality of rule changes in \textit{past versions}. Further properties should be analysed concerning rationality. Famously, the AGM belief revision postulates \cite{Alchourron1985} define a notion of rationality for belief revision operators and a related set of postulates has been proposed for norm revision \cite{Boella2009a}. It is important to analyse whether our formalisation of ascriptive legal rule change meets these postulates. However, since we are dealing with rules that ascribe the legality of changing other legal rules and not just `straightforward' norm change, it is possible that the postulates are not met and moreover that a slightly different set of postulates may be more appropriate that do hold for our formalisation. Investigating properties relating to the legality of rule change, temporal dynamics and rationality would give a better understanding of what our formal definitions really mean at a deeper level.

In addition to analysing the behaviour of each individual formalisation, it is also important to formally compare with other formalisations. One possible way to do this would be to abstract the semantics. For example, by using Tosatto et al's \cite{Tosatto2012a,Tosatto2012b} graph-based formalism where obligations, permissions and institutional facts are nodes and rules are edges between those nodes. The idea would be to take existing abstractions of formalisms using Tosatto et al's abstraction and compare them to an abstraction of the formal semantics contributed by this dissertation. The benefit would be not just a formal comparison between the PARAGon framework and other frameworks, but also a comparison where patterns of reasoning and differences between them are visualised. However, this type of analysis requires further investigation since there, to the best of our knowledge, appears to be no work on abstracting systems of the `type' we formalise. In particular in Chapter~\ref{chapter_6} where an institution evolves over two timelines and there can be complex interactions between past, present and future institutional facts. In order to analyse differences between our formalism and others, a means to abstract our semantics should be found, which is a significant undertaking and left for future work.

\subsection{Improvements}
\label{secChap8FutureWorkImpro}

\subsubsection{Putting Automated Multi--level Governance into Practice}

In Chapter~\ref{chapter_3} and Chapter~\ref{chapter_4} we looked at reasoning about compliance in multi--level governance. A focus was placed on reasoning about abstraction at different levels of governance. This aspect of the PARAGon framework was assessed against a real--world case--study for the governance of human societies and governments.

An avenue for future work is to determine if the same framework can be applied to artificial societies. This would result in an formalised and automated operationalisation of two proposals. First, the proposal of what Pitt et al. \cite{Pitt2015} call polycentric governance for socio--technical systems. Here, they argue that in complex artificial MAS a single one--size--fits--all institution is inadequate, since different localised parts of the MAS may need different regulations. Therefore, in our own wording, separate lower level institutions should be designed appealing to subsidiarity. These institutions should be compliant with overarching institutions abstractly prescribing what regulations should be implemented. Secondly, Dignum et al. propose the OMNI design methodology framework for institutions/organizations governing artificial MASs \cite{Dignum2005}. Here, the design methodology is focussed on a regulation abstraction hierarchy where at the most abstract level statutes comprising values, objectives and contexts should be designed. Then, abstract norms implementing these statutes. Implementing the abstract norms are concrete norms. In this methodology an assumed function that concretises abstract norms is used. PARAgon offers a semantics for reasoning about abstract norms that could potentially be used to weaken or remove the assumption of a concretising function. Future work should extend and apply the PARAGon framework to further operationalise and develop the ideas presented in these papers.

\subsubsection{Enhancing Institution Abstraction Reasoning}

Another important future development is extending the semantics for abstract regulation reasoning presented theoretically in Chapter~\ref{chapter_3} to support ontological alignment of institutions. In our proposal lower--level institutions' regulatory effects were re--interpreted at the abstract level defined by higher--level institutions. For example, obliging storing communications metadata in a lower level institution is abstracted to obliging storing personal data in a higher--level institution. In one sense, this means lower level institutions' regulations were \textit{aligned} with the abstractions defined by higher--level institutions.

However, we assumed that the concrete terminology shared by lower and higher--level institutions have the same meaning and are already ontologically aligned. For example, storing metadata in the lower--level institution coincides with storing metadata in the higher--level institution. Thus, if the lower--level obliges storing metadata, and the higher--level views storing metadata as storing personal data, then from the higher--level institution's perspective the lower--level is obliging that personal data is stored. In general, ontologically aligning institutions is an important problem that needs addressing for comparing regulations between institutions.

Aside from alignment, abstraction reasoning should also be improved to address further deontic concepts. We focussed on how to abstract obligations and prohibitions, based on counts--as rules describing the abstraction of events and fluents. However, there are many other deontic concepts, as described by Hohfeld \cite{Hohfeld1913}, that play an important role in institutions. They include accounting for the concepts of right and liberty. For example, a right to life correlates with a prohibition on others to take life away. The liberty to choose who you love depends on there being no prohibition on choosing who you love. Each of these might be involved in multi--level governance in the form of obliged/prohibited rights and liberties, or rights and liberties represented in constitutions which require obligations/prohibitions to (not) be imposed in lower--level institutions. Such Hohfeldian notions of rights and liberties should be investigated for abstracting and governing.

Likewise, concepts of legal power should also be accounted for. Legal power, is the ability to (legally) affect an institution according to ascriptions provided by constitutive rules \cite{Jones1996,Makinson1986}. To give an example in a multi--level governance context, a higher--level institution governing a marriage institution could state that ``it is obliged that newly weds have the power to annul their marriage''. From the perspective of this dissertation, an interesting aspect is determining from a relatively concrete lower--level institution (such as a marriage institution) whether concrete rules and regulations constitute an abstract power such as newly weds having the legal power to annul their marriage. Moreover, concepts related to power should also be accounted for. Namely, disability (the opposite of power), liability (you are only legally liable if you have legal power) and immunity (without liability you are immune from not fulfilling norms you have no legal power to do so). 

One possibility to incorporate all of these concepts is to build on the logics of Hohfeldian rights and liberties \cite{Kanger1966,Kanger1971,Kanger1972,Kanger1985}, and institutional power \cite{Jones1996}. That is, we could try to combine these formalisations of Hohfeldian legal concepts with multi--level governance formalisation in order to govern instances of Hohfeldian concepts.

\subsubsection{Putting the Ascription of Legal Rule Changes into Practice}

The formalisation in Chapter~\ref{chapter_6} provides a fixed point characterisation for ascribing legal and valid rule change. The fixed point characterisation provides the mechanism to find models, which can be implemented in any sufficiently expressive language. Yet, there remain practical concerns for the formalism to be usable in a real--world setting.

An obvious improvement is to address the lack of implementation and complexity analysis. Many of the constructs in the formalism are similar to Event Calculus constructs \cite{Kowalski1986} (e.g. a fluent holds if it is initiated until it is terminated). An axiomatisation for an Event Calculus (version) exists in ASP \cite{Kim2009}. Hence, one way to implement the formalism would be to provide an Event Calculus--like axiomatisation in ASP. A benefit of this approach is that there are already existing complexity results for ASP \cite{Eiter2007,Faber2007} which can be used to assess the complexity of an axiomatisation.

Another problem that needs addressing is the assumption that, in the formal framework, all rules which can and/or will be added/removed from an institution are fully characterised by an institution's event generation and state consequence functions. It makes more sense in practical terms to allow the functions describing rules or a database of rules to be updateable with new rules. Then, with new rules added to the database, rule enactment actions can be undertaken by agents to ensure the rules become a valid and recognised part of the institution. If an ASP approach is taken to implementing the formalism, then a logic program would need to be updateable with new rules during run--time. One possibility for updating a rule database is to build on recent advances in answer--set programming for dynamic domains \cite{Gebser2015}.

\subsubsection{Legal Interpretation}

In this dissertation we looked at the legality of institution designs and rule change but we did not investigate legal interpretation, which is closely related to the topics we did investigate as we will discuss. Legal interpretation is the act of courts deciding what various concepts mean, such as to determine the applicability of a rule or whether actions of an agent constitute legally obliged behaviour, even if the written law does not make explicit recommendations for such cases. Legal interpretation is important, since it allows legal norms to be adapted to unforeseen circumstances and changes to the understanding of various concepts by society not represented in the written law. For example, a categorical prohibition on storing private data might have previously meant preventing the storage of communications content between people, but as time went on it became clear that storing communications meta--data is also an infringement on privacy -- hence a court would then update the notion of private data to include meta-data. 

As Boella et al. show \cite{Boella2010a,Boella2010b} legal interpretation is, formally, a case of rule change. Specifically, whilst legislators have the legal power to change ontological and regulative rules at ``design--time'', courts have the legal power to change only ontological rules at ``run--time'' (i.e. as cases are brought before the court). Boella et al. provide a formal mechanism for re--interpreting the ontological meaning of literals in the antecedent of regulations (the applicability of the rule) and the consequent (the obliged actions or state of affairs) based on interpreting the original \textit{goal} of the legislator (i.e. the intention behind the law). Hence, legal interpretation is an important aspect of rule change that ensures laws, at run-time, can be correctly applied and understood according to the goals behind enacting those laws at design--time.

Relating to these aspects, in this dissertation we contributed formalisms that can be used to reason about the \textit{effects} of legal interpretation. In Chapter~\ref{chapter_3} and Chapter~\ref{chapter_4} we looked at interpreting the abstract meaning of detached deontological statements based on ontological interpretations provided by courts. But, the ontological interpretations were formalised \textit{a--priori} as counts--as rules (e.g. meta--data count-as personal data) in the case study on EU multi-level governance. 

We did, on the other hand, investigate the generation of hypotheses, using ILP, for explaining and rectifying non--compliant institution designs in Chapter~\ref{chapter_5}. Effectively, ILP synthesised changes to ontological and deontological rules in order to ensure the regulatory effects of an institution do not violate the regulations of other institutions. Courts synthesise interpretations, which are effectively changes to the existing set of ontological rules. But, suitable legal interpretation ensures the law is applied and interpreted correctly with respect to its intention, which is different from our purpose of resolving non-compliant institution designs.

Later, we investigated rule change legality and its effects on how an institution evolves in the past/present/future in Chapter~\ref{chapter_6}. Since legal interpretation is a case of rule change, the formalisation of when rule changes are legal could be used, with some enhancements, to reason about when legal interpretations are legal. For example, if we want to reason about a statement such as ``a court can introduce a new counts-as rule so long as it is not regulatory and by introducing that rule the goal of the law being interpreted is promoted''. By combining the work on interpreting the abstract meaning of regulations in Chapter~\ref{chapter_3} and Chapter~\ref{chapter_4} and the work on legal rule change in Chapter~\ref{chapter_6} we can potentially capture the effects of legally-valid legal interpretation, which is the addition of ontological rules ascribed by rule change regulating counts-as rules, on the abstract meanings of regulatory effects. Our work is related to reasoning about legal interpretation in terms of its effects and its legality when viewed as a type of rule change. 

However, this dissertation did not investigate how legal interpretations could be generated in order to support the original goal of a law when it is unsupported according to the run--time interpretation of legal concepts. Potentially, existing approaches by Boella et al. \cite{Boella2010a,Boella2010b}, which use defeasible logic to interpret the law, could be adopted to provide a more complete system for reasoning about institutions, their interpretation and governance.

\subsection{New Research Lines}
\label{secChap8FutureWorkNewResLines}

\subsubsection{Social Commitments}

Whilst this dissertation looks at prescriptions from the perspective of when institutional rule enactment and institution designs are legal, a similar type of prescription is social commitments. Social commitments, such as promises, emerge between agents. Social commitments establish relationships between parties and counter--parties and expectations between agents. In a social commitment, when one agent, called the debtor, commits to bring about X for the benefit of another agent, called the creditor, then the creditor has the \textit{expectation} that X will be brought about for their benefit and the debtor knows that the creditor has such an expectation \cite{Castelfranchi1995,Castelfranchi1998}. The creditor, has flexibility and autonomy to bring about X in any way they choose including through delegation to another agent. To give an example from \cite{Yolum2004}, a social protocol states whereupon a potential buyer requests a quote for electronic goods, the potential seller is committed to providing the quote and the buyer \textit{expects} to receive a quote.

Social commitments should be formalised since they provide a powerful and ubiquitous means for agents to cooperate in a flexible way on the basis of mutual understanding, which supports social interaction planning whilst preserving participants' autonomy. Furthermore, social commitments allow social endeavours' progress to be assessed with in--built key performance indicators (KPIs), such as `fulfilled' and `violated'. Hence, social commitments have seen formal work in the normative reasoning sphere for creating new social protocols from existing commitment--based protocols \cite{Baldoni2015a,Gerard2015}, planning the fulfilment of commitments on the basis of expectations and mutually held beliefs \cite{Yolum2004}, addressing conflicting commitments \cite{Ajmeri2016}, and converging on a mutual understanding of commitments and expectations \cite{Chopra2015a}, to name a few.

In this dissertation we looked at how to synthesise institution changes to rectify non--compliance and how to determine when rule changes are valid, such as to support planning valid legal rule changes. As we saw in Chapter~\ref{chapter_6}, creating and altering rules makes it (im)possible to create and alter rules. This is just as true for social commitments. To give one example, in order to plan a larger project a project manager needs to get various project participants to commit to giving estimated work hours required for deliverables. When participants fulfil commitments, a project manager is able to plan further commitments. Or, to give another example, by creating a social commitment for the trading of goods and services it is possible for each participant to assess the trust they should place in the counter--party after the commitment is fulfilled or violated. Consequently, creating and fulfilling/violating social commitments makes it possible to determine whether further commitments should be created on the basis of new information or increased trust. However, whilst commitment creation and fulfilment is entwined with further commitment creation and planning, planning the synthesis of commitments has not been addressed taking into account this interdependence as far as we can see.

This dissertation offers ideas which can be built on to create such plans. Chapter~\ref{chapter_6}'s formalism shows that rule creation (e.g. commitments) is non--trivial given pre--existing rules which ascribe rule changes, where rule changes enable/disable further rule changes. An ASP axiomatisation of Chapter~\ref{chapter_6}'s formalism as outlined previously would potentially support planning rule changes. ILP applied to rule synthesis for compliant institution designs contributed by Chapter~\ref{chapter_5} can potentially be used as a basis for synthesising the right commitments for a given problem. But, the difficulty is determining how to synthesise commitments given that setting up expectations and/or agents fulfilling those expectations is necessary to build trust and meet sub--goals for further commitments to be planned for meeting an overarching social goal.

\subsubsection{Formalising the Subsidiarity Principle in Multi--level Governance} 

This dissertation looked at formalising compliance in multi--level governance. Subsidiarity, the principle that what can be legislated at the local level should be left up to the local level, is also an aspect of multi--level governance. This dissertation contributes reasoning for abstract regulations which are used to realise subsidiarity, but this dissertation does not investigate subsidiarity itself.

Yet, subsidiarity is an important principle. Ostrom \cite[p.90]{Ostrom1990} argues that regulations must be based on local conditions. Whilst regulations should be coordinated across groups/jurisdictions (e.g. carbon emission targets), Wilson argues \cite{Wilson2013} implementation should be left to the lowest--level jurisdiction where implementation is effective. By involving local lay knowledge, as opposed to a one--size--fits--all approach, policies are more effective by being enhanced with learning from past policy decisions \cite{Newig2008,Pellizzoni2003,Yearley2003}. In general, a one--size--fits--all approach is argued as resulting in failure \cite{Ostrom2009}, which the subsidiarity principle aims to address. 

The same arguments in favour of subsidiarity for human society governance are made for the governance and design of sociotechnical systems, such as smart grids. Frey et al. \cite{Frey2013} and Pitt et al. \cite{Pitt2014}, propose a design/governance framework for socio--technical systems, where higher--levels provide over--arching aims but, adhering to the principle of subsidiarity, delegate to lower--levels for implementation or further delegation to even lower--levels. For example, Pitt et al. \cite{Pitt2014} talk about decentralised community energy systems. These are small residential communities comprising residences which both produce energy (e.g. with wind turbines) and consume energy, and supply energy to one another. Institutions regulate the communities to manage the appropriation of common energy resources. Scaling involves composing a system of decentralised community energy sub--systems, each sub--system demanding and supplying energy from one another. In order to govern a system of systems, hierarchical governance is proposed, where sub--system's institutions are coordinated (e.g. to ensure energy is shared between sub--systems) but appealing to the principle of subsidiarity different sub--system institutions comprise different regulations as deemed appropriate for the community being governed (e.g. one operates a market economy and another operates a communal economy). Subsidiarity uses local lay knowledge to govern disparate societies or socio--technical systems towards a common aim.

In Chapter~\ref{chapter_3} and Chapter~\ref{chapter_4} we looked at the abstraction aspect of subsidiarity, where abstract regulations leave some institutional design decisions up to the local level. However, subsidiarity itself has not been formalised in terms of determining \textit{what} should be left up to the local level. Nor does subsidiarity appear to have been given a precise measure. The question remains, subsidiarity is the principle that what can be done at the local level should be left up to the local level through delegation, but how far down should the delegation go and how much freedom should be given to lower--levels of institution designers? In particular, what level of subsidiarity is suitable taking into account a cross--institutional objective, and lower--level institution designer's capabilities and local lay knowledge?

Arguably, the lack of formalisation is problematic since subsidiarity is a structural optimisation principle and therefore if not measured multi--level governance is not optimisable. We can conceive that if subsidiarity is not realised through over--delegation and overly--abstract regulations at higher--levels, then designers of lower--level institutions have too much room to design compliant institutions. Therefore, coordination between individual sub--systems/societies to meet a global aim is not guaranteed. Moreover, if subsidiarity is not realised due to too little delegation and too concrete regulations in higher--level institutions then lower--level institution designers have too little room to optimise regulations for local lay knowledge to meet global aims. Consequently, subsidiarity needs to be formalised, according to these assumed problems, in order to promote measurement, optimisation, compliance and therefore meeting global goals, and automation.

There are many aspects to address when formalising subsidiarity. A metric should be defined for the \textit{abstractness} level a particular institution possesses at a certain governance level. In Chapter~\ref{chapter_3} and Chapter~\ref{chapter_4} we saw how abstract regulations can be complied with but did not define how abstract each level is in a qualitative or quantitative sense. Abstraction in this context is viewed as providing many ways with which to comply, the shortcoming of this measure is that compliance and therefore `abstractness' is highly dependent on the ontology used. Consequently, a novel measure of abstractness is required that does not depend on the number of ways abstract concepts are ascribed.

A metric should be defined to measure how disparate lower--level systems are and whether they should be viewed as separate systems in the first place, or whether there is a `one--size--fits--all' institution for a particular goal and multiple jurisdictions. That is, we need a system disparity metric to calculate the level of abstractness each regulation and each governance level should be issued at. There already exist metrics for differences (inconsistencies) between propositions \cite{Lafage2001}, knowledge bases \cite{Grant2016} and separate agents' beliefs \cite{Hunter2014}. But it is unclear whether techniques used for such metrics can be extended to assessing system disparity. 

Finally, it should be possible to determine what capability local levels have to regulate more appropriately than higher--levels using local lay knowledge for a particular objective. One possibility is to define reasoning about whether there is appropriate local--level knowledge such that lower--levels are more capable than higher--levels in regulating towards global aims. Such reasoning should be from the perspective of higher--levels that do not necessarily know exactly \textit{what} the local knowledge is. However, how to extract information for whether local--level institution designers have appropriate knowledge is unclear. Abstractness, system disparity and local capabilities to self--govern appropriately should be measurable and combined to formalise a subsidiarity metric for a given objective, yet as we have speculated, each measurement is non--trivial to define.

\section{Closing Remarks}
\label{secChap8Reflection}

We finish with a few closing remarks regarding the aspects of governing governance this dissertation focuses on. This dissertation contributes formal and practical reasoning for governing institutional design and change with the PARAGon framework. The contribution is split into formalising institution design and enactment governance. Chapter~\ref{chapter_3}, Chapter~\ref{chapter_4} and Chapter~\ref{chapter_5} contributed reasoning for how institutions \textit{should} be designed. Chapter~\ref{chapter_6} contributed reasoning for how institutional change \textit{can} be enacted. This raises the question, does institutional design and enactment governance involve deontic shoulds or alethic cans?

From the \textit{should} perspective, higher--level institutional rules prescribe the design of lower--level institutions in multi--level governance and are formalised as soft constraints in Chapters~\ref{chapter_3}, Chapter~\ref{chapter_4} and Chapter~\ref{chapter_5}. In our case, higher--level institutions comprise primary rules that abstractly prescribe lower--level institutional designs. Such primary rules preserve the autonomy of lower--level institution designers since they can be violated by designing a non--compliant institution.

From the \textit{can} perspective, institutional change enactment is governed by Hart's secondary legal rules \cite[p.81]{Hart1961}. These secondary rules define how institutional rules are created and changed, formalised as hard \textit{institutional} constraints, making (im)possible rule changes in an institution, in Chapter~\ref{chapter_6}. Specifically, secondary rules were formalised as Searle's constitutive rules. As discussed in the background section, constitutive rules \textit{constitute} an institution. According to our conceptualisation an institution's secondary rules \textit{constitute} the ways in which the institutional rule changes can be enacted. The formalism in Chapter~\ref{chapter_6} ensures only valid rule changes recognised by the institution can be made representing the \textit{actual} state of an institution's rules according to the its own secondary rules.

Parallels can be drawn between institutions governing social behaviour, such as formalised in the literature, and the notions of institutions that govern governance defined by this dissertation's formalisations. In the institution of chess, according to Ricciardi \cite{Ricciardi1997}, the rules constitute the game. Moves can be physically made outside of chess' rules, but then the game of chess is not being played. In other words, you cannot cheat at chess. An institution comprising prescriptions (ought statements), such as the law, may be considered as `cheatable' if cheating is defined as the act of violating those prescriptions. An instance of an institution formalised in this dissertation, focussing on institutions involved in governing governance, comprises rules that \textit{constitute} that institution. Institution designers governed by instances of our institutions can `cheat' with respect to the institutions governing them in one sense and not in another. In our formalisation, you can cheat at institution design in the sense of designing a non--compliant institution (Chapter~\ref{chapter_3}, Chapter~\ref{chapter_4} and \ref{chapter_5}), but you cannot cheat at institution design in the sense of enacting rule changes invalidly (Chapter~\ref{chapter_6}). This dissertation formalises both of these kinds of governance. This dissertation argues that it has formalised institutional governing of governance, where regulations regulate how institutions \textit{should} be designed and how institutional change enactments \textit{can} be made.


\titleformat{\chapter}[display]
{\flushright\titlestyle}
{\color{black}\fontsize{96pt}{96pt}\selectfont\bfseries\thechapter}
{0pt}
{\Huge\color{black}}
\titlespacing{\chapter}{0pt}{0pt}{2\baselineskip}
\dottedcontents{appendix}[1.5em]{\vspace{0.5\baselineskip}\titlefont\bfseries}{1.5em}{0pc}

\newpage
\chapter{Appendices}

\begin{sloppypar}
Appendices are available online at \url{http://www.thomascking.com/dissertation/ThomasKingPhDAppendices.pdf} (accessed at time of writing on 14-09-2016).
\end{sloppypar}

\thumbfalse
\chapter*{Acknowledgements}
\setheader{Acknowledgements}
\addcontentsline{toc}{chapter}{Acknowledgements}
\label{acknowledgements}

This dissertation came about through periods of frenzy and pauses for thought, apathy and persistence, ignorance and exploration, disorientation and discovery, and disagreement and acquiescence. I cannot say I found the process particularly easy. However, ``omelets are not made without breaking eggs''. Although not something I completely agree with, the man who coined the phrase (Fran\c{c}ois de Charette) was executed in 1796 for breaking too many eggs, I sympathise with short--term difficulties being necessary for certain ends.

Those end results are not just papers, presentations and propositions, but also personal progress in acquiring more practical perspectives. One such perspective was becoming tolerant of my work being ``na\"{i}ve'', such as by not capturing all aspects of a particular concept. Na\"{i}ve work can help our understanding, as the field of Deontic Logic has shown through decades of conceptual changes and refinements. The Japanese aesthetic of Wabi Sabi encapsulates such a view, where transience and imperfection are sought and not fought. That outlook extends to people, including myself, as much as it applies to technical work. Moreover, it may be that if the response to a difficult position is not fight then it is flight, persistently tolerating  issues is important -- perhaps this all sounds clich\'{e}d, but the grass is always greener on the other side. Acknowledging and eventually feeling that not every problem is to be (re)solved, meant that the final work and paper were a walk in the park. 

Getting to that point, it must be emphasised, was not a solitary effort. There are many people that, in ways small and big, shaped my personal and research outlook in constructive ways and helped to ensure that the outcome was overall positive. I would therefore like to segue to thanking some of those people that helped me along the way.

Firstly, I would like to thank, in no particular order, my supervisory team comprising my promotor Prof. dr. Catholijn M. Jonker, and copromotors Dr. Birna van Riemsdijk and Dr. Virginia Dignum for their support, discussion and useful comments. Catholijn, thank you for giving me the opportunity to study at Interactive Intelligence -- a research group that is both positively diverse in people and research topics -- for having confidence when things were not going well, and for helping me with career advice and the next steps towards the end. Birna, thank you for giving me time to discover my research interests, putting together a diverse and open ended project, emphasising simplicity (but not too simple!) over complexity and helping me through the early stages of learning formal work. Virginia, thank you for helping me in understanding where and when to stop solving a large problem and start focussing on solving a small problem well, ensuring different aspects of my research were more modular than iterative, and especially, thank you for understanding and resolving seemingly insurmountable difficulties. Of course, there is more to thank for, but I feel these aspects were at the core of getting me to the end.

Moreover, I would like to extend my thanks to my mentor, Dr. Klaas Pieter Hart. Although we met infrequently, Klaas listened during times of strife, gave practical advice and, much to my reassurance, validated my concerns.  

Special thanks to the examination committee: Prof. dr. Katie Atkinson, Prof. dr. Frances Brazier, Prof. dr. Jeremy Pitt and Prof. dr. Leon van der Torre. Thank you for reading my dissertation. Moreover, thank you Katie Atkinson for the remarks, they helped with clarification and it is good to know there is agreement on the dissertation's contributions and shortcomings. Thank you Leon van der Torre for the detailed comments, they helped to better position the work and gave useful insights into nascent research topics.

I would also like to thank  Dr. Tingting Li, Dr. Julian Padget and Dr. Marina de Vos for the collaborative research and co--authorships. In particular, thank you for being so welcoming when I visited Bath, introducing me to your research on institutions and your overall approach. Our co--authorships greatly influenced my dissertation and research.

I'd also like to thank everyone who was involved in the SHINE multi--disciplinary project under which this research was conducted. Especially, I'd like to thank the SHINE PhD/postdoc researchers -- Qingzhi Liu, Dr. Paolo Palmieri, Gleb Polevoy, Venkat Roy and Sergio Velasco -- for the enjoyable project meetings and finding ways to work together.

I was between two departments during my PhD. First, I would like to thank all of my wonderful colleagues at the ICT section of TBM and the Interactive Intelligence group of EWI, there are too many to mention so thank you all. Both groups were diverse in research topics and people, so thank you for widening my personal and research perspectives.

I would specifically like to thank my TBM office--mates: Jie Jiang and Janneke van der Zwaan for discussing common problems encountered, being so welcoming and thank you Jie for those early days explaining the area as a whole and what you thought I needed to do to get started. I would also like to thank Virginia's latest PhD students (at the time of writing), Klara Pigmans and Rijk Mercuur for the interesting discussions about their own research and perspectives, which I look forward to continuing. I would also like to thank my II office mates: Alex Kayal for his anecdotes and banter, Marieke Peeters for listening to my cynicism and countering it with research rigour, and Wenxin Wang for conversing about family and showing it can coincide with work. Moreover, I would like to thank the following inspiring individuals for helping with the softer problems encountered: Reyhan Aydogan for her positivity and emphasis on work--life balance, Chris Detweiler for his dry humour and relateability, Maaike Harbers for her outgoing enthusiasm and compassion, Max Knobbout for his insightful knowledge and openness, Arman Noroozian for his academic enthusiasm and rational calmness, and Vanessa Vakili for her grounded outlook and sympathetic listening.

Moreover, I would also like to thank from the II group Fawad Khan for helping with implementation, and Wouter Pasman with helping me to resolve code problems and for always having an alternative take on things. Thank you Anita Hoogmoed and Caroline de Wit for always being willing to help with organisational matters and maintaining a cheerful state. I'd like to thank Ruud de Jong and Bart Vastenhouw for happily assisting in technical matters (and also Ruud for the vast coffee supplies).

My long--time friend Sonal Sodha/Keay was there for me during the difficult times. Moreover, long long before I even embarked on my PhD she gave me a sharp push that helped me get to where I am now. Thank you, Sonal.

Finally, I would like to thank my wonderful life partner, R$\overline{\mbox{u}}$ta, for always being there for me, providing both joy and support. Thank you for all of the adventures helping me to get away from work and for being understanding when all I could do was work. Thank you for your love and support -- I could not have done it without you R$\overline{\mbox{u}}$ta, I love you insanely.
\chapter*{Curriculum Vitae}
\addcontentsline{toc}{chapter}{Curriculum Vitae}
\setheader{Curriculum Vitae}

Thomas C. King was born on June 18, 1987 in London, United Kingdom. He began studying for a Bachelors in Computer Science in 2008 at King's College London (KCL). He postponed his Bachelors' studies, after two years whereupon he worked as a software developer and architect for UBS. Thomas resumed his studies one year later during which he became interested in AI, in particular on logic based approaches to topics such as belief and change. He completed his Bachelors at KCL, First Class with Honours, and was awarded the best KCL Computer Science dissertation of 2012. He then decided to pursue his interests in AI further by undertaking a PhD. Having considered a PhD in sub--symbolic AI such as genetic algorithms or symbolic AI choosing symbolic AI, he chose the latter focusing on normative reasoning, having become interested in ascribing meaning with tangible and precise conceptualisations.

His PhD research was conducted from 2012 until 2016 at TU Delft, on formalising institutions, in the philosophical and legal sense (e.g. conventions, laws, contracts and promises), with a focus on institutions that ascribe and regulate the creation and behaviour of other institutional rules. Thomas enjoyed attending and presenting his work at numerous conferences and workshops, as well as making research visits to the University of Bath, gaining an interest in the combination of practical and conceptual work on normative reasoning as a result. Hence -- in order to pursue the combination of practical and formal work further -- from September 2016, Thomas started working at the School of Computing and Communications of Lancaster University, United Kingdom, as a post--doctoral Senior Research Associate researching social commitment reasoning as an abstraction for: databases, decentralising software and verifying potential agent interactions.
\chapter*{SIKS Dissertation Series}
\setheader{SIKS Dissertation Series}
\newcommand{\SIKSdiss}[3]{{\bf #1}\hspace*{1ex}#2, {\it #3.}\\}

\noindent
\SIKSdiss{2016-41}{Thomas C King (TUD)}{Governing Governance: A Formal Framework for Analysing Institutional Design and Enactment Governance}
\SIKSdiss{2016-40}{Christian Detweiler (TUD)}{Accounting for Values in Design}
\SIKSdiss{2016-39}{Merijn Bruijnes (UT)}{Believable Suspect Agents; Response and Interpersonal Style Selection for an Artificial Suspect}
\SIKSdiss{2016-38}{Andrea Minuto (UT)}{MATERIALS THAT MATTER  -  Smart Materials meet Art \& Interaction Design}
\SIKSdiss{2016-37}{Giovanni Sileno (UvA)}{Aligning Law and Action - a conceptual and computational inquiry}
\SIKSdiss{2016-36}{Daphne Karreman (UT)}{Beyond R2D2: The design of nonverbal interaction behavior optimized for robot-specific morphologies}
\SIKSdiss{2016-35}{Zhaochun Ren}{Monitoring Social Media: Summarization, Classification and Recommendation}
\SIKSdiss{2016-34}{Dennis Schunselaar (TUE)}{Title:  Configurable Process Trees: Elicitation, Analysis, and Enactment}
\SIKSdiss{2016-33}{Peter Bloem (UVA)}{Single Sample Statistics, exercises in learning from just one example}
\SIKSdiss{2016-32}{Eelco Vriezekolk (UT)}{Assessing Telecommunication Service Availability Risks for Crisis Organisations}
\SIKSdiss{2016-31}{Mohammad Khelghati (UT)}{Deep web content monitoring}
\SIKSdiss{2016-30}{Ruud Mattheij (UvT)}{The Eyes Have It}
\SIKSdiss{2016-29}{Nicolas Höning (TUD)}{Peak reduction in decentralised electricity systems -Markets and prices for flexible planning}
\SIKSdiss{2016-28}{Mingxin Zhang (TUD)}{Large-scale Agent-based Social Simulation - A study on epidemic prediction and control}
\SIKSdiss{2016-27}{Wen Li (TUD)}{Understanding Geo-spatial Information on Social Media}
\SIKSdiss{2016-26}{Dilhan Thilakarathne (VU)}{In or Out of Control: Exploring Computational Models to Study the Role of Human Awareness and Control in Behavioural Choices, with Applications in Aviation and Energy Management Domains}
\SIKSdiss{2016-25}{Julia Kiseleva (TU/e)}{Using Contextual Information to Understand Searching and Browsing Behavior}
\SIKSdiss{2016-24}{Brend Wanders (UT)}{Repurposing and Probabilistic Integration of Data; An Iterative and data model independent approach}
\SIKSdiss{2016-23}{Fei Cai (UVA)}{Query Auto Completion in Information Retrieval}
\SIKSdiss{2016-22}{Grace Lewis (VU)}{Software Architecture Strategies for Cyber-Foraging Systems}
\SIKSdiss{2016-21}{Alejandro Moreno Célleri (UT)}{From Traditional to Interactive Playspaces: Automatic Analysis of Player Behavior in the Interactive Tag Playground}
\SIKSdiss{2016-20}{Daan Odijk (UVA)}{Context \& Semantics in News \& Web Search}
\SIKSdiss{2016-19}{Julia Efremova (Tu/e)}{Mining Social Structures from Genealogical Data}
\SIKSdiss{2016-18}{Albert Meroño Peñuela}{Refining Statistical Data on the Web}
\SIKSdiss{2016-17}{Berend Weel (VU)}{Towards Embodied Evolution of Robot Organisms}
\SIKSdiss{2016-16}{Guangliang Li (UVA)}{Socially Intelligent Autonomous Agents that Learn from Human Reward}
\SIKSdiss{2016-15}{Steffen Michels (RUN)}{Hybrid Probabilistic Logics - Theoretical Aspects, Algorithms and Experiments}
\SIKSdiss{2016-14}{Ravi Khadka (UU)}{Revisiting Legacy Software System Modernization}
\SIKSdiss{2016-13}{Nana Baah Gyan (VU)}{The Web, Speech Technologies and Rural Development in West Africa - An}
\SIKSdiss{2016-12}{Max Knobbout (UU)}{Logics for Modelling and Verifying Normative Multi-Agent Systems}
\SIKSdiss{2016-11}{Anne Schuth (UVA)}{Search Engines that Learn from Their Users}
\SIKSdiss{2016-10}{George Karafotias (VUA)}{Parameter Control for Evolutionary Algorithms}
\SIKSdiss{2016-09}{Archana Nottamkandath (VU)}{Trusting Crowdsourced Information on Cultural Artefacts}
\SIKSdiss{2016-08}{Matje van de Camp (TiU)}{A Link to the Past: Constructing Historical Social Networks from Unstructured Data}
\SIKSdiss{2016-07}{Jeroen de Man (VU)}{Measuring and modeling negative emotions for virtual training}
\SIKSdiss{2016-06}{Michel Wilson (TUD)}{Robust scheduling in an uncertain environment}
\SIKSdiss{2016-05}{Evgeny Sherkhonov (UVA)}{Expanded Acyclic Queries: Containment and an Application in Explaining Missing Answers}
\SIKSdiss{2016-04}{Laurens Rietveld (VU)}{Publishing and Consuming Linked Data}
\SIKSdiss{2016-03}{Maya Sappelli (RUN)}{Knowledge Work in Context: User Centered Knowledge Worker Support}
\SIKSdiss{2016-02}{Michiel Christiaan Meulendijk (UU)}{Optimizing medication reviews through decision support: prescribing a better pill to swallow}
\SIKSdiss{2016-01}{Syed Saiden Abbas (RUN)}{Recognition of Shapes by Humans and Machines}
\SIKSdiss{2015-35}{Jungxao Xu (TUD)}{Affective Body Language of Humanoid Robots: Perception and Effects in Human Robot Interaction}
\SIKSdiss{2015-34}{Victor de Graaf (UT)}{Gesocial Recommender Systems}
\SIKSdiss{2015-33}{Frederik Schadd (TUD)}{Ontology Mapping with Auxiliary Resources}
\SIKSdiss{2015-32}{Jerome Gard(UL)}{Corporate Venture Management in SMEs}
\SIKSdiss{2015-31}{Yakup Ko\c{c} (TUD)}{On the robustness of Power Grids}
\SIKSdiss{2015-30}{Kiavash Bahreini(OU)}{Real-time Multimodal Emotion Recognition in E-Learning}
\SIKSdiss{2015-29}{Hendrik Baier (UM)}{Monte-Carlo Tree Search Enhancements for One-Player and Two-Player Domains}
\SIKSdiss{2015-28}{Janet Bagorogoza (TiU)}{KNOWLEDGE MANAGEMENT AND HIGH PERFORMANCE; The Uganda Financial Institutions Model for HPO}
\SIKSdiss{2015-27}{S\'{a}ndor H\'{e}man (CWI)}{Updating compressed colomn stores}
\SIKSdiss{2015-26}{Alexander Hogenboom (EUR)}{Sentiment Analysis of Text Guided by Semantics and Structure}
\SIKSdiss{2015-25}{Steven Woudenberg (UU)}{Bayesian Tools for Early Disease Detection}
\SIKSdiss{2015-24}{Richard Berendsen (UVA)}{Finding People, Papers, and Posts: Vertical Search Algorithms and Evaluation}
\SIKSdiss{2015-23}{Luit Gazendam (VU)}{Cataloguer Support in Cultural Heritage}
\SIKSdiss{2015-21}{Sibren Fetter (OUN)}{Using Peer-Support to Expand and Stabilize Online Learning}
\SIKSdiss{2015-20}{Lo\"{i}s Vanh\'{e}e (UU)}{Using Culture and Values to Support Flexible Coordination}
\SIKSdiss{2015-19}{Bernardo Tabuenca (OUN)}{Ubiquitous Technology for Lifelong Learners}
\SIKSdiss{2015-18}{Holger Pirk (CWI)}{Waste Not, Want Not! - Managing Relational Data in Asymmetric Memories}
\SIKSdiss{2015-17}{Andr\'{e} van Cleeff (UT)}{Physical and Digital Security Mechanisms: Properties, Combinations and Trade-offs}
\SIKSdiss{2015-16}{Changyun Wei (UT)}{Cognitive Coordination for Cooperative Multi-Robot Teamwork}
\SIKSdiss{2015-15}{Klaas Andries de Graaf (VU)}{Ontology-based Software Architecture Documentation}
\SIKSdiss{2015-14}{Bart van Straalen (UT)}{A cognitive approach to modeling bad news conversations}
\SIKSdiss{2015-13}{Giuseppe Procaccianti(VU)}{Energy-Efficient Software}
\SIKSdiss{2015-12}{Julie M. Birkholz (VU)}{Modi Operandi of Social Network Dynamics: The Effect of Context on Scientific Collaboration Networks}
\SIKSdiss{2015-11}{Yongming Luo(TUE)}{Designing algorithms for big graph datasets: A study of computing bisimulation and joins}
\SIKSdiss{2015-10}{Henry Hermans (OUN)}{OpenU: design of an integrated system to support lifelong learning}
\SIKSdiss{2015-09}{Randy Klaassen(UT)}{HCI Perspectives on Behavior Change Support Systems}
\SIKSdiss{2015-08}{Jie Jiang (TUD)}{Organizational Compliance: An agent-based model for designing and evaluating organizational interactions}
\SIKSdiss{2015-07}{Maria-Hendrike Peetz(UvA)}{Time-Aware Online Reputation Analysis}
\SIKSdiss{2015-06}{Farideh Heidari (TUD)}{Business Process Quality Computation - Computing Non-Functional Requirements to Improve Business Processes}
\SIKSdiss{2015-05}{Christoph B\"{o}sch (UT)}{Cryptographically Enforced Search Pattern Hiding}
\SIKSdiss{2015-04}{Howard Spoelstra (OUN)}{Collaborations in Open Learning Environments}
\SIKSdiss{2015-03}{Twan van Laarhoven (RUN)}{Machine learning for network data}
\SIKSdiss{2015-02}{Faiza Bukhsh (UvT)}{Smart auditing: Innovative Compliance Checking in Customs Controls}
\SIKSdiss{2015-01}{Niels Netten (UvA)}{Machine Learning for Relevance of Information in Crisis Response}
\SIKSdiss{2014-47}{Shangsong Liang (UVA)}{Fusion and Diversification in Information Retrieval}
\SIKSdiss{2014-46}{Ke Tao (TUD)}{Social Web Data Analytics: Relevance, Redundancy, Diversity}
\SIKSdiss{2014-45}{Birgit Schmitz (OUN)}{Mobile Games for Learning: A Pattern-Based Approach}
\SIKSdiss{2014-44}{Paulien Meesters (UvT)}{Intelligent Blauw. Met als ondertitel: Intelligence-gestuurde politiezorg in gebiedsgebonden eenheden.}
\SIKSdiss{2014-43}{Kevin Vlaanderen (UU)}{Supporting Process Improvement using Method Increments}
\SIKSdiss{2014-42}{Carsten Eijckhof (CWI/TUD)}{Contextual Multidimensional Relevance Models}
\SIKSdiss{2014-41}{Frederic Hogenboom (EUR)}{Automated Detection of Financial Events in News Text}
\SIKSdiss{2014-40}{Walter Omona (RUN)}{A Framework for Knowledge Management Using ICT in Higher Education}
\SIKSdiss{2014-39}{Jasmina Maric (UvT)}{Web Communities, Immigration, and Social Capital}
\SIKSdiss{2014-38}{Danny Plass-Oude Bos (UT)}{Making brain-computer interfaces better: improving usability through post-processing.}
\SIKSdiss{2014-37}{Maral Dadvar (UT)}{Experts and Machines United Against Cyberbullying}
\SIKSdiss{2014-36}{Joos Buijs (TUE)}{Flexible Evolutionary Algorithms for Mining Structured Process Models}
\SIKSdiss{2014-35}{Joost van Ooijen (UU)}{Cognitive Agents in Virtual Worlds: A Middleware Design Approach}
\SIKSdiss{2014-34}{Christina Manteli(VU)}{The Effect of Governance in Global Software Development: Analyzing Transactive Memory Systems.}
\SIKSdiss{2014-33}{Tesfa Tegegne (RUN)}{Service Discovery in eHealth}
\SIKSdiss{2014-32}{Naser Ayat (UvA)}{On Entity Resolution in Probabilistic Data}
\SIKSdiss{2014-31}{Leo van Moergestel (UU)}{Agent Technology in Agile Multiparallel Manufacturing and Product Support}
\SIKSdiss{2014-30}{Peter de Cock (UvT)}{Anticipating Criminal Behaviour}
\SIKSdiss{2014-29}{Jaap Kabbedijk (UU)}{Variability in Multi-Tenant Enterprise Software}
\SIKSdiss{2014-28}{Anna Chmielowiec (VU)}{Decentralized k-Clique Matching}
\SIKSdiss{2014-27}{Rui Jorge Almeida (EUR)}{Conditional Density Models Integrating Fuzzy and Probabilistic Representations of Uncertainty}
\SIKSdiss{2014-26}{Tim Baarslag (TUD)}{What to Bid and When to Stop}
\SIKSdiss{2014-25}{Martijn Lappenschaar (RUN)}{New network models for the analysis of disease interaction}
\SIKSdiss{2014-24}{Davide Ceolin (VU)}{Trusting Semi-structured Web Data}
\SIKSdiss{2014-23}{Eleftherios Sidirourgos (UvA/CWI)}{Space Efficient Indexes for the Big Data Era}
\SIKSdiss{2014-22}{Marieke Peeters (UU)}{Personalized Educational Games - Developing agent-supported scenario-based training}
\SIKSdiss{2014-21}{Kassidy Clark (TUD)}{Negotiation and Monitoring in Open Environments}
\SIKSdiss{2014-20}{Mena Habib (UT)}{Named Entity Extraction and Disambiguation for Informal Text: The Missing Link}
\SIKSdiss{2014-19}{Vinicius Ramos (TUE)}{Adaptive Hypermedia Courses: Qualitative and Quantitative Evaluation and Tool Support}
\SIKSdiss{2014-18}{Mattijs Ghijsen (UVA)}{Methods and Models for the Design and Study of Dynamic Agent Organizations}
\SIKSdiss{2014-17}{Kathrin Dentler (VU)}{Computing healthcare quality indicators automatically: Secondary Use of Patient Data and Semantic Interoperability}
\SIKSdiss{2014-16}{Krystyna Milian (VU)}{Supporting trial recruitment and design by automatically interpreting eligibility criteria}
\SIKSdiss{2014-15}{Natalya Mogles (VU)}{Agent-Based Analysis and Support of Human Functioning in Complex Socio-Technical Systems: Applications in Safety and Healthcare}
\SIKSdiss{2014-14}{Yangyang Shi (TUD)}{Language Models With Meta-information}
\SIKSdiss{2014-13}{Arlette van Wissen (VU)}{Agent-Based Support for Behavior Change: Models and Applications in Health and Safety Domains}
\SIKSdiss{2014-12}{Willem van Willigen (VU)}{Look Ma, No Hands: Aspects of Autonomous Vehicle Control}
\SIKSdiss{2014-11}{Janneke van der Zwaan (TUD)}{An Empathic Virtual Buddy for Social Support}
\SIKSdiss{2014-10}{Ivan Salvador Razo Zapata (VU)}{Service Value Networks}
\SIKSdiss{2014-09}{Philip Jackson (UvT)}{Toward Human-Level Artificial Intelligence: Representation and Computation of Meaning in Natural Language}
\SIKSdiss{2014-08}{Samur Araujo (TUD)}{Data Integration over Distributed and Heterogeneous Data Endpoints}
\SIKSdiss{2014-07}{Arya Adriansyah (TUE)}{Aligning Observed and Modeled Behavior}
\SIKSdiss{2014-06}{Damian Tamburri (VU)}{Supporting Networked Software Development}
\SIKSdiss{2014-05}{Jurriaan van Reijsen (UU)}{Knowledge Perspectives on Advancing Dynamic Capability}
\SIKSdiss{2014-04}{Hanna Jochmann-Mannak (UT)}{Websites for children: search strategies and interface design - Three studies on children's search performance and evaluation}
\SIKSdiss{2014-03}{Sergio Raul Duarte Torres (UT)}{Information Retrieval for Children: Search Behavior and Solutions}
\SIKSdiss{2014-02}{Fiona Tuliyano (RUN)}{Combining System Dynamics with a Domain Modeling Method}
\SIKSdiss{2014-01}{Nicola Barile (UU)}{Studies in Learning Monotone Models from Data}
\SIKSdiss{2013-43}{Marc Bron (UVA)}{Exploration and Contextualization through Interaction and Concepts}
\SIKSdiss{2013-42}{L\'{e}on Planken (TUD)}{Algorithms for Simple Temporal Reasoning}
\SIKSdiss{2013-41}{Jochem Liem (UVA)}{Supporting the Conceptual Modelling of Dynamic Systems: A Knowledge Engineering Perspective on Qualitative Reasoning}
\SIKSdiss{2013-40}{Pim Nijssen (UM)}{Monte-Carlo Tree Search for Multi-Player Games}
\SIKSdiss{2013-39}{Joop de Jong (TUD)}{A Method for Enterprise Ontology based Design of Enterprise Information Systems}
\SIKSdiss{2013-38}{Eelco den Heijer (VU)}{Autonomous Evolutionary Art}
\SIKSdiss{2013-37}{Dirk B\"{o}rner (OUN)}{Ambient Learning Displays}
\SIKSdiss{2013-36}{Than Lam Hoang (TUe)}{Pattern Mining in Data Streams}
\SIKSdiss{2013-35}{Abdallah El Ali (UvA)}{Minimal Mobile Human Computer Interaction}
\SIKSdiss{2013-34}{Kien Tjin-Kam-Jet (UT)}{Distributed Deep Web Search}
\SIKSdiss{2013-33}{Qi Gao (TUD)}{User Modeling and Personalization in the Microblogging Sphere}
\SIKSdiss{2013-32}{Kamakshi Rajagopal (OUN)}{Networking For Learning; The role of Networking in a Lifelong Learner's Professional Development}
\SIKSdiss{2013-31}{Dinh Khoa Nguyen (UvT)}{Blueprint Model and Language for Engineering Cloud Applications}
\SIKSdiss{2013-30}{Joyce Nakatumba (TUE)}{Resource-Aware Business Process Management: Analysis and Support}
\SIKSdiss{2013-29}{Iwan de Kok (UT)}{Listening Heads}
\SIKSdiss{2013-28}{Frans van der Sluis (UT)}{When Complexity becomes Interesting: An Inquiry into the Information eXperience}
\SIKSdiss{2013-27}{Mohammad Huq (UT)}{Inference-based Framework Managing Data Provenance}
\SIKSdiss{2013-26}{Alireza Zarghami (UT)}{Architectural Support for Dynamic Homecare Service Provisioning}
\SIKSdiss{2013-25}{Agnieszka Anna Latoszek-Berendsen (UM)}{Intention-based Decision Support. A new way of representing and implementing clinical guidelines in a Decision Support System}
\SIKSdiss{2013-24}{Haitham Bou Ammar (UM)}{Automated Transfer in Reinforcement Learning}
\SIKSdiss{2013-23}{Patricio de Alencar Silva(UvT)}{Value Activity Monitoring}
\SIKSdiss{2013-22}{Tom Claassen (RUN)}{Causal Discovery and Logic}
\SIKSdiss{2013-21}{Sander Wubben (UvT)}{Text-to-text generation by monolingual machine translation}
\SIKSdiss{2013-20}{Katja Hofmann (UvA)}{Fast and Reliable Online Learning to Rank for Information Retrieval}
\SIKSdiss{2013-19}{Renze Steenhuizen (TUD)}{Coordinated Multi-Agent Planning and Scheduling}
\SIKSdiss{2013-18}{Jeroen Janssens (UvT)}{Outlier Selection and One-Class Classification}
\SIKSdiss{2013-17}{Koen Kok (VU)}{The PowerMatcher: Smart Coordination for the Smart Electricity Grid}
\SIKSdiss{2013-16}{Eric Kok (UU)}{Exploring the practical benefits of argumentation in multi-agent deliberation}
\SIKSdiss{2013-15}{Daniel Hennes (UM)}{Multiagent Learning - Dynamic Games and Applications}
\SIKSdiss{2013-14}{Jafar Tanha (UVA)}{Ensemble Approaches to Semi-Supervised Learning Learning}
\SIKSdiss{2013-13}{Mohammad Safiri(UT)}{Service Tailoring: User-centric creation of integrated IT-based homecare services to support independent living of elderly}
\SIKSdiss{2013-12}{Marian Razavian(VU)}{Knowledge-driven Migration to Services}
\SIKSdiss{2013-11}{Evangelos Pournaras(TUD)}{Multi-level Reconfigurable Self-organization in Overlay Services}
\SIKSdiss{2013-10}{Jeewanie Jayasinghe Arachchige(UvT)}{A Unified Modeling Framework for Service Design.}
\SIKSdiss{2013-09}{Fabio Gori (RUN)}{Metagenomic Data Analysis: Computational Methods and Applications}
\SIKSdiss{2013-08}{Robbert-Jan Merk(VU)}{Making enemies: cognitive modeling for opponent agents in fighter pilot simulators}
\SIKSdiss{2013-07}{Giel van Lankveld (UvT)}{Quantifying Individual Player Differences}
\SIKSdiss{2013-06}{Romulo Goncalves(CWI)}{The Data Cyclotron: Juggling Data and Queries for a Data Warehouse Audience}
\SIKSdiss{2013-05}{Dulce Pumareja (UT)}{Groupware Requirements Evolutions Patterns}
\SIKSdiss{2013-04}{Chetan Yadati(TUD)}{Coordinating autonomous planning and scheduling}
\SIKSdiss{2013-03}{Szymon Klarman (VU)}{Reasoning with Contexts in Description Logics}
\SIKSdiss{2013-02}{Erietta Liarou (CWI)}{MonetDB/DataCell: Leveraging the Column-store Database Technology for Efficient and Scalable Stream Processing}
\SIKSdiss{2013-01}{Viorel Milea (EUR)}{News Analytics for Financial Decision Support}
\SIKSdiss{2012-51}{Jeroen de Jong (TUD)}{Heuristics in Dynamic Sceduling; a practical framework with a case study in elevator dispatching}
\SIKSdiss{2012-50}{Steven van Kervel (TUD)}{Ontologogy driven Enterprise Information Systems Engineering}
\SIKSdiss{2012-49}{Michael Kaisers (UM)}{Learning against Learning - Evolutionary dynamics of reinforcement learning algorithms in strategic interactions}
\SIKSdiss{2012-48}{Jorn Bakker (TUE)}{Handling Abrupt Changes in Evolving Time-series Data}
\SIKSdiss{2012-47}{Manos Tsagkias (UVA)}{Mining Social Media: Tracking Content and Predicting Behavior}
\SIKSdiss{2012-46}{Simon Carter (UVA)}{Exploration and Exploitation of Multilingual Data for Statistical Machine Translation}
\SIKSdiss{2012-45}{Benedikt Kratz (UvT)}{A Model and Language for Business-aware Transactions}
\SIKSdiss{2012-44}{Anna Tordai (VU)}{On Combining Alignment Techniques}
\SIKSdiss{2012-43}{Withdrawn}{N/A}
\SIKSdiss{2012-42}{Dominique Verpoorten (OU)}{Reflection Amplifiers in self-regulated Learning}
\SIKSdiss{2012-41}{Sebastian Kelle (OU)}{Game Design Patterns for Learning}
\SIKSdiss{2012-40}{Agus Gunawan (UvT)}{Information Access for SMEs in Indonesia}
\SIKSdiss{2012-39}{Hassan Fatemi (UT)}{Risk-aware design of value and coordination networks}
\SIKSdiss{2012-38}{Selmar Smit (VU)}{Parameter Tuning and Scientific Testing in Evolutionary Algorithms}
\SIKSdiss{2012-37}{Agnes Nakakawa (RUN)}{A Collaboration Process for Enterprise Architecture Creation}
\SIKSdiss{2012-36}{Denis Ssebugwawo (RUN)}{Analysis and Evaluation of Collaborative Modeling Processes}
\SIKSdiss{2012-35}{Evert Haasdijk (VU)}{Never Too Old To Learn -- On-line Evolution of Controllers in Swarm- and Modular Robotics}
\SIKSdiss{2012-34}{Pavol Jancura (RUN)}{Evolutionary analysis in PPI networks and applications}
\SIKSdiss{2012-33}{Rory Sie (OUN)}{Coalitions in Cooperation Networks (COCOON)}
\SIKSdiss{2012-32}{Wietske Visser (TUD)}{Qualitative multi-criteria preference representation and reasoning}
\SIKSdiss{2012-31}{Emily Bagarukayo (RUN)}{A Learning by Construction Approach for Higher Order Cognitive Skills Improvement, Building Capacity and Infrastructure}
\SIKSdiss{2012-30}{Alina Pommeranz (TUD)}{Designing Human-Centered Systems for Reflective Decision Making}
\SIKSdiss{2012-29}{Almer Tigelaar (UT)}{Peer-to-Peer Information Retrieval}
\SIKSdiss{2012-28}{Nancy Pascall (UvT)}{Engendering Technology Empowering Women}
\SIKSdiss{2012-27}{Hayrettin Gurkok (UT)}{Mind the Sheep! User Experience Evaluation \& Brain-Computer Interface Games}
\SIKSdiss{2012-26}{Emile de Maat (UVA)}{Making Sense of Legal Text}
\SIKSdiss{2012-25}{Silja Eckartz (UT)}{Managing the Business Case Development in Inter-Organizational IT Projects: A Methodology and its Application}
\SIKSdiss{2012-24}{Laurens van der Werff (UT)}{Evaluation of Noisy Transcripts for Spoken Document Retrieval}
\SIKSdiss{2012-23}{Christian Muehl (UT)}{Toward Affective Brain-Computer Interfaces: Exploring the Neurophysiology of Affect during Human Media Interaction}
\SIKSdiss{2012-22}{Thijs Vis (UvT)}{Intelligence, politie en veiligheidsdienst: verenigbare grootheden?}
\SIKSdiss{2012-21}{Roberto Cornacchia (TUD)}{Querying Sparse Matrices for Information Retrieval}
\SIKSdiss{2012-20}{Ali Bahramisharif (RUN)}{Covert Visual Spatial Attention, a Robust Paradigm for Brain-Computer Interfacing}
\SIKSdiss{2012-19}{Helen Schonenberg (TUE)}{What's Next? Operational Support for Business Process Execution}
\SIKSdiss{2012-18}{Eltjo Poort (VU)}{Improving Solution Architecting Practices}
\SIKSdiss{2012-17}{Amal Elgammal (UvT)}{Towards a Comprehensive Framework for Business Process Compliance}
\SIKSdiss{2012-16}{Fiemke Both (VU)}{Helping people by understanding them - Ambient Agents supporting task execution and depression treatment}
\SIKSdiss{2012-15}{Natalie van der Wal (VU)}{Social Agents. Agent-Based Modelling of Integrated Internal and Social Dynamics of Cognitive and Affective Processes.}
\SIKSdiss{2012-14}{Evgeny Knutov(TUE)}{Generic Adaptation Framework for Unifying Adaptive Web-based Systems}
\SIKSdiss{2012-13}{Suleman Shahid (UvT)}{Fun and Face: Exploring non-verbal expressions of emotion during playful interactions}
\SIKSdiss{2012-12}{Kees van der Sluijs (TUE)}{Model Driven Design and Data Integration in Semantic Web Information Systems}
\SIKSdiss{2012-11}{J.C.B. Rantham Prabhakara (TUE)}{Process Mining in the Large: Preprocessing, Discovery, and Diagnostics}
\SIKSdiss{2012-10}{David Smits (TUE)}{Towards a Generic Distributed Adaptive Hypermedia Environment}
\SIKSdiss{2012-09}{Ricardo Neisse (UT)}{Trust and Privacy Management Support for Context-Aware Service Platforms}
\SIKSdiss{2012-08}{Gerben de Vries (UVA)}{Kernel Methods for Vessel Trajectories}
\SIKSdiss{2012-07}{Rianne van Lambalgen (VU)}{When the Going Gets Tough: Exploring Agent-based Models of Human Performance under Demanding Conditions}
\SIKSdiss{2012-06}{Wolfgang Reinhardt (OU)}{Awareness Support for Knowledge Workers in Research Networks}
\SIKSdiss{2012-05}{Marijn Plomp (UU)}{Maturing Interorganisational Information Systems}
\SIKSdiss{2012-04}{Jurriaan Souer (UU)}{Development of Content Management System-based Web Applications}
\SIKSdiss{2012-03}{Adam Vanya (VU)}{Supporting Architecture Evolution by Mining Software Repositories}
\SIKSdiss{2012-02}{Muhammad Umair(VU)}{Adaptivity, emotion, and Rationality in Human and Ambient Agent Models}
\SIKSdiss{2012-01}{Terry Kakeeto (UvT)}{Relationship Marketing for SMEs in Uganda}
\SIKSdiss{2011-49}{Andreea Niculescu (UT)}{Conversational interfaces for task-oriented spoken dialogues: design aspects influencing interaction quality}
\SIKSdiss{2011-48}{Mark Ter Maat (UT)}{Response Selection and Turn-taking for a Sensitive Artificial Listening Agent}
\SIKSdiss{2011-47}{Azizi Bin Ab Aziz(VU)}{Exploring Computational Models for Intelligent Support of Persons with Depression}
\SIKSdiss{2011-46}{Beibei Hu (TUD)}{Towards Contextualized Information Delivery: A Rule-based Architecture for the Domain of Mobile Police Work}
\SIKSdiss{2011-45}{Herman Stehouwer (UvT)}{Statistical Language Models for Alternative Sequence Selection}
\SIKSdiss{2011-44}{Boris Reuderink (UT)}{Robust Brain-Computer Interfaces}
\SIKSdiss{2011-43}{Henk van der Schuur (UU)}{Process Improvement through Software Operation Knowledge}
\SIKSdiss{2011-42}{Michal Sindlar (UU)}{Explaining Behavior through Mental State Attribution}
\SIKSdiss{2011-41}{Luan Ibraimi (UT)}{Cryptographically Enforced Distributed Data Access Control}
\SIKSdiss{2011-40}{Viktor Clerc (VU)}{Architectural Knowledge Management in Global Software Development}
\SIKSdiss{2011-39}{Joost Westra (UU)}{Organizing Adaptation using Agents in Serious Games}
\SIKSdiss{2011-38}{Nyree Lemmens (UM)}{Bee-inspired Distributed Optimization}
\SIKSdiss{2011-37}{Adriana Burlutiu (RUN)}{Machine Learning for Pairwise Data, Applications for Preference Learning and Supervised Network Inference}
\SIKSdiss{2011-36}{Erik van der Spek (UU)}{Experiments in serious game design: a cognitive approach}
\SIKSdiss{2011-35}{Maaike Harbers (UU)}{Explaining Agent Behavior in Virtual Training}
\SIKSdiss{2011-34}{Paolo Turrini (UU)}{Strategic Reasoning in Interdependence: Logical and Game-theoretical Investigations}
\SIKSdiss{2011-33}{Tom van der Weide (UU)}{Arguing to Motivate Decisions}
\SIKSdiss{2011-32}{Nees-Jan van Eck (EUR)}{Methodological Advances in Bibliometric Mapping of Science}
\SIKSdiss{2011-31}{Ludo Waltman (EUR)}{Computational and Game-Theoretic Approaches for Modeling Bounded Rationality}
\SIKSdiss{2011-30}{Egon van den Broek (UT)}{Affective Signal Processing (ASP): Unraveling the mystery of emotions}
\SIKSdiss{2011-29}{Faisal Kamiran (TUE)}{Discrimination-aware Classification}
\SIKSdiss{2011-28}{Rianne Kaptein(UVA)}{Effective Focused Retrieval by Exploiting Query Context and Document Structure}
\SIKSdiss{2011-27}{Aniel Bhulai (VU)}{Dynamic website optimization through autonomous management of design patterns}
\SIKSdiss{2011-26}{Matthijs Aart Pontier (VU)}{Virtual Agents for Human Communication - Emotion Regulation and Involvement-Distance Trade-Offs in Embodied Conversational Agents and Robots}
\SIKSdiss{2011-25}{Syed Waqar ul Qounain Jaffry (VU))}{Analysis and Validation of Models for Trust Dynamics}
\SIKSdiss{2011-24}{Herwin van Welbergen (UT)}{Behavior Generation for Interpersonal Coordination with Virtual Humans On Specifying, Scheduling and Realizing Multimodal Virtual Human Behavior}
\SIKSdiss{2011-23}{Wouter Weerkamp (UVA)}{Finding People and their Utterances in Social Media}
\SIKSdiss{2011-22}{Junte Zhang (UVA)}{System Evaluation of Archival Description and Access}
\SIKSdiss{2011-21}{Linda Terlouw (TUD)}{Modularization and Specification of Service-Oriented Systems}
\SIKSdiss{2011-20}{Qing Gu (VU)}{Guiding service-oriented software engineering - A view-based approach}
\SIKSdiss{2011-19}{Ellen Rusman (OU)}{The Mind ' s Eye on Personal Profiles}
\SIKSdiss{2011-18}{Mark Ponsen (UM)}{Strategic Decision-Making in complex games}
\SIKSdiss{2011-17}{Jiyin He (UVA)}{Exploring Topic Structure: Coherence, Diversity and Relatedness}
\SIKSdiss{2011-16}{Maarten Schadd (UM)}{Selective Search in Games of Different Complexity}
\SIKSdiss{2011-15}{Marijn Koolen (UvA)}{The Meaning of Structure: the Value of Link Evidence for Information Retrieval}
\SIKSdiss{2011-14}{Milan Lovric (EUR)}{Behavioral Finance and Agent-Based Artificial Markets}
\SIKSdiss{2011-13}{Xiaoyu Mao (UvT)}{Airport under Control. Multiagent Scheduling for Airport Ground Handling}
\SIKSdiss{2011-12}{Carmen Bratosin (TUE)}{Grid Architecture for Distributed Process Mining}
\SIKSdiss{2011-11}{Dhaval Vyas (UT)}{Designing for Awareness: An Experience-focused HCI Perspective}
\SIKSdiss{2011-10}{Bart Bogaert (UvT)}{Cloud Content Contention}
\SIKSdiss{2011-09}{Tim de Jong (OU)}{Contextualised Mobile Media for Learning}
\SIKSdiss{2011-08}{Nieske Vergunst (UU)}{BDI-based Generation of Robust Task-Oriented Dialogues}
\SIKSdiss{2011-07}{Yujia Cao (UT)}{Multimodal Information Presentation for High Load Human Computer Interaction}
\SIKSdiss{2011-06}{Yiwen Wang (TUE)}{Semantically-Enhanced Recommendations in Cultural Heritage}
\SIKSdiss{2011-05}{Base van der Raadt (VU)}{Enterprise Architecture Coming of Age - Increasing the Performance of an Emerging Discipline.}
\SIKSdiss{2011-04}{Hado van Hasselt (UU)}{Insights in Reinforcement Learning; Formal analysis and empirical evaluation of temporal-difference}
\SIKSdiss{2011-03}{Jan Martijn van der Werf (TUE)}{Compositional Design and Verification of Component-Based Information Systems}
\SIKSdiss{2011-02}{Nick Tinnemeier(UU)}{Organizing Agent Organizations. Syntax and Operational Semantics of an Organization-Oriented Programming Language}
\SIKSdiss{2011-01}{Botond Cseke (RUN)}{Variational Algorithms for Bayesian Inference in Latent Gaussian Models}
\SIKSdiss{2010-53}{Edgar Meij (UVA)}{Combining Concepts and Language Models for Information Access}
\SIKSdiss{2010-52}{Peter-Paul van Maanen (VU)}{Adaptive Support for Human-Computer Teams: Exploring the Use of Cognitive Models of Trust and Attention}
\SIKSdiss{2010-51}{Alia Khairia Amin (CWI)}{Understanding and supporting information seeking tasks in multiple sources}
\SIKSdiss{2010-50}{Bouke Huurnink (UVA)}{Search in Audiovisual Broadcast Archives}
\SIKSdiss{2010-49}{Jahn-Takeshi Saito (UM)}{Solving difficult game positions}
\SIKSdiss{2010-48}{Withdrawn}{}
\SIKSdiss{2010-47}{Chen Li (UT)}{Mining Process Model Variants: Challenges, Techniques, Examples}
\SIKSdiss{2010-46}{Vincent Pijpers (VU)}{e3alignment: Exploring Inter-Organizational Business-ICT Alignment}
\SIKSdiss{2010-45}{Vasilios Andrikopoulos (UvT)}{A theory and model for the evolution of software services}
\SIKSdiss{2010-44}{Pieter Bellekens (TUE)}{An Approach towards Context-sensitive and User-adapted Access to Heterogeneous Data Sources, Illustrated in the Television Domain}
\SIKSdiss{2010-43}{Peter van Kranenburg (UU)}{A Computational Approach to Content-Based Retrieval of Folk Song Melodies}
\SIKSdiss{2010-42}{Sybren de Kinderen (VU)}{Needs-driven service bundling in a multi-supplier setting - the computational e3-service approach}
\SIKSdiss{2010-41}{Guillaume Chaslot (UM)}{Monte-Carlo Tree Search}
\SIKSdiss{2010-40}{Mark van Assem (VU)}{Converting and Integrating Vocabularies for the Semantic Web}
\SIKSdiss{2010-39}{Ghazanfar Farooq Siddiqui (VU)}{Integrative modeling of emotions in virtual agents}
\SIKSdiss{2010-38}{Dirk Fahland (TUE)}{From Scenarios to components}
\SIKSdiss{2010-37}{Niels Lohmann (TUE)}{Correctness of services and their composition}
\SIKSdiss{2010-36}{Jose Janssen (OU)}{Paving the Way for Lifelong Learning; Facilitating competence development through a learning path specification}
\SIKSdiss{2010-35}{Dolf Trieschnigg (UT)}{Proof of Concept: Concept-based Biomedical Information Retrieval}
\SIKSdiss{2010-34}{Teduh Dirgahayu (UT)}{Interaction Design in Service Compositions}
\SIKSdiss{2010-33}{Robin Aly (UT)}{Modeling Representation Uncertainty in Concept-Based Multimedia Retrieval}
\SIKSdiss{2010-32}{Marcel Hiel (UvT)}{An Adaptive Service Oriented Architecture: Automatically solving Interoperability Problems}
\SIKSdiss{2010-31}{Victor de Boer (UVA)}{Ontology Enrichment from Heterogeneous Sources on the Web}
\SIKSdiss{2010-30}{Marieke van Erp (UvT)}{Accessing Natural History - Discoveries in data cleaning, structuring, and retrieval}
\SIKSdiss{2010-29}{Stratos Idreos(CWI)}{Database Cracking: Towards Auto-tuning Database Kernels}
\SIKSdiss{2010-28}{Arne Koopman (UU)}{Characteristic Relational Patterns}
\SIKSdiss{2010-27}{Marten Voulon (UL)}{Automatisch contracteren}
\SIKSdiss{2010-26}{Ying Zhang (CWI)}{XRPC: Efficient Distributed Query Processing on Heterogeneous XQuery Engines}
\SIKSdiss{2010-25}{Zulfiqar Ali Memon (VU)}{Modelling Human-Awareness for Ambient Agents: A Human Mindreading Perspective}
\SIKSdiss{2010-24}{Dmytro Tykhonov}{Designing Generic and Efficient Negotiation Strategies}
\SIKSdiss{2010-23}{Bas Steunebrink (UU)}{The Logical Structure of Emotions}
\SIKSdiss{2010-22}{Michiel Hildebrand (CWI)}{End-user Support for Access to\\ Heterogeneous Linked Data}
\SIKSdiss{2010-21}{Harold van Heerde (UT)}{Privacy-aware data management by means of data degradation}
\SIKSdiss{2010-20}{Ivo Swartjes (UT)}{Whose Story Is It Anyway? How Improv Informs Agency and Authorship of Emergent Narrative}
\SIKSdiss{2010-19}{Henriette Cramer (UvA)}{People's Responses to Autonomous and Adaptive Systems}
\SIKSdiss{2010-18}{Charlotte Gerritsen (VU)}{Caught in the Act: Investigating Crime by Agent-Based Simulation}
\SIKSdiss{2010-17}{Spyros Kotoulas (VU)}{Scalable Discovery of Networked Resources: Algorithms, Infrastructure, Applications}
\SIKSdiss{2010-16}{Sicco Verwer (TUD)}{Efficient Identification of Timed Automata, theory and practice}
\SIKSdiss{2010-15}{Lianne Bodenstaff (UT)}{Managing Dependency Relations in Inter-Organizational Models}
\SIKSdiss{2010-14}{Sander van Splunter (VU)}{Automated Web Service Reconfiguration}
\SIKSdiss{2010-13}{Gianluigi Folino (RUN)}{High Performance Data Mining using Bio-inspired techniques}
\SIKSdiss{2010-12}{Susan van den Braak (UU)}{Sensemaking software for crime analysis}
\SIKSdiss{2010-11}{Adriaan Ter Mors (TUD)}{The world according to MARP: Multi-Agent Route Planning}
\SIKSdiss{2010-10}{Rebecca Ong (UL)}{Mobile Communication and Protection of Children}
\SIKSdiss{2010-09}{Hugo Kielman (UL)}{A Politiele gegevensverwerking en Privacy, Naar een effectieve waarborging}
\SIKSdiss{2010-08}{Krzysztof Siewicz (UL)}{Towards an Improved Regulatory Framework of Free Software. Protecting user freedoms in a world of software communities and eGovernments}
\SIKSdiss{2010-07}{Wim Fikkert (UT)}{Gesture interaction at a Distance}
\SIKSdiss{2010-06}{Sander Bakkes (UvT)}{Rapid Adaptation of Video Game AI}
\SIKSdiss{2010-05}{Claudia Hauff (UT)}{Predicting the Effectiveness of Queries and Retrieval Systems}
\SIKSdiss{2010-04}{Olga Kulyk (UT)}{Do You Know What I Know? Situational Awareness of Co-located Teams in Multidisplay Environments}
\SIKSdiss{2010-03}{Joost Geurts (CWI)}{A Document Engineering Model and Processing Framework for Multimedia documents}
\SIKSdiss{2010-02}{Ingo Wassink (UT)}{Work flows in Life Science}
\SIKSdiss{2010-01}{Matthijs van Leeuwen (UU)}{Patterns that Matter}
\SIKSdiss{2009-46}{Loredana Afanasiev (UvA)}{Querying XML: Benchmarks and Recursion}
\SIKSdiss{2009-45}{Jilles Vreeken (UU)}{Making Pattern Mining Useful}
\SIKSdiss{2009-44}{Roberto Santana Tapia (UT)}{Assessing Business-IT Alignment in Networked Organizations}
\SIKSdiss{2009-43}{Virginia Nunes Leal Franqueira (UT)}{Finding Multi-step Attacks in Computer Networks using Heuristic Search and Mobile Ambients}
\SIKSdiss{2009-42}{Toine Bogers (UvT)}{Recommender Systems for Social Bookmarking}
\SIKSdiss{2009-41}{Igor Berezhnyy (UvT)}{Digital Analysis of Paintings}
\SIKSdiss{2009-40}{Stephan Raaijmakers (UvT)}{Multinomial Language Learning: Investigations into the Geometry of Language}
\SIKSdiss{2009-39}{Christian Stahl (TUE, Humboldt-Universitaet zu Berlin)}{Service Substitution -- A Behavioral Approach Based on Petri Nets}
\SIKSdiss{2009-38}{Riina Vuorikari (OU)}{Tags and self-organisation: a metadata ecology for learning resources in a multilingual context}
\SIKSdiss{2009-37}{Hendrik Drachsler (OUN)}{Navigation Support for Learners in Informal Learning Networks}
\SIKSdiss{2009-36}{Marco Kalz (OUN)}{Placement Support for Learners in Learning Networks}
\SIKSdiss{2009-35}{Wouter Koelewijn (UL)}{Privacy en Politiegegevens; Over geautomatiseerde normatieve informatie-uitwisseling}
\SIKSdiss{2009-34}{Inge van de Weerd (UU)}{Advancing in Software Product Management: An Incremental Method Engineering Approach}
\SIKSdiss{2009-33}{Khiet Truong (UT)}{How Does Real Affect Affect Affect Recognition In Speech?}
\SIKSdiss{2009-32}{Rik Farenhorst (VU) and Remco de Boer (VU)}{Architectural Knowledge Management: Supporting Architects and Auditors}
\SIKSdiss{2009-31}{Sofiya Katrenko (UVA)}{A Closer Look at Learning Relations from Text}
\SIKSdiss{2009-30}{Marcin Zukowski (CWI)}{Balancing vectorized query execution with bandwidth-optimized storage}
\SIKSdiss{2009-29}{Stanislav Pokraev (UT)}{Model-Driven Semantic Integration of Service-Oriented Applications}
\SIKSdiss{2009-28}{Sander Evers (UT)}{Sensor Data Management with Probabilistic Models}
\SIKSdiss{2009-27}{Christian Glahn (OU)}{Contextual Support of social Engagement and Reflection on the Web}
\SIKSdiss{2009-26}{Fernando Koch (UU)}{An Agent-Based Model for the Development of Intelligent Mobile Services}
\SIKSdiss{2009-25}{Alex van Ballegooij (CWI)}{"RAM: Array Database Management through Relational Mapping"}
\SIKSdiss{2009-24}{Annerieke Heuvelink (VUA)}{Cognitive Models for Training Simulations}
\SIKSdiss{2009-23}{Peter Hofgesang (VU)}{Modelling Web Usage in a Changing Environment}
\SIKSdiss{2009-22}{Pavel Serdyukov (UT)}{Search For Expertise: Going beyond direct evidence}
\SIKSdiss{2009-21}{Stijn Vanderlooy (UM)}{Ranking and Reliable Classification}
\SIKSdiss{2009-20}{Bob van der Vecht (UU)}{Adjustable Autonomy: Controling Influences on Decision Making}
\SIKSdiss{2009-19}{Valentin Robu (CWI)}{Modeling Preferences, Strategic Reasoning and Collaboration in Agent-Mediated Electronic Markets}
\SIKSdiss{2009-18}{Fabian Groffen (CWI)}{Armada, An Evolving Database System}
\SIKSdiss{2009-17}{Laurens van der Maaten (UvT)}{Feature Extraction from Visual Data}
\SIKSdiss{2009-16}{Fritz Reul (UvT)}{New Architectures in Computer Chess}
\SIKSdiss{2009-15}{Rinke Hoekstra (UVA)}{Ontology Representation - Design Patterns and Ontologies that Make Sense}
\SIKSdiss{2009-14}{Maksym Korotkiy (VU)}{From ontology-enabled services to service-enabled ontologies (making ontologies work in e-science with ONTO-SOA)}
\SIKSdiss{2009-13}{Steven de Jong (UM)}{Fairness in Multi-Agent Systems}
\SIKSdiss{2009-12}{Peter Massuthe (TUE, Humboldt-Universitaet zu Berlin)}{Operating Guidelines for Services}
\SIKSdiss{2009-11}{Alexander Boer (UVA)}{Legal Theory, Sources of Law \& the Semantic Web}
\SIKSdiss{2009-10}{Jan Wielemaker (UVA)}{Logic programming for knowledge-intensive interactive applications}
\SIKSdiss{2009-09}{Benjamin Kanagwa (RUN)}{Design, Discovery and Construction of Service-oriented Systems}
\SIKSdiss{2009-08}{Volker Nannen (VU)}{Evolutionary Agent-Based Policy Analysis in Dynamic Environments}
\SIKSdiss{2009-07}{Ronald Poppe (UT)}{Discriminative Vision-Based Recovery and Recognition of Human Motion}
\SIKSdiss{2009-06}{Muhammad Subianto (UU)}{Understanding Classification}
\SIKSdiss{2009-05}{Sietse Overbeek (RUN)}{Bridging Supply and Demand for Knowledge Intensive Tasks - Based on Knowledge, Cognition, and Quality}
\SIKSdiss{2009-04}{Josephine Nabukenya (RUN)}{Improving the Quality of Organisational Policy Making using Collaboration Engineering}
\SIKSdiss{2009-03}{Hans Stol (UvT)}{A Framework for Evidence-based Policy Making Using IT}
\SIKSdiss{2009-02}{Willem Robert van Hage (VU)}{Evaluating Ontology-Alignment Techniques}
\SIKSdiss{2009-01}{Rasa Jurgelenaite (RUN)}{Symmetric Causal Independence Models}







\end{document}